\documentclass[letterpaper,twocolumn,10pt]{article}
\usepackage{usenix}
\usepackage{xurl}

\newcommand{\Orca}{{\scshape Orca}\xspace}
\newcommand{\Sys}{{\scshape AccelGen}\xspace}

\usepackage{tabularx}
\usepackage{url}
\usepackage{array}
\usepackage{booktabs}
\usepackage{epsfig,endnotes}
\usepackage[ruled, lined, linesnumbered, commentsnumbered, longend]{algorithm2e}
\usepackage{multirow}
\usepackage{url}
\usepackage{subfig}
  \usepackage{wrapfig}
\newcommand{\DELF}[1]{\iffalse #1 \fi}
\newcommand{\Con}[1]{\iffalse #1 \fi}

\newcommand{\DELMayNeed}[1]{\iffalse #1 \fi}

\newcommand\tsr[1]{{\color{red}#1}}

\usepackage{amsthm}
\theoremstyle{definition}

\usepackage{color}
\usepackage{amsfonts}
\usepackage{comment}

\newcommand{\sh}[1]{\textcolor{blue}{[??: #1]}}

\newcommand{\DEL}[1]{\iffalse #1 \fi}

\newcommand\tanbin[1]{{\color{red}#1}}

\newcommand\tannsdi[1]{{\color{red}#1}}

\renewcommand{\linespread}{1}

\usepackage{mathtools}          

\newcommand{\squishlist}{
\begin{list}{$\bullet$}
  { \setlength{\itemsep}{0pt}
     \setlength{\parsep}{0pt}
     \setlength{\topsep}{0pt}
     \setlength{\partopsep}{0pt}
     \setlength{\leftmargin}{0.7em}
     \setlength{\labelwidth}{0.7em}
     \setlength{\labelsep}{0.2em} } }

\newcommand{\squishlisttwo}{
\begin{list}{$\bullet$}
  { \setlength{\itemsep}{0pt}
     \setlength{\parsep}{0pt}
    \setlength{\topsep}{0pt}
    \setlength{\partopsep}{0pt}
    \setlength{\leftmargin}{2em}
    \setlength{\labelwidth}{1.5em}
    \setlength{\labelsep}{0.5em} } }

\newcommand{\squishlistthree}{
\begin{list}{$\bullet$}
  { \setlength{\itemsep}{0pt}
     \setlength{\parsep}{0pt}
    \setlength{\topsep}{0pt}
    \setlength{\partopsep}{0pt}
    \setlength{\leftmargin}{1em}
    \setlength{\labelwidth}{1.5em}
    \setlength{\labelsep}{0.5em} } }

\newcommand{\squishend}{
  \end{list}  }

\newtheorem{thm}{\textbf{Observation}}
\usepackage[most]{tcolorbox}
\definecolor{light-gray}{gray}{0.95}
\tcolorboxenvironment{thm}{
   center,
   boxsep=1.0pt,
   top=0.0pt,
   bottom=0.1pt,
    width=\linewidth,
    colframe=light-gray,
    colback=light-gray
}

\newtheorem{thm1}{\textbf{Proposition}}
\usepackage[most]{tcolorbox}
\definecolor{light-gray}{gray}{0.95}
\tcolorboxenvironment{thm1}{
   center,
   boxsep=1.0pt,
   top=0.0pt,
   bottom=0.0pt,
    width=\linewidth,
    colframe=light-gray,
    colback=light-gray
}

\def\BibTeX{{\rm B\kern-.05em{\sc i\kern-.025em b}\kern-.08em
    T\kern-.1667em\lower.7ex\hbox{E}\kern-.125emX}}

\usepackage{cleveref}
\crefname{section}{\S}{\SS}
\crefformat{section}{\S#2#1#3} 
\crefformat{subsection}{\S#2#1#3}
\crefformat{subsubsection}{\S#2#1#3}

\usepackage{enumitem}
\setlist{itemsep=0pt,parsep=0pt,topsep=0pt}

\usepackage[compact]{titlesec}
\usepackage[font=small,labelfont=bf,textfont=bf,aboveskip=8pt]{caption}
\usepackage[all]{nowidow}
\usepackage[normalem]{ulem}
\usepackage{zref-totpages}
\newcommand*\circled[1]{\tikz[baseline=(char.base)]{
		\node[shape=circle,draw,inner sep=0.55pt] (char) {#1};}}

\begin{document}

\title{\Sys: Heterogeneous SLO-Guaranteed High-Throughput LLM Inference Serving for Diverse Applications}
\author{Haiying Shen and Tanmoy Sen\\ University of Virginia}
\maketitle



\DEL{\newpage
1/9: our paper said that saturated capacity is usually lower than the marked capacity, is there a reference to support it. what is the marked capacity of our previous tested machine, which has pivot/saturated capacity 126.96 TFLOPS.\\

1/8: for the dataset that has 100k prompt length, can you draw the same fig as Figure 11. in vLLM paper now? \\

1/9:
show me the fig, then I will decide if the following figs need to be drawn:\\
O5.9 to support O6
1.similar fig as Fig11 in vLLM paper, but X is the # of blocks for the initial prompt\\

Run vLLM on the 100k dataset:
The following figs are for O4:
2. x: # of tokens beyond pivot forward size, y: percentage of batches\\
3. x:  # of tokens beyond pivot forward size, y: PP latency for the batch\\
4. x: prompt length, y: CDF of \% of prompts that fail to receive KVC for the prompt\\
5. x: max prompt length of a request in the cache, y: CDF of \% of KVC allocation failure\\
6. x: ave prompt length of a request in the cache, y: CDF of \% of KVC allocation failure\\

The following figure is for O6:
7. x: forward size, y: CDF of batches\\
8. x: each iteration, y: several lines: GPU compute utilization, KVC allocated amount, KVC actual use amount (8.1),\\ forward length, newly added total prompt length (8.2)\\
using our methods (remove iteration-level SLO), \\
9. a similar fig as Fig11 in vLLM paper, but X is cache size (in token or in block) needed for each chunk (if a prompt is not long, we also call it chunk, so a long prompt as several chunks and a short prompt has one chunk)
some chunks do not need cache allocation if their cache is allocated along with the previous chunk\\

10. a similar fig as Fig11 in vLLM paper, but X is the chunk size in tokens\\

11.for the above 2 figs, use time sequence as X, then draw the value of each chunk in Y \\

12. also, draw another fig:
X: after each iteration, Y: target Stf-Sf (current), available cache size (in token), i.e., non-allocated cache size\\

add another fig:
x: input length range
y: average, 5th and 95th percentile of the output length

{\sh{1/20: you need to put both 13B and 175B results in the figs in Analysis section and Exp. section. Put both results in the same fig.
did you add number of OOM figs in exp. section?}}

{\sh{13.excel: for vLLM and Orca, also record the number of PP tasks, and number of TG tasks in each batch. draw a CDF fig: x=percentage of PP tasks, y=CDF of batches (i.e., iterations)}}

??Referring to the vLLM paper and recently published papers, add more related work to our related work section.

refer to vLLM paper to re-write the implementation section, you need to add more details of our proposed methods-done ??

1/10
the 3 colors in exp. section should be exactly the same as the 3 colors in fig 16 in the analysis section: blue, gray and orange
1/10/2023
Run vLLM and draw the following figs.

using our methods (remove iteration-level SLO),
then draw
a similar fig as Fig11 in vLLM paper, but X is cache size (in token or in block) needed for each chunk (if a prompt is not long, we also call it chunk, so a long prompt as several chunks and a short prompt has one chunk)
some chunks do not need cache allocation if their cache is allocated along with the previous chunk

a similar fig as Fig11 in vLLM paper, but X is the chunk size in tokens

2.for the above 2 figs, use time sequence as X, then draw the value of each chunk in Y

3. also, draw another fig:
X: after each iteration, Y: target Stf-Sf (current), available cache size (in token), i.e., non-allocated cache size
4. In our experiment, record how many preemptions occurred

\newpage
}
\setcounter{page}{1}
\begin{abstract}

\DEL{In this paper, we study a mixed-prompt
scenario for large-language model (LLM) inference systems, which comprises a majority of short
prompts and a subset of long prompts (with $\geq$4k tokens). Our experiment measurements show that in such a scenario, due to imbalanced prompt lengths, relying on first come and first serve (FCFS), existing serving systems often generate long iteration time, under-utilization and over-utilization of GPU, and key-value cash allocation failures, thereby limiting achieved throughput, request latency, and user experience. In online LLM applications, users' experience often depends on iteration time but existing work usually pays attention to service-level-objective (SLO) on job completion time (JCT). Thus, we propose a \underline{Mu}lti-\underline{R}esource-\underline{a}ware batching based serving system (\Sys). 
\Sys introduces innovations in three key aspects: 1) Iteration-level SLO, 2) SLO-guaranteed temporal GPU load balancing, 3) Iteration-level key-value cache guarantee, and 4) Multi-resource-aware batching.} 



In this paper, we consider a mixed-prompt scenario for a large language model (LLM) inference serving system that supports diverse applications with both short prompts and long prompts and heterogeneous SLOs for iteration time. 
To improve throughput when handling long prompts, previous research introduces a chunking method, but has not addressed heterogeneous SLOs. To address the limitation, we propose \Sys, a high-throughput LLM inference serving system with heterogeneous SLO guarantees for diverse applications. \Sys introduces three core components: (1) SLO-guaranteed dynamic chunking, which dynamically adjusts chunk sizes to maximize GPU compute utilization while meeting iteration-level SLOs; (2) Iteration-level SLO-based task prioritization, which prioritizes tight-SLO requests and batches requests with similar SLOs; (3) Multi-resource-aware batching, which selects queued requests to maximize the utilizations of both GPU compute resource and key-value cache (KVC). 
Trace-driven real experiments demonstrate that \Sys achieves 1.42-11.21$\times$ higher throughput, 1.43-13.71$\times$ higher goodput, 37-90\% higher SLO attainment, 
and 1.61-12.22$\times$ lower response latency compared to the state-of-the-art approaches. 
It achieves performance near the \emph{Oracle}, which optimally maximizes goodput.




\end{abstract}

\DEL{\sh{vLLM has long prompt chunking method now. Pls read their code and document, and then write here about the method-done}
{\color{red}vLLM employs a technique known as "chunked prefill" to manage long prompts efficiently. This method divides extensive prompts into smaller segments, or "chunks," which are processed sequentially. By doing so, vLLM reduces memory usage and enhances throughput during inference. Multiple prefill chunks can be added with the decoding. DUring scheduling, the decoding is prioritized over the prefill chunks. During preemption, the prefill chunk are prioritized. There is also scope to change the priority.

$enable_chunked_prefill$ needs to be set to true to enable this process. In the chunked prefill process, the prompt and context are divided into manageable lengths, with the context's length stored in $cum_prompt_context_lens$. There is a $max_num_batched_tokens$ parameter. This parameter sets the maximum number of tokens that can be processed in a single batch, encompassing both prefill and decode tokens. When the total number of tokens in a prefill request exceeds this limit, vLLM divides the prompt into smaller chunks to fit within the specified batch size. This chunking allows vLLM to interleave prefill processing with decoding tasks, optimizing GPU compute utilization and reducing latency. 

By adjusting the $max_num_batched_tokens parameter$, users can control the balance between prefill and decode operations in each batch, tailoring the system's performance to specific workload requirements. This approach allows vLLM to handle prompts that exceed typical length limitations by processing them in parts, thereby optimizing performance and resource utilization.} 
}

\section{Introduction}\label{sec:introduction}

 \DEL{At the core of these generative models lies the Transformer model, which uses an attention mechanism. By using the attention mechanism, Transformer models can learn better representations where each token (or word) of the sequence may have a direct connection with every other token, which was not possible in the previous recurrent models such as Long Short-Term Memory (LSTM).}

Transformer-based generative large-language models (LLMs) have garnered considerable attention in Natural Language Processing (NLP) applications such as text generation~\cite{Zhang2022OPTOP}, text summarization~\cite{textsummary}, code generation~\cite{codegeneration}, and chatbot~\cite{chatbot}. In the model inference, each request begins with a user-entered prompt,  comprising a sequence of tokens. The model processes this prompt in the first iteration to generate the initial token, and subsequently generates tokens in an autoregressive manner~\cite{280922} in each iteration until the entire response text is produced. Requests are processed in batches, including the prompt processing (PP) tasks and the token generation (TG) tasks.

\DELMayNeed{That is, the server needs to run the LLM model as many times as the number of tokens to generate.} \DELMayNeed{This is different from other models such as ResNet~\cite{he2016deep} for image  and BERT~\cite{devlin2018bert} for text classification that process an inference request by running the model only once.} 
\DEL{in which the requests are classified from the pre-trained model as they do not have dependency on the previous tokens to generate a new one.}


\DEL{Inference servers, such as Triton~\cite{fang2021deployment} and TensorRT~\cite{olston2017tensorflow}, 
provide configuration options for the number of inference requests in a batch (batch size) and schedule requests within a batch. They wait to schedule the next batch after the current one is executed. However, this request-level scheduling approach results in long request latency since requests that finish earlier within a batch cannot return to the client, while newly arrived requests must wait until the current batch completes. To address this issue, the iteration-level scheduling system \Orca~\cite{280922} schedules requests at the iteration level, utilizing the first-come-first-serve (FCFS) policy to form a batch. 
\Orca pre-allocates key-value cache (KVC) for the maximum sequence length, which can range from 8K to 32K~\cite{OpenAI5}, for each prompt. However, this approach wastes memory~\cite{jin2023s}, limits the batch size and hence GPU compute utilization (e.g., 0.4\%~\cite{jin2023s}) and throughput. vLLM~\cite{vllm}, employing FCFS, was introduced to address the limitations of \Orca by dividing the KVC into blocks and allocating blocks to a request, which are not necessarily stored in a contiguous space. In cases of KVC overflow, it preempts requests and evicts their KV values from the KVC to the CPU or recomputes the KV values.
}


Given the diverse applications running on an LLM inference serving system, prompt lengths can vary significantly, ranging from just a few tokens for chat applications to 4K-100K tokens for more extensive content like electronic books used in text summarization, code generation, and legal document analysis~\cite{soskkobayashi2018bookcorpus}. 
Additionally, user experience is critical; users may have different iteration time service-level-objectives (SLOs) in online applications based on their reading speed and application requirements, with a typical reading speed around 0.1875 seconds per token~\cite{timetoken,OpenAIAPI,jin2023s}. Offline applications, on the other hand, may impose job completion time (JCT) SLOs. Thus, in this paper, we consider a mixed-prompt scenario, consisting of short prompts and long
prompts (containing $\geq$4K tokens) and heterogeneous SLOs.

To improve throughput when handling long prompts, previous research introduces chunking~\cite{Agrawal2023SARATHIEL,fastgen,vllm,vLLM-long}. 
The approaches select chunks or requests from the waiting queue until the target forward size (or token budget) -- defined as the desired total number of tokens in a batch -- is reached, to fully utilize the GPU compute resource. 
However, our trace-driven measurement analysis (\cref{sec:analysis}) reveals that these systems fail to meet heterogeneous iteration-level SLOs, potentially degrade user experience, and fall short of maximizing GPU compute and key-value cache (KVC) utilization. Specifically, our findings indicate that:
\squishlist 


\item[(1)] Heterogeneous iteration-level SLOs cannot be met in current LLM systems, making it necessary to account for SLO diversity in scheduling.  

\item[(2)] Focusing solely on maximizing GPU compute resources does not guarantee maximum throughput for a workload. It is essential to jointly maximize both GPU compute utilization and KVC utilization at each iteration to improve throughput.\looseness=-1

\item[(3)] We should limit the number of concurrently running long-prompt requests to free up KVC resources more quickly to improve throughput.

\item[(4)] 
The first-come-first-serve (FCFS) policy used in current LLM systems~\cite{298679,vllm} underutilizes GPU resources when meeting iteration-level SLOs, but batching requests with similar SLOs enhances throughput.

\item[(5)] The varying available GPU compute resource and unallocated KVC space after each iteration, coupled with the diverse GPU and KVC demands of different prompts, provide an opportunity to identify prompts or prompt chunks to be added to the batch to maximize both GPU compute and KVC utilizations, thus improving throughput. 

\squishend


\DEL{in such a scenario, due to the imbalance in
prompt lengths, existing serving systems~\cite{280922,vllm} employing FCFS often exhibit prolonged
iteration time, GPU underutilization and overutilization, as
well as KVC allocation failures of long prompts hence KVC underutilization (since FCFS stops
adding prompts to the batch if the first prompt cannot fit
into the KVC). 
These issues limit the achieved
throughput, job completion time (JCT), and potentially deteriorate user experience.
}

\DEL{The prolonged iteration times potentially deteriorate user experience in online LLM applications, yet existing research predominantly focuses on service-level objectives (SLOs) concerning JCT.}

\DEL{a long prompt generates long PP time and delays the PP of other requests within the same batch. This delay increases the user waiting time for a token, thereby degrading overall user experience. For instance, authors submitting prompts to ChatOPT often experience prolonged waiting times for responses. 
Furthermore, long prompts occupy a significant portion of the KVC during their TG phases, exacerbating the challenge posed by the limited KVC.}

\DEL{Indeed, the prolonged iteration time can significantly impact user experience, particularly in online LLM applications, given that the normal reading speed is approximately 0.1875 seconds per token~\cite{timetoken,OpenAIAPI,jin2023s}.
In previous research, the service-level-objective (SLO) typically defines the JCT or the time it takes for a request to be completely processed by the LLM model~\cite{Jin2023S3IG}. However, even if the JCT SLO is met (e.g., 2s for a response with 10 tokens), users may still experience considerable wait time for individual tokens (e.g., 0.5s). 
}

\DEL{An intriguing challenge arises: \emph{Can we transform long prompts from foes to friends by mitigating their adverse effects and simultaneously addressing the inherent problem of low throughput in the mixed-prompt scenario?} Despite previous research efforts~\cite{280922,vllm, Jin2023S3IG,Zheng2023ResponseLP,sheng2023high} aimed at enhancing performance, to the best of our knowledge, there has been no dedicated research specifically addressing the effective handling of long prompts or leveraging them to improve overall system performance.}

\DELMayNeed{\Orca parallelizes the batch execution in non-attention layers, and processes the requests in a batch sequentially in the attention layer due to their different sequence lengths.}

\DEL{As a result, transformer-based LLMs are often
limited by GPU memory (memory in short) capacity and bandwidth, resulting in underutilization of GPU compute resources (GPU in short) (e.g., 0.4\%)~\cite{jin2023s}.
To address this problem, vLLM~\cite{vllm} was proposed to that divides the
KVC into blocks and allocate blocks to a request are not necessarily stored in contiguous space.}

\DELMayNeed{{\color{gray}{On the other hand, transformer-based LLMs have inherent problems of failing to utilize multi-resources concurrently, thus constraining throughput.
PP tasks are computation heavy while TG tasks are memory heavy. Using FCFS, \Orca may have most PP tasks in a batch (resulting in 19.7\% memory utilization and 81.3\% GPU compute utilization), or have most TG tasks in a batch (resulting in 27.4\% GPU compute utilization and 87.6\% memory utilization) (O\ref{comb}). vLLM's statistics...??}}}


\DELMayNeed{{\color{gray}{\noindent\textbf{Exacerbate the memory bottleneck.}
\Orca pre-allocates key/value (KV) cache for the maximum sequence length (which can be from 8K to 32K~\cite{OpenAI5}) for each prompt. It wastes memory~\cite{jin2023s}, limits batch size and throughput. The KVC allocation is 79.6\%, but 70\% will not be used on average (Observation(O)\ref{kvcache2}).

\noindent\textbf{Fail to tackle long prompts.}
First, long-prompt processing delays all other requests in the same batch by {3$\times$} (O\ref{other}). Second, a long prompt can exacerbate the memory bottleneck as it can increase the allocated KVC space by {25}\%.
Third, after a long prompt is processed and becomes a TG task with one token input, 
it can change GPU compute utilization from {60}\% to {26}\%. However, PP of one long prompt can help increase GPU compute utilization (O\ref{other}).

\noindent\textbf{Fail to fully utilize multi-resources concurrently.}
PP tasks are computation heavy while TG tasks are memory heavy. Using FCFS to form a batch reaching the configured batch size, existing transformer-based LLMs may have most PP tasks in a batch (resulting in 19.7\% memory utilization and 81.3\% GPU compute utilization), or have most TG tasks in a batch (resulting in 27.4\% GPU compute utilization and 87.6\% memory utilization) (O\ref{comb}). 
}}}

\DEL{To seek the problem solution, we first conducted and measurement analysis based on real traces and observed that:

\squishlist

\item The throughput increases with the forward size until a certain point but the majority batches have short forward sizes (O\ref{pivot}). 

\item Existing LLM systems fall short in handling long prompts, leading to delayed iteration time (and hence degraded user experience) and overloading the KVC (O\ref{longprompt} and O\ref{kvcache2}).

\DEL{\item Long prompts not only overload GPU but also tend to be failed to allocate to KVC and also overflow KVC.

\item Chunking a long-prompt can reduce KVC use (e.g., 16$\times$) and processing latency, and meanwhile help increase the GPU compute utilization and throughput (e.g., 1.8$\times$) by batching with TG tasks while avoiding overflowing the KVC. 
}
\squishend

}

\DEL{\item Long prompts make ??\% waiting times of a token longer than 0.1875s.}

\DEL{{\color{gray}{\item
Combining PP requests (e.g., a long-prompt chunk) and TG requests in a batch can fully utilize GPU and GPU memory, thus improving throughput and request latency. Therefore, long-prompt chunks become TG requests' friends by 
helping increase GPU compute utilization (O\ref{comb}).

\item Iteration-level cache guarantee can help reduce cache bottleneck while still guaranteeing the KVC need from all requests in a batch (O\ref{cache}).
}}}

Building on these key insights, based on the chunking method of FastGen, we propose \Sys, a high-throughput LLM inference serving system designed to meet heterogeneous SLOs while maximizing throughput for diverse applications. \Sys incorporates the following methods to achieve the goal: 

\squishlist



\item[(1)] \textbf{SLO-guaranteed dynamic chunking.} \Sys determines the token budget and batches requests with
similar SLOs to maximize GPU compute utilization as much as possible while satisfying the iteration-level 
 SLOs of the requests in the batch. In addition, it limits concurrently running long-prompt requests to improve throughput. 


\item[(2)] \textbf{Iteration-level SLO-based task prioritization.}
\Sys enables users to specify the SLOs for Time-To-First-Token (TTFT) in PP and for Time-Between-Tokens (TBT) in TG of a job, or specify the JCT SLO for a job. 
It converts a JCT SLO into iteration-level SLOs. Based on iteration-level SLOs, it orders the waiting requests to enhance SLO compliance and facilitate batching requests with similar SLOs to enhance throughput.



\DEL{\Sys eliminates the batch size constraint and focuses on making the token size in a batch to maximize GPU compute utilization while avoiding overloading GPU and meeting the iteration-level SLO for inference time. 

It dynamically partitions long prompts to achieve this objective.
matching the forward size ($S_{f}$) to the token budget ($S_b$), maximizing GPU compute utilization while meeting the iteration-level SLO for inference time. It dynamically partitions long prompts to achieve this objective.
}


\DEL{\item[(3)] \textbf{Multi-resource-aware batching.} 
Instead of using FCFS, \Sys selectively chooses prompts that fit within the KVC, skips those exceeding the capacity, and adjusts prompt 
chunks to ensure that the KVC is fully allocated. 
}





\item[(3)] \textbf{Multi-resource-aware batching.} After each iteration, \Sys selects requests and adjusts prompt chunk lengths from the prioritized waiting queue to maximize both GPU compute utilization and KVC utilization in the subsequent iteration.  




\DEL{\item \textbf{Parallelism in the attention layer}. \Sys batches the requests at different iterations in such a way so that the attention operation can be parallelized.(??briefly explain this method)}


\squishend
Our tace-driven real experiments show that \Sys achieves 1.42-11.21$\times$ higher throughput, 1.43-13.71$\times$ higher goodput, 37-90\% higher SLO attainment, 
and 1.61-12.22$\times$ lower response latency compared to the state-of-the-art approaches. This is the first paper on heterogeneous iteration-level SLOs, a topic of interest to industry for practical applications. 

\section{Background and Theoretical Founcation}\label{subsec:background}

\subsection{Background}

\DEL{\begin{table}
    \centering
\caption{Notations}
\label{tab:NotationTable}
    \footnotesize
    \linespread{1.1}\selectfont
    \begin{tabular}{c|m{6.cm}}
    \hline
    \multicolumn{1}{c|}{\textbf{Variable}} & \multicolumn{1}{c}{\textbf{Definition}} \\
  \hline
  $I_k^j$ & the $k^{th}$ request in the $j^{th}$ iteration \\
  \hline
  $M(I_j^k)$ & KVC demand of $I_k^j$\\
  \hline
  $L_k^j$& the iteration-level SLO of $I_k^j$\\
  \hline
  $b_k^j$ & indicate whether $I_k^j$ is selected\\
  \hline
  $c_k^j$ & indicate whether $I_k^j$'s SLO is satisfied\\
  \hline
  $S_f$ & forward size (num of tokens in the input sequence)\\
  \hline
  $S_b$ & token budget\\
  \hline
  $S_{b}$ & pivot forward size\\
  \hline
  $S_g$ & size of generated tokens\\
  \hline
  $B$ & batch size (num of requests in a batch)\\
  \hline
   $S_b$ & num of tokens a block in KVC\\
  \hline
  $\bar{T}_{pf}$& execution time of a batch with $S_{b}$\\
  \hline
 $S_{l}$/$S_{l}^r$& sequence size of a request $r$\\
  \hline
 $S_{I}^r$&Input sequence length of a request $r$\\
  \hline
  $N_{ck}$ & num. of chunks of an input sequence\\
  \hline
  $T_e$ & a PP or TG task's execution time\\
  \hline
  $T_w$ & a task's waiting time\\
  \hline
  $T_r$ & remaining time of the task in an iteration\\
  \hline
  $P_{max}$ & maximum time length of a preemption\\
  \hline
  $P$ & probability for a request to be preempted\\
  \hline
  $N_r$ & num of remaining chunks and predicted tokens to generate\\
  \hline
  $\textbf{W}$ & waiting queue ordered by remaining time \\
  \hline
  $\textbf{U}$ & urgent requests in queue\\
  \hline
  $\textbf{G}$ & a group of selected requests\\
  \hline
  $A_{KV}$ & available KVC space\\
   \hline
  $D_{KV}$& KVC space demand\\
   \hline
  $A_{GPU}$ & available GPU computation resource\\
   \hline
  $D_{GPU}$ & GPU computation resource demand\\
  \hline
\end{tabular}
\end{table}
}

\begin{figure}
    \centering
\includegraphics[width=0.85\columnwidth,height=0.18\textheight]{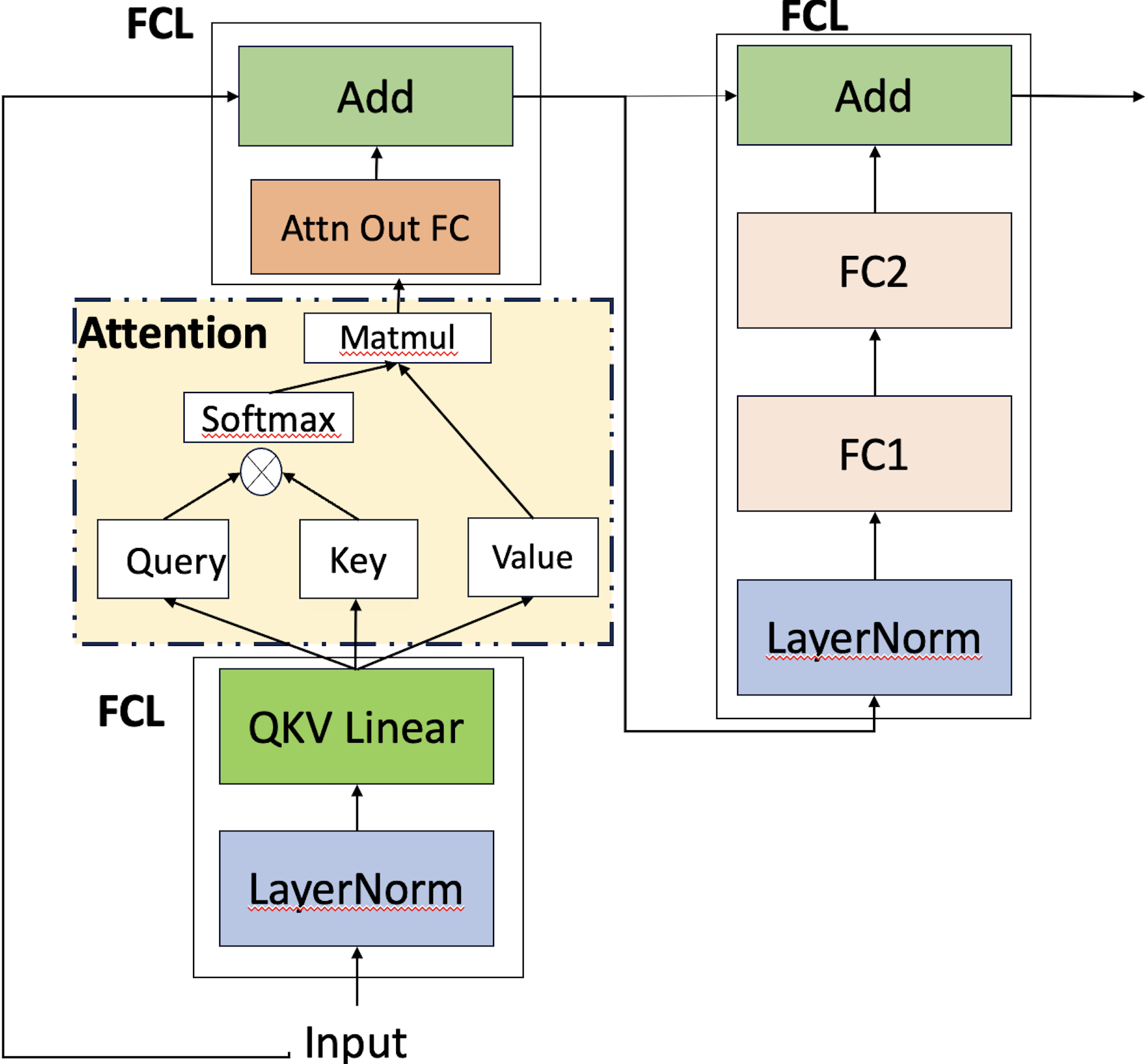}
    \caption{Transformer layer of OPT. }
    \label{fig:transformer}
\end{figure}

 As a representative of the transformer model, we illustrate the OPT structure in Figure~\ref{fig:transformer}. OPT is a decoder-only transformer model and its layers fall into two categories: 1) forward computation layers (FCL), and 2) attention layer. 
\DEL{As shown in Figure~\ref{fig:transformer}, before the attention layer, OPT transformer consists of the layer normalization operation (LayerNorm) and the QKV Linear (linear and split operations to get the query, key, and value). Operations performed after Attention are, in order, a linear operation (Attn Out Linear), an add operation for residual connection (Add), layer normalization operation (LayerNorm), the fully connect layers (FC) operations, and the other residual connection operation (Add). These layers can be divided into two categories: 1) forward computation layers (FCL), and 2) attention layer (marked within a dashed box) in Figure~\ref{fig:transformer}. 
}The attention layer takes three input components: Query ($\mathbf{Q}$), Key ($\mathbf{K}$), and Value ($\mathbf{V}$). These components are obtained by projecting the embeddings ($\mathbf{E}$) using three learned matrices $\mathbf{W_q}$, $\mathbf{W_k}$, and $\mathbf{W_v}$ in the transformer: \DEL{The embedded matrix for the whole token sequence input is $\mathbf{E}$, where $e_i$ is the embedding of each token in the sequence. 
Then, the projection is created by}\vspace{-0.05in}
\begin{equation*}
    \mathbf{Q} = \mathbf{W_q} \mathbf{E} \text{, }
    \mathbf{K} = \mathbf{W_K} \mathbf{E} \text{, }
    \mathbf{V} = \mathbf{W_v} \mathbf{E} \vspace{-0.05in}
\end{equation*}
At the layer level, each matrix multiplication is a fully connected layer with input and output neurons. Then its number of operations equals:\vspace{-0.15in}
\begin{equation}\label{equ:opeNum1}
2 \times S_f \times H^2, \vspace{-0.05in}
\end{equation}where $S_f$ is forward size and $H$ is the model dimension. 

The attention module includes the dot products of the query tokens in $\mathbf{Q}$  with all the keys of previous tokens in $\mathbf{K}$ to measure the similarity of previous tokens from the new token's perspective: $Similarity(\mathbf{Q,K}) = \mathbf{QK}^T$. The attention operation 
produces the output: \vspace{-0.1in}
\begin{equation}
    \label{eq:6}
    Self-attention(\mathbf{Q,K,V}) = softmax(\frac{\mathbf{QK}^T}{\sqrt{d_k}}) \mathbf{V}, \vspace{-0.05in}
\end{equation} where 
$d_k$ is the dimension of the $\mathbf{K}$ or $\mathbf{Q}$. 
The number of operations at the layer level equals: \vspace{-0.05in}
\begin{equation}\label{equ:opeNum2}\vspace{-0.01in}
4 \times S_f^2 \times H. \vspace{-0.05in}
\end{equation}

The FC2 layer outputs logits, which are passed through the Add to generate the probabilities of words in the vocabulary. 
In the Transformer model, the above operations are executed by each attention head in parallel. Finally, the results from all attention heads are concatenated. 


\DEL{\begin{figure}[htb]
\centering
    \subfloat[Iteration time.\vspace{-0.05in}\label{fig:time-long-seq}]{{\includegraphics[width=0.48\linewidth,height=0.15\textheight]{Fig/time-iter.png} }}
    \hfill
    \subfloat[GPU compute utilization.\vspace{-0.05in}\label{fig:gpu-long-seq}]{{\includegraphics[width=0.48\linewidth,height=0.15\textheight]{Fig/gpu-util-iter.png} }}
    \hfill
   \DEL{  \subfloat[Allocated memory.\vspace{-0.05in} \label{fig:long-seq-mem-alloc}]{{\includegraphics[width=0.24\linewidth,height=0.15\textheight]{Fig/gpu-util-iter-up-2.png} }}%
 \hfill
  \subfloat[Used memory.\vspace{-0.05in} \label{fig:long-seq-mem}]{{\includegraphics[width=0.24\linewidth,height=0.15\textheight]{Fig/used-memory-iter.png} }}%
 \hfill}

\vspace{-0.05in}   \caption{\small{(Fake)Performance for different prompt lengths for BookCorpus.\vspace{-0.1in}}}
    \label{fig:long-seq-13b-bc}
\end{figure}}

\begin{figure*}[t]
\centering
    \subfloat[W/o prompt chunking.\vspace{-0.05in}\label{fig:prompt-to-token-1}]{{\includegraphics[width=0.41\linewidth,height=0.14\textheight]{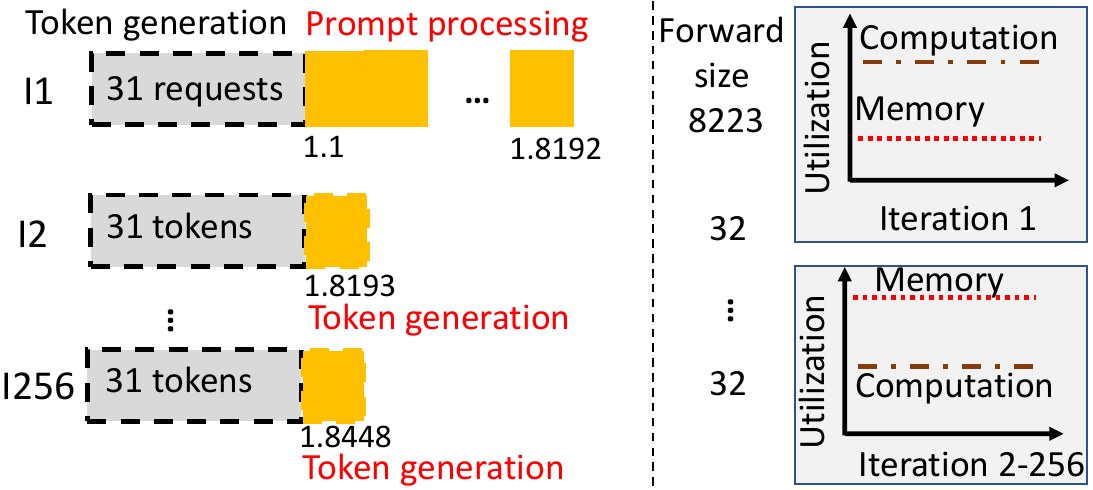} }}%
    \hfill
\subfloat[W/ prompt chunking.\vspace{-0.05in}\label{fig:prompt-to-token-2}]{{\includegraphics[width=0.41\linewidth,height=0.14\textheight]{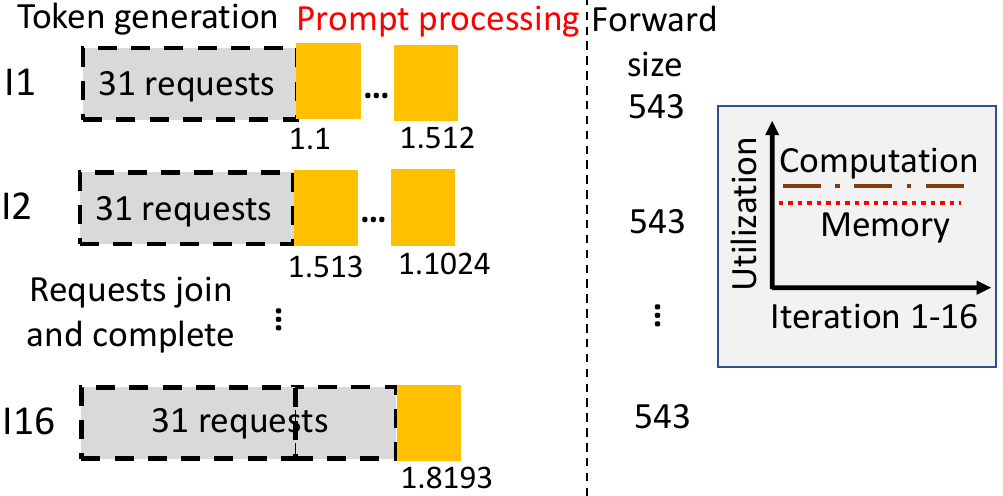} }}%
    \hfill
\vspace{-0.00in}
   \caption{\small{Illustration of chunking a long prompt. \vspace{-0.2in}}}%
    \label{fig:motivation}
\end{figure*}

\subsection{Theoretical Foundation}

\noindent\textbf{Throughput}. For the Transformer layer, the primary throughput contributors are the FCL layers (QKV, Atten Out, FC1, FC2) and Attention. Neglecting the negligible operations of other layers like Add and LayerTransform, an FCL layer's number operations is given by Equation~\eqref{equ:opeNum1}. Since each layer is based on a multiple of the model dimension, the total impact of each layer is ($3 (QKV) + 1 (Atten out) + 4 (FC1) + 4 (FC2)) = 12 \times 2 \times H^2$.
\DEL{Attention is a little more complex, particularly for generations. However, for a single sequence, attention has 2 batch matrix multiplications with heads matrix multiplications and with size, $head\_size \times SL \times SL$,  where $SL$ defines the input sequence length, i.e., number of tokens, $tokens$. This will give $4 \times hidden\_size \times SL \times SL$. So}
Based on Equations~\eqref{equ:opeNum1} and~\eqref{equ:opeNum2}, the total operations for a single transformer layer (assuming only one sequence) is ($24 \times S_f \times H^2 + 4 \times H \times S_f^2$), which can be written as~\cite{narayanan2021efficient}:
\vspace{-0.15in}
\begin{equation}
\label{eq:totop}
24 \times S_f \times H^2 \times (1 + \dfrac{S_f} {(6 \times H)}).  \vspace{-0.05in}
\end{equation}
The first addend comes from the FCL operations, and the second number is derived from the attention operations. 
For short sequence lengths ($S_{f}$$<<$$6 \times H$), it is reasonable to approximate the total operations by considering only this first addend. 

\DEL{\begin{thm1}
\label{Guidance1} The throughput almost linearly depends on forward size. 
\end{thm1}}

\DEL{Decoding is the cost of a single matrix multiplication, but since the vocab size can be very large, it can also be costly: $2 \times S_f \times H \times vocab\_dim$, where $vocab\_dim$ represents the size of the vocabulary (??which part in the fig has this cost- we don't show the vocabulary size in the figure, the output probability is generated for every word in the vocabulary.). {\sh{do we need this paragraph?}}
}

\noindent\textbf{KVC allocation}. In an iteration, the KV values of all the tokens from a step are calculated and stored in the KVC. In the subsequent iteration, only the KV tensors of the newly generated token need computation, while others are loaded from the KVC. 
The byte storage per token equals~\cite{kv-cache}: \vspace{-0.05in}
\begin{equation}\label{equ:cache} 2\times 2\times \# \text{ of transformer layers} \times H. \vspace{-0.15in}\end{equation}

A request $i$'s sequence length (denoted by $S_l^i$) is the sum of its prompt length and the number of generated tokens so far. The vLLM approach~\cite{vllm} mitigates the KVC bottleneck problem in \Orca by using a block-based approach.
Assume the block size is $b=128$.
When a token is generated, a block is created for this token and the next 127 tokens. Next, when the $129^{th}$ token is generated, another 128-token block is created.
The allocated KVC space for a batch $\mathbf{B}$ in \Orca equals: $|\mathbf{B}|\cdot S_{l_{max}}$, in which $S_{l_{max}}$ is the maximum sequence length and $|\mathbf{B}|$ is batch size. In vLLM, it equals: $\sum_{i\in {\mathbf{B}}} \lceil \frac{S_l^i}{b} \rceil \cdot b$.
\Sys is built based on the vLLM system.

\DEL{\vspace{-0.1in}
\begin{thm1} \label{Guidance2} 
The used KVC space of a request depends on its current sequence length. 
\end{thm1}
\vspace{-0.1in}
}

\DEL{\begin{figure*}[t]
\centering
    \subfloat[No chunking.\label{fig:prompt-to-token-1}]{{\includegraphics[width=0.48\linewidth,height=0.145\textheight]{Fig/no-chunk-2.png} }}%
    \hfill
    \subfloat[Chunking.\label{fig:prompt-to-token-2}]{{\includegraphics[width=0.48\linewidth,height=0.145\textheight]{Fig/chunk-2.png} }}%
    \hfill

   \caption{\small{Long prompts become foe by increasing request latency and reducing throughput (?need to think whether add squares).}}%
    \label{fig:motivation}
\end{figure*}}

\subsection{Chunking on Long Prompts}
Let's see an example to undertand the detrimental effects of long prompts and benefits of chunking. 
Figure~\ref{fig:motivation} (a) illustrates the processing of long-prompt requests for an 8192-token prompt with 256 generation steps. In the figure, $x.y$ means the $x^{th}$ prompt's $y^{th}$ token. The batch size is 32. In the batch, one request is PP, and the other 31 requests are either short PP or TG steps. The long prompt processing task consumes a significant amount of the GPU resource and takes a long time to complete, as per Equation~\eqref{eq:totop}. Consequently, the iteration time is prolonged, delaying other requests in the batch. Also, after the PP task becomes a TG task, each iteration has 32 forward size, leading to significant underutilization of GPU. 
Figure~\ref{fig:prompt-to-token-2} illustrates the scenario where we divide the long prompt into multiple 512-token chunks. The first batch takes around $\times 16$ less time to process. The remaining chunks are added to the batches in subsequent iterations. The forward size of each batch is 543. Therefore, in each iteration, the GPU computation resource is more fully utilized compared to the no-chunking case. Chunking distributes the GPU load across 256 iterations, reducing iteration time for long prompts and increasing throughput.

\begin{figure*}[t]
\centering
\subfloat[CDF of batches vs. iteration time.\vspace{-0.01in}\label{fig:14-3}]{{\includegraphics[width=0.24\linewidth,height=0.145\textheight]{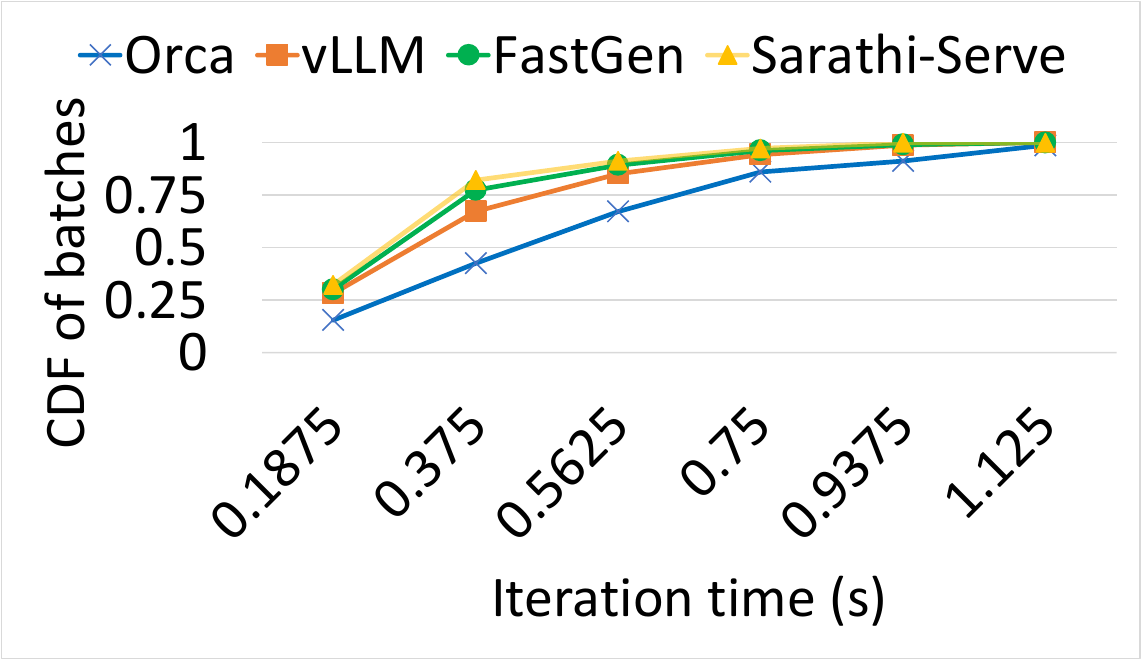} }}
    \hfill
    \subfloat[GPU compute utilization.\vspace{-0.01in}\label{fig:batch-capacity}]{{\includegraphics[width=0.24\linewidth,height=0.145\textheight]{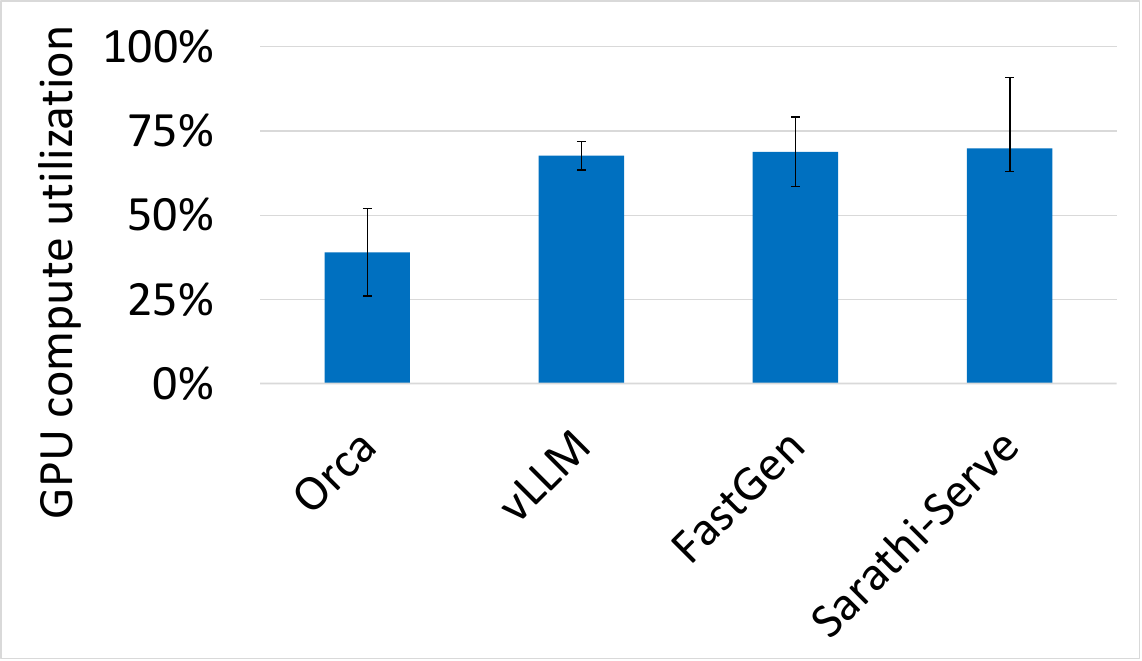} }}
    \hfill
    \subfloat[KVC utilization.\vspace{-0.01in}\label{fig:kvc-capacity}]{{\includegraphics[width=0.24\linewidth,height=0.145\textheight]{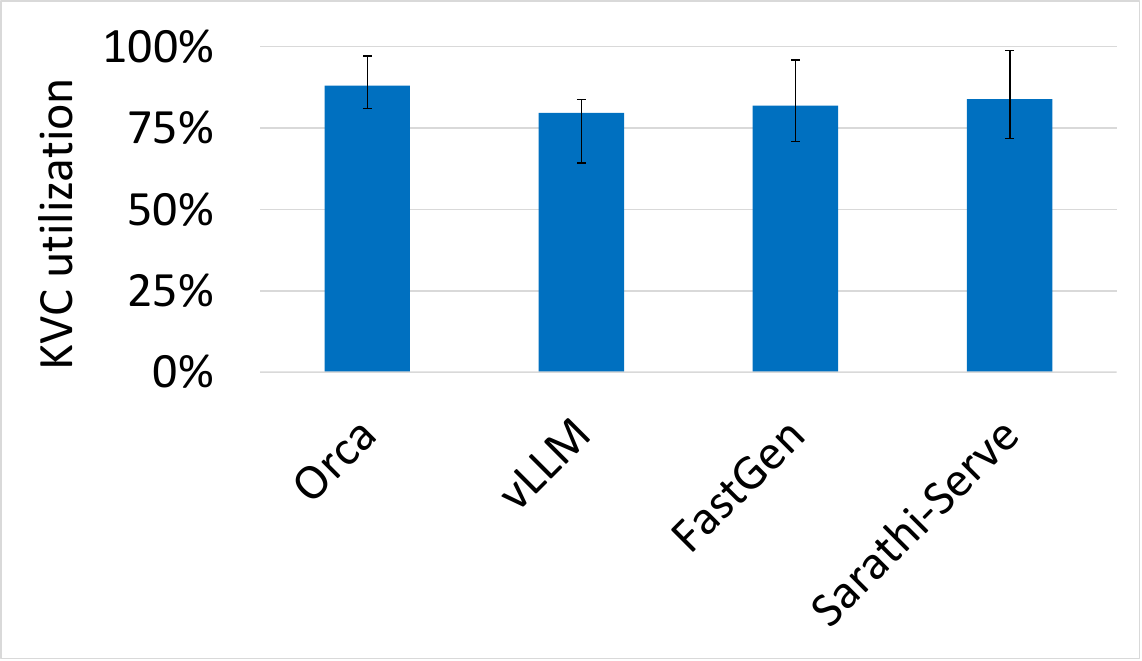} }}
    \hfill
    \subfloat[SLO attainment.\vspace{-0.01in}\label{fig:slo-methods}]{{\includegraphics[width=0.24\linewidth,height=0.145\textheight]{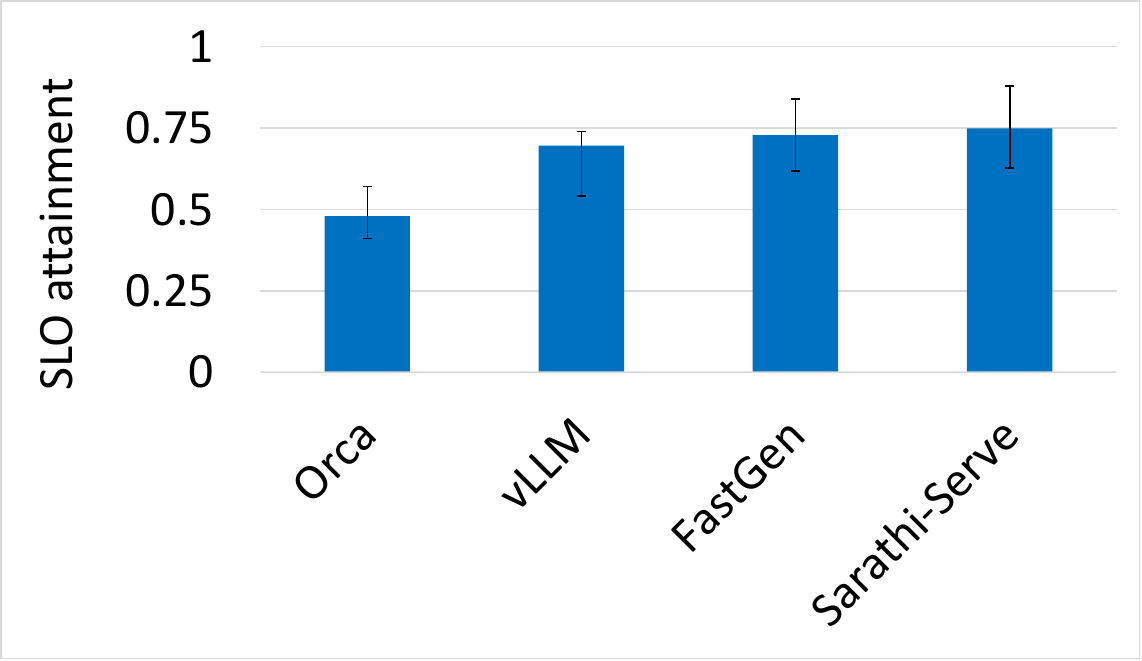} }}
    \hfill
    \DEL{\subfloat[Iteration time vs. forward size.\vspace{-0.01in}\label{fig:forward-size-it}]{{\includegraphics[width=0.32\linewidth,height=0.145\textheight]{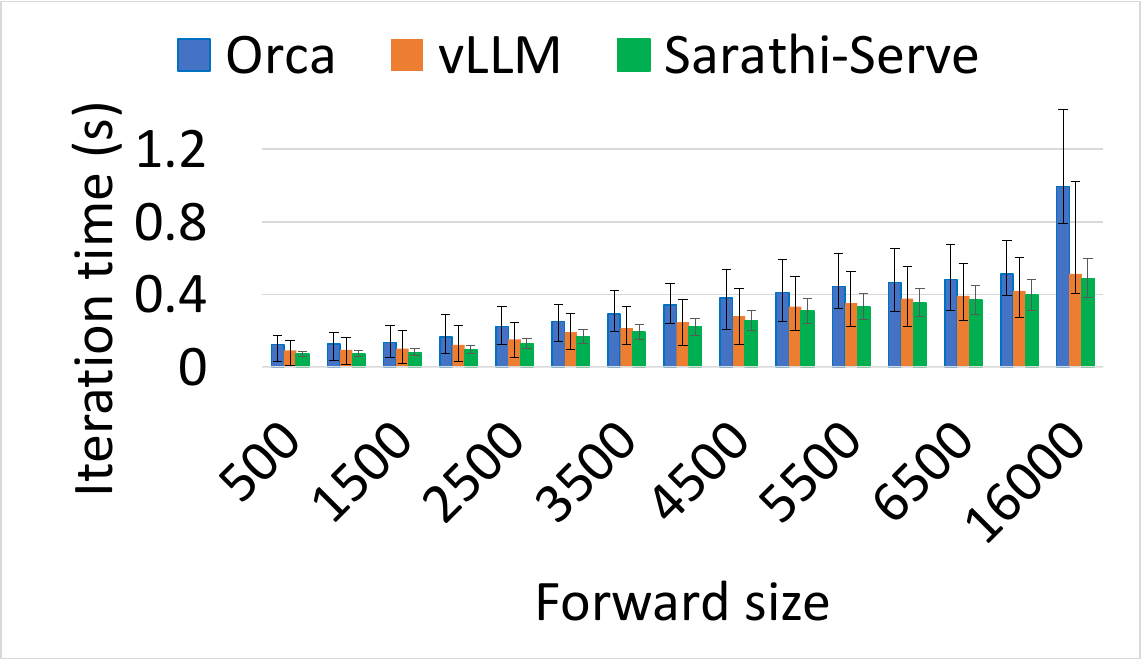} }}
    \hfill}
\DEL{\subfloat[CDF of batches vs. forward size.\vspace{-0.01in}\label{fig:14-1}]{{\includegraphics[width=0.32\linewidth,height=0.15\textheight]{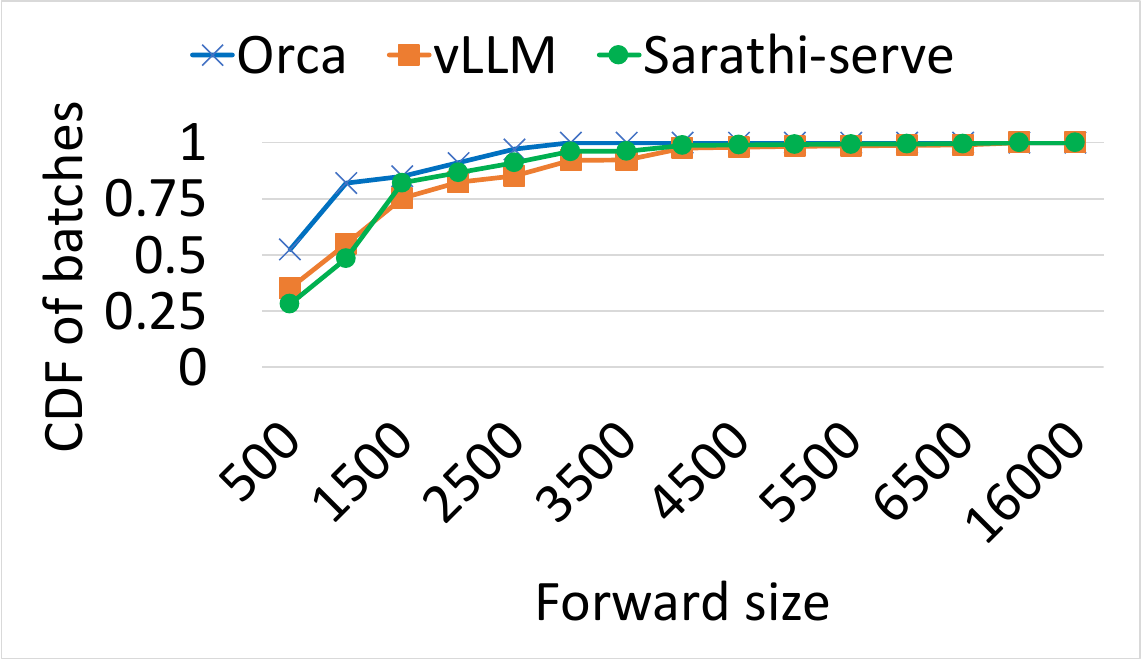} }}
    \hfill}
    \DEL{\subfloat[Prompt length distribution.\vspace{-0.01in}\label{fig:14-2}]{{\includegraphics[width=0.32\linewidth,height=0.145\textheight]{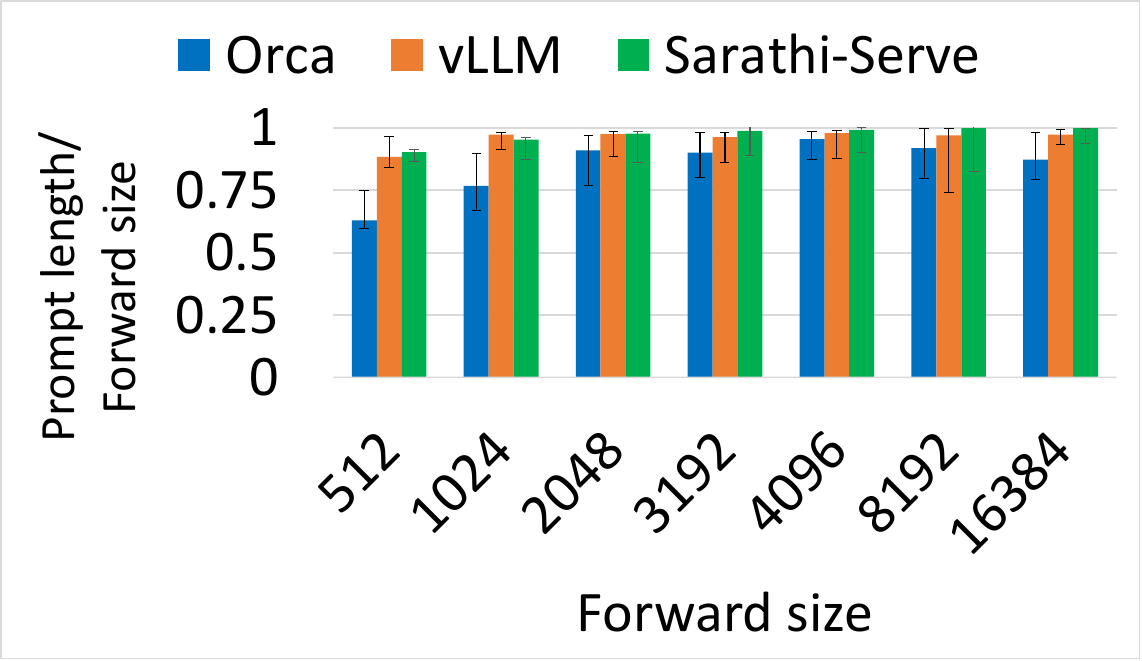} }}
    \hfill}
    \DEL{\subfloat[SLO attainment (SSR).\vspace{-0.01in}\label{fig:slo-methods}]{{\includegraphics[width=0.32\linewidth,height=0.145\textheight]{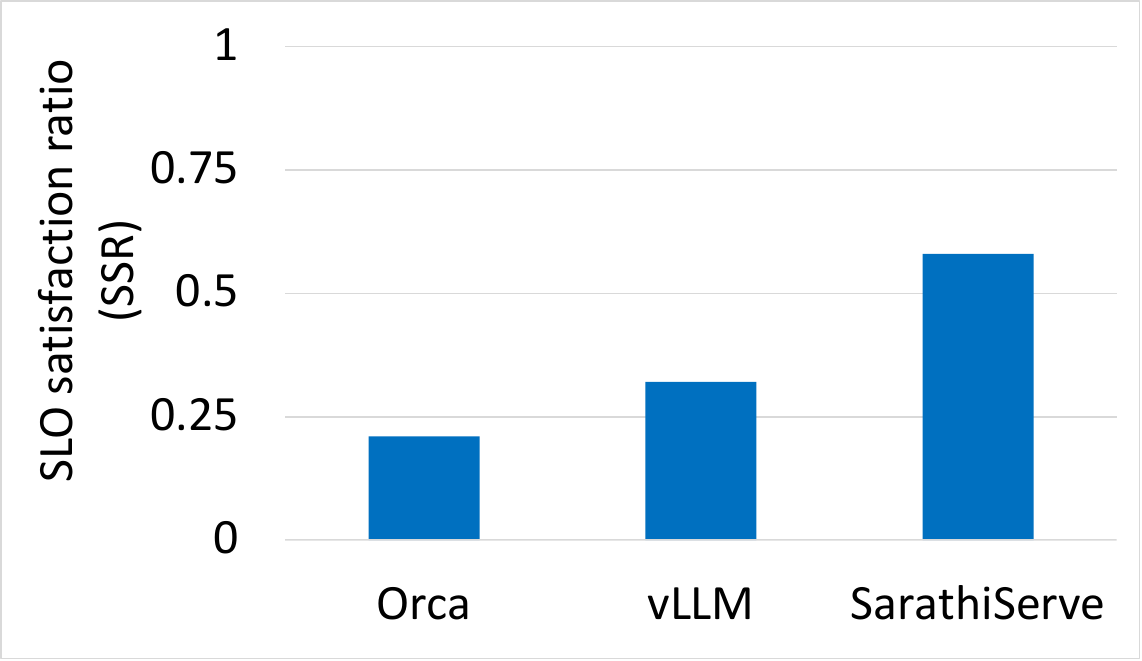} }}
    \hfill
    \subfloat[forward size/Capacity.\vspace{-0.01in}\label{fig:batch-capacity}]{{\includegraphics[width=0.32\linewidth,height=0.145\textheight]{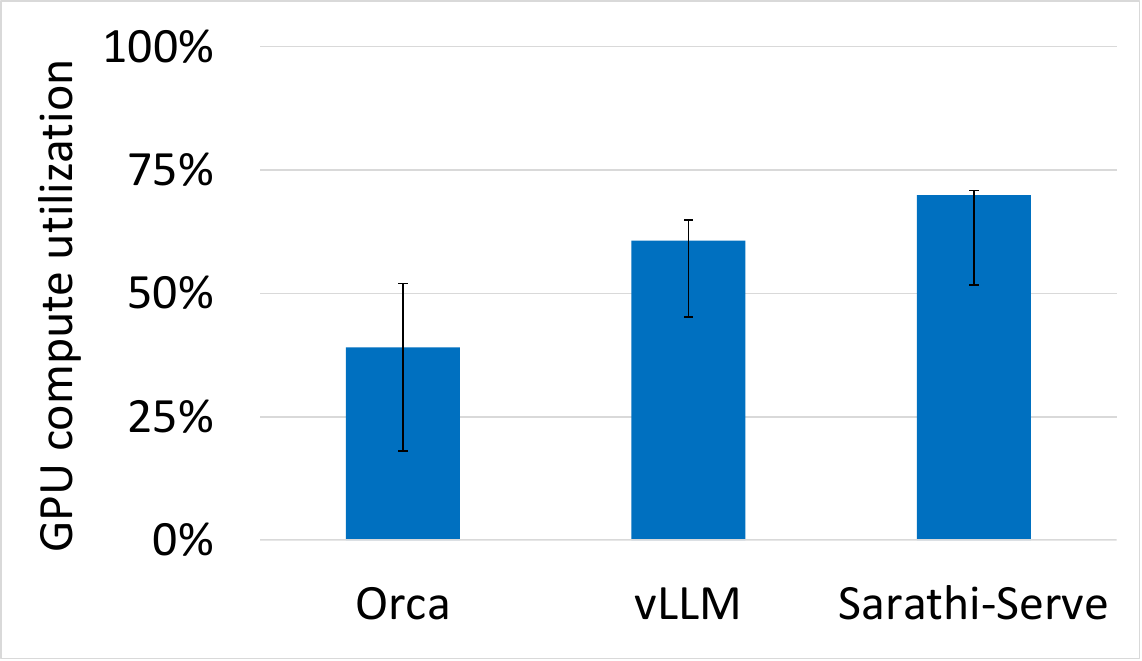} }}
    \hfill
    \subfloat[Num. of iterations of the occations/Total
Num. of iterations.\vspace{-0.01in}\label{fig:case-analysis}]{{\includegraphics[width=0.32\linewidth,height=0.15\textheight]{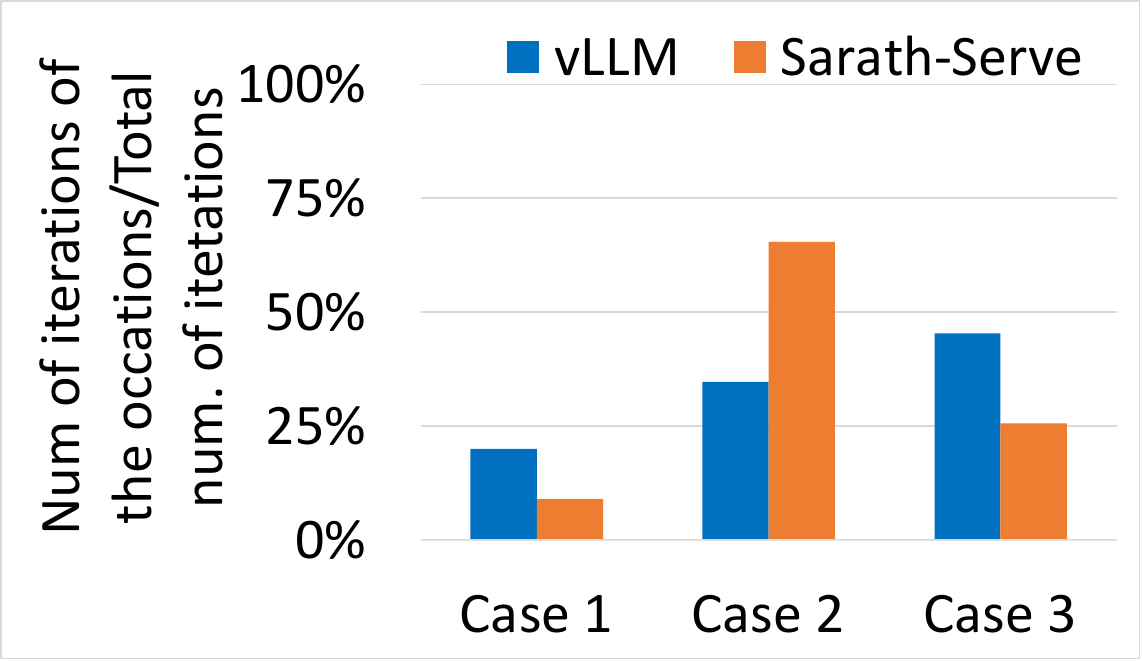} }}
    \hfill
    \subfloat[CDF of difference of chunk size and available GPU/KVC.\vspace{-0.01in}\label{fig:case-analysis-2}]{{\includegraphics[width=0.32\linewidth,height=0.145\textheight]{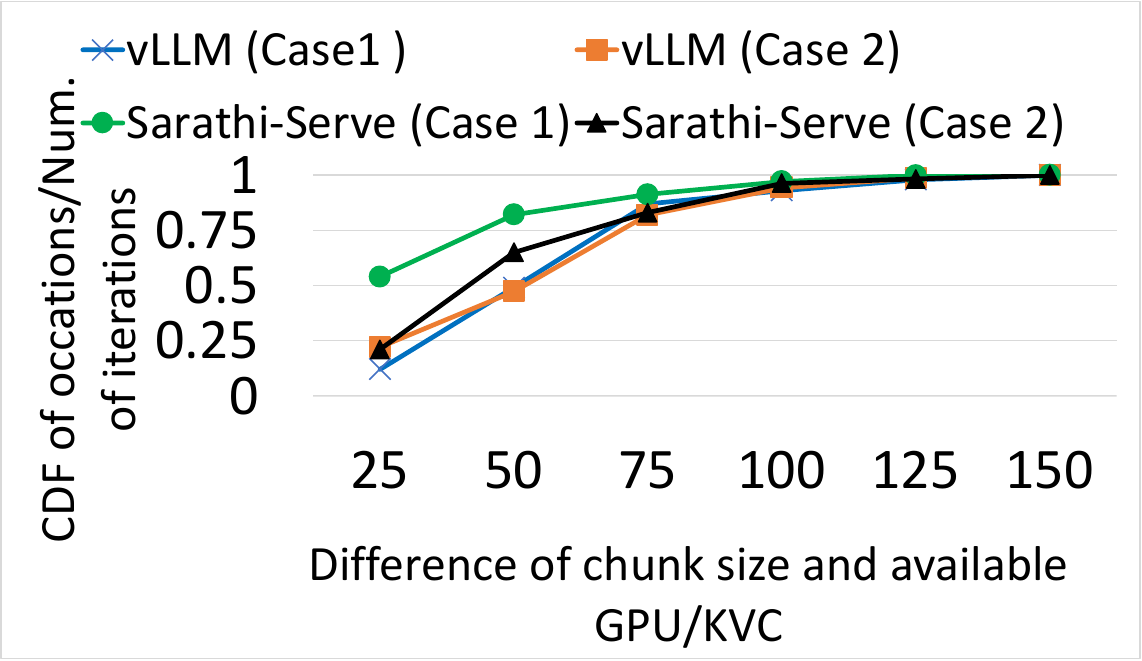} }}
    \hfill
    \subfloat[Average occupied KVC.\vspace{-0.01in}\label{fig:prompt-batch}]{{\includegraphics[width=0.32\linewidth,height=0.145\textheight]{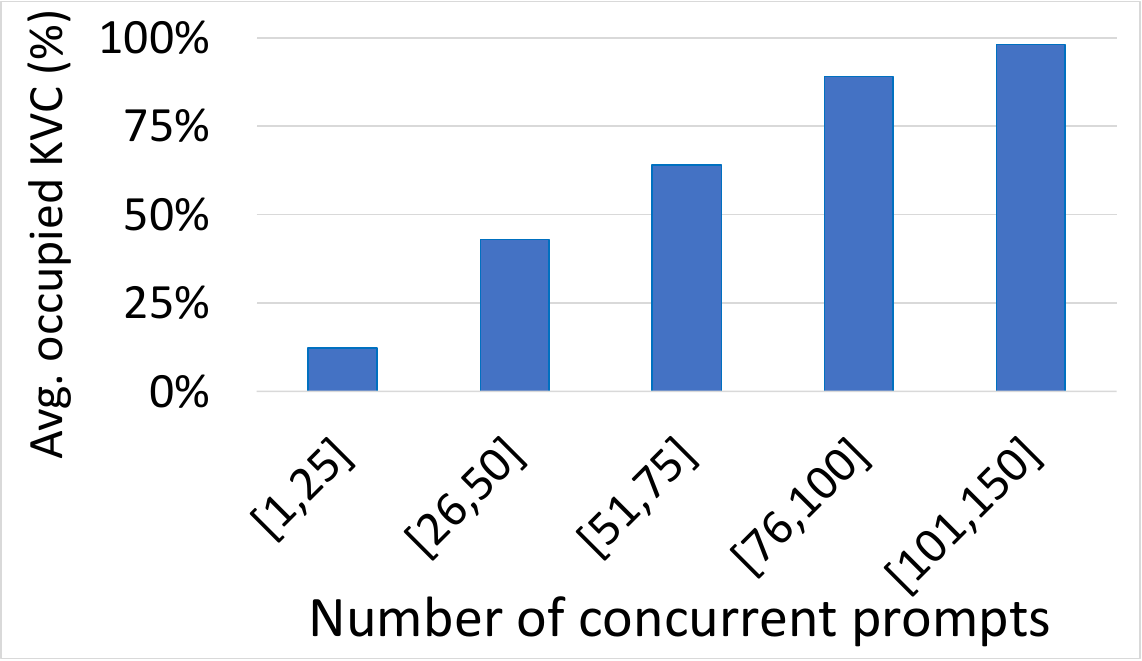} }}
    \hfill}
    \DEL{\subfloat[Percentage of prompt tasks.\vspace{-0.01in}\label{fig:14-4}]{{\includegraphics[width=0.24\linewidth,height=0.145\textheight]{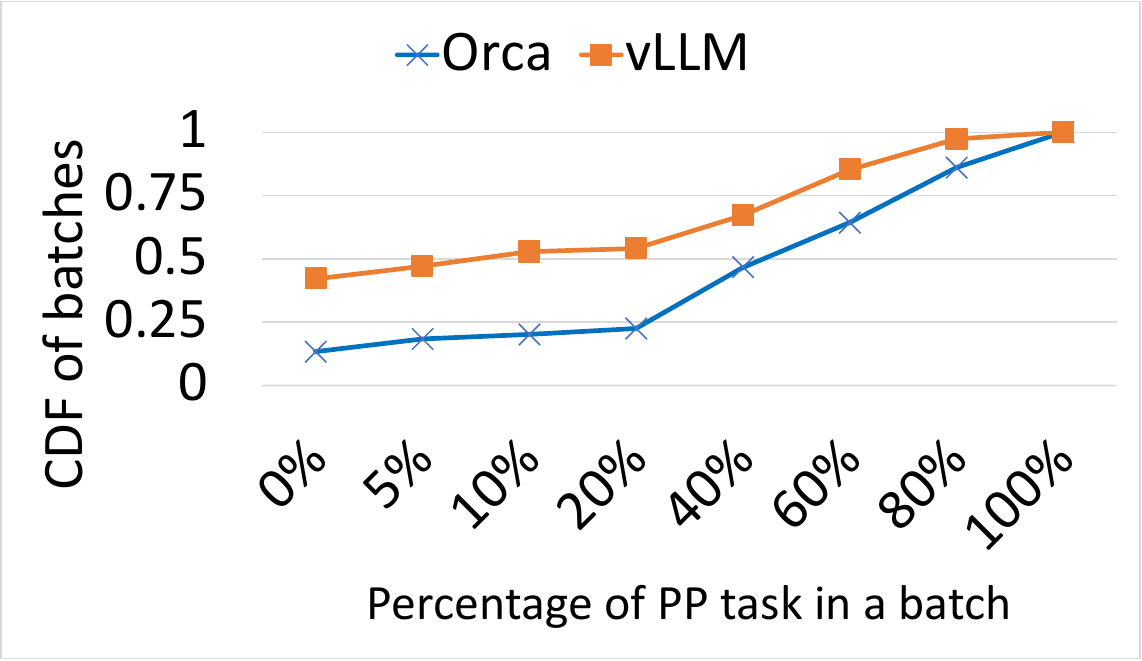} }}
    \hfill}
   \caption{{Performance of different LLM systems for OPT-13B.\vspace{-0.2in}}}
    \label{fig:pp-impact}
\end{figure*}

\begin{figure*}[t]
\centering
\subfloat[CDF of batches vs. iteration time.\vspace{-0.01in}\label{fig:14-3-opt-175b}]{{\includegraphics[width=0.24\linewidth,height=0.145\textheight]{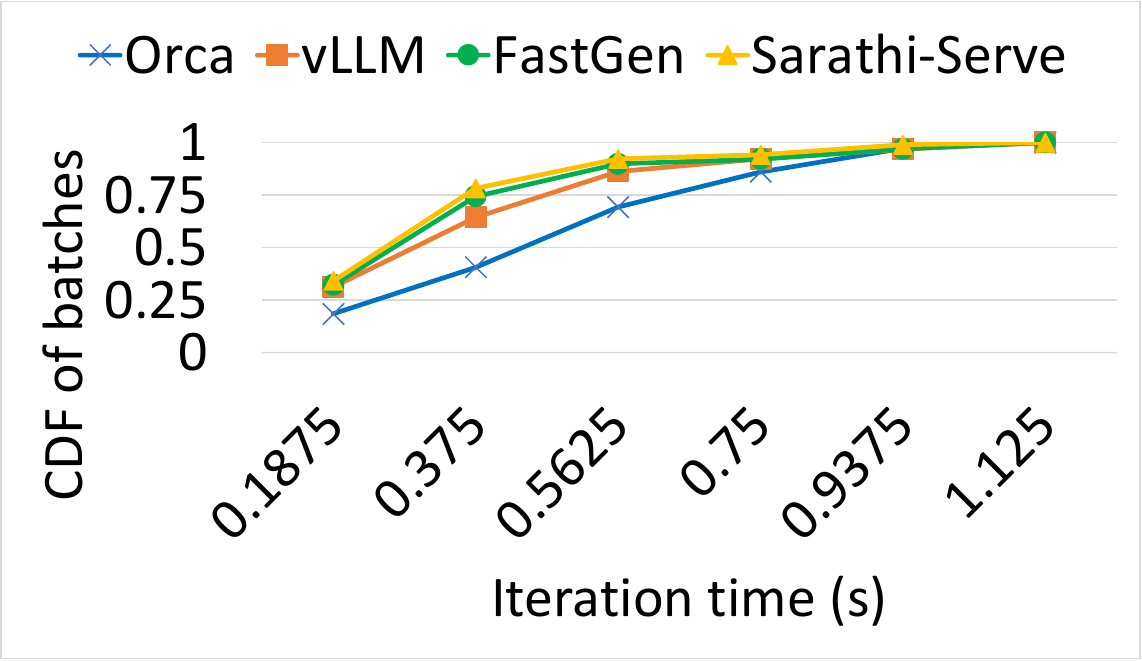} }}
    \hfill
    \subfloat[GPU compute utilization.\vspace{-0.01in}\label{fig:batch-capacity-175b}]{{\includegraphics[width=0.24\linewidth,height=0.145\textheight]{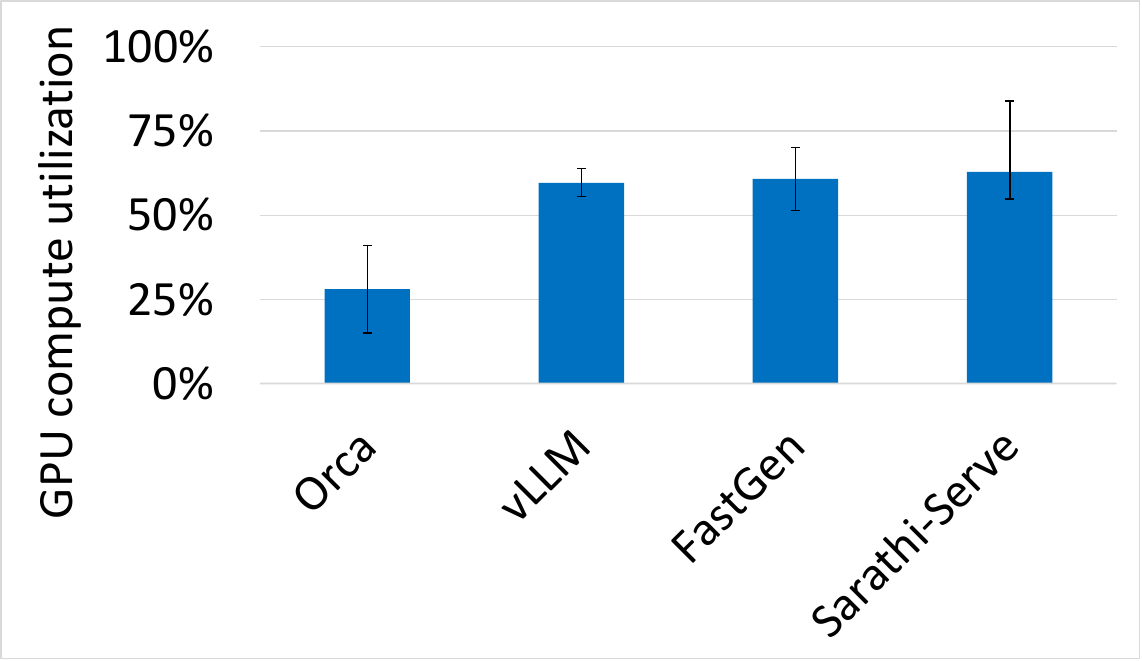} }}
    \hfill
    \subfloat[KVC utilization.\vspace{-0.01in}\label{fig:kvc-utilization-175b}]{{\includegraphics[width=0.24\linewidth,height=0.145\textheight]{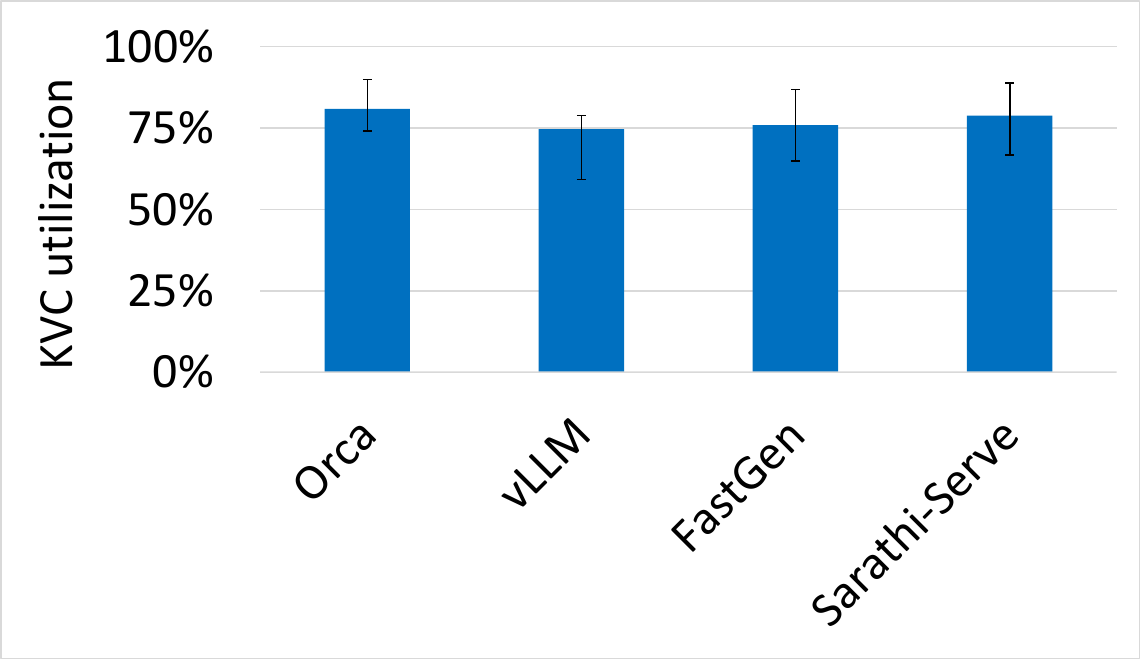} }}
    \hfill
    \subfloat[SLO attainment.\vspace{-0.01in}\label{fig:slo-methods-175}]{{\includegraphics[width=0.24\linewidth,height=0.145\textheight]{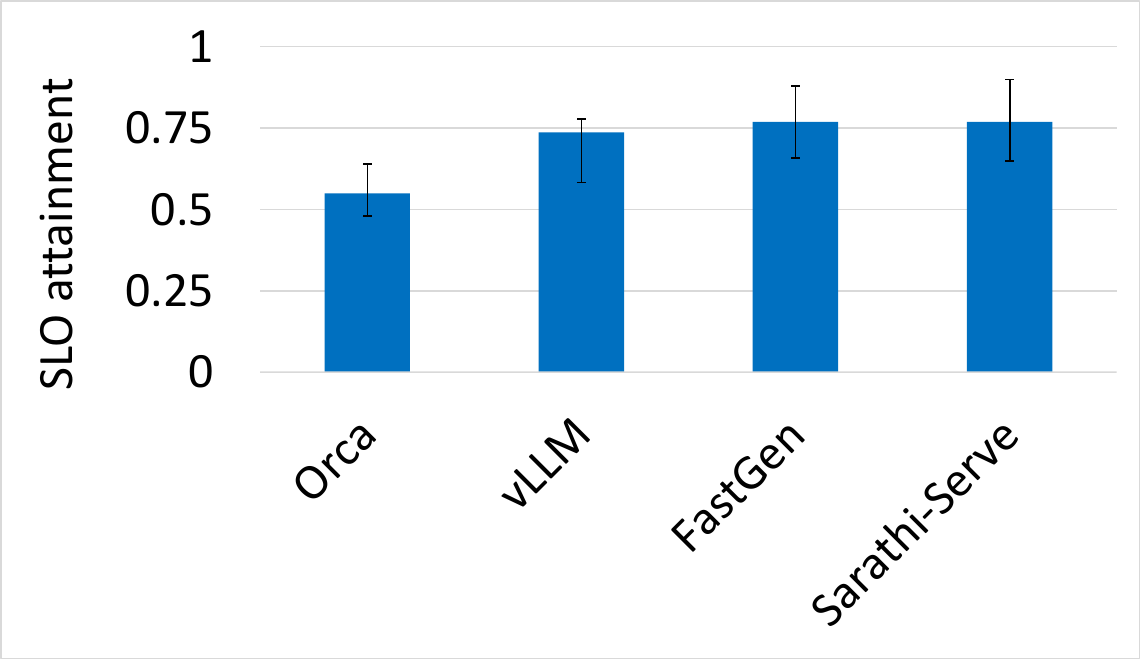} }}
    \hfill
    \DEL{\subfloat[Iteration time vs. forward size.\vspace{-0.01in}\label{fig:forward-size-itopt-175b}]{{\includegraphics[width=0.32\linewidth,height=0.145\textheight]{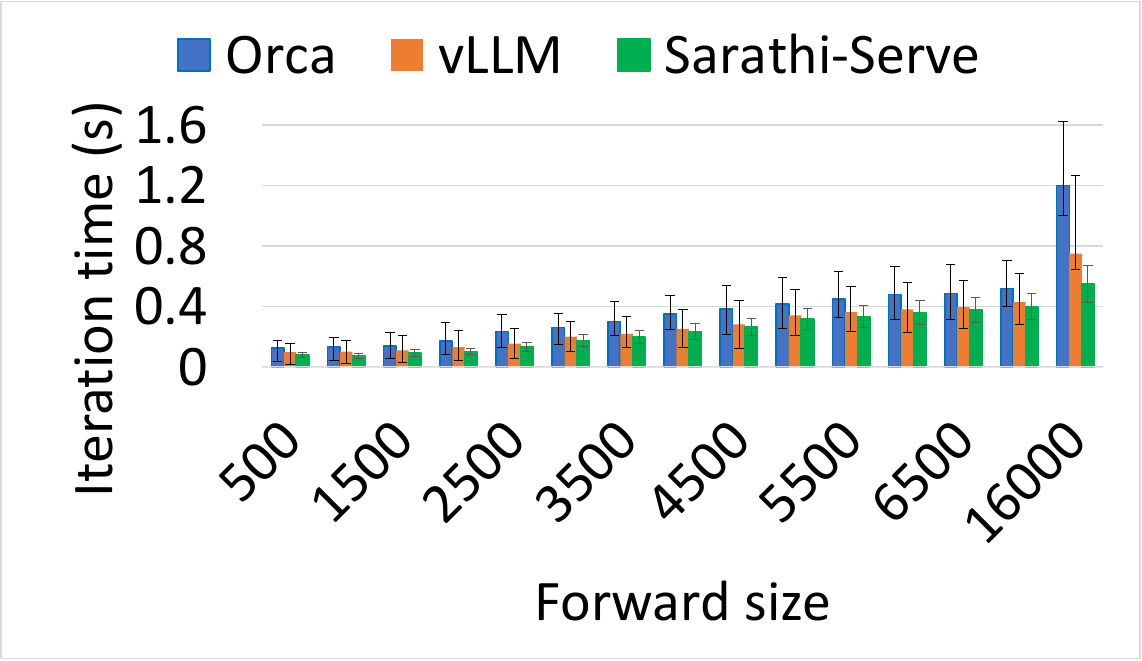} }}
    \hfill}
\DEL{\subfloat[CDF of batches vs. forward size.\vspace{-0.01in}\label{fig:14-1-opt-175v}]{{\includegraphics[width=0.32\linewidth,height=0.145\textheight]{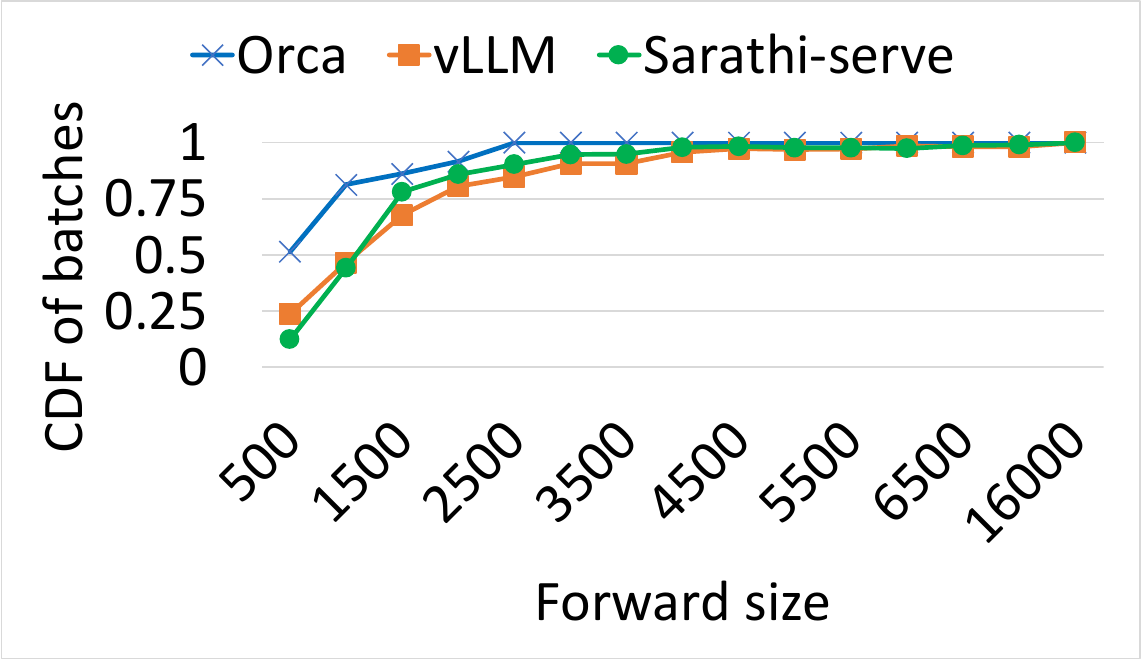} }}
    \hfill}
    \DEL{\subfloat[Prompt length distribution.\vspace{-0.01in}\label{fig:14-2}]{{\includegraphics[width=0.32\linewidth,height=0.145\textheight]{Figures/14-2-up-4.pdf} }}
    \hfill}
    \DEL{\subfloat[SLO attainment (SSR).\vspace{-0.01in}\label{fig:slo-methods}]{{\includegraphics[width=0.32\linewidth,height=0.145\textheight]{Figures/SLO.pdf} }}
    \hfill
    \subfloat[forward size/Capacity.\vspace{-0.01in}\label{fig:batch-capacity}]{{\includegraphics[width=0.32\linewidth,height=0.145\textheight]{Figures/batch-tokens-capacity.pdf} }}
    \hfill
    \subfloat[Num. of iterations of the occations/Total
Num. of iterations\sh{change it to "Percentage of iterations for each case". Also change Y name to this name"}.\vspace{-0.01in}\label{fig:case-analysis}]{{\includegraphics[width=0.32\linewidth,height=0.15\textheight]{Figures/case-analysis.pdf} }}
    \hfill
    \subfloat[CDF of difference of chunk size and available GPU/KVC.\vspace{-0.01in}\label{fig:case-analysis-2}]{{\includegraphics[width=0.32\linewidth,height=0.145\textheight]{Figures/cdf-case-wise.pdf} }}
    \hfill
    \subfloat[Average occupied KVC.\vspace{-0.01in}\label{fig:prompt-batch}]{{\includegraphics[width=0.32\linewidth,height=0.145\textheight]{Figures/prompt-batch.pdf} }}
    \hfill}
    \DEL{\subfloat[Percentage of prompt tasks.\vspace{-0.01in}\label{fig:14-4}]{{\includegraphics[width=0.24\linewidth,height=0.145\textheight]{Fig/14-4-up-3.pdf} }}
    \hfill}
   \caption{Performance of different LLM systems for OPT-175B. \vspace{-0.2in} }
    \label{fig:pp-impact-175b}
\end{figure*}

\section{Experiment Analysis}\label{sec:analysis}\vspace{-0.05in}

In this section, we experimentally analyze the performance of existing LLM systems. Our observations serve as motivation for our work and provide insights into the design of \Sys. 

\DEL{\sh{What I always told you is for each parameter setting, must have a reference to support – did you follow it? You told me that except SLO, all other parameters are from other papers, which paper supports 512 and why the reviewers still have the comments about why you set a parameter to a certain value?}}


\vspace{-0.02in}
\subsection{Experiment Settings}\vspace{-0.05in}

\noindent{\textbf{Model settings.} } 
We ran the OPT-13B model~\cite{Zhang2022OPTOP} on one GPU, and the OPT-175B~\cite{Zhang2022OPTOP} model on eight GPUs, configuring the pipeline and tensor parallelism degrees to 8 and 2, respectively, as in~\cite{Agrawal2023SARATHIEL}. Both used fp16-formatted model parameters and intermediate activations. 

\DEL{following Sarathi-Serve we combine pipeline parallelism to distribute layers across GPUs with tensor parallelism to split computation within each layer across multiple GPUs in the same pipeline stage, ensuring efficient utilization of GPU resources. The degrees of pipeline and tensor parallelism are 8 and 2, respectively. \sh{we will need the pipelining, so more GPUs and a large model as in Saranthi-done}}

\noindent{\textbf{Machine settings.}} We conducted our experiments on an AWS p4d.24xlarge instance, equipped with 8 NVIDIA A100 GPUs, with each GPU having 80GB of memory. The GPUs are connected with a 600 GB/s NVSwitch. 

\noindent{\textbf{LLM systems.}} We used the source code of Sarathi-Serve~\cite{298679}, FastGen~\cite{fastgen} and vLLM~\cite{vllm}, and implemented \Orca~\cite{280922} by ourselves using FasterTransformer since its source code is not available. 
As in~\cite{vllm}, we 
set the block size to 32 tokens. As in~\cite{280922}, we set the batch size to 8 and set the maximum sequence length to 8K in \Orca. We set KVC size to 12GB in the 
OPT-13B case and to 33.75GB in each GPU in the OPT-175B case. 


\noindent \textbf{Datasets.} To create a mix-prompt scenario, we combined the Alpaca~\cite{alpaca}, ShareGPT~\cite{shareGPT} and BookCorpus~\cite{soskkobayashi2018bookcorpus} traces. Alpaca and ShareGPT have 52K and 90K requests, with up to 500 and 2K token prompt lengths, respectively. BookCorpus has 11K unpublished books, with longer prompt lengths up to 100K tokens. We used random selection with replacement to create a dataset of 20K requests from the traces, ensuring that 35\% of the requests originated from BookCorpus.




\noindent{\textbf{Request settings.}} As in~\cite{vllm}, each request can generate up to 2048 tokens. The requests arrive following a Poisson distribution with an arrival rate of 8 requests/s~\cite{280922}. 

The resource utilization profiling was performed using the gpustat library~\cite{gpustat} with a 0.1s time interval. 
\emph{Token budget} (denoted as $S_b$) is referred to as the target forward size. In our experiment, as the forward size ($S_f$) increases, throughput improves and nearly saturates at a forward size of 768 and 1280 for OPT-13B and OPT-175B, respectively~\cite{Wang2019BenchmarkingTG}, referred to as \emph{pivot forward size} (denoted by $S_{pf}$), $S_{pf}$ is used as the token budget unless otherwise specified in our experiments. This allows near-full GPU compute utilization without substantially increasing batch processing latency. 
In this paper, 
the figures with error bars report the average, the 5th and the 95th percentiles of the result values.




\DEL{add "Sarathi-serve" to all the figures. set SLO to 0.1875s. In the first figure, set one X point to 0.1875 and one X point to $>$0.1875, remove (d), add figs: 1) Y: SLO attainment (SSR), X: three methods, 2) Y: forward size/capacity (capacity is token budget), average, p5 and p95, X: three methods-done }

\subsection{Measurement Results}




\DEL{    
\includegraphics[width=\linewidth,height=0.15\textheight]{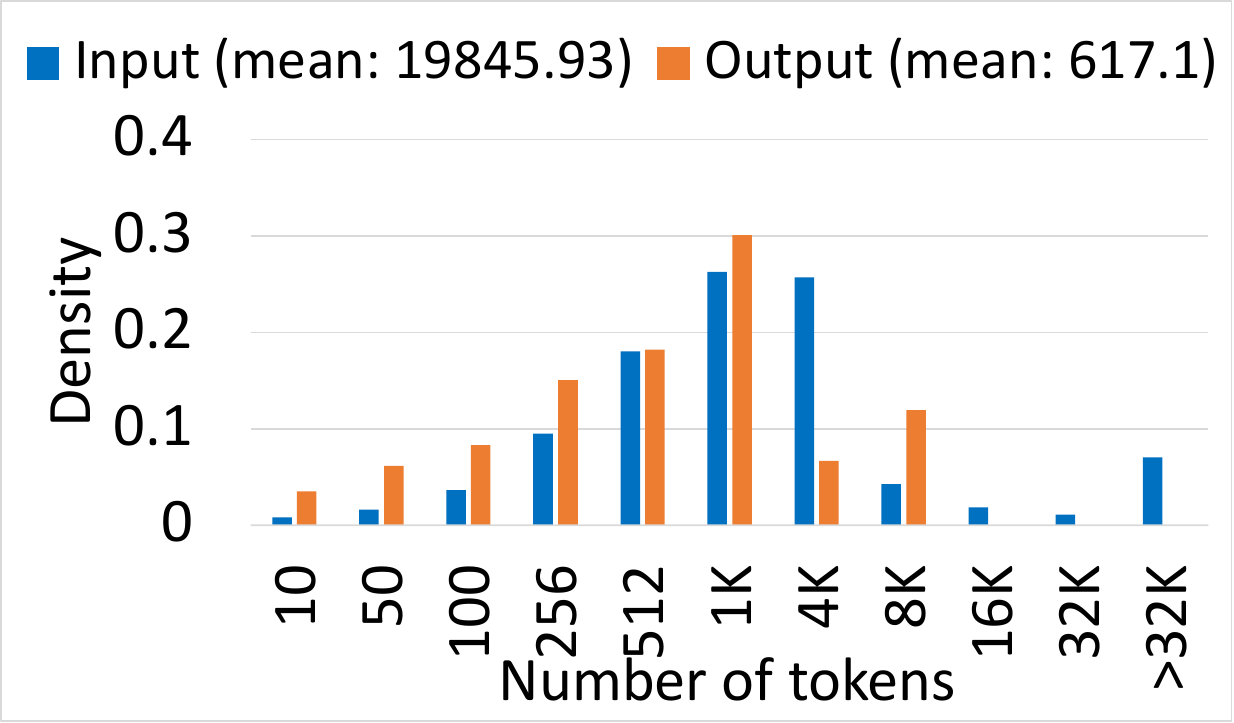}
\vspace{-0.1in} \caption{Features of the dataset.\vspace{-0.00in} }
    \label{fig:dataset-denisty}
\end{minipage}%
   \hfill%
   }

\DEL{\begin{figure}[t]
\centering
    \subfloat[SLO attainment (SSR).\vspace{-0.01in}\label{fig:slo-methods}]{{\includegraphics[width=0.48\linewidth,height=0.145\textheight]{Figures/SLO.pdf} }}
    \hfill
    \subfloat[forward size/Capacity.\vspace{-0.01in}\label{fig:batch-capacity}]{{\includegraphics[width=0.48\linewidth,height=0.145\textheight]{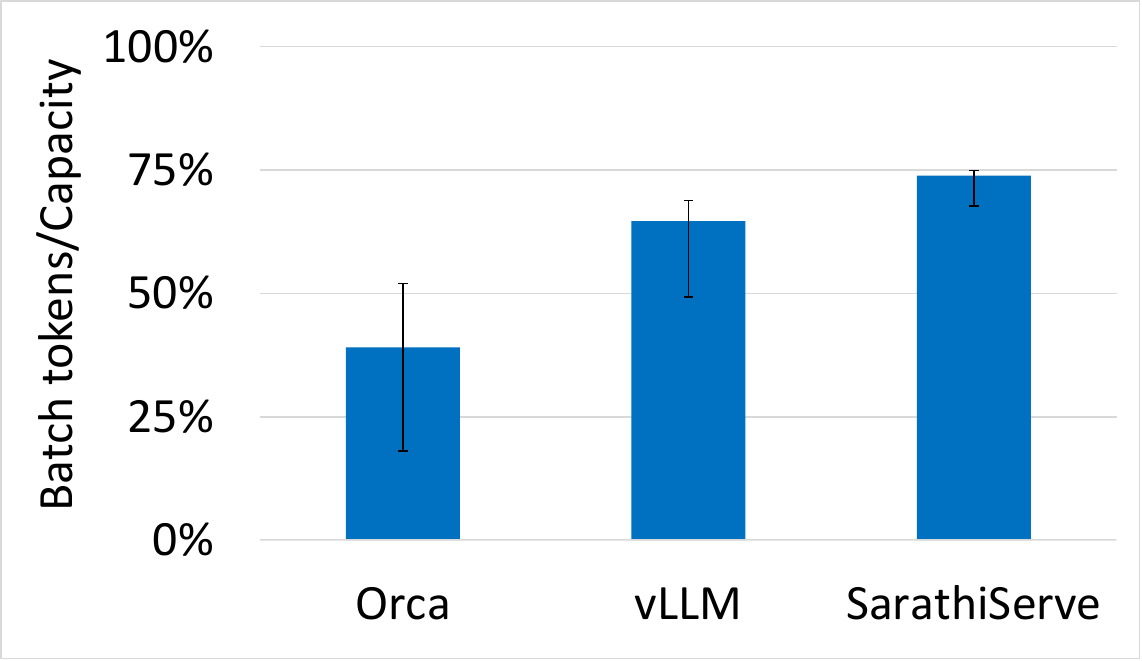} }}
    \hfill
   \caption{Performance of methods in SLO and capacity for OPT-13B.\vspace{-0.0in}}
    \label{fig:pp-slo}
\end{figure}}

\DEL{\begin{figure}[t]
\centering
    \subfloat[SLO attainment (SSR).\vspace{-0.01in}\label{fig:slo-methods-175}]{{\includegraphics[width=0.48\linewidth,height=0.15\textheight]{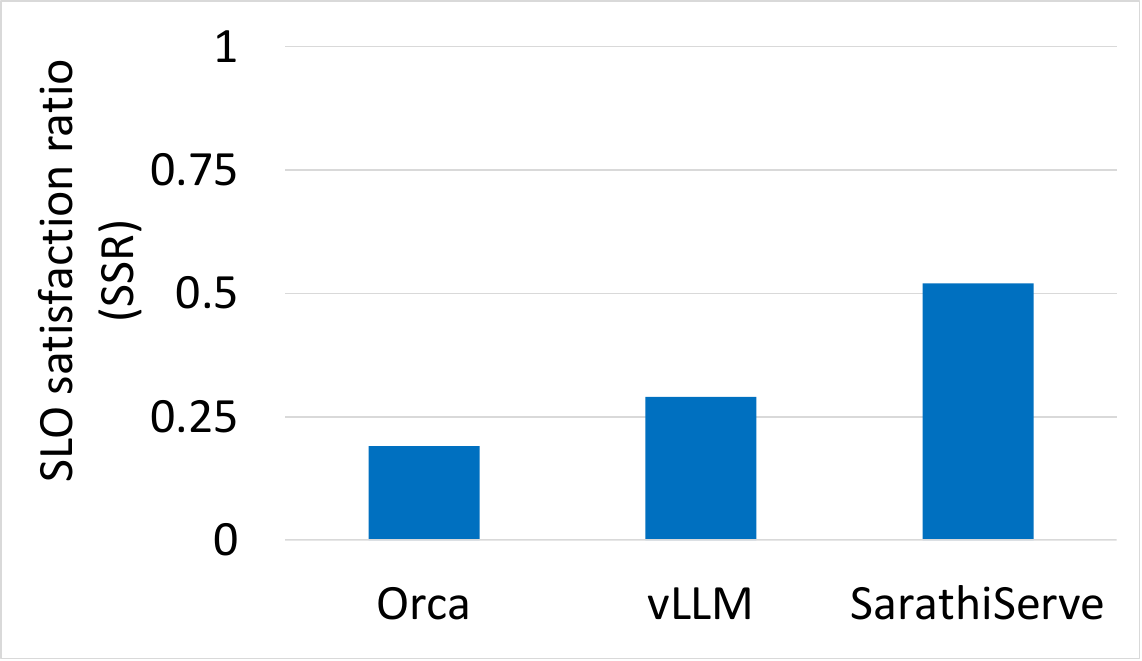} }}
    \hfill
    \subfloat[forward size/Capacity.\vspace{-0.01in}\label{fig:batch-capacity-175b}]{{\includegraphics[width=0.48\linewidth,height=0.15\textheight]{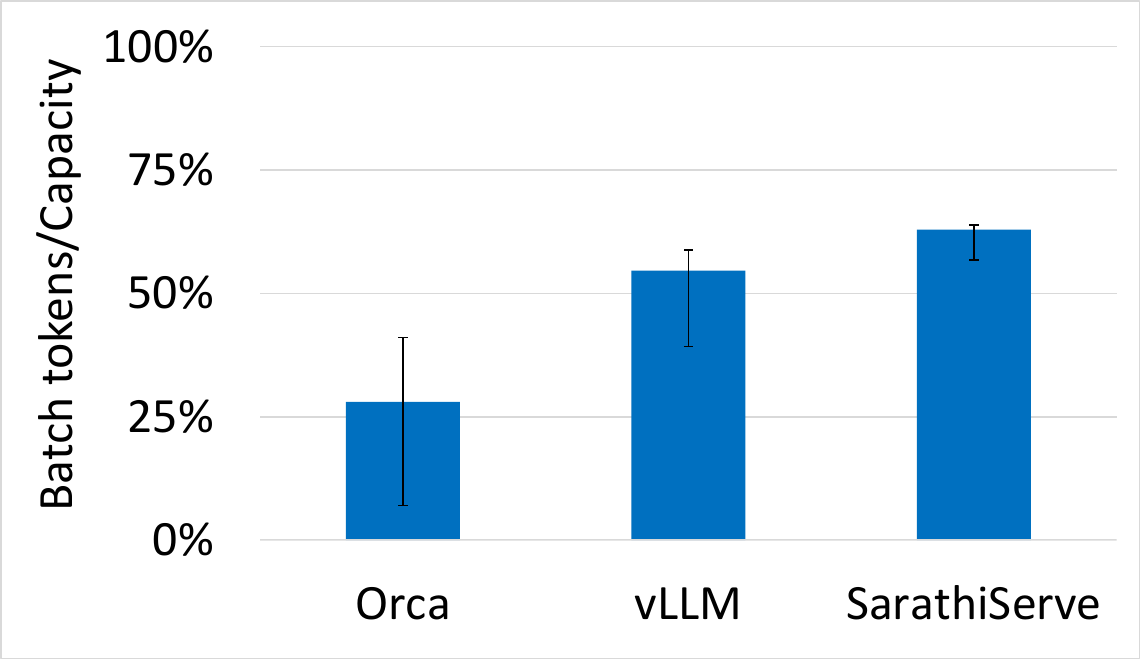} }}
    \hfill
   \caption{\small{Performance of methods in SLO and capacity for OPT-175B.\vspace{-0.0in}}}
    \label{fig:pp-slo-175}
\end{figure}}

\DEL{We observe that the iteration time for the first iteration
time  (i.e., PP) is much longer than those in the subsequent iterations. As the prompt length increases, the first
iteration time grows roughly in a linear manner, while
the iteration times in the subsequent iterations do not change due to constant one input sequence. The observation is consistent with that in \cite {\sh{add the archived paper with the same results in Fig}}.
}

\DEL{We conducted another experiment which limits the forward size to a certain value by chunking prompts. We plotted these metrics versus each forward size varying from 64 to 7168 in Figure~\ref{fig:throughput} in Appendix?, and obtained the same observations.}

\DEL{\noindent\textbf{Throughput Performance.} 
In Figure~\ref{fig:throughput_new}, the average throughput in TFLOP/s and iteration time are presented for each forward size category in vLLM. As forward size ($S_f$) increases, the throughput rises and nearly saturates at a forward size of 768
, which we refer to as the pivot forward size ($S_{b}$)~\cite{Wang2019BenchmarkingTG}. The throughput increase in TFLOP/s becomes less than 3\% beyond this value. These results align with theoretical expectations in Equation~\eqref{eq:totop}. Additionally, the iteration time also increases with growing $S_{f}$.

\begin{wrapfigure}{c}{4cm}\vspace{-0.2in}
\centering
\includegraphics[width=\linewidth,height=0.15\textheight]{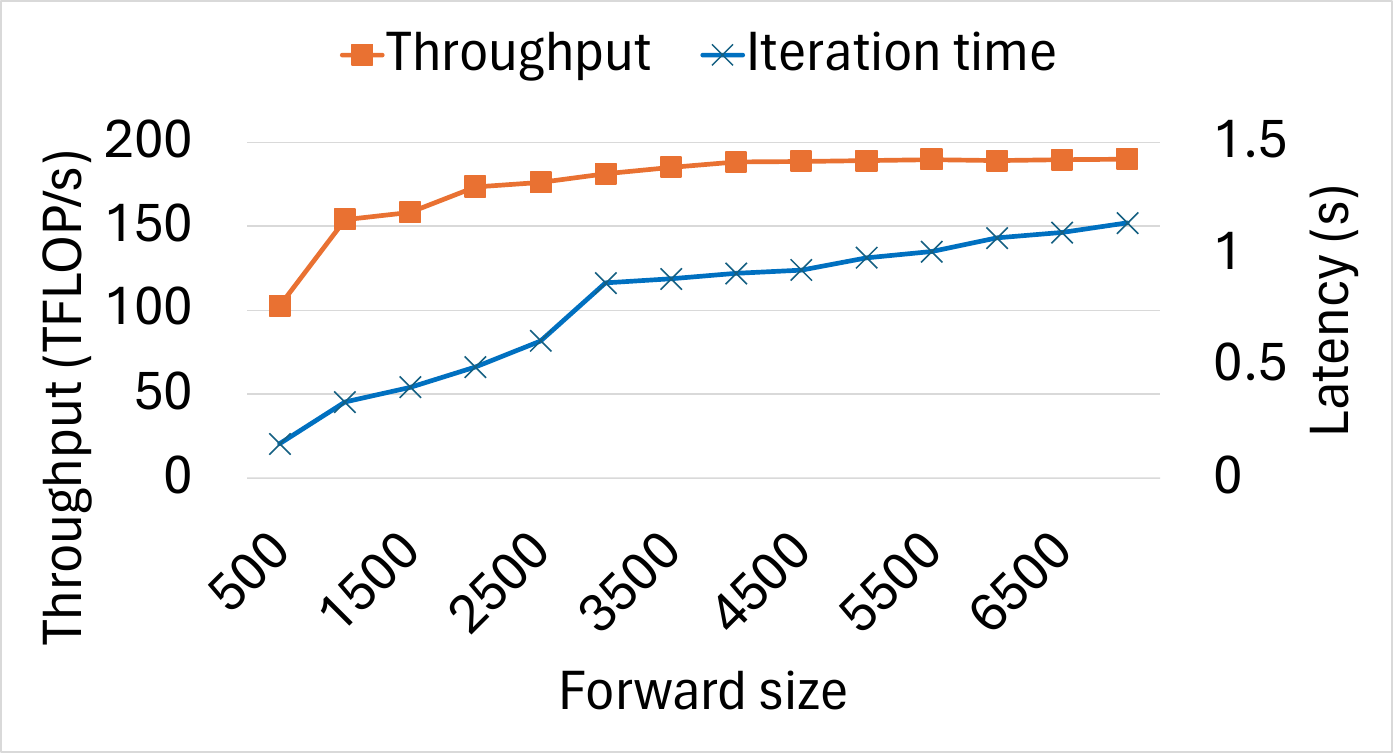}
\vspace{-0.05in}    \caption{Impact of forward size in vLLM.\vspace{-0.1in} }
    \label{fig:throughput_new}
\end{wrapfigure}
}



\noindent\textbf{Resource Utilization and SLO Performance.} 
Figures~\ref{fig:14-3} and~\ref{fig:14-3-opt-175b} depict the cumulative distribution function (CDF) of batches (or iterations) versus iteration time 
for OPT-13B and OPT-175B, respectively. Only 16\%-34\% of batches on average across the two models in \Orca, vLLM, {FastGen} and Sarathi-Serve complete within 0.1875s, indicating that many iterations exceed the normal reading speed.


\DEL{Removed due to removing previous figures}


\DEL{Removed due to removal of previous figure c}

Figures~\ref{fig:batch-capacity} and~\ref{fig:batch-capacity-175b} show the average GPU compute utilization per iteration, calculated as the ratio of the actual forward size to the token budget.  \Orca, vLLM, {FastGen}, Sarathi-Serve achieve GPU compute utilization of 33\%, 61\%, {62\%} and 64\%, respectively, on average for both models. KVC constraints and long prompt chunks hinder the full utilization of the remaining token budget. That is, when the KVC cannot host a chunk, the chunk won't be added to the batch. The results indicate significant potential for improvement in GPU compute utilization. 

\DEL{\Orca shows low GPU compute utilization for max allocation method, while vLLM and Sarathi-Serve demonstrate low GPU compute allocation because of their static chunking method while not taking the token budget or the available KVC into account.} 

Figures~\ref{fig:kvc-capacity} and~\ref{fig:kvc-utilization-175b} show the average KVC utilization per iteration, defined as the ratio of allocated KVC space to total KVC space in tokens. \Orca, vLLM, {FastGen} and Sarathi-Serve achieve KVC utilization of 88\%, 79\%, {80\%} and 83\%, respectively, on average for both models. Although the systems add requests to the batch until the batch size is reached in \Orca or the KVC space is fully allocated, residual KVC space often remains that cannot accommodate the next request or chunk, resulting in incomplete KVC utilization. 

A prompt or prompt chunk that fully utilizes both the remaining GPU compute and KVC resources is preferable over one that only fully utilizes the remaining GPU compute resource. This is because if the former is not selected for execution, it may later be unable to run due to insufficient KVC resources. Fully utilizing both resources better achieves the goal of maximizing throughput of a workload. However, the systems' exclusive focus on GPU compute utilization, combined with its FIFO policy, prevents it from prioritizing such optimal choices. Therefore, they fail to maximize both GPU compute and KVC utilizations. 

We then measure the performance of the systems in meeting the SLO requirements. We grouped the prompts within the length range of 512 tokens, and measured the average prompt processing latency for each group using the Triton Inference Server with the Faster Transformer backend. As~\cite{Li2023AlpaServeSM} that uses SLO-scale to determine SLO, we set a TTFT SLO as the product of the average prompt processing latency of its group and an SLO-scale~\cite{Li2023AlpaServeSM} randomly selected from the range of [0.5,1.5]. The TBT SLO was set to 0.1875s $\times$ SLO~scale, which was randomly selected from $\{0.25,0.5,1,2\}$ for each request. Figures~\ref{fig:slo-methods} and~\ref{fig:slo-methods-175} show the iteration SLO attainment, defined as the percentage of iterations that meet their iteration-level SLOs. We see that the systems fail to achieve high SLO attainment since they do not address heterogeneous SLOs. 

We then set the system-wide SLO in Sarathi-Serve from 0.2s to 1s, increasing by 0.2s at each step as in~\cite{298679}. Figures~\ref{fig:slo-wise} and~\ref{fig:slo-wise-175} show the iteration-SLO attainment and GPU compute utilization versus the system-wide SLO in Sarathi-Serve. A stringent system-wide SLO, while ensuring compliance with heterogeneous SLOs, reduces throughput. In contrast, a more relaxed system-wide SLO avoids throughput reduction but results in lower SLO attainment.





\DEL{removed, because asked to remove Figure 5 and 6}

\DEL{Additionally, Figures~\ref{fig:batch-capacity} and~\ref{fig:batch-capacity-175b} plot the percentage of forward size relative to the capacity reached by each method for the OPT-13B and OPT-175B models, respectively. The error bars in these figures represent the 5th and 95th percentiles. We observe that both Sarathi-Serve and vLLM demonstrate high percentages due to their efficient chunking methods.}




\DEL{Since Sarathi-Serve outperforms \Orca and vLLM~\cite{vllm}, we present Sarathi-Serve's results in the rest of this section unless otherwise stated. } 
\DEL{For each batch, we calculated the ratio of the number of prompt processing (PP) tasks in a batch. Figure~\ref{fig:14-4} illustrates the cumulative distribution function (CDF) of batches versus the ratio. Notably, 28\% of batches have no PP tasks, and 4.9\%, 3.7\%, 1.8\%, 18.6\%, 17.8\%, 16.9\%, and 8.4\% have ratios of 5\%, 10\%, 20\%, 40\%, 60\%, 80\%, and 100\%. These results suggest that in most cases, the batch has no or few PP tasks, and in 25.3\% of batches, PP tasks occupy no less than 80\% of the tasks, potentially leading to under-utilization or over-utilization of GPU compute resources.}



\DEL{\begin{figure}[t]
\centering
    \subfloat[OPT-13B.\vspace{-0.01in}\label{fig:case-analysis}]{{\includegraphics[width=0.48\linewidth,height=0.15\textheight]{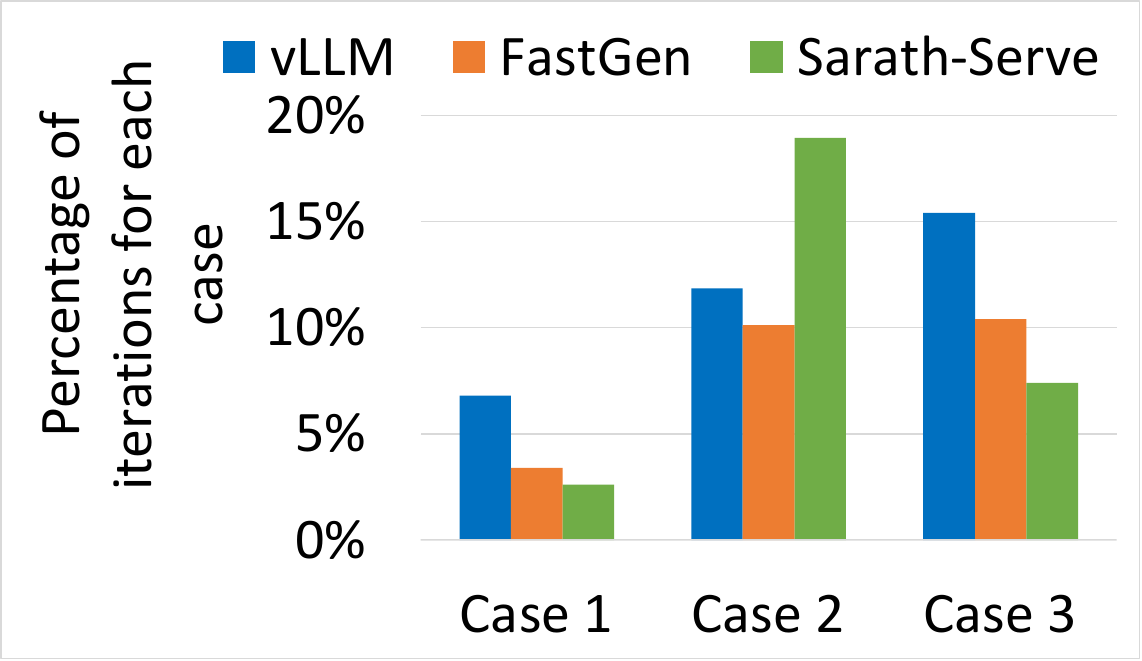} }}
    \hfill
    \DEL{\subfloat[CDF of difference of chunk size and available GPU/KVC.\vspace{-0.01in}\label{fig:case-analysis-2}]{{\includegraphics[width=0.32\linewidth,height=0.15\textheight]{Figures/cdf-case-wise.pdf} }}
    \hfill}
    \DEL{\subfloat[OPT-175B.\vspace{-0.01in}\label{fig:case-analysis-175}]{{\includegraphics[width=0.48\linewidth,height=0.15\textheight]{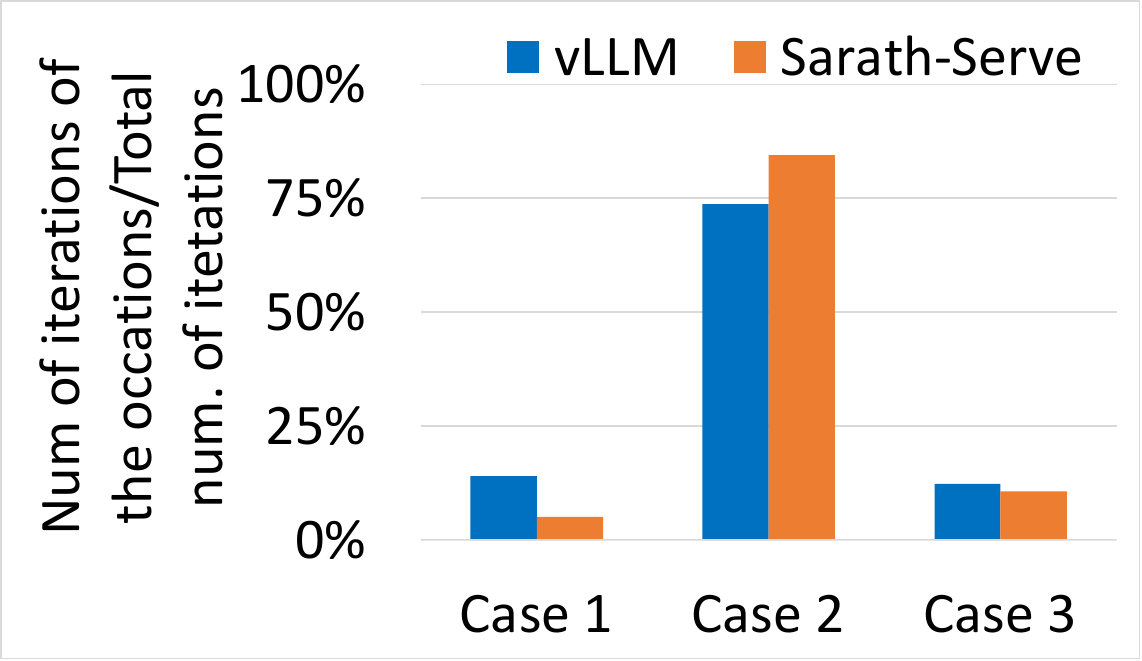} }}}
    \subfloat[OPT-175B.\vspace{-0.01in}\label{fig:case-analysis-175}]{{\includegraphics[width=0.48\linewidth,height=0.15\textheight]{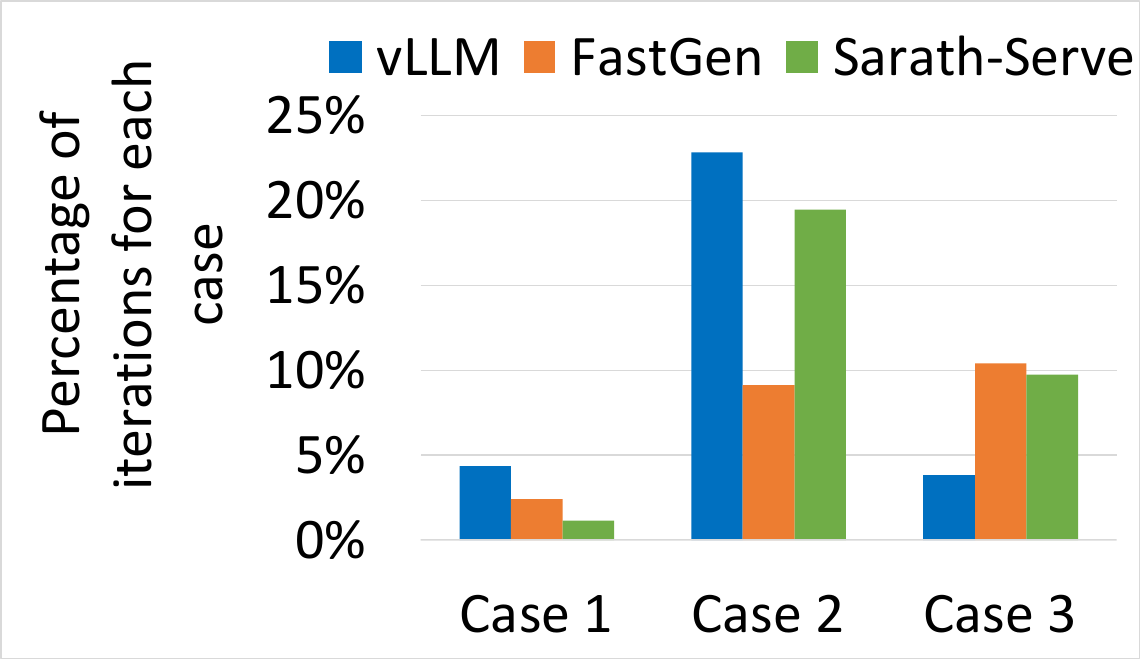} }}
    \hfill
    
    \DEL{\subfloat[Percentage of prompt tasks.\vspace{-0.01in}\label{fig:14-4}]{{\includegraphics[width=0.24\linewidth,height=0.15\textheight]{Fig/14-4-up-3.pdf} }}
    \hfill}
   \caption{Chunk sizes exceeding the remaining token budget, combined with KVC constraints, prevent the maximization of GPU compute and KVC utilization.\vspace{-0.0in}}
    \label{fig:pp-token}
\end{figure}}

\DEL{
\begin{figure}[t]
\centering
    \subfloat[Average occupied KVC.\vspace{-0.01in}\label{fig:prompt-batch}]{{\includegraphics[width=0.48\linewidth,height=0.15\textheight]{Figures/prompt-batch.pdf} }}
    \hfill
    \DEL{\subfloat[CDF of difference of chunk size and available GPU/KVC.\vspace{-0.01in}\label{fig:case-analysis-175-cdf}]{{\includegraphics[width=0.32\linewidth,height=0.15\textheight]{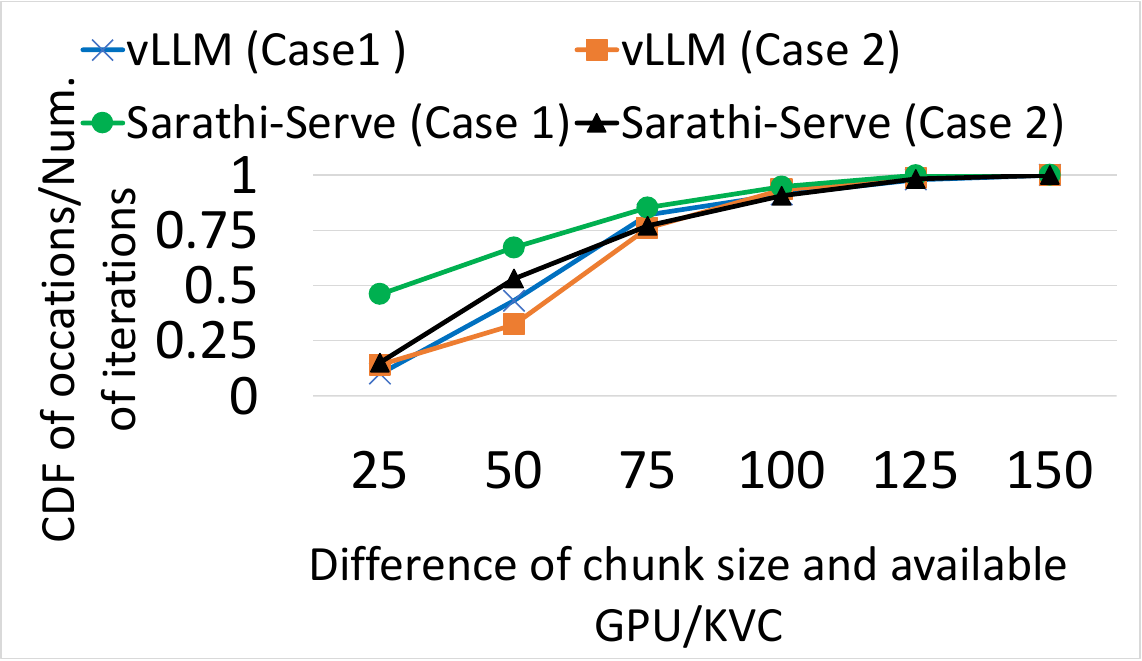} }}
    \hfill}
    \subfloat[Average occupied KVC.\vspace{-0.01in}\label{fig:prompt-batch-175}]{{\includegraphics[width=0.48\linewidth,height=0.15\textheight]{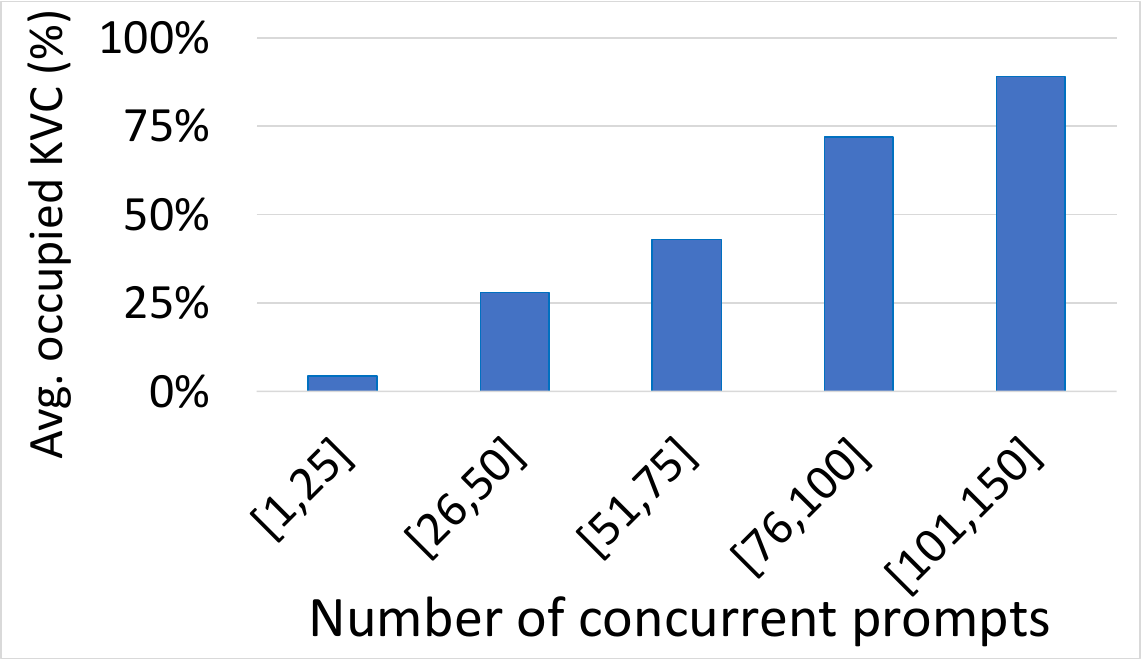} }}
    \hfill
    \DEL{\subfloat[Percentage of prompt tasks.\vspace{-0.01in}\label{fig:14-4-175}]{{\includegraphics[width=0.24\linewidth,height=0.15\textheight]{Fig/14-4-up-3.pdf} }}
    \hfill}
   \caption{Average occupied KVC for Sarathi-Serve for different number of concurrent prompts in a batch\sh{inference-done }.\vspace{-0.0in}}
    \label{fig:pp-token-175}
\end{figure}
}


\DEL{\noindent\textbf{Illustration of low utilization case.} We see the detrimental results of long prompts in the above. If not appropriately handled, long prompts become foe. 
Figure~\ref{fig:motivation} shows how the requests are processed for a 8192-token prompt. The batch size is 32. In the batch, one request is PP, other 31 requests are either short PP or TG steps. The long PP task takes a significant amount of the GPU and a long time to be processed ((P\ref{Guidance1})), which delays the other requests. 
After the PP task becomes a TG task, each iteration has 32 forward size, so the GPU computation resource is significantly under-utilized (P\ref{Guidance1}).
}


\DEL{\vspace{-0.1in}
\begin{thm}\label{overallPerformance}
Sarathi-Serve reduces long iteration time caused by long prompts, 
but still cannot maximize GPU compute utilization or KVC utilization.   
}


\DEL{In existing systems, most batches, particularly those with long prompts, encounter iteration times longer than the typical reading speed (i.e., 0.1875s/token).}

\DEL{In existing systems, most batches, particularly those with long prompts, encounter iteration times longer than the typical reading speed (i.e., 0.1875s/token).}


\begin{figure}[t]
\centering
    \subfloat[OPT-13B.\vspace{-0.01in}\label{fig:slo-wise}]{{\includegraphics[width=0.48\linewidth,height=0.145\textheight]{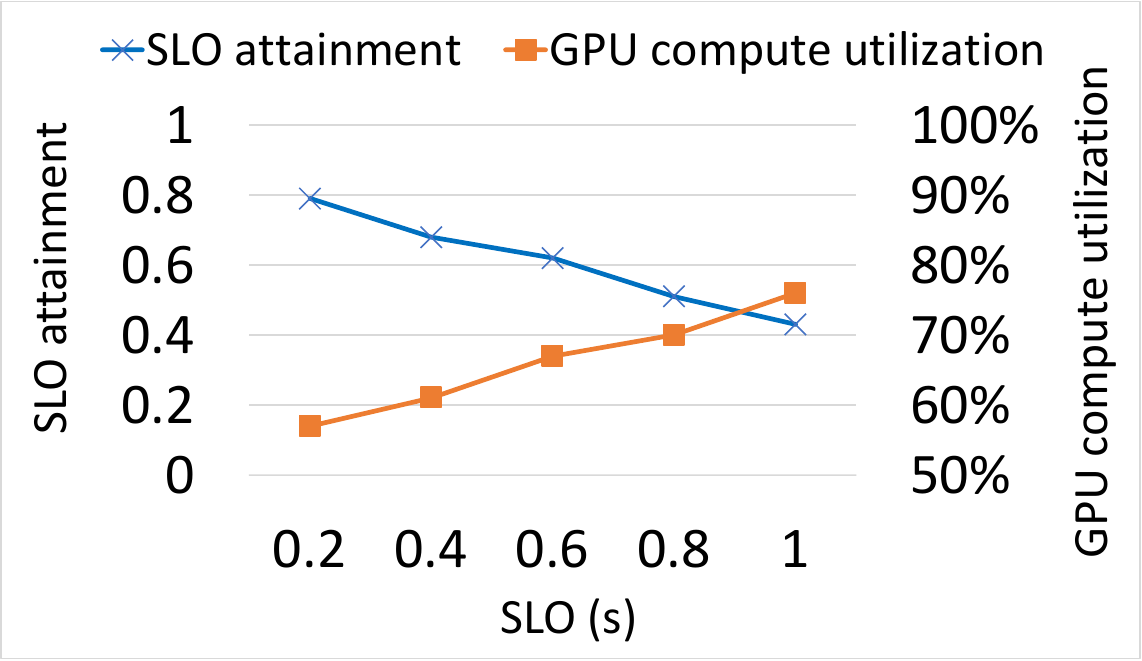} }}
    \hfill
    \DEL{\subfloat[CDF of difference of chunk size and available GPU/KVC.\vspace{-0.01in}\label{fig:case-analysis-2}]{{\includegraphics[width=0.32\linewidth,height=0.145\textheight]{Figures/cdf-case-wise.pdf} }}
    \hfill}
    \DEL{\subfloat[OPT-175B.\vspace{-0.01in}\label{fig:jct-175}]{{\includegraphics[width=0.48\linewidth,height=0.145\textheight]{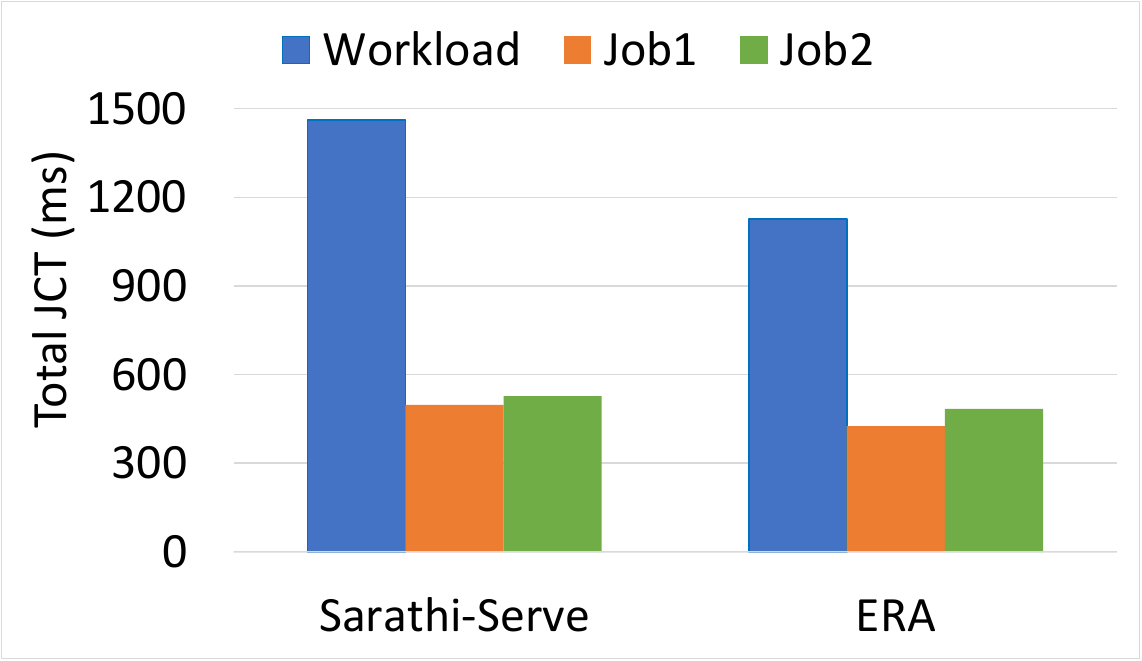} }}}
    \subfloat[OPT-175B.\vspace{-0.01in}\label{fig:slo-wise-175}]{{\includegraphics[width=0.48\linewidth,height=0.145\textheight]{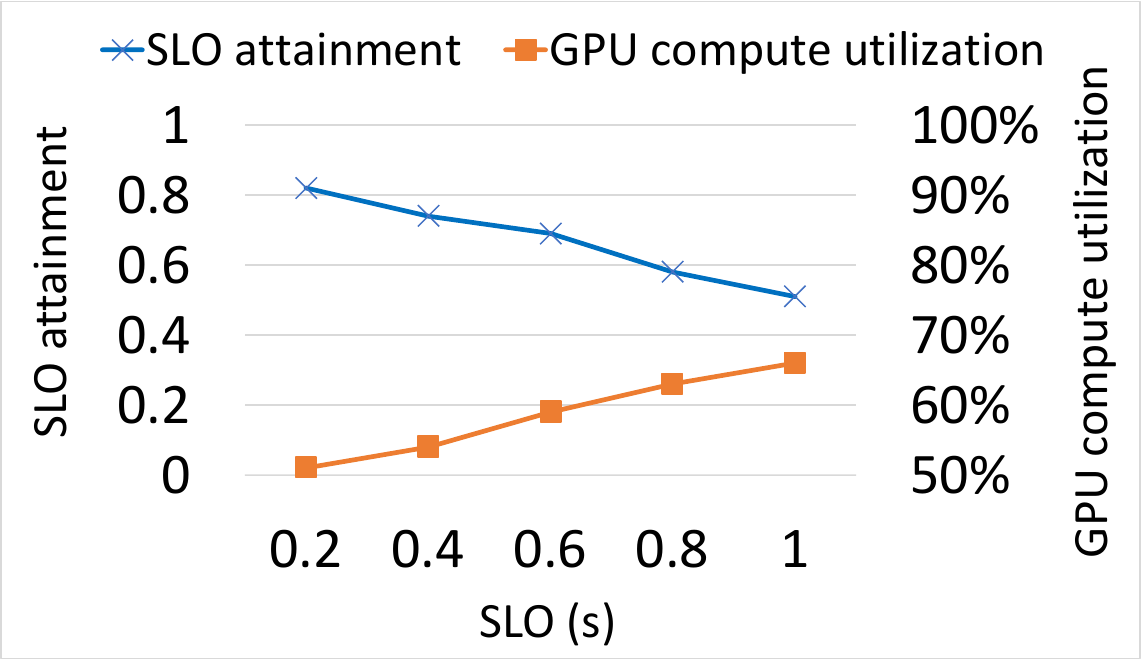} }}
    \hfill
    \DEL{\subfloat[Percentage of prompt tasks.\vspace{-0.01in}\label{fig:14-4}]{{\includegraphics[width=0.24\linewidth,height=0.145\textheight]{Fig/14-4-up-3.pdf} }}
    \hfill}
   \caption{Performance of meeting SLOs in Sarathi-Serve.\vspace{-0.2in}}
    \label{fig:combined-175}
\end{figure}

\DEL{Then, we investigate the causes for not maximizing GPU compute utilization: 1) static chunking, 2) KVC limit, 3) keep more concurrent prompts in the KVC, 4) SLO setting, 


(?there is remaining resource in the following cases/unused KVC or unused token budget. ....)When there is an available token budget:
Occasion 1. min chunk size is larger than it
Occation, 2, min chunk size is less than it but there will be residual avaialble KVC. 3. min chunk size is no larger than it, but available KVC is not enough

\sh{add a fig: Y: Num. of iterations of the occations/Total Num. of iterations, X: Case 1, Case 2 and Case 3-done}}

\DEL{\begin{figure}. 
    \centering
\includegraphics[width=0.85\columnwidth,height=0.15\textheight]{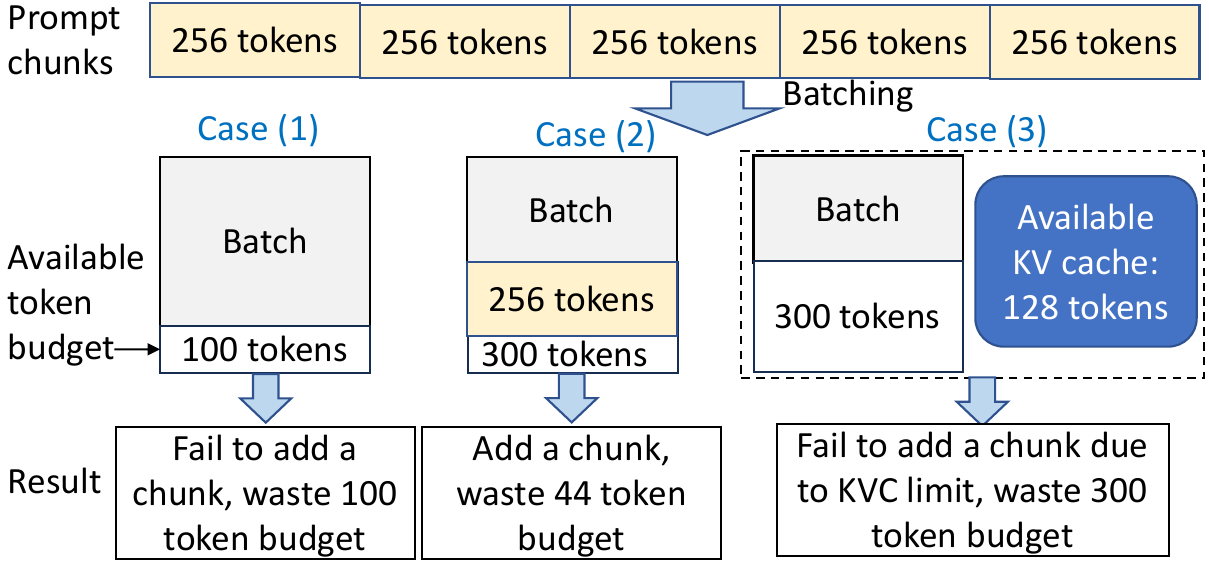}
    \caption{Causes for GPU compute and KVC underutilization. }
    \label{fig:evenchunk}
\end{figure}}

\begin{figure}[t]
\centering
    \subfloat[OPT-13B.\vspace{-0.01in}\label{fig:jct}]{{\includegraphics[width=0.48\linewidth,height=0.145\textheight]{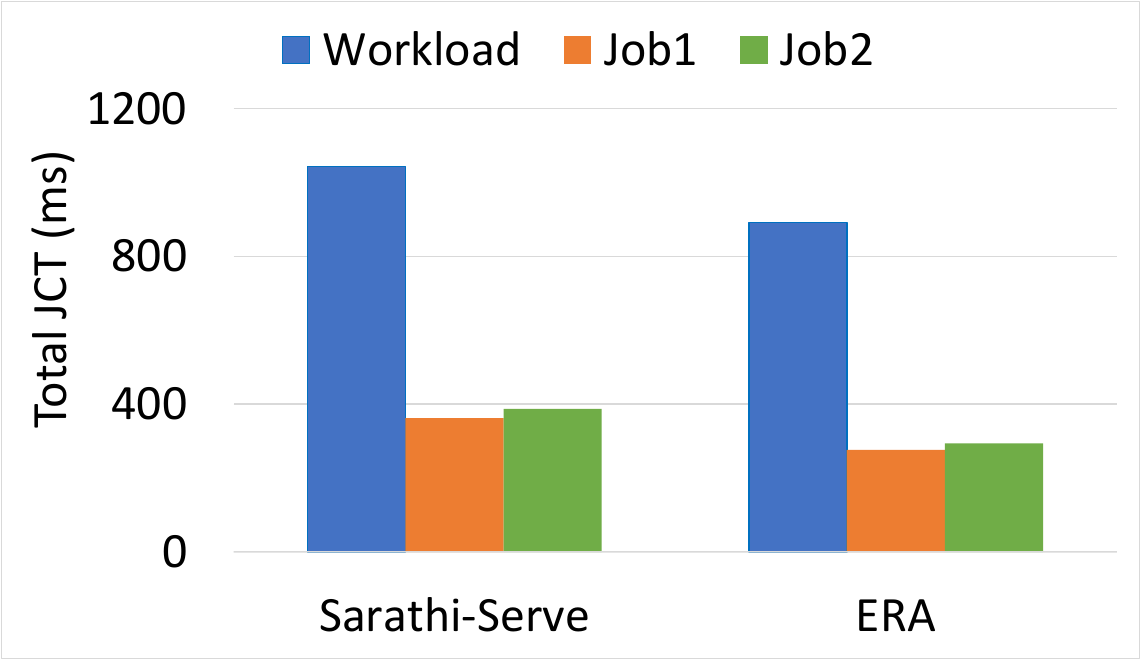} }}
    \hfill
    \DEL{\subfloat[CDF of difference of chunk size and available GPU/KVC.\vspace{-0.01in}\label{fig:case-analysis-2}]{{\includegraphics[width=0.32\linewidth,height=0.145\textheight]{Figures/cdf-case-wise.pdf} }}
    \hfill}
    \DEL{\subfloat[OPT-175B.\vspace{-0.01in}\label{fig:jct-175}]{{\includegraphics[width=0.48\linewidth,height=0.145\textheight]{Figures/total-jct-175.pdf} }}}
    \subfloat[OPT-175B.\vspace{-0.01in}\label{fig:jct-175}]{{\includegraphics[width=0.48\linewidth,height=0.145\textheight]{Figures/total-jct-175.pdf} }}
    \hfill
    
    \DEL{\subfloat[Percentage of prompt tasks.\vspace{-0.01in}\label{fig:14-4}]{{\includegraphics[width=0.24\linewidth,height=0.15\textheight]{Fig/14-4-up-3.pdf} }}
    \hfill}
   \caption{Exclusive Request Allocation (ERA) vs. Sarathi-Serve.}
    \label{fig:combined}
\end{figure}

\DEL{\begin{figure}[t]
    \centering
    \begin{minipage}[!t]{0.235\textwidth}
    \centering
\includegraphics[width=0.96\columnwidth,height=0.15\textheight]{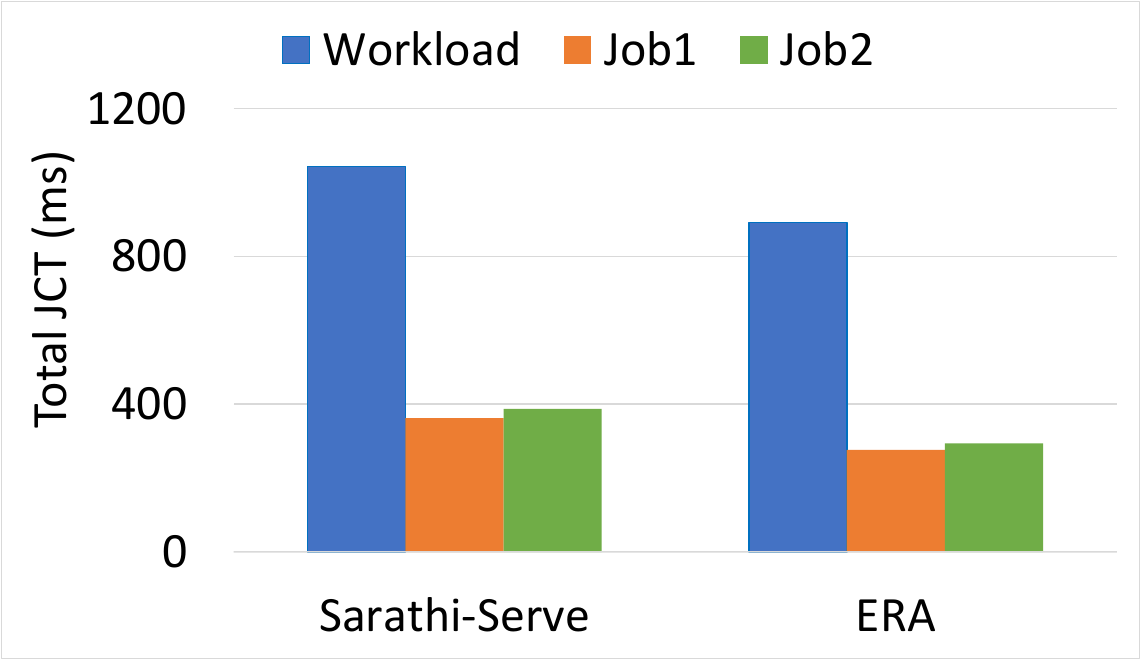}
\vspace{-0.05in}\caption{JCT of Sarathi-Serve and prioritizing chunks of one request for OPT-13B.}
\label{fig:jct}\vspace{-0.05in}
\end{minipage}
 \begin{minipage}[!t]{0.235\textwidth}
    \centering
\includegraphics[width=0.96\columnwidth,height=0.15\textheight]{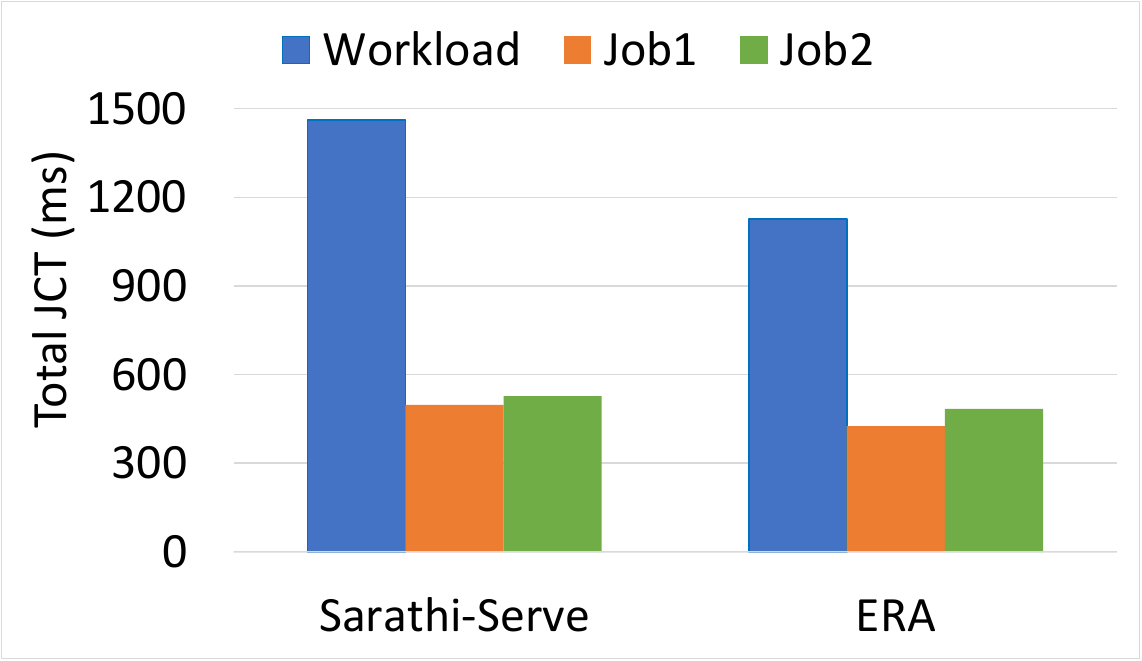}
\vspace{-0.05in}\caption{JCT of Sarathi-Serve and prioritizing chunks of one request for OPT-175B.\sh{reorganize figs as indicated in the email}}
\label{fig:jct-175}\vspace{-0.05in}
\end{minipage}
    \label{fig:combined}
\end{figure}}

\DEL{We then analyze the causes for not maximizing GPU
compute utilization (i.e., not fully utilizing token budget) and KVC utilization in vLLM, \tsr{FastGen} and Sarathi-Serve. It can be caused by three cases as illustrated in Figure~\ref{fig:evenchunk}. In the figure, the long prompt is partitioned into 256-token chunks. In case 1, the minimum chunk size (256 tokens) is larger than the remaining token budget (100 tokens). In case 2, the minimum chunk size (256 tokens) is smaller than the remaining token budget (300 tokens), but adding a chunk leads to remaining token budget (44 tokens) that cannot be used. In case 3, the minimum chunk size is not larger than the available resources, but the available (i.e., unallocated) KVC (128 tokens) is insufficient to host the minimum chunk (256 tokens).
Figures~\ref{fig:case-analysis} and~\ref{fig:case-analysis-175} show the percentage of iterations corresponding to each case. In vLLM, the percentage ranges between 6\% and 17\%, \tsr{in FastGen the percentage ranges between 3\% and 11\%} and in Sarathi-Serve, the percentage ranges between 2\% and 19\%. The results confirm the existence of these cases, attributed to chunk sizes exceeding the remaining token budget and KVC constraints. Dynamically adjusting the chunk size could enable better utilization of the remaining token budget and KVC space.}



\DEL{Add a fig with 2 lines: Y: CDF of Occation1/Num. of iterations, X: min chunk size-available GPU or available KVC, another line: Y: CDF of Occation2/Num. of iterations-done, removed as asked} 



\DEL{As a result, static chunking leads to GPU and KVC wastes as shown in Figure~\ref{fig:gap}.}

\DEL{\begin{wrapfigure}{c}{4cm}\vspace{-0.15in}
\centering
{{\includegraphics[width=1\linewidth,height=0.15\textheight]{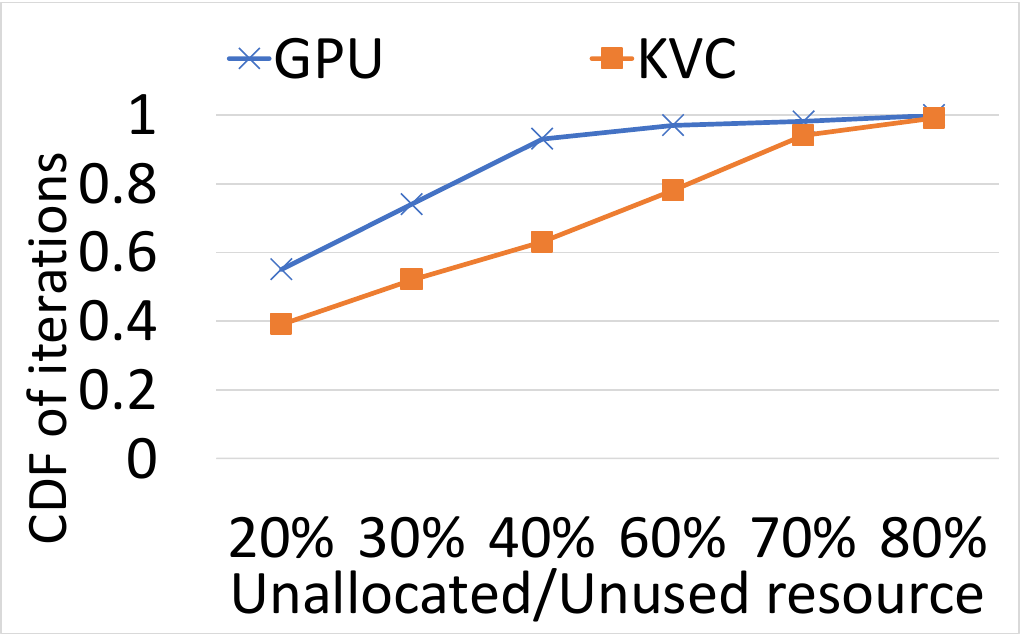} }}\vspace{-0.2in}
\caption{Available resources after each iteration.\vspace{-0.05in}}
    \label{fig:gap}
\end{wrapfigure}}

\DEL{Removed due to removal of figure 10...\noindent\textbf{Resource Utilization.} Figure~\ref{fig:gap} shows the CDF of iterations across all requests versus a certain amount of unused GPU or unallocated KVC after each iteration, respectively. 
Unused GPU is measured by ($(S_{b}-S_{f})/S_{b}$). 
The unused GPU and unallocated KVC exhibit variance after each iteration. It ranges from [18\%, 82\%] and [17\%, 89\%], respectively. 
We observe that 55\%  and 39\% of the iterations have 20\% of unused GPU and unallocated KVC, respectively. Further, 2\% and 5\% of the iterations have at least 70\% of unused GPU and unallocated KVC, respectively.}


\begin{thm}\label{DynamicChunking}
Heterogeneous SLOs cannot be met in current LLM systems, making it necessary to account for SLO diversity in scheduling.  
\end{thm}

\begin{thm}\label{EvenChunking}
An FCFS approach focused solely on maximizing GPU compute utilization underutilizes KVC resources, reducing throughput. Jointly optimizing GPU compute and KVC utilization per iteration is essential to improve throughput.
\end{thm}


\DEL{\begin{figure}
    \centering
\includegraphics[width=0.85\columnwidth,height=0.15\textheight]{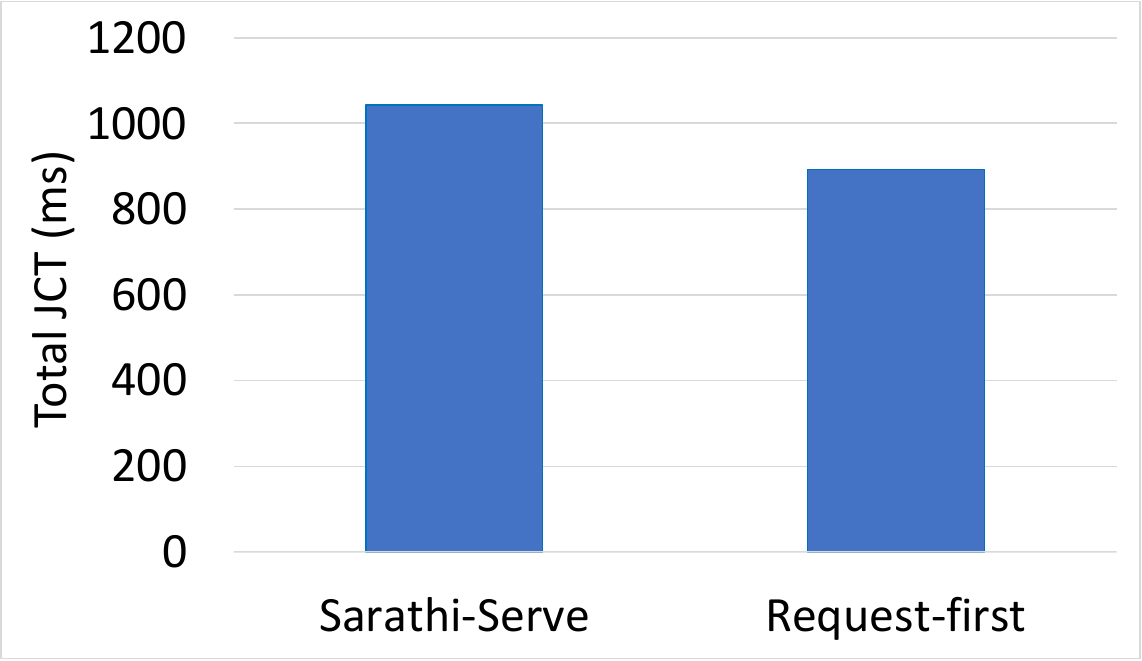}
    \caption{JCT of Sarathi-Serve and prioritizing chunks of one request. }
    \label{fig:total-jct}
\end{figure}}



\DEL{\begin{figure}[htbp]
\begin{minipage}[t]{0.48\linewidth}
\includegraphics[width=\linewidth,height=0.15\textheight]{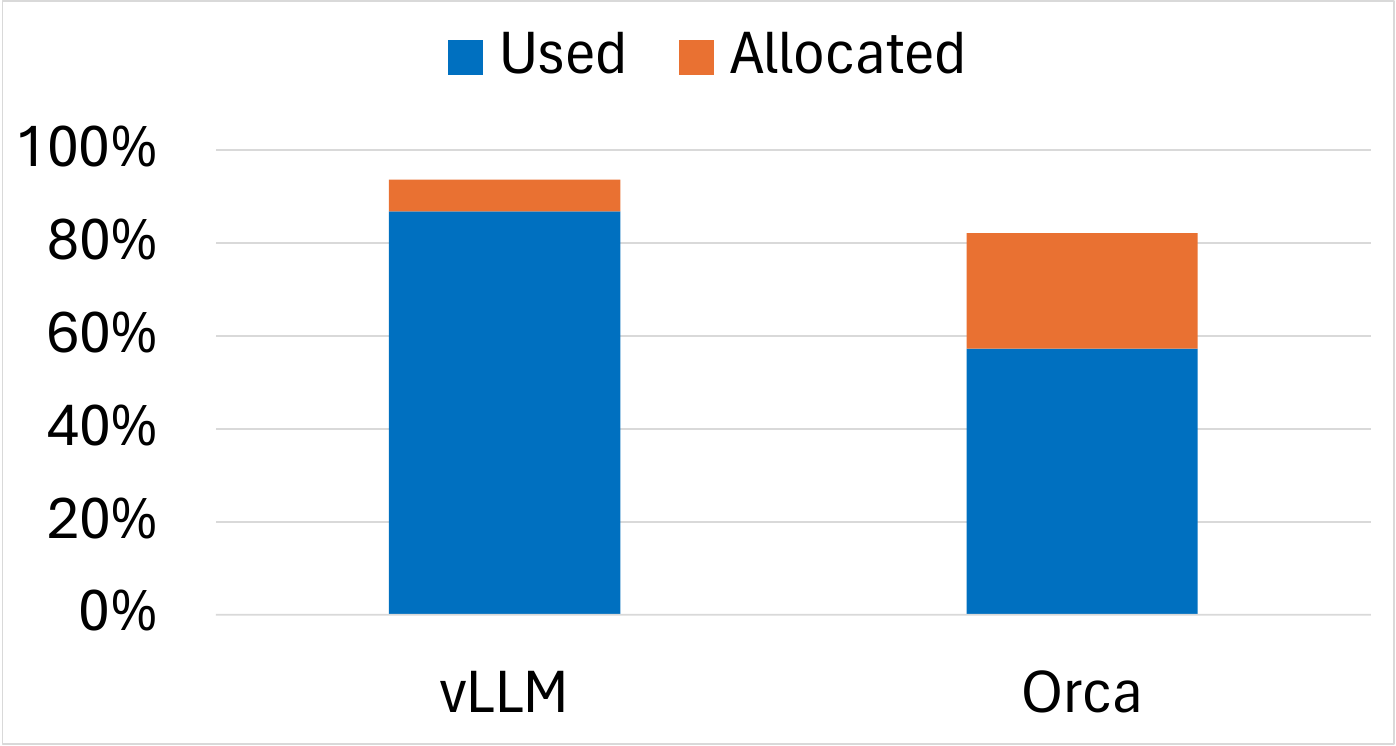}
\vspace{-0.01in}    \caption{Cache wastage.\vspace{-0.01in} }
    \label{fig:cache-wastage}
\end{minipage}%
    \hfill%
\begin{minipage}[t]{0.48\linewidth}
\includegraphics[width=\linewidth,height=0.15\textheight]{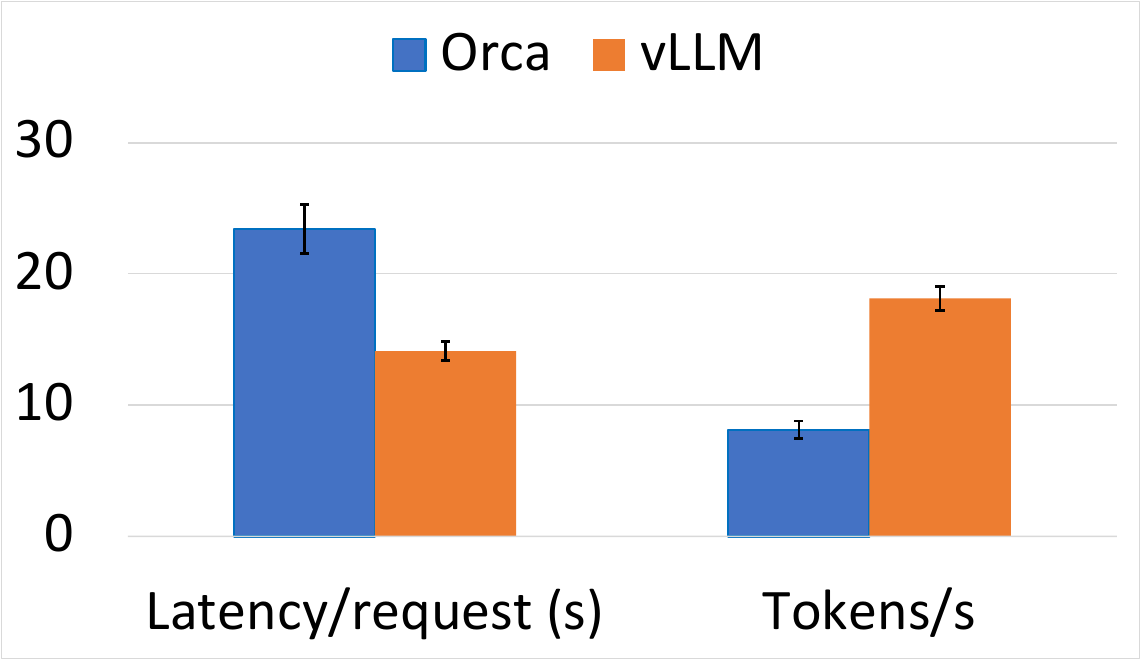}
\vspace{-0.01in} \caption{Latency and Throughput.\vspace{-0.05in} }
    \label{fig:throughput-token}
\end{minipage}
\end{figure}
}

\DEL{{\color{gray}{\noindent\textbf{Iteration-level SLO.} We ran 32 jobs and set the prompt-processing SLO to a value randomly chosen from  [0.2,2]s, and set the iteration-SLO as 0.1875s. Their sum for a job is its JCT SLO. We measured the percentage of jobs that meet their JCT SLOs. Then,
we replaced five randomly selected prompts with five 6144-token prompts and measured the results again. Figure~\ref{fig:slo-cdf} shows the CDF of the results. We observe that 
only 82\%,  73\% and 61\% of the jobs meet their JCT SLO for the batch size of 2, 4 and 8, respectively, when there are no long prompts, When there are long prompts,  the SLO attainment drops by approximately 35\% for all the batch sizes. Moreover, the average time to generate tokens for the jobs, which met their JCT SLOs is 0.47 seconds; approximately 3$\times$ higher than the usual 0.1875s/token speed.
}}}

\DEL{{\color{gray}However, when we add prompts to the batch to reach token budget in order to maximize GPU compute utilization, we need to consider the constraint from the KVC. The added prompts should be able to receive enough KVC for their prompts and also won't exacerbate the KVC bottleneck in their token generation. Therefore, we study the performance on KVC allocation of LLM systems. }}


\DEL{{\color{gray}{We then conducted an experiment to verify the influence of long prompts on other requests in the mix-prompt scenario where the majority of requests are short prompts. 
We tested on two scenarios, denoted by $S_1$ and $S_2$. In $S_1$, we chose the prompt length of the requests randomly from [64, 2048]. In $S_2$, in all the created batches, we replaced the shortest-prompt with one prompt of length 6k. 
Figure~\ref{fig:long-batch} shows the average
GPU compute utilization, memory utilization, and response latency (for PP and 1 TG) for different batch sizes.
We observe that as the batch size increases, all metric values increase. In $S_2$, the response latency increases by approximately 3 $\times$, and the GPU compute utilization increases by approximately 3\% and memory utilization also increases by approximately 25\% caused by the long prompt.}}}


 \DEL{\begin{thm}\label{thm9}
  When different requests have different JCT SLOs, \Orca's JCT SLO attainment is low (71.8\%), which reduces by 35\% after adding long prompts. Also, even though JCT SLOs are satisfied, a user may wait 0.47 seconds for a token, which impairs user experience. [Figure~\ref{fig:slo-cdf}]
\end{thm}}  

\DEL{\vspace{-0.1in}
\begin{thm}\label{thm9}
\Orca's JCT SLO attainment is low and becomes even lower with long prompts.
Also, even though a job's JCT SLO is satisfied, a user may wait long for a token.
\end{thm}
\vspace{-0.1in}
}

\DEL{\sh{draw a fig: X: Num. of concurrent prompts in a batch, Y: Avg. occupied KVC of these prompts-done}
Figures~\ref{fig:prompt-batch} and~\ref{fig:prompt-batch-175} show \tsr{the average occupied KVC by batches for the numbers of concurrent prompts over all the iterations.} \sh{It is the previous expeirment and you collect the data from the experimant. You do not do a separate exp. and vary the X yourself-done, we just group the batches based on on number of prompts .} in a batch. For this analysis, the prompts are grouped into equal bin sizes of 25, covering a range of [1, 125].
From the figures, we observe that as the number of concurrent prompts in the batch increases, the occupied KVC also increases, potentially leading to overflow.}

\noindent\textbf{Order of Chunks in KVC Allocation.} Sarathi-Serve allocates the token budget across chunks of different long prompts, increasing concurrent long-prompt requests. We argue that reducing concurrency can alleviate KVC limitations by fully allocating the token budget to one request before moving to the next, allowing for quicker KVC release. We call this method \emph{Exclusive Request Allocation (ERA)}. To validate this, we compared ERA and Sarathi-Serve using two long prompts (10,214 and 10,252 tokens) and 20 short prompts (128-384 tokens). Figures~\ref{fig:jct} and~\ref{fig:jct-175} show that ERA reduces the JCT for both long-prompt jobs and the overall workload via minimizing concurrent KVC occupancy by long-prompt requests.

\DEL{To validate this, we conducted an experiment to compare ERA and Sarathi-Serve. We selected two long prompts with prompt lengths of 10,214 and 10,252, along with 20 short prompts ranging from 128 to 384 tokens. Figures~\ref{fig:jct} and~\ref{fig:jct-175} illustrate the total JCT for the entire workload and the JCT for each long-prompt job. Compared to Sarathi-Serve, ERA reduces the JCT for both the long-prompt jobs and the overall workload via minimizing the concurrent occupancy of the KVC by the two long prompts.}

\begin{figure}[t]
\centering
    \subfloat[OPT-13B.\vspace{-0.01in}\label{fig:slo-order-13b}]{{\includegraphics[width=0.48\linewidth,height=0.145\textheight]{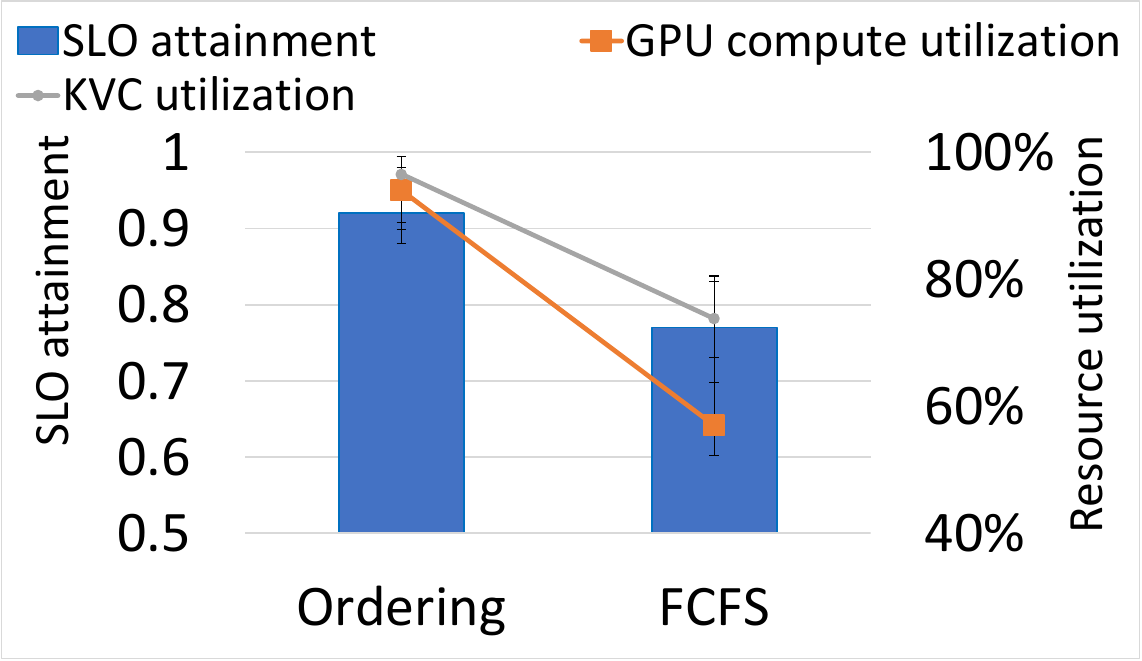} }}
    \hfill
    \DEL{\subfloat[CDF of difference of chunk size and available GPU/KVC.\vspace{-0.01in}\label{fig:case-analysis-2}]{{\includegraphics[width=0.32\linewidth,height=0.145\textheight]{Figures/cdf-case-wise.pdf} }}
    \hfill}
    \DEL{\subfloat[OPT-175B.\vspace{-0.01in}\label{fig:jct-175}]{{\includegraphics[width=0.48\linewidth,height=0.145\textheight]{Figures/total-jct-175.pdf} }}}
    \subfloat[OPT-175B.\vspace{-0.01in}\label{fig:slo-order-175b}]{{\includegraphics[width=0.48\linewidth,height=0.145\textheight]{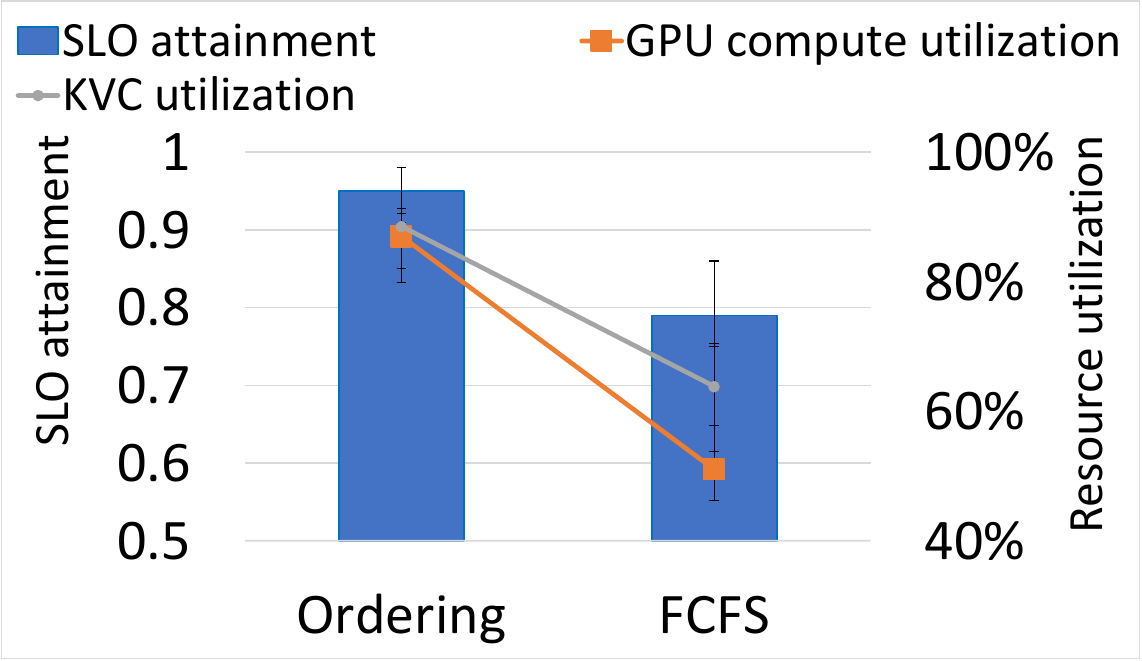} }}
    \hfill
    
    \DEL{\subfloat[Percentage of prompt tasks.\vspace{-0.01in}\label{fig:14-4}]{{\includegraphics[width=0.24\linewidth,height=0.15\textheight]{Fig/14-4-up-3.pdf} }}
    \hfill}
   \caption{Importance of SLO-based ordering. \vspace{-0.0in}}
    \label{fig:slo-ordering}
\end{figure}

\DEL{\begin{figure}[t]
\centering
    \subfloat[OPT-13B.\vspace{-0.01in}\label{fig:slo-wise}]{{\includegraphics[width=0.48\linewidth,height=0.15\textheight]{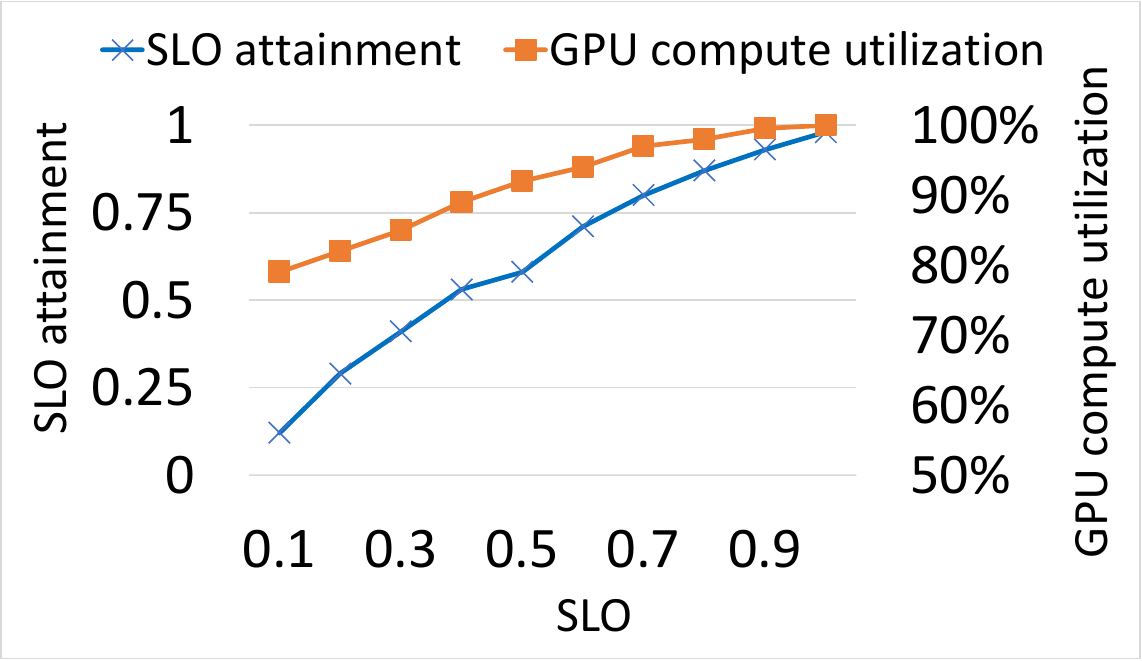} }}
    \hfill
    \DEL{\subfloat[CDF of difference of chunk size and available GPU/KVC.\vspace{-0.01in}\label{fig:case-analysis-2}]{{\includegraphics[width=0.32\linewidth,height=0.15\textheight]{Figures/cdf-case-wise.pdf} }}
    \hfill}
    \DEL{\subfloat[OPT-175B.\vspace{-0.01in}\label{fig:jct-175}]{{\includegraphics[width=0.48\linewidth,height=0.15\textheight]{Figures/total-jct-175.pdf} }}}
    \subfloat[OPT-175B.\vspace{-0.01in}\label{fig:slo-wise-175}]{{\includegraphics[width=0.48\linewidth,height=0.15\textheight]{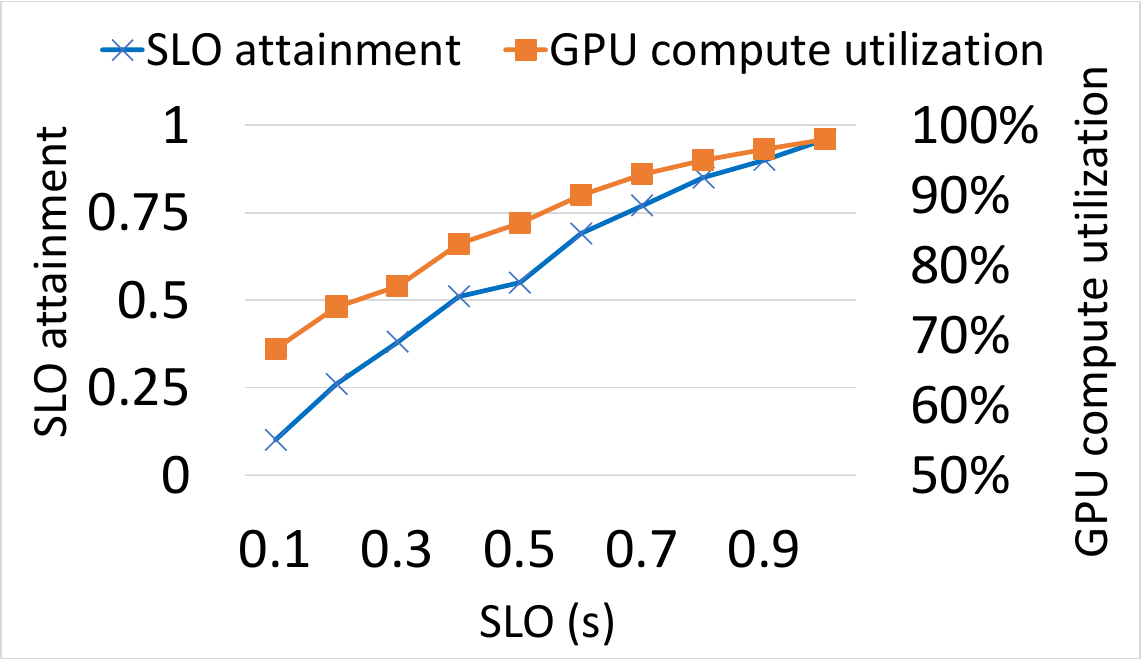} }}
    \hfill
    
    \DEL{\subfloat[Percentage of prompt tasks.\vspace{-0.01in}\label{fig:14-4}]{{\includegraphics[width=0.24\linewidth,height=0.15\textheight]{Fig/14-4-up-3.pdf} }}
    \hfill}
   \caption{SLO-wise performance.\vspace{-0.0in}}
    \label{fig:combined-175}
\end{figure}}

\DEL{\begin{figure}[t]
   \begin{minipage}[!t]{0.235\textwidth}
    \centering
\includegraphics[width=0.96\columnwidth,height=0.15\textheight]{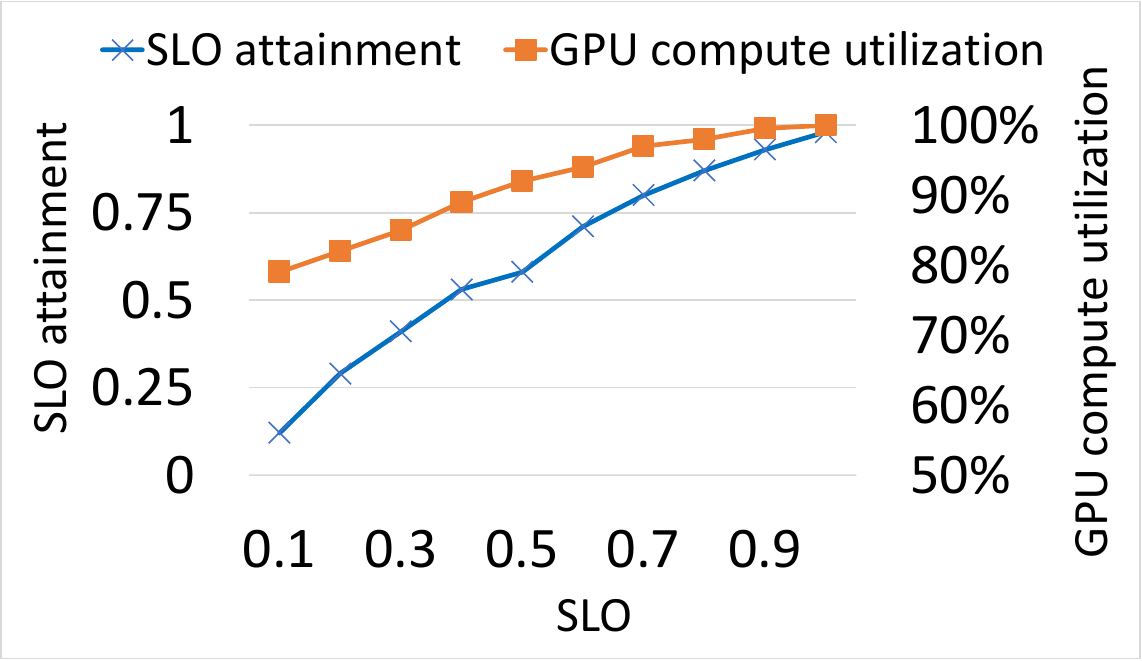}
   \vspace{-0.05in} \caption{SLO-wise performance for OPT-13B.}
    \label{fig:slo-wise}\vspace{-0.05in}
 \end{minipage}
    \begin{minipage}[!t]{0.235\textwidth}
    \centering
\includegraphics[width=0.96\columnwidth,height=0.15\textheight]{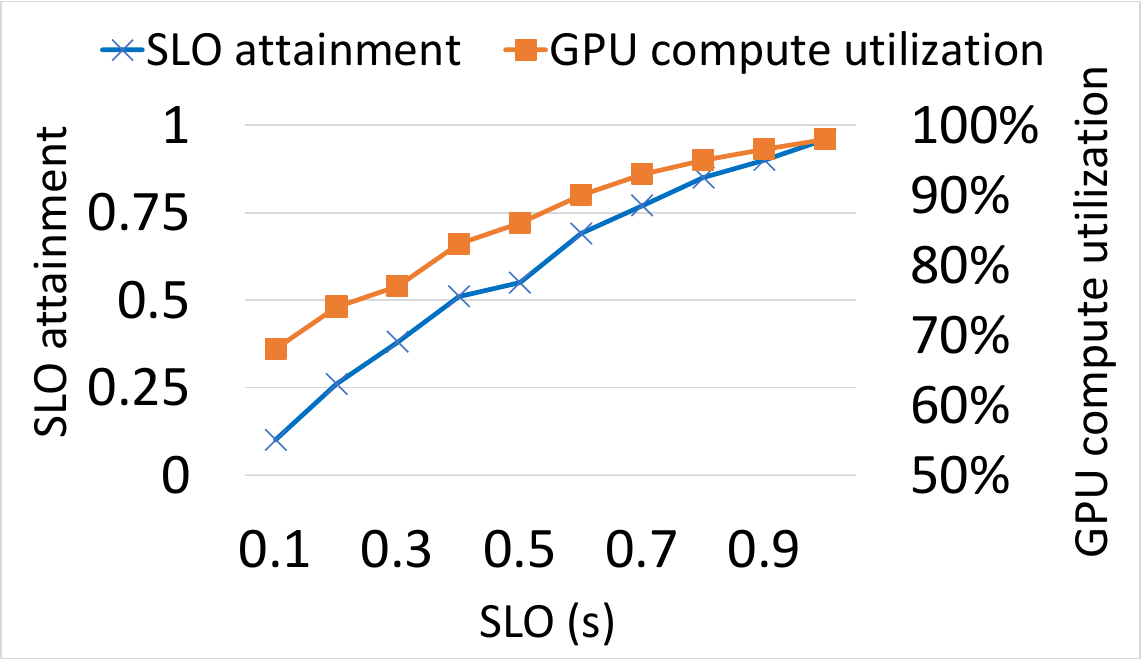}
   \vspace{-0.05in} \caption{SLO-wise performance for OPT-175B.}
    \label{fig:slo-wise-175}\vspace{-0.05in}
    \end{minipage}
    \label{fig:combined-175}
\end{figure}}

\DEL{\begin{figure}[t]
\centering
    \subfloat[OPT-13B.\vspace{-0.01in}\label{fig:overhead}]{{\includegraphics[width=0.48\linewidth,height=0.15\textheight]{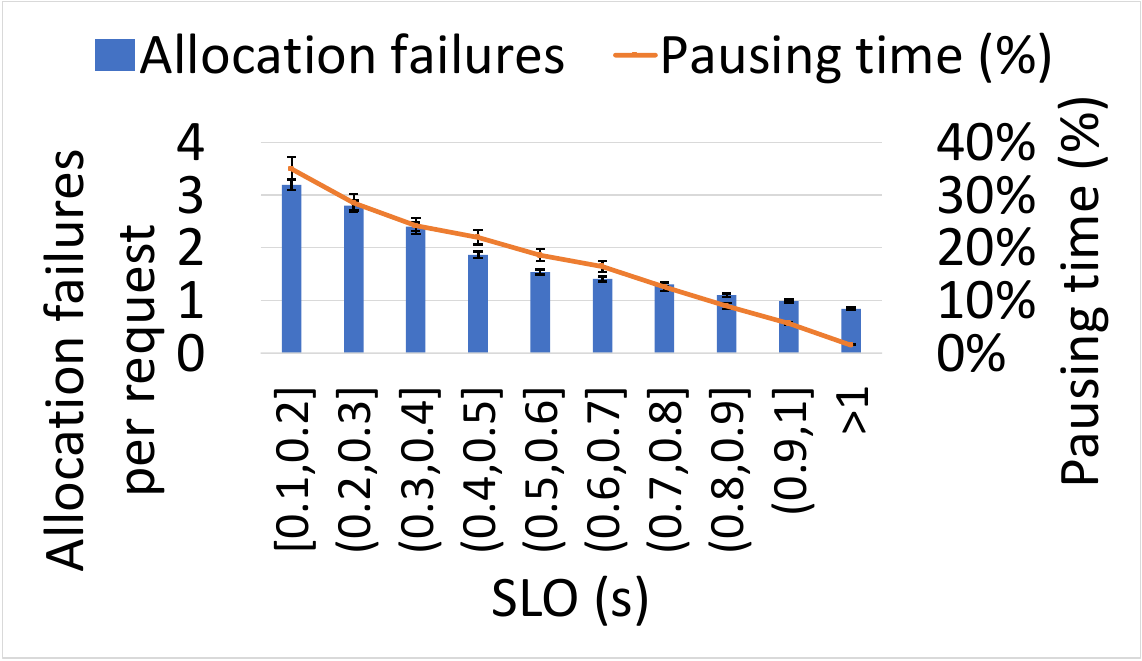} }}
    \hfill
    \DEL{\subfloat[CDF of difference of chunk size and available GPU/KVC.\vspace{-0.01in}\label{fig:case-analysis-2}]{{\includegraphics[width=0.32\linewidth,height=0.15\textheight]{Figures/cdf-case-wise.pdf} }}
    \hfill}
    \DEL{\subfloat[OPT-175B.\vspace{-0.01in}\label{fig:jct-175}]{{\includegraphics[width=0.48\linewidth,height=0.15\textheight]{Figures/total-jct-175.pdf} }}}
    \subfloat[OPT-175B.\vspace{-0.01in}\label{fig:overhead-175b}]{{\includegraphics[width=0.48\linewidth,height=0.15\textheight]{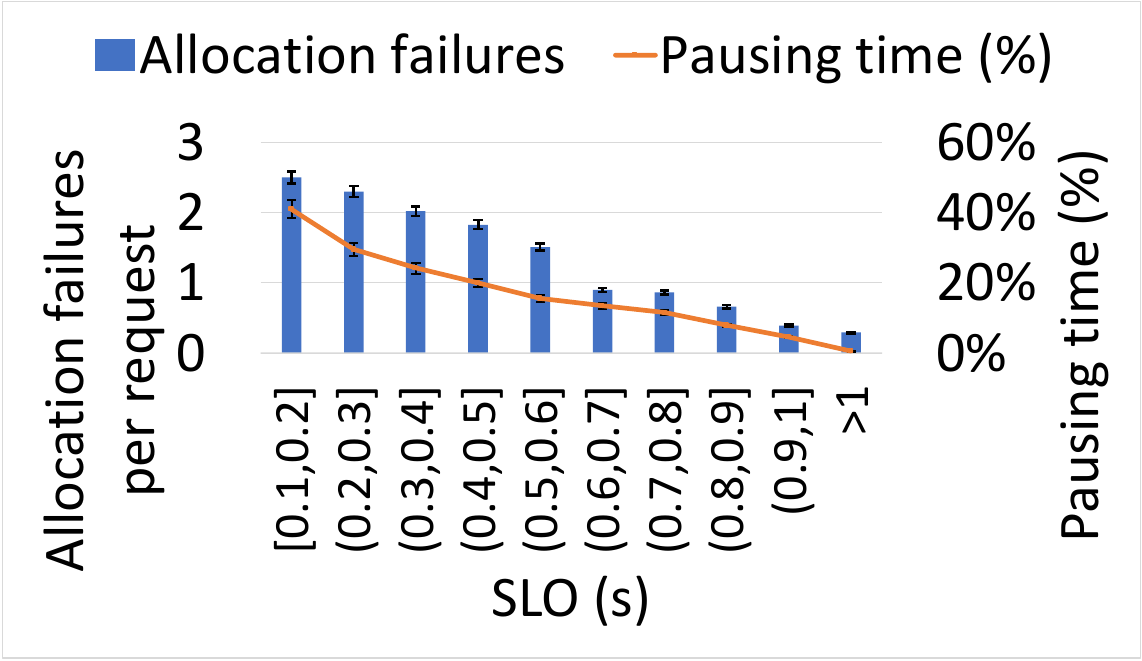} }}
    \hfill
    
    \DEL{\subfloat[Percentage of prompt tasks.\vspace{-0.01in}\label{fig:14-4}]{{\includegraphics[width=0.24\linewidth,height=0.15\textheight]{Fig/14-4-up-3.pdf} }}
    \hfill}
   \caption{Allocation failures and pausing time.\vspace{-0.0in}}
    \label{fig:combined-3}
\end{figure}}

\DEL{\begin{figure}[t]
    \centering
    \begin{minipage}[!t]{0.235\textwidth}
    \centering
\includegraphics[width=0.96\columnwidth,height=0.15\textheight]{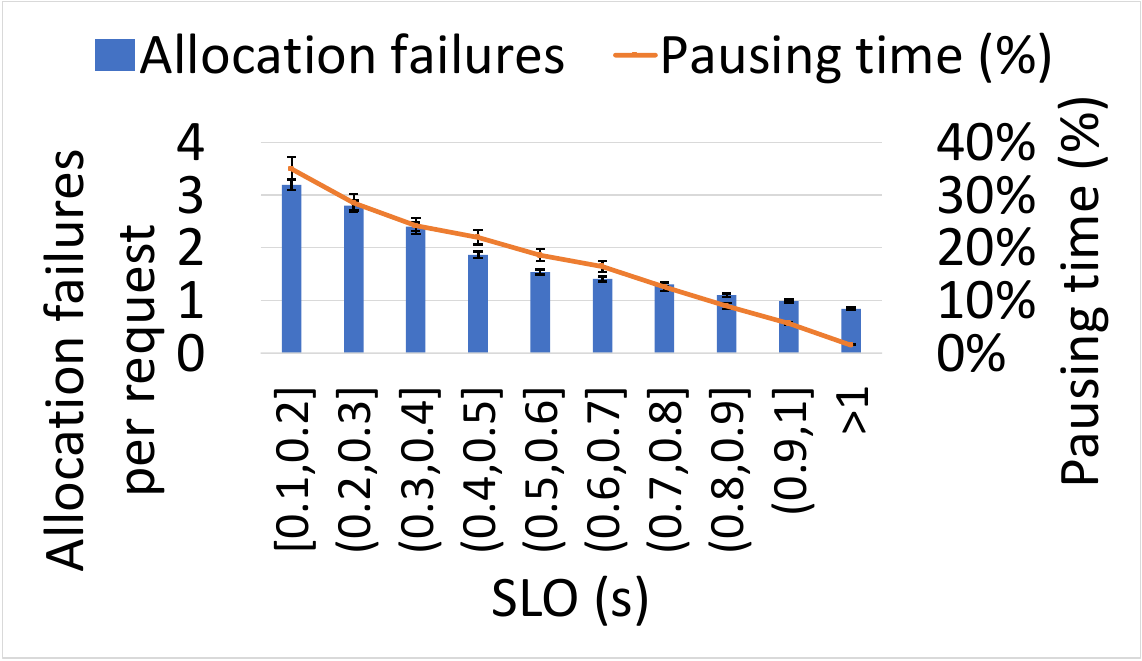}
\vspace{-0.05in}\caption{Allocation failures and pausing time for OPT-13B.}
\label{fig:overhead}\vspace{-0.05in}
\end{minipage}
    \begin{minipage}[!t]
    {0.235\textwidth}\centering
\includegraphics[width=0.96\columnwidth,height=0.15\textheight]{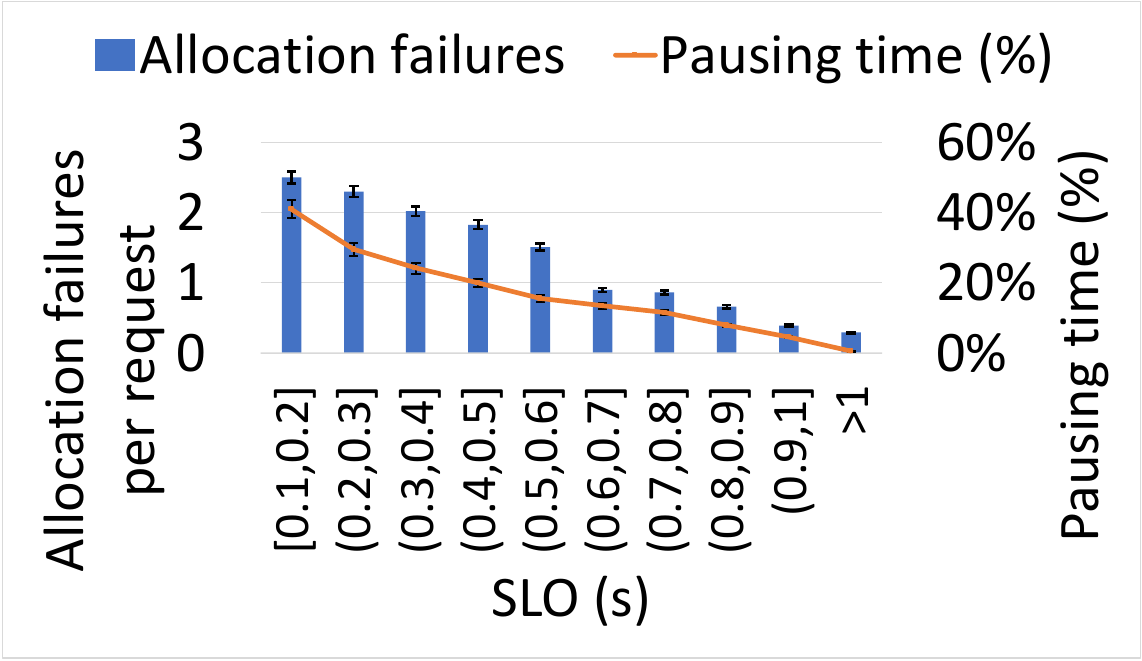}
\vspace{-0.05in}\caption{Allocation failures and pausing time for OPT-175B.}
\label{fig:overhead-175b}\vspace{-0.05in}    
    \end{minipage}
    \label{fig:combined-3}
\end{figure}}

\DEL{\begin{figure}[t]
\centering
    \subfloat[OPT-13B.\vspace{-0.01in}\label{fig:slo-waiting}]{{\includegraphics[width=0.48\linewidth,height=0.15\textheight]{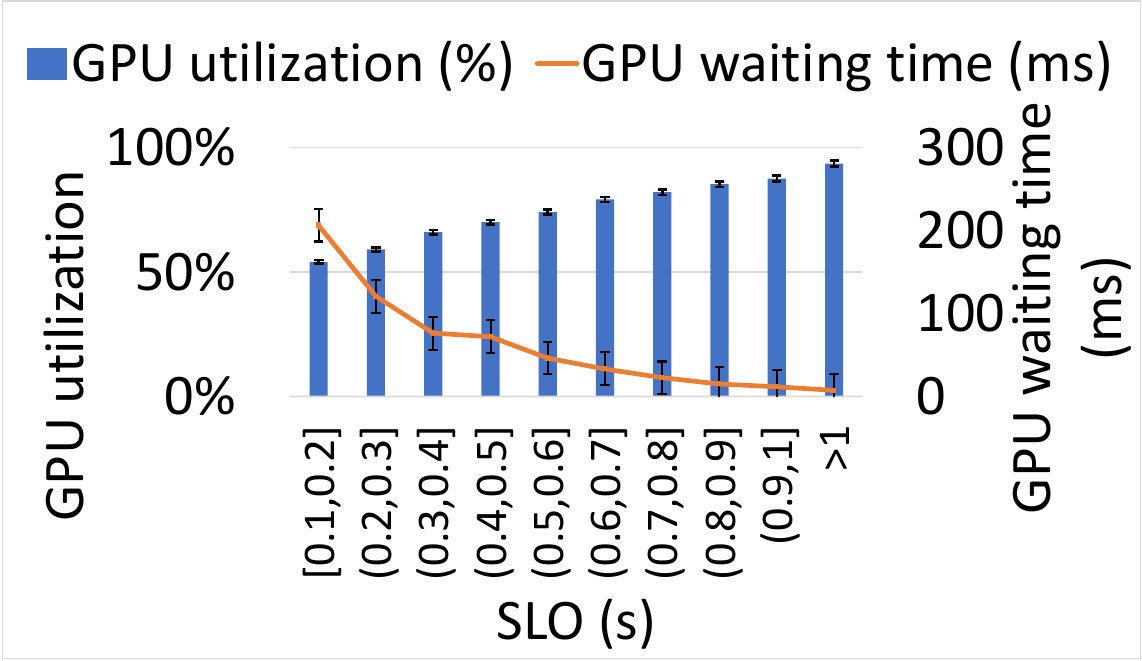} }}
    \hfill
    \DEL{\subfloat[CDF of difference of chunk size and available GPU/KVC.\vspace{-0.01in}\label{fig:case-analysis-2}]{{\includegraphics[width=0.32\linewidth,height=0.15\textheight]{Figures/cdf-case-wise.pdf} }}
    \hfill}
    \DEL{\subfloat[OPT-175B.\vspace{-0.01in}\label{fig:jct-175}]{{\includegraphics[width=0.48\linewidth,height=0.15\textheight]{Figures/total-jct-175.pdf} }}}
    \subfloat[OPT-175B.\vspace{-0.01in}\label{fig:slo-waiting-175b}]{{\includegraphics[width=0.48\linewidth,height=0.15\textheight]{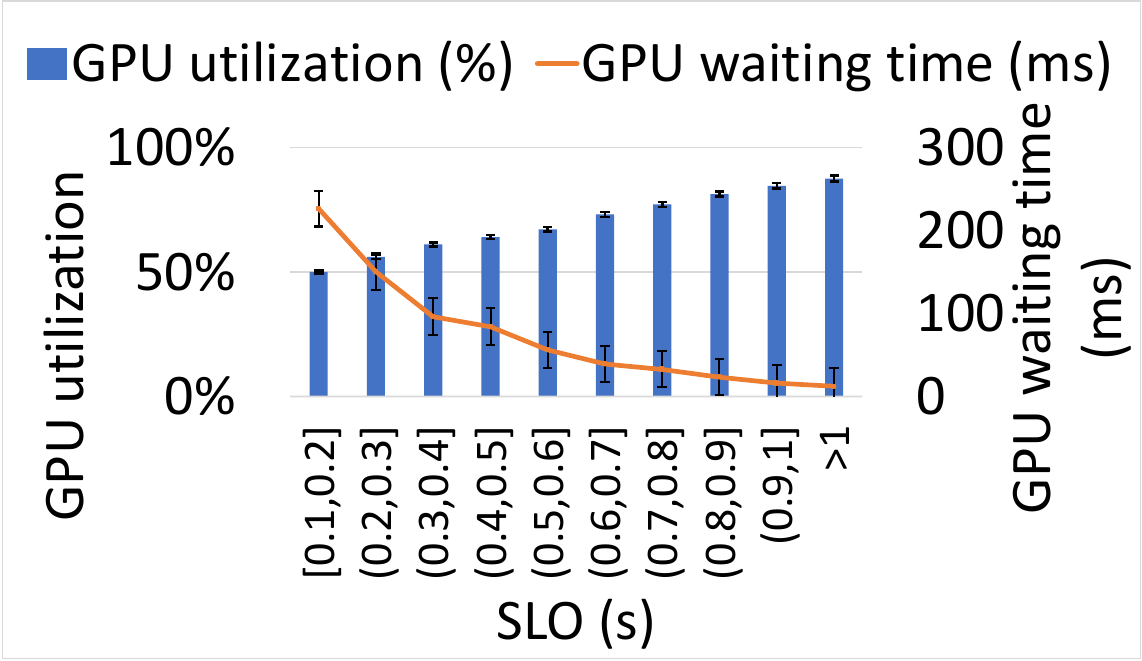} }}
    \hfill
    
    \DEL{\subfloat[Percentage of prompt tasks.\vspace{-0.01in}\label{fig:14-4}]{{\includegraphics[width=0.24\linewidth,height=0.15\textheight]{Fig/14-4-up-3.pdf} }}
    \hfill}
   \caption{GPU compute utilization and GPU waiting time.\vspace{-0.0in}}
    \label{fig:combined-3-175}
\end{figure}}

\DEL{\begin{figure}[t]
    \centering
    \begin{minipage}[!t]{0.235\textwidth}
\centering
\includegraphics[width=0.96\columnwidth,height=0.15\textheight]{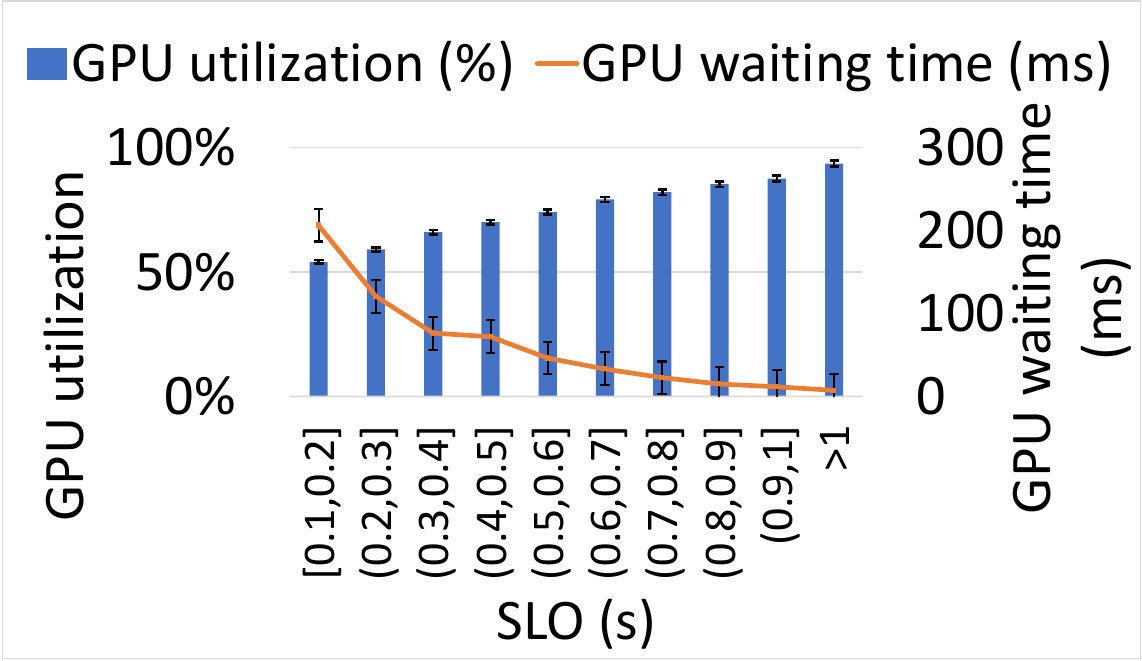}
   \vspace{-0.05in} \caption{GPU compute utilization and GPU waiting time for OPT-13B.}
    \label{fig:slo-waiting}\vspace{-0.05in}
\end{minipage}
    \begin{minipage}[!t]{0.235\textwidth}
    \centering
\includegraphics[width=0.96\columnwidth,height=0.15\textheight]{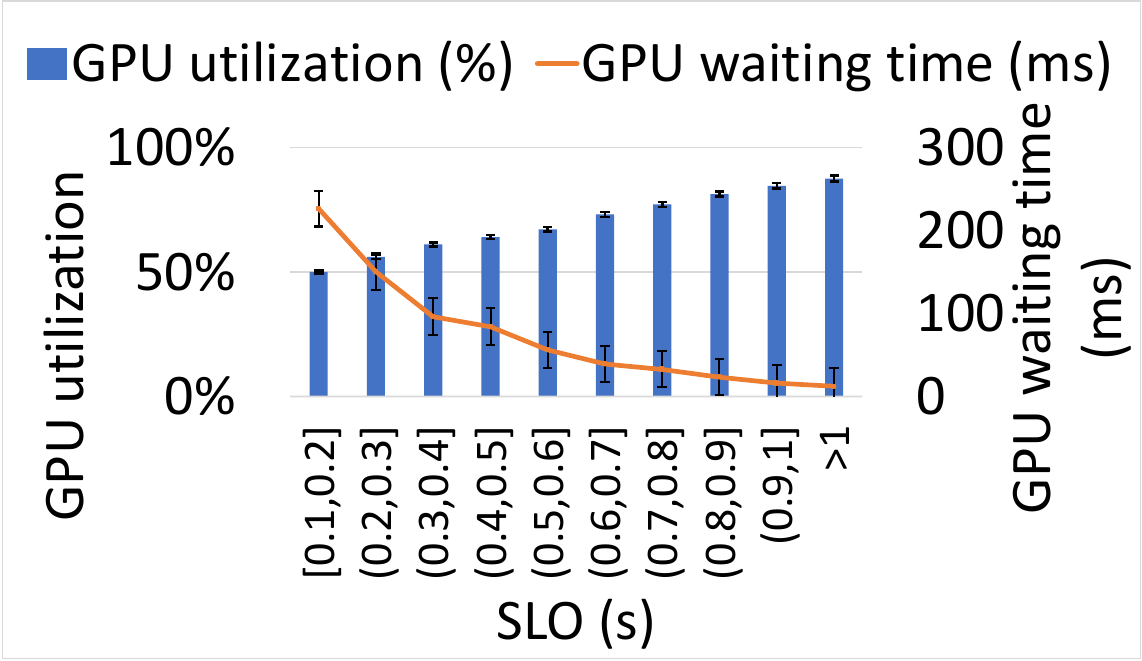}
   \vspace{-0.05in} \caption{GPU compute utilization and GPU waiting time for OPT-175B.}
    \label{fig:slo-waiting-175b}\vspace{-0.05in}
    \end{minipage}
    \label{fig:combined-3-175}
\end{figure}}

\DEL{While Sarathi-Serve optimizes by filling the token budget with alternate prompts, the alternative method prioritizes chunks from the first long prompt until it is fully processed before selecting chunks from the second prompt. We observe that this method of prioritizing chunks within a single request reduces the overall JCT of the prompts.}


\begin{thm}\label{kvcache2}
Picking up chunks from different long-prompts increases their KVC occupying time and thus reduces throughput (reqs/s) compared to ERA. 
\end{thm}

\DEL{A KVC allocation failure occurs when the first queuing request or a running request cannot receive its demanded KVC space. \DEL{Figure~\ref{fig:fail-total-2} shows the density of failures for waiting requests during batching versus the allocated KVC percentage after each iteration. As the allocated KVC increases, the density of failures also increases. When the KVC has a smaller available space, more failures occur. When queuing requests cannot be added to the batch, it increases the waiting time, and limits the batch size and throughput. 
Figure~\ref{fig:fail-total-4} shows the batch size and throughput, measured in requests/s, versus allocated KVC \% after each iteration. It is evident that as the allocated KVC increases, the batch size and throughput decrease, indicating that long prompts reduce throughput by occupying the KVC. Therefore, there is a need to reduce the number of concurrently running long-prompt jobs.
KVC failures arise when either a running request exhausts its allocated block but cannot receive a new block or the first waiting request cannot be allocated due to insufficient KVC.} Figure~\ref{fig:block-density} illustrates the distribution of KVC allocation failures, distinguishing between running requests and waiting requests corresponding to the allocated KVC. We see that higher allocated KVC leads to more failures, and waiting requests are the primary contributors to KVC allocation failures. Additionally, Figure~\ref{fig:cdf-delay} depicts the CDF of KVC allocation failures of waiting requests versus the delay from the occurrence of failure to the time when the required KVC is received. These failures result in significant waiting times, ranging from 0.25s to 3.5s.}

\noindent\textbf{Tradeoff Between Throughput and Heterogeneous SLO Compliance.} Figure~\ref{fig:slo-ordering} compares iteration SLO attainment, GPU compute utilization, and KVC utilization in Sarathi-Serve using FIFO and request ordering based on the remaining time to their SLOs. When batching requests, we ensure the iteration time doesn't exceed the most stringent SLO in the batch, with iteration time estimated from profiled data. Compared to FIFO, the ordering method achieves 37\% and 34\% higher GPU compute and KVC utilization, respectively, and 16\% higher SLO attainment. In FIFO, requests with large SLO variance can end up in the same batch, where stringent-SLO requests reduce batch size though loose-SLO requests allow for larger batch sizes. In contrast, the SLO-based ordering policy groups requests with similar SLOs, improving resource utilization ad throughput.

\begin{thm}\label{pivot}
\DEL{In Sarathi-serve, the system-wide SLO setting 
may fail to satisfy more stringent SLO requirements. 
Thus, we need to consider heterogeneous SLOs of different requests.
(?focus more on heterogeneous SLOs at the beginning, Section 2-3)}

The FCFS policy underutilizes GPU resources when meeting iteration time SLOs, but batching requests with similar SLOs enhances throughput.

\DEL{Consistent with the theoretical expectations in Equation~\eqref{eq:totop}, the throughput increases with the forward size until it reaches the pivot forward size. 
\DEL{Many batches exhibit a forward size smaller than the pivot forward size ($S_{b}$), while others surpass $S_{b}$ due to lengthy prompts. This suggests that} 
GPU resources are underutilized most of the time and overutilized by long prompts, leading to high iteration times.} 

\DEL{\sh{add a fig: avg, p5 and p95 of KVC allocation failures per requst, avg, p5 and p95 of pausing time \% due to KVC allocation failures-done}
\sh{[??: --different requests has different SLOs. Your figs show all system has the same SLO — it is wrong. The purpose is to show it is important to order requests based on iteration-based SLOs-done, drew for range of SLO.}

\sh{use model parallelism,  different SLOs, we have different token budgets. Do not order requests based on SLOs. Add figures for GPU compute utilization, and bubbles, i.e., a GPU's waiting time, two bars for the two metrics, each bar as p5 and p95-done.}
}


\end{thm}

\DEL{\vspace{-0.05in}
(??can remove this Observation)\begin{thm}
By employing the block-based KVC allocation, the KVC allocation failures for generation tasks lead to high JCT delay for a request...?
\end{thm}
}


\DEL{\begin{figure}[]
\centering
\DEL{\subfloat[Density of prompt length upon allocation failure.\vspace{-0.01in}\label{fig:fail-overall}]{{\includegraphics[width=0.32\linewidth,height=0.15\textheight]{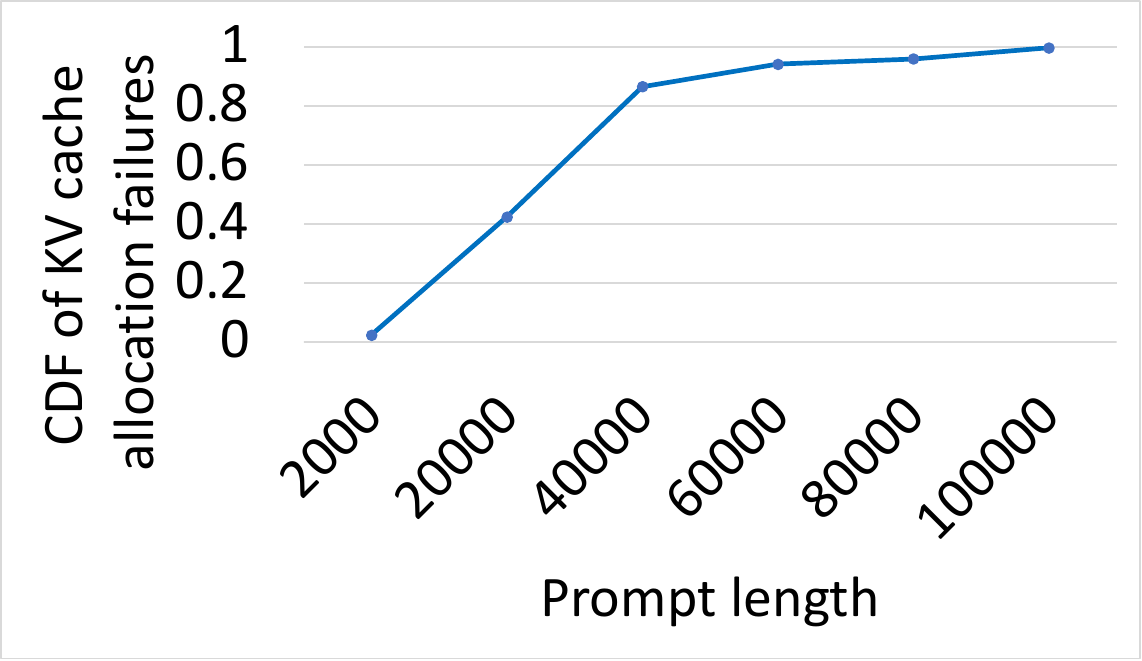} }}
    \hfill}
\subfloat[Density of prompt length upon  failures.\vspace{-0.01in}\label{fig:fail-overall}]{{\includegraphics[width=0.32\linewidth,height=0.15\textheight]{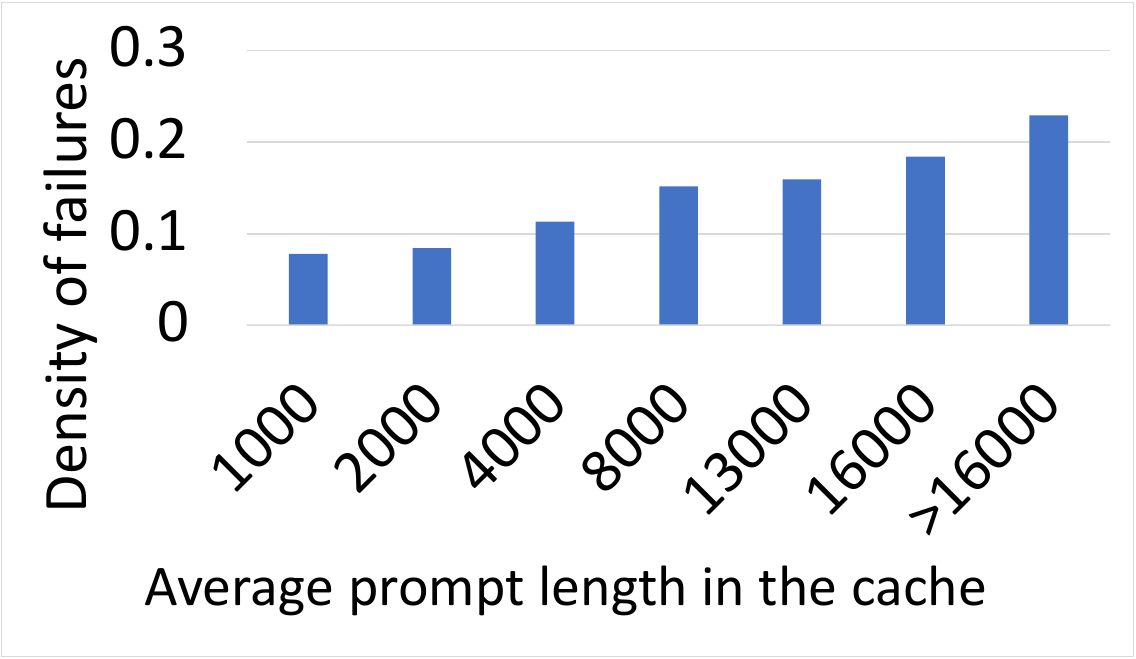} }}
    \hfill
\subfloat[Density of prompt length upon  failures in waiting.\vspace{-0.01in}\label{fig:fail-waiting}]{{\includegraphics[width=0.32\linewidth,height=0.15\textheight]{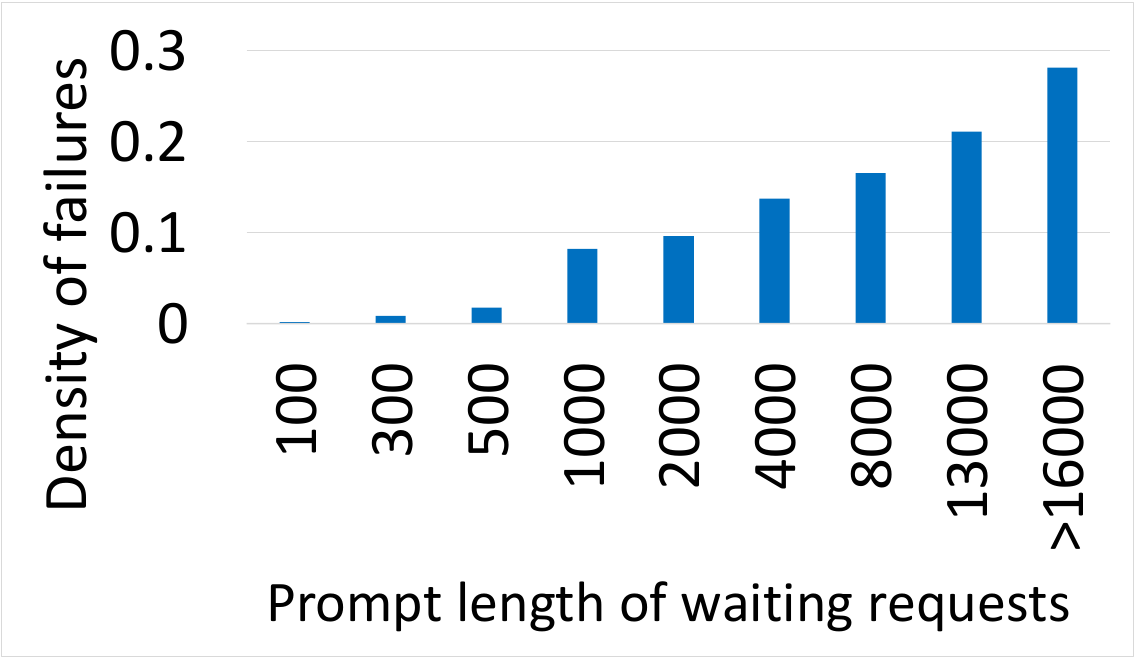} }}
    \hfill
    \subfloat[Batch size and throughput upon failures.\vspace{-0.01in}\label{fig:bt-overall}]{{\includegraphics[width=0.32\linewidth,height=0.15\textheight]{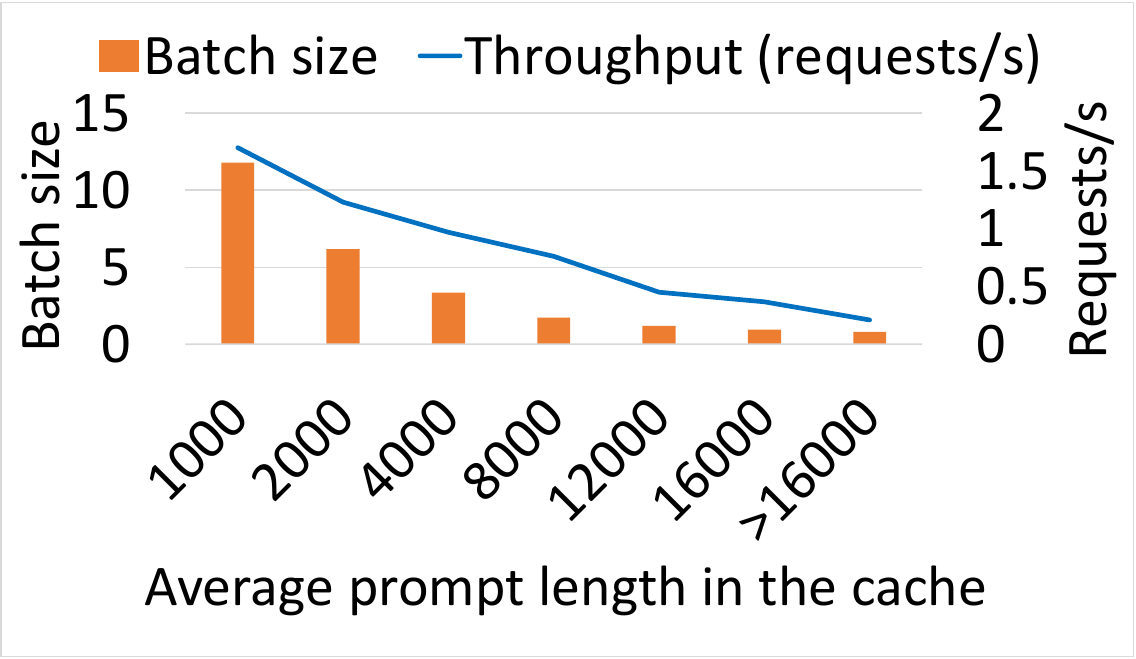} }}
    \hfill
  %
\vspace{-0.05in}
   \caption{\small{Impact of long prompts on KVC allocation.\sh{keep these figs}\vspace{-0.1in}}}%
    \label{fig:pp-impact} \vspace{-0.1in}
\end{figure}

\begin{figure}[]
\centering
\DEL{\subfloat[Density of prompt length upon allocation failure.\vspace{-0.01in}\label{fig:fail-overall}]{{\includegraphics[width=0.32\linewidth,height=0.15\textheight]{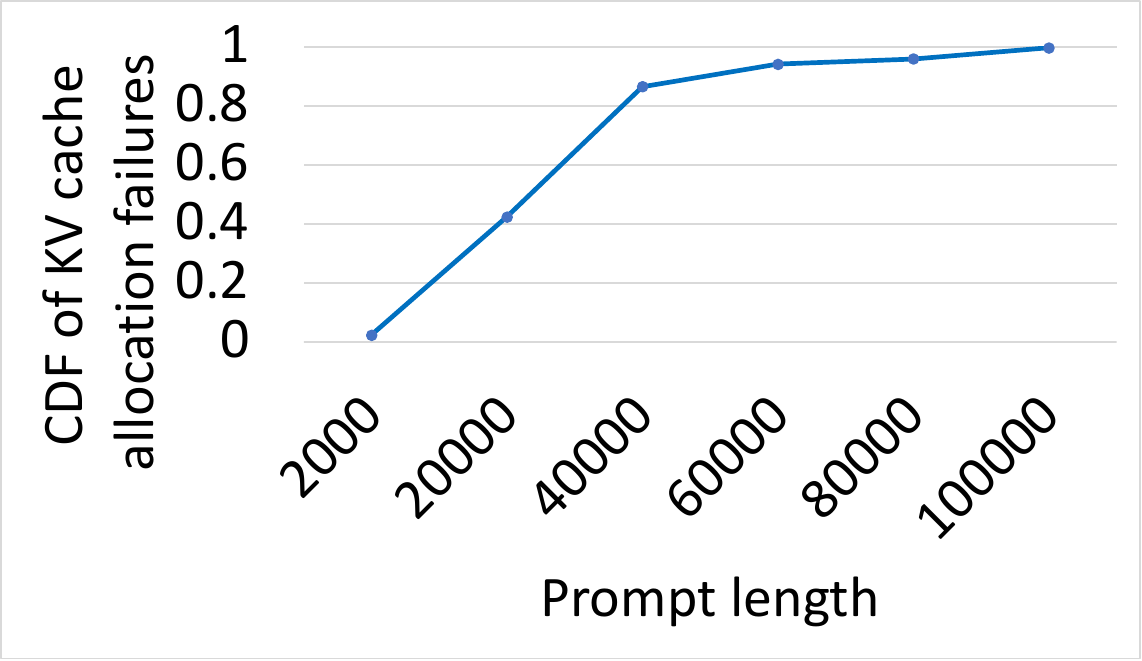} }}
    \hfill}
\subfloat[Density of prompt length upon  failures for current X-value.\vspace{-0.01in}\label{fig:fail-overall-2}]{{\includegraphics[width=0.48\linewidth,height=0.15\textheight]{NewFigs/total-8a.pdf} }}
    \hfill
\DEL{\subfloat[Density of prompt length upon  failures in waiting for 3-new category.\vspace{-0.01in}\label{fig:fail-waiting}]{{\includegraphics[width=0.32\linewidth,height=0.15\textheight]{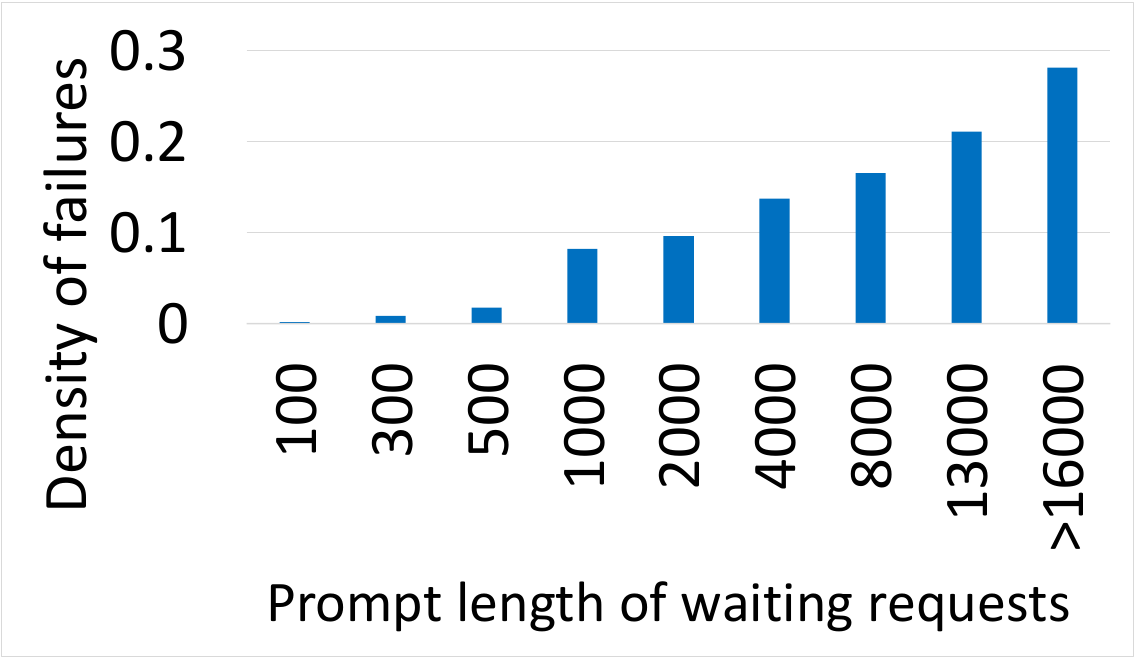} }}
    \hfill}
    \subfloat[Batch size and throughput upon failures for current X-value.\vspace{-0.01in}\label{fig:bt-overall-2}]{{\includegraphics[width=0.48\linewidth,height=0.15\textheight]{NewFigs/total-batch-8b.pdf} }}
    \hfill
  %
\vspace{-0.05in}
   \caption{\small{Impact of long prompts on KVC allocation.\sh{keep these figs}\vspace{-0.1in}}}%
    \label{fig:pp-impact} \vspace{-0.1in}
\end{figure}
}

\DEL{\begin{figure*}[!t]
\centering
\subfloat[Density of total prompt length upon allocation failure in cache.\vspace{-0.01in}\label{fig:fail-total-1}]{{\includegraphics[width=0.24\linewidth,height=0.15\textheight]{NewFigs/desity-failure-total.pdf} }}
    \hfill
\subfloat[Density of allocated KVC \% upon  failures\sh{need to check if it is prompt or all occupied}.\vspace{-0.01in}\label{fig:fail-total-2}]{{\includegraphics[width=0.24\linewidth,height=0.15\textheight]{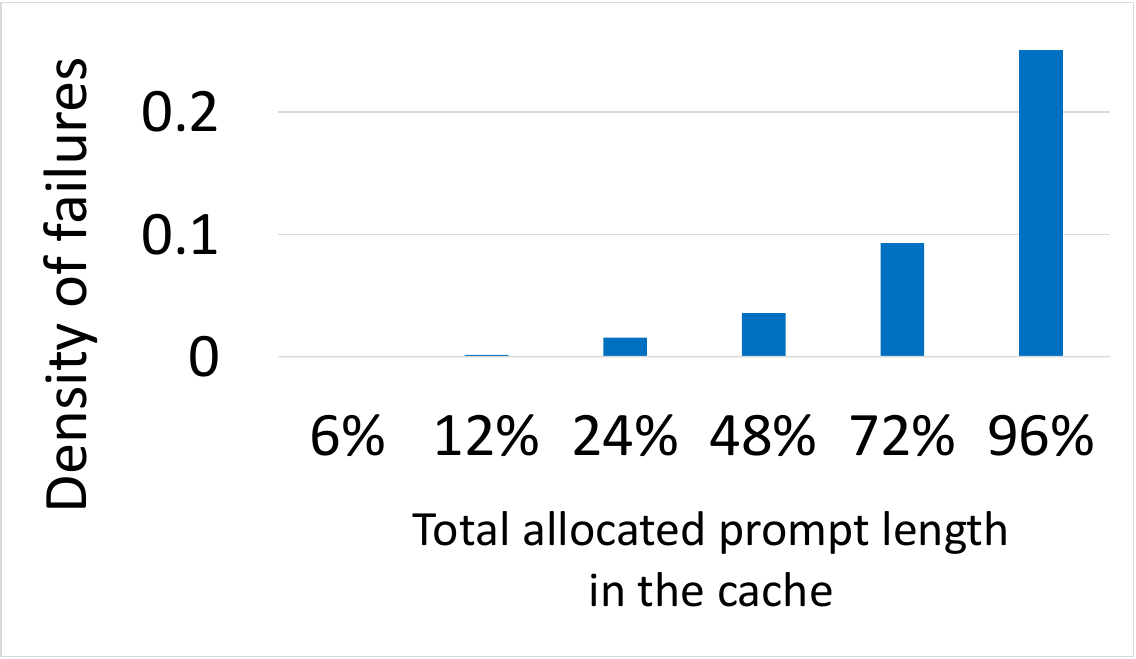} }}
    \hfill
    \subfloat[Batch size and throughput for total prompt length upon failures.\vspace{-0.01in}\label{fig:fail-total-3}]{{\includegraphics[width=0.24\linewidth,height=0.15\textheight]{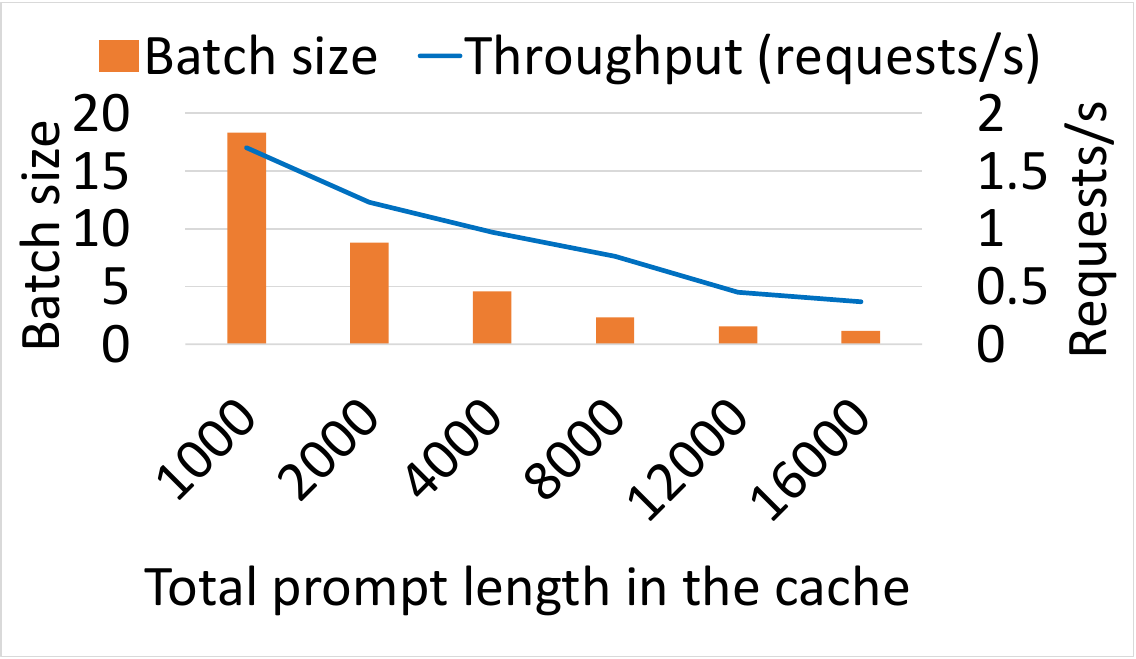} }}
    \hfill
    \subfloat[Batch size and throughput for allocated KVC\% upon failures.\vspace{-0.01in}\label{fig:fail-total-4}]{{\includegraphics[width=0.24\linewidth,height=0.15\textheight]{NewFigs/6-batch-allocated.pdf} }}
    \hfill
  %
\vspace{-0.05in}
   \caption{\small{Impact of long prompts on total KVC allocation.\vspace{-0.1in}}}%
    \label{fig:pp-impact-2} \vspace{-0.1in}
\end{figure*}}

\DEL{\begin{figure}[!t]
\centering
\DEL{\subfloat[Density of total length upon allocation failure in cache.\vspace{-0.01in}\label{fig:fail-total-1}]{{\includegraphics[width=0.24\linewidth,height=0.15\textheight]{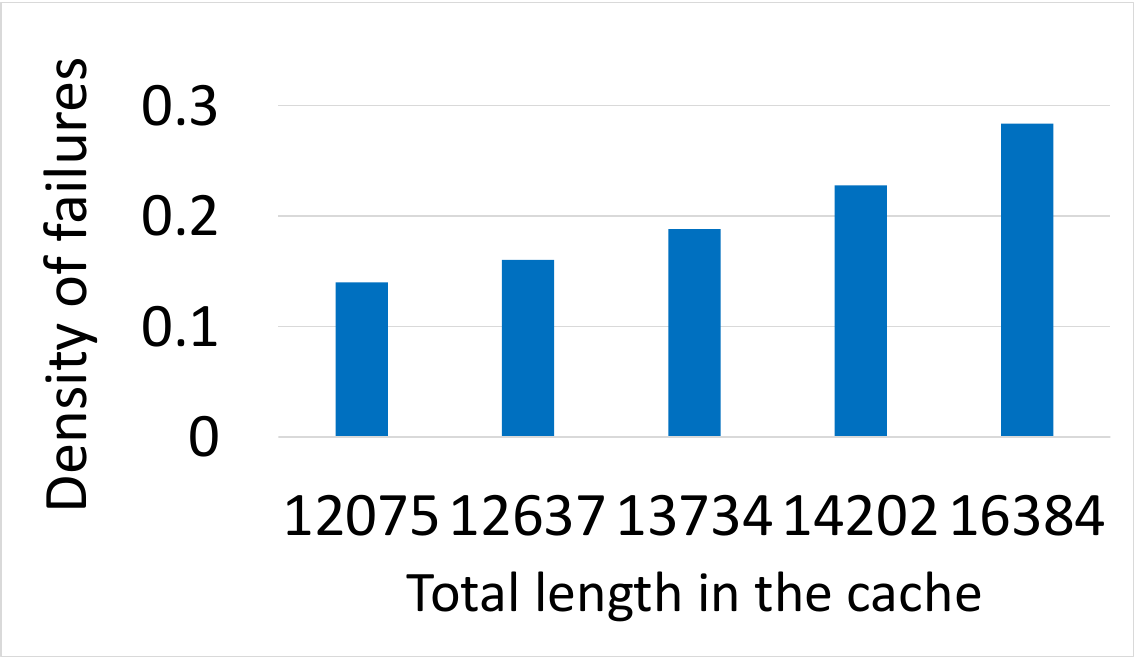} }}
    \hfill}
\subfloat[Density of allocated KVC \% upon  failures.\vspace{-0.01in}\label{fig:fail-total-2}]{{\includegraphics[width=0.48\linewidth,height=0.15\textheight]{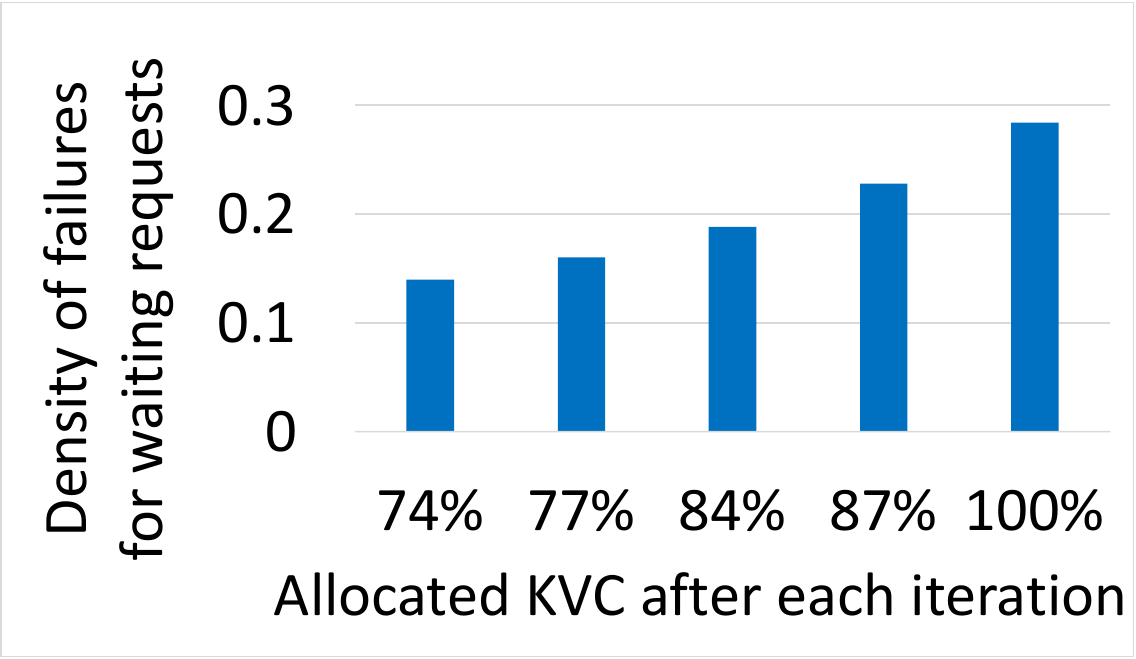} }}
    \hfill
   \DEL{ \subfloat[Batch size and throughput for total prompt length upon failures.\vspace{-0.01in}\label{fig:fail-total-3}]{{\includegraphics[width=0.24\linewidth,height=0.15\textheight]{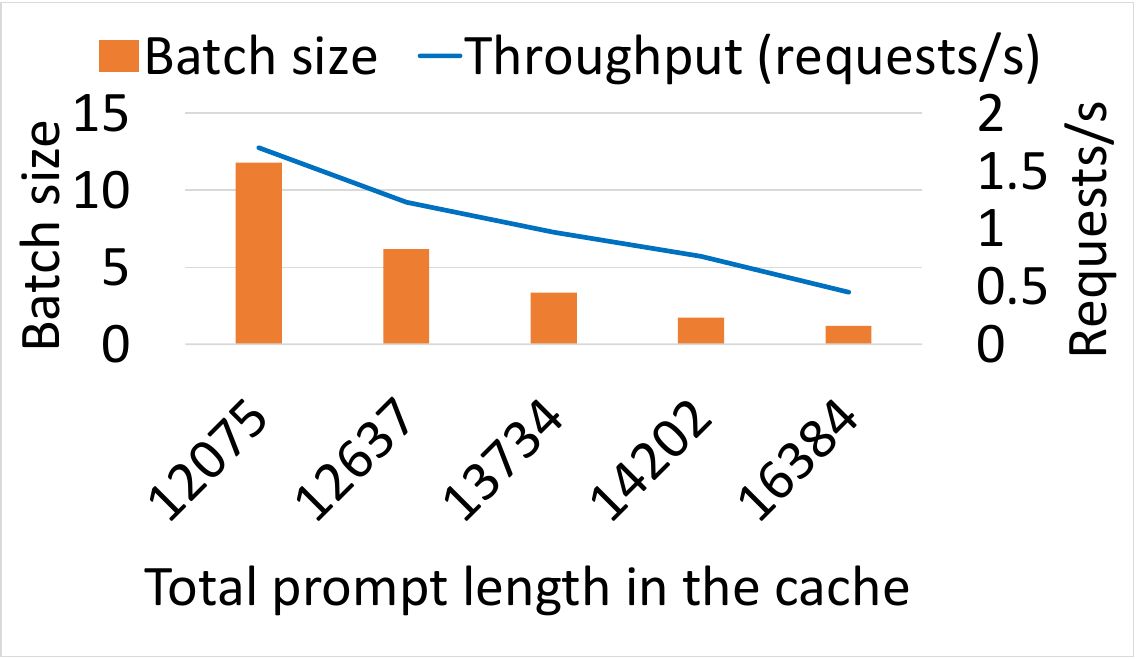} }}
    \hfill}
   \DEL{ \subfloat[Batch size and throughput for allocated KVC\% upon failures\sh{make X name format the same as the left fig}.\vspace{-0.01in}\label{fig:fail-total-4}]{{\includegraphics[width=0.48\linewidth,height=0.15\textheight]{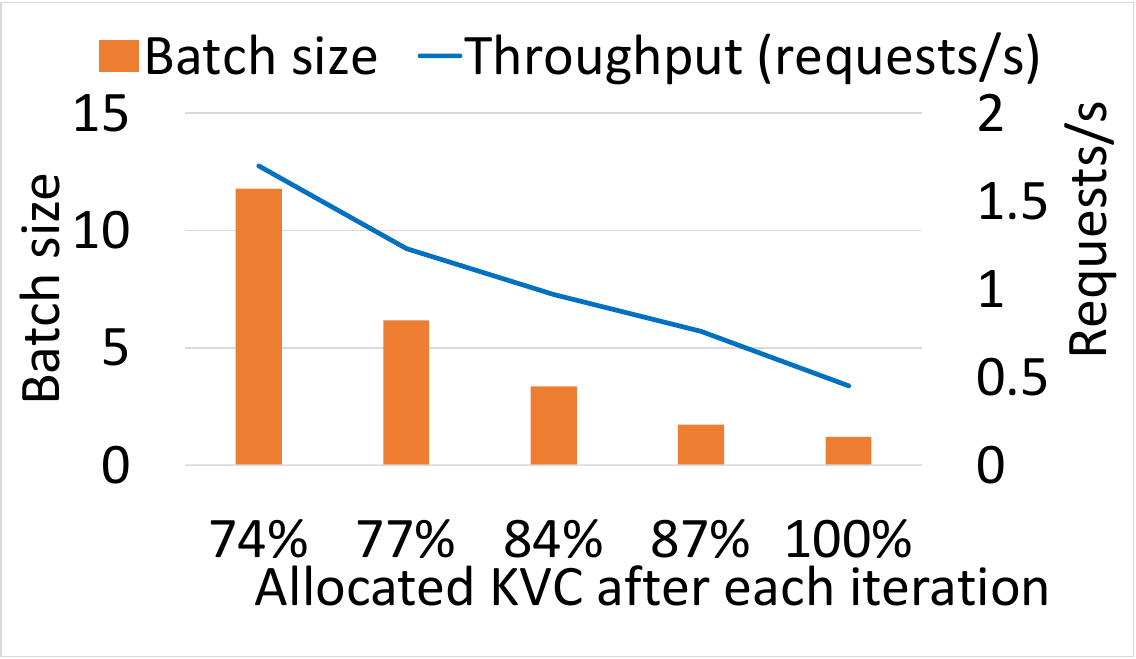} }}
    \hfill}
  %
\vspace{-0.05in}
   \caption{\small{Impact of total length on KVC allocation\sh{can you add 5\% and 95\%? }.\vspace{-0.1in}}}%
    \label{fig:pp-impact-2} \vspace{-0.1in}
\end{figure}
}


\DEL{\begin{figure}[t]\vspace{0.1in}
    \begin{minipage}[t]{0.48\linewidth}
\includegraphics[width=\linewidth,height=0.15\textheight]{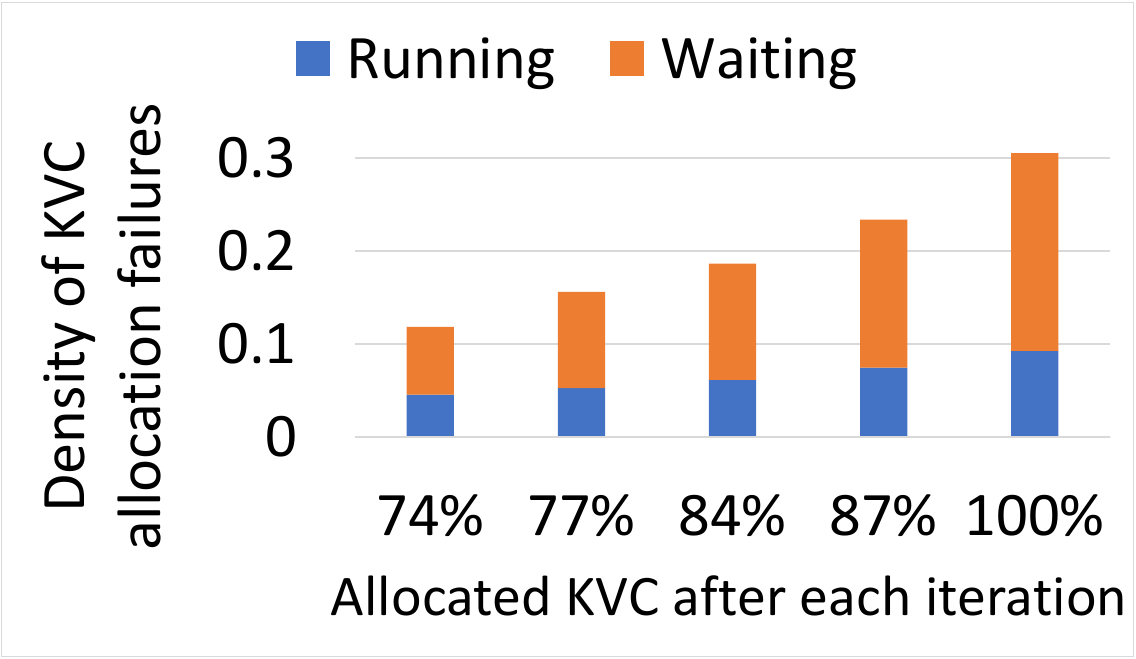}
\vspace{-0.01in} \caption{Failures vs. allocated KVC.\vspace{-0.05in} }
    \label{fig:block-density}
\end{minipage}%
   \hfill
   \begin{minipage}[t]{0.48\linewidth}
\includegraphics[width=\linewidth,height=0.15\textheight]{NewFigs/cdf-delay-com.pdf}
\vspace{-0.17in} \caption{Waiting time due to KVC allocation failures.\vspace{-0.05in} }
    \label{fig:cdf-delay}
\end{minipage}
\end{figure}

\begin{figure}[]\vspace{-0.05in}
\subfloat[Density of failures vs. prompt length upon  failures.\vspace{-0.01in}\label{fig:fail-waiting-2}]{{\includegraphics[width=0.48\linewidth,height=0.15\textheight]{NewFigs/prompt-density-waiting.pdf} }}
 \hfill
\subfloat[Unallocated KVC vs. prompt length upon  failures.\vspace{-0.01in}\label{fig:fail-waiting}]{{\includegraphics[width=0.48\linewidth,height=0.15\textheight]{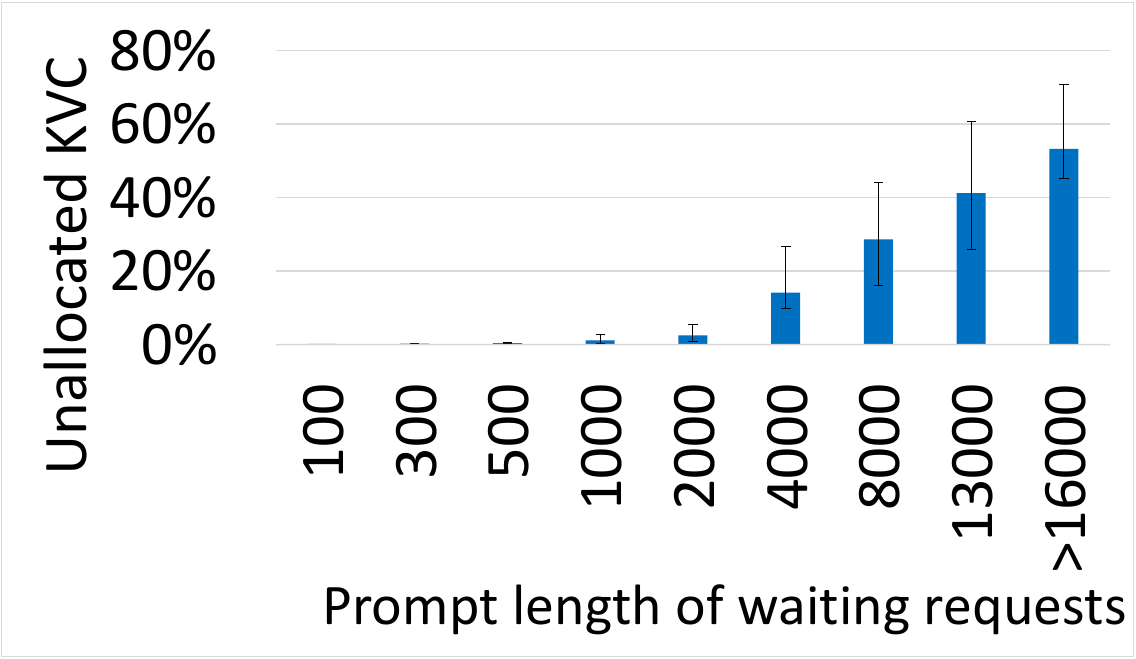} }}
\vspace{-0.05in}
\caption{KVC allocation failures for long prompts and the resultant KVC underutilization.}
 \label{fig:pp-impact} \vspace{-0.15in}
\end{figure}
}

\DEL{Figure~\ref{fig:fail-waiting} illustrates the unallocated KVC when a KVC allocation failure occurs for a prompt with a specific length. Notably, longer prompts tend to lead to higher unallocated KVC space. 
}

\DEL{Figure~\ref{fig:fail-waiting-2} shows the density of failures versus the prompt length of waiting requests that failed to be allocated. It shows that longer prompts in the waiting queue have a higher density of failures because they need larger KVC space. Figure~\ref{fig:fail-waiting} shows the unallocated KVC after a batch is formed versus the prompt length of the first waiting request that failed to be allocated. Notably, longer prompts lead to larger unallocated KVC space since the FCFS policy stops adding prompts to the batch if the first prompt in the waiting queue cannot be allocated. 
}
\DEL{As the prompt length of the first waiting request increases, the unallocated KVC \% keeps increasing because long prompts cannot be allocated on the remaining available KVC. This leads to significant KVC wastage.}

\DEL{\vspace{-0.1in}
\begin{thm}\label{kvcache2}

In vLLM, longer prompts in the waiting queue encounter more KVC allocation failures, leading to higher waiting times and KVC underutilization.

\DEL{vLLM results in more KVC allocation failures for longer prompts in the waiting queue. Longer waiting prompts lead to higher waiting times and more KVC underutilization.}

\DEL{ ??Long prompts tend to overflow the KVC, leading to KVC allocation failures and consequently reducing throughput.

vLLM frequently encounters KVC allocation failures due  to insufficient KVC space or lengthy prompts, which increases iteration times.}
 
\end{thm}
\vspace{-0.1in}
}

\DEL{\begin{figure}[]
\centering
\DEL{\subfloat[Forward length and added prompt.\vspace{-0.01in}\label{fig:iteration-time-overall}]{{\includegraphics[width=0.48\linewidth,height=0.15\textheight]{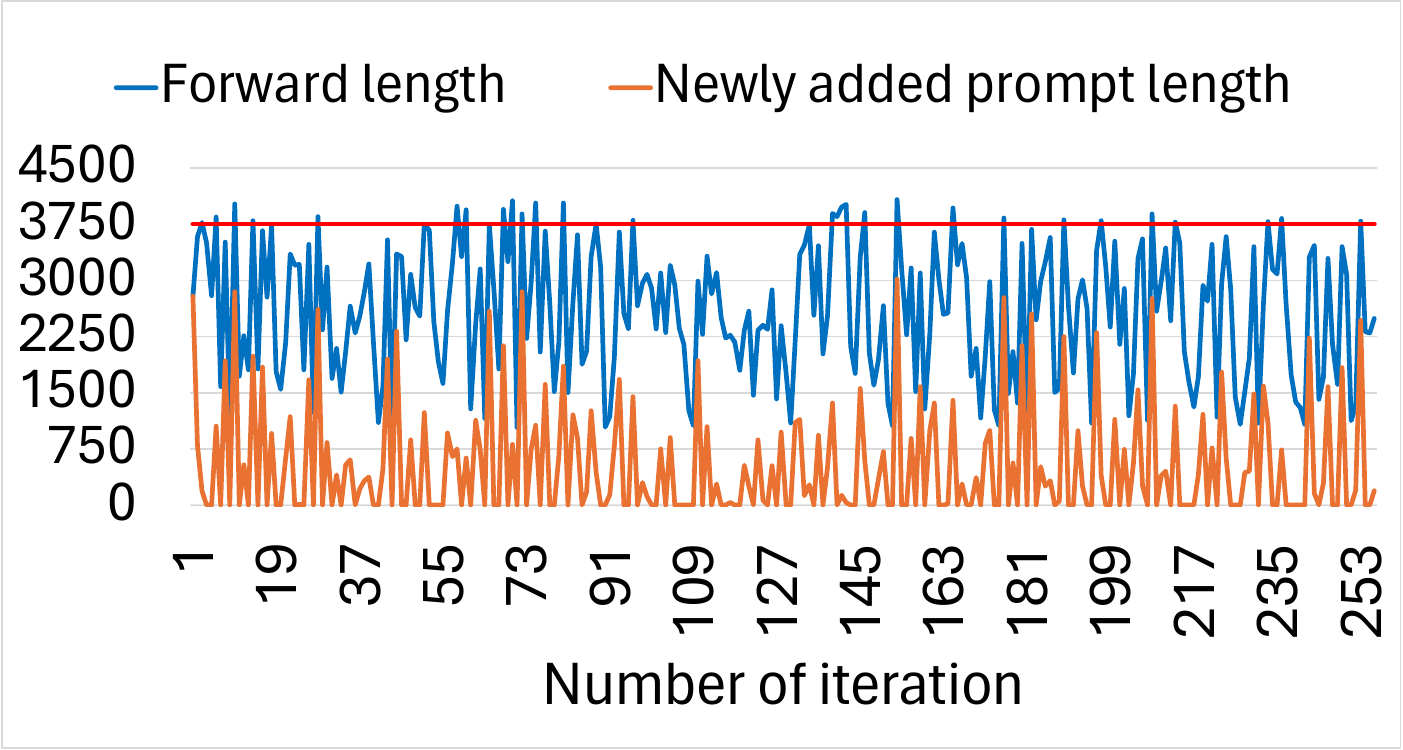} }}
    \hfill
    }
    \subfloat[GPU and KVC usage over time.\vspace{-0.01in}\label{fig:jct-overall1}]{{\includegraphics[width=0.48\linewidth,height=0.15\textheight]{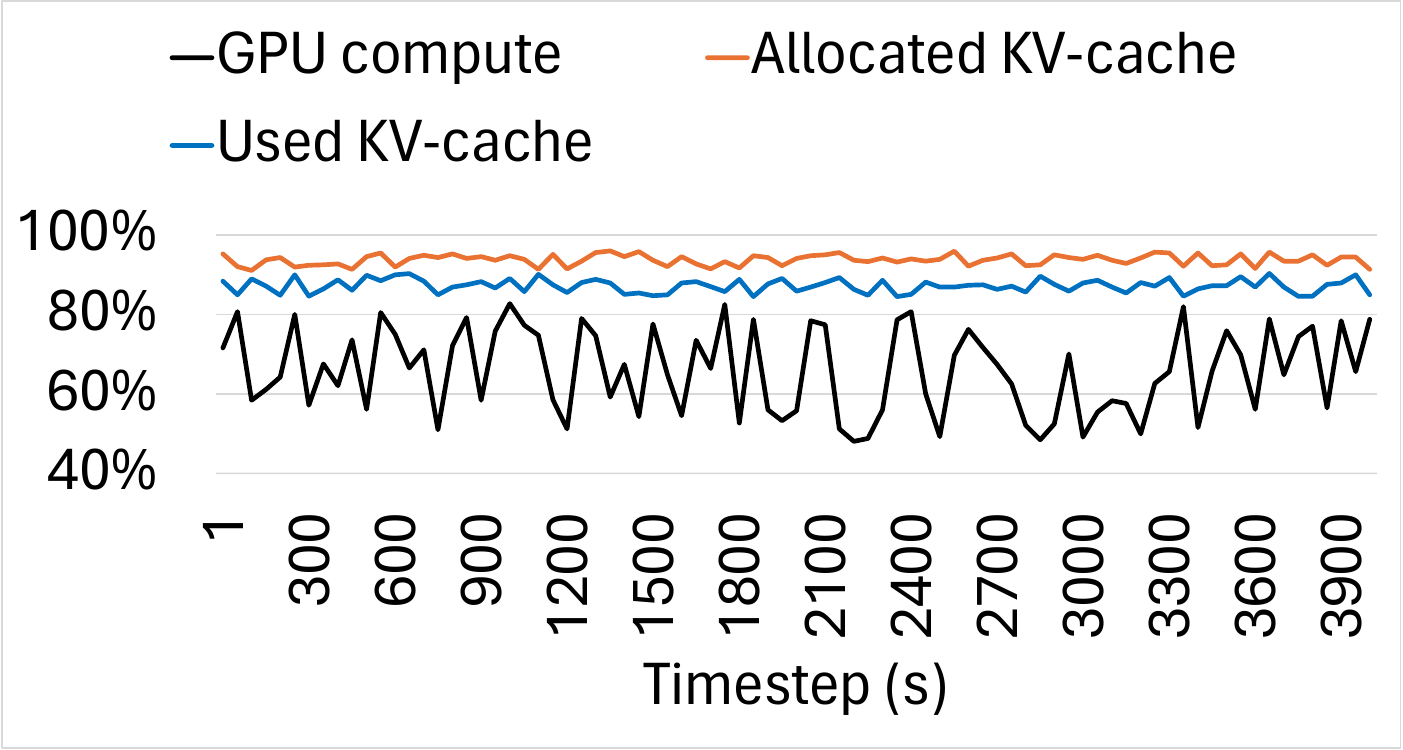} }}
    \hfill
   \DEL{  \subfloat[GPU and KVC usage over time.\vspace{-0.01in}\label{fig:fsi-2}]{{\includegraphics[width=0.48\linewidth,height=0.15\textheight]{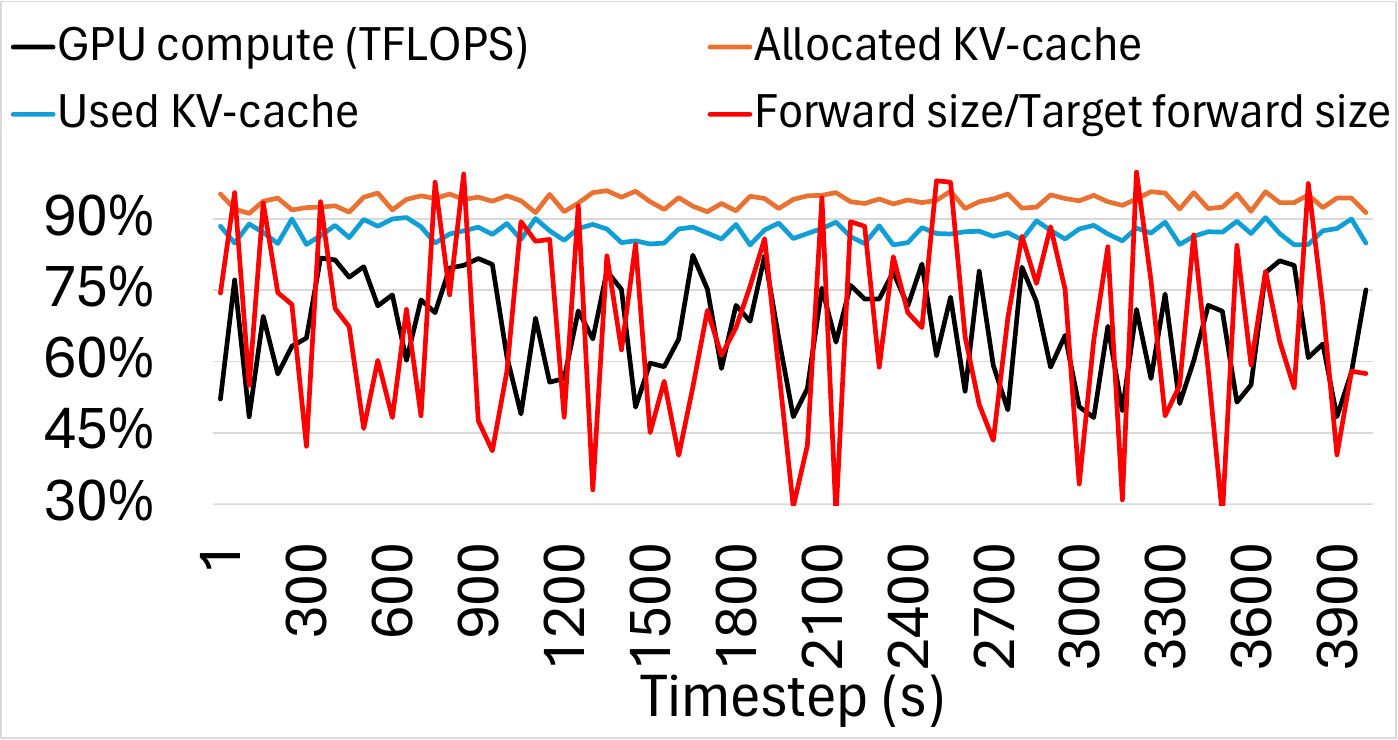} }}
    \hfill
  \subfloat[GPU and KVC usage after each iteration.\vspace{-0.01in}\label{fig:fsi}]{{\includegraphics[width=0.48\linewidth,height=0.15\textheight]{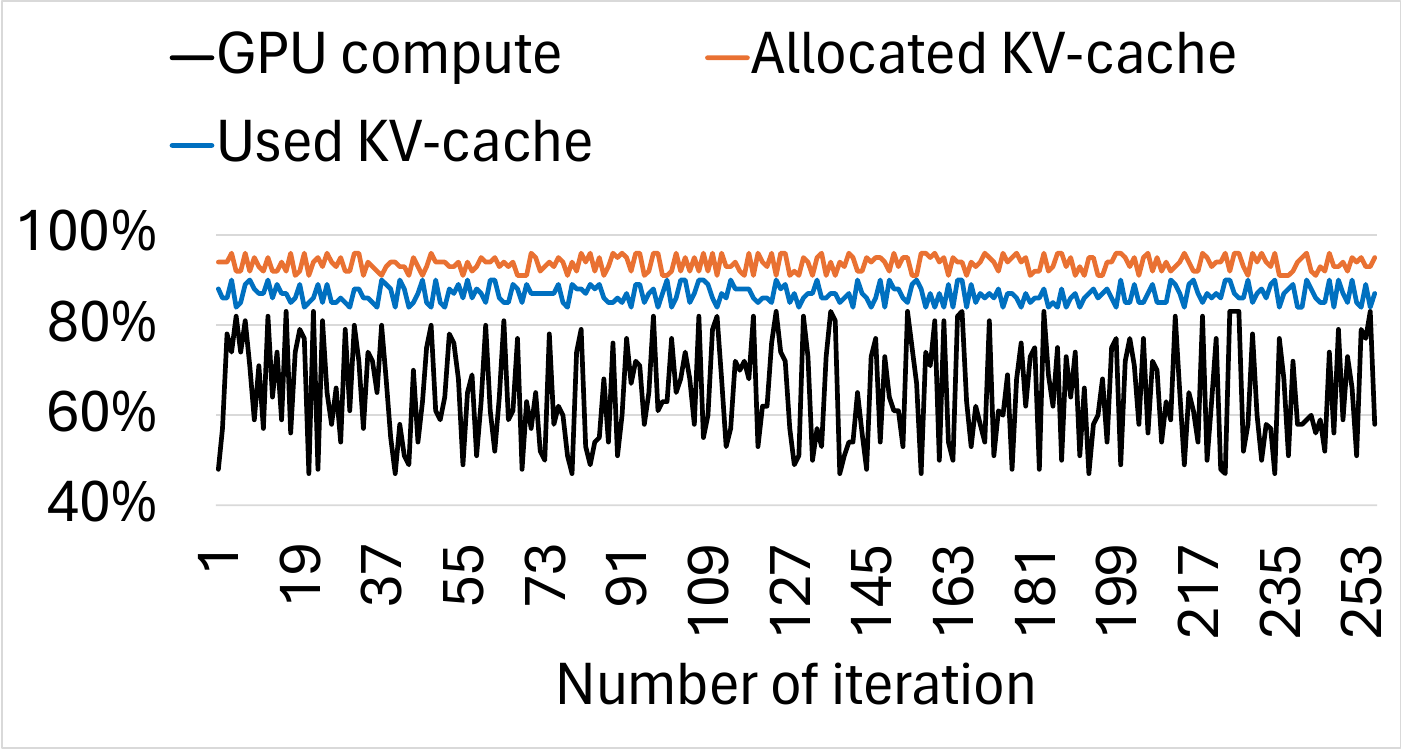} }}
    \hfill}
    \subfloat[Available resource after each iteration\sh{change Y from number of percentage in this fig. can you draw CDF for this fig? this available KVC is unallocated or unused?}.\vspace{-0.01in}\label{fig:gap}]{{\includegraphics[width=0.48\linewidth,height=0.15\textheight]{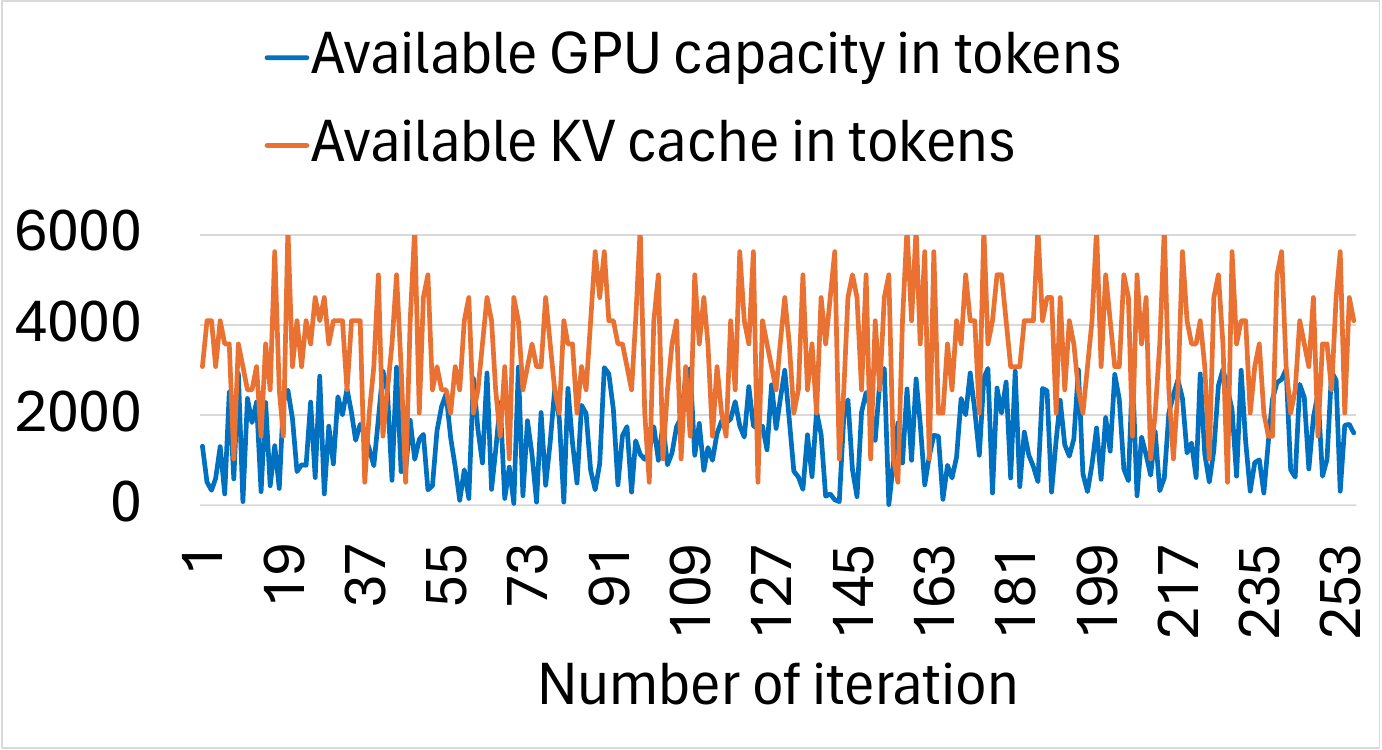} }}
   \DEL{\hfill
    \subfloat[Available resource over time .\vspace{-0.01in}\label{fig:gap2}]{{\includegraphics[width=0.24\linewidth,height=0.15\textheight]{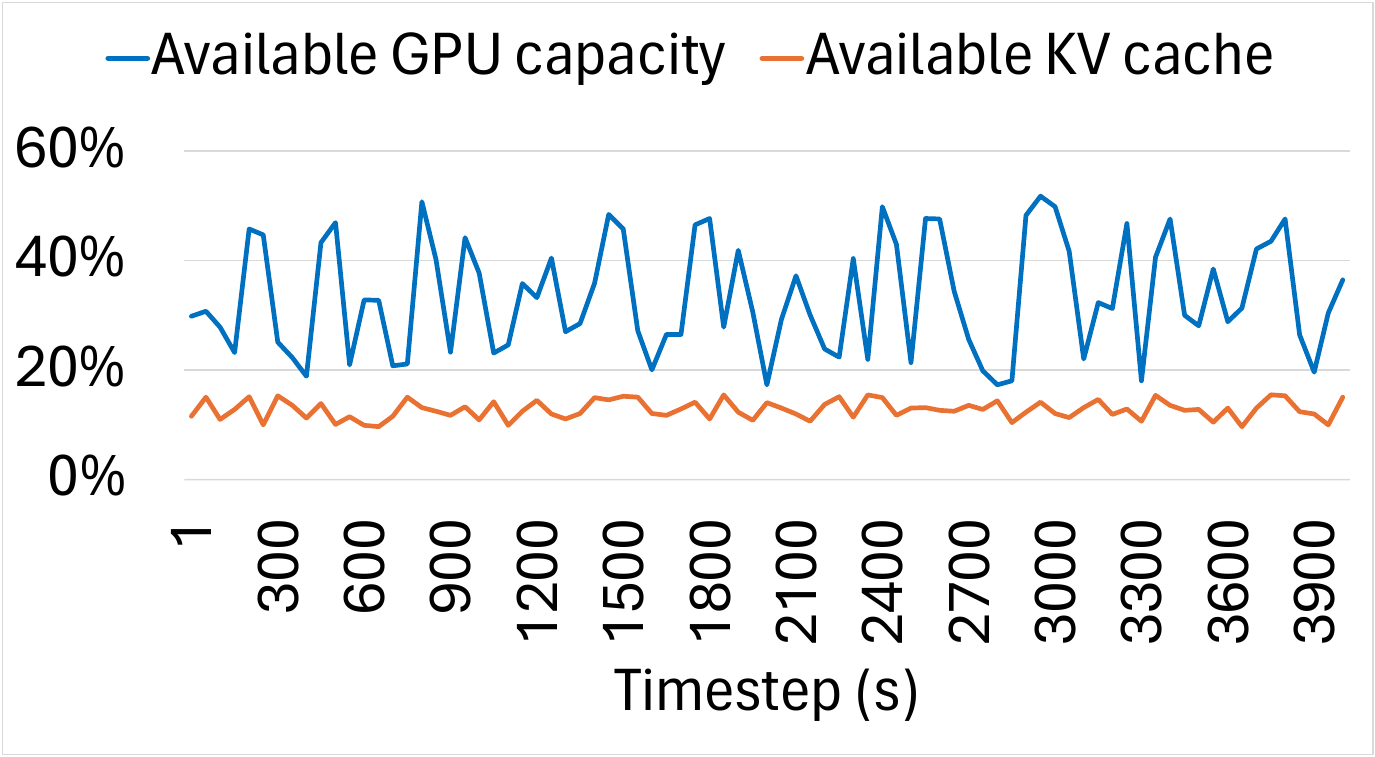} }}
    \hfill
    }
\vspace{-0.1in}
   \caption{\small{Resource utilization over time and available resources after each iteration.\vspace{-0.05in}}}%
    \label{fig:pp-impact}
\end{figure}
}

\begin{figure}[t]\vspace{-0.15in}
\subfloat[Input and output length distribution.\vspace{-0.01in}\label{fig:distribution}]{{\includegraphics[width=0.48\linewidth,height=0.145\textheight]{Fig/1-up.pdf} }}
    \hfill
\subfloat[Output length vs. prompt length.\vspace{-0.1in}\label{fig:output-length}]{{\includegraphics[width=0.48\linewidth,height=0.145\textheight]{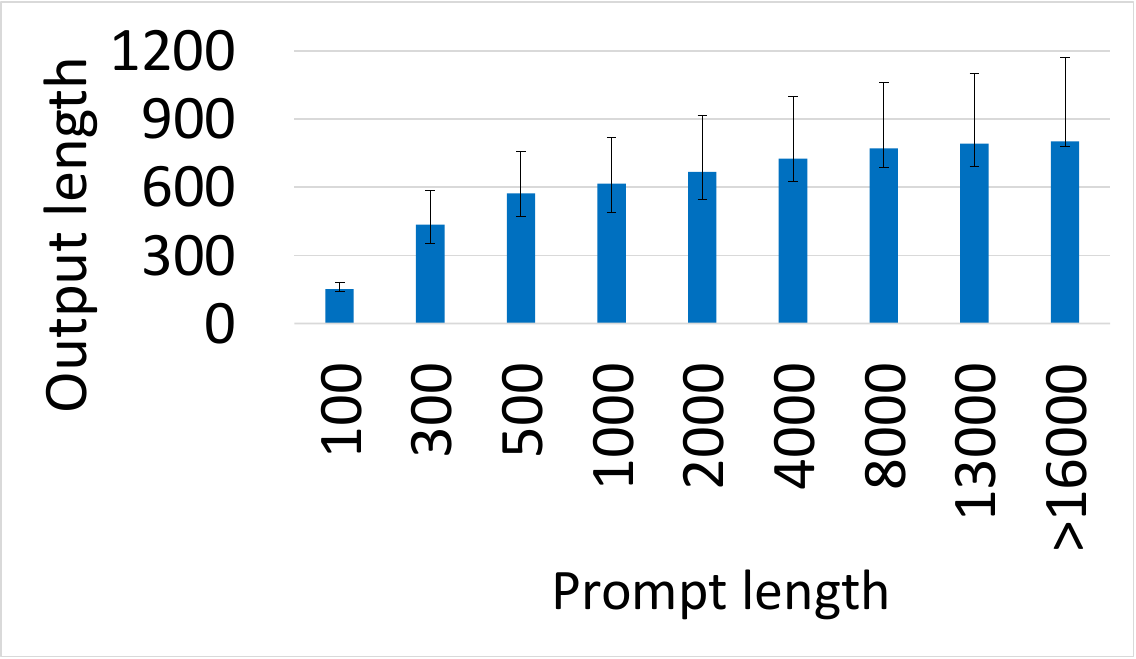} }}
\vspace{-0.05in}\caption{Features of the requests.\vspace{-0.05in} }
\label{fig:dataset-denisty}
\end{figure}

\DEL{Figure~\ref{fig:jct-overall1} shows the average GPU compute utilization, allocated KVC amount, and used KVC amount at each time interval of 5s. We see that the GPU compute resource is not fully utilized most of the time, with values ranging from [48\%, 82\%], with the average value of 75\%. The ranges of the allocated and used KVC  are [88\%, 96\%] and [84\%,90\%], respectively. The average values of them are 93\% and 86\%. \looseness=-1}



\noindent\textbf{Different Resource Demands of Requests.} Figure~\ref{fig:distribution} shows the density of the lengths of inputs and outputs, representing the probability of input or output lengths falling within specific ranges. We see that 34\% of prompts fall within [10, 512], 52\% within [512, 4K], and 14\% exceed 4K. This highlights the diversity in prompt lengths and, consequently, in GPU compute and KVC resource demands. For outputs, 51\% are in the [10, 512] range, 36\% in [512, 4K], and 13\% exceed 4K. The varying lengths of outputs contribute to different KVC demands for distinct requests. Figure~\ref{fig:output-length} shows the output length corresponding to each input length range. We see that the input and output lengths vary across different requests.

\vspace{-0.1in}
\begin{thm}\label{imbalance}
The varying availability of GPU compute and KVC resources (Figures~\ref{fig:batch-capacity},~\ref{fig:batch-capacity-175b},~\ref{fig:kvc-capacity} and~\ref{fig:kvc-utilization-175b}) after each iteration, 
coupled with the diverse GPU compute and KVC demands of different prompts (Figure~\ref{fig:dataset-denisty}), provide an opportunity to identify prompts or prompt chunks to be added to the batch to maximize both GPU compute and KVC utilizations, thus improving throughput. 
\end{thm}
\vspace{-0.1in}

\DEL{ GPU sometimes is overloaded while underloaded at other times, \DEL{depending on the lengths of the added prompts,} and KVC available size vary over time. 
 Based on O\ref{thm:promptfeature}, we should select prompts to ensure that the GPU and memory are neither overloaded nor underloaded at each iteration.
}



\DEL{The results can also be explained by the GPU compute utilization in each iteration shown Figure~\ref{fig:gpu-long-seq}. 
The first iteration generates a very high GPU compute utilization. While in the following
iterations, only the key-value tensors of the newly generated
token requires computation, resulting in low GPU compute utilizations. 
}

\DEL{\begin{figure*}[!t]
\centering
    \subfloat[Alpaca.\vspace{-0.05in}\label{fig:gpu-u}]{{\includegraphics[width=0.32\linewidth,height=0.15\textheight]{Fig/utilization-single.png} }}
    \hfill
    \subfloat[ShareGPT.\vspace{-0.05in}\label{fig:memory-a}]{{\includegraphics[width=0.32\linewidth,height=0.15\textheight]{Fig/utilization-single.png} }}
    \hfill
    \subfloat[BookCorpus.\vspace{-0.05in}\label{fig:memory-u}]{{\includegraphics[width=0.32\linewidth,height=0.15\textheight]{Fig/utilization-single.png} }}
    \hfill
\vspace{-0.05in}
   \caption{\small{(Fake) Resource utilization on a single request.\vspace{-0.1in}}}%
    \label{fig:utilization1}
\end{figure*}}

\DEL{\begin{figure}[htbp]
\begin{minipage}[t]{0.48\linewidth}
\includegraphics[width=\linewidth,height=0.15\textheight]{Fig/long-non-chunk-19.png}
\vspace{-0.01in}    \caption{Impact of a long prompt in a batch.\vspace{-0.01in} }
    \label{fig:long-batch}
\end{minipage}%
    \hfill%
\begin{minipage}[t]{0.48\linewidth}
\includegraphics[width=\linewidth,height=0.15\textheight]{Fig/long-chunk-19.png}
\vspace{-0.01in} \caption{Impact of chunking.\vspace{-0.05in} }
    \label{fig:chunk}
\end{minipage}
\end{figure}}

\DEL{\begin{figure}[t]
\DEL{\begin{minipage}[t]{0.48\linewidth}
    \includegraphics[width=\linewidth,height=0.15\textheight]{Fig/single-batch.png}
    \caption{Resource utilization on a single request. }
    \label{fig:utilization4}
\end{minipage}%
    \hfill}
\begin{minipage}[t]{0.48\linewidth}
    \includegraphics[width=\linewidth,height=0.15\textheight]{Fig/prompt.png}
    \caption{Resource utilization on a batch (3 prompt 1 token tasks).\vspace{-0.1in} }
    \label{fig:utilization-prompt}
\end{minipage}
\DEL{\begin{minipage}[t]{0.48\linewidth}
    \includegraphics[width=\linewidth,height=0.15\textheight]{Fig/prompt-single.png}
    \caption{Resource utilization on a batch size of 4 (3 token 1 prompt. }
    \label{fig:utilization-token}
\end{minipage}
}
\begin{minipage}[t]{0.48\linewidth}
    \includegraphics[width=\linewidth,height=0.15\textheight]{Fig/mixed.png}
    \caption{Resource utilization for mixed tasks.\vspace{-0.1in} }
    \label{fig:utilization}
    \end{minipage}
\end{figure}}

\DEL{Therefore, the available GPU compute resource and memory resource vary over time. Different prompts have different prompt lengths and hence different demands on GPU compute resources and memory resources {\DEL{show the density of the prompt length fig at the first}}. The resources are sometimes over-utilized?? and sometimes under-utilized. Some long prompts overload both resources??. The results in the figures show the heterogeneous demands from the prompts, so we can use prompts to enable reach the token budget. A long prompt can help limit the batch size (or the KVC demand) while reaching the token budget. This is because though adding $m$ shorter prompts can reach the same length as the one prompt, these prompts generate $m$ sequences in KVC that generate tokens. }

\DEL{To solve the problem in \cref{sec:introduction}, two problems arise: 1) how to avoid the adverse effects of long prompts (O\ref{longprompt} and O\ref{kvcache2}), and 2) how to fully utilize two resources while avoiding overloading the two resources concurrently (O\ref{pivot} and O\ref{imbalance}). We propose novel methods to simultaneously handle the two problems to change long prompts from foe to friend. 
we dynamically chunk a long prompt to help a batch reach the $S_{f}=S_b$ to fully utilize GPU while reducing iteration time, its own KVC allocation failure and the KVC overflow. 
Regarding avoiding overflowing
the KVC, using a long prompt chunk is more effective than using multiple prompts to reach $S_{f}=S_b$ since the latter approach will generate more request sequences or cache demands in KVC. {\color{gray}{For example, the current batch has $S_f=1500$ so we need to add prompts with length equalling 3750-1500=2250. We could add 5 prompts with 512 prompt length each, or add 1 prompt with 2250 prompt length. Initially, both cases need 5 KVC blocks. But after 512 TG steps, the former case needs 5 blocks KVC in total, while the latter only needs 1 block KVC.}} 

Based on O\ref{imbalance}, to fully utilize the KVC and avoid overflowing KVC, and avoid allocation failures, we propose the iteration-level cache guarantee method that guarantees that the prompts that can be allocated to KVC are selected in an iteration.

Instead of using JCT SLO, we propose the iteration-level SLO method to help enhance user experience especially in the presence of long prompts.

Considering various lengths or GPU and memory demands of prompts (Figure~\ref{fig:dataset-denisty}), we should select
prompts to ensure that the GPU and memory are
neither overloaded nor underloaded at each iteration. Thus, we propose multi-Resource-aware batch formation.
}


\DEL{Long prompts make the existing long iteration time and KVC bottleneck problems even more severe ((O\ref{longprompt} and O\ref{kvcache2})). To address the problem outlined in \cref{sec:introduction}, we propose novel methods to handle long prompts and leverage their potential benefits. The key challenges are avoiding the adverse effects of long prompts (O\ref{longprompt} and O\ref{kvcache2}) and utilizing long prompts to fully exploit GPU and memory resources without overloading them (O\ref{pivot} and O\ref{imbalance}). Our proposed solutions are as follows. ?}

\DEL{\section{Design Foundation and Rationales}\label{sec:motivation}

In the mixed-prompt scenario, the imbalance in prompt lengths leads to GPU underutilization and overutilization (O\ref{pivot}). Additionally, long prompts increase iteration time ((O\ref{longprompt}) and may fail to be allocated to KVC, leading to KVC underutilization (with unallocated space) (O\ref{kvcache2}). O\ref{imbalance} provides guidance on  addressing these issues.}

\DEL{\begin{figure}[h]
    \centering
\includegraphics[width=1\columnwidth,height=0.15\textheight]{Fig/chunk-motivation-up-2.png}
    \caption{Long prompts become foe by overflow KVC.}
    \label{fig:chunk-4}
\end{figure}
}

\DEL{Let's consider another example to illustrate how long prompts can potentially overflow the KVC. In Figure~\ref{fig:chunk-motivation-2-1}, a sequence of 8192-token prompts with generation steps of 256 needs to be processed. Without chunking, in each iteration, an 8192-token prompt is added to the batch. Then, when the first prompt is at the last iteration, 256 requests are being processed concurrently, requiring KVC for the prompts at that moment to be $\text{8k} \times 256 = 2M$ tokens. With prompt chunking, in each iteration, a 512-token prompt is added to the batch. Then, when the first prompt is at the last iteration, 256/16 requests are being processed concurrently.
Therefore, the KV cache requirement at this time for the prompts is
$ \text{8k} \times (256/ 16)  = 128K$  tokens. Thus, chunking can transform long prompts into friends by not only improving time-efficiency and GPU compute utilization, but also avoiding KVC overflow.}

\section{System Design of \Sys}


\DEL{In the mixed-prompt scenario, the imbalance in prompt lengths leads to GPU underutilization and overutilization (O\ref{pivot}). Additionally, long prompts increase iteration time ((O\ref{longprompt}) and may fail to be allocated to KVC, leading to KVC underutilization (with unallocated space) (O\ref{kvcache2}). O\ref{imbalance} provides guidance on  addressing these issues.}

\DEL{When the prompt lengths of requests are similar and not long, they do not differ greatly in GPU and KVC resource consumption, as per Equations~\eqref{eq:totop} and \eqref{equ:cache}. Then, configuring the batch size to fully utilize KVC and using FCFS to form a batch~\cite{vllm,280922} are effective. However, in the mixed-prompt scenario, where the prompt lengths vary greatly and some are long, these methods become less effective. 
Using FCFS, the imbalance in prompt lengths leads to GPU underutilization and overutilization (O\ref{pivot}). Additionally, long prompts increase iteration time ((O\ref{longprompt}) and may fail to be allocated to KVC, leading to KVC underutilization (with unallocated space) (O\ref{kvcache2}).
}

\DEL{O\ref{overallPerformance} shows that the previous method for handling long prompts falls short in maximizing throughput, motivating the approach in this work. O\ref{DynamicChunking} recommends dynamically adjusting the chunk size for long prompts rather than using a fixed size. O\ref{EvenChunking} emphasizes the need to fully utilize both GPU and KVC resources concurrently. O\ref{kvcache2} suggests completing one long-prompt request before starting the next. O\ref{pivot} indicates the importance of meeting diverse SLOs rather than applying a uniform SLO across the system. O\ref{imbalance} provides guidance for addressing these challenges.
}

\subsection{Overview}

\DEL{\textbf{Leveraging observations for designing \Sys.} LLM inference systems already suffer from memory bound, resultant low GPU compute utilization (hence lower throughput).
Long prompts will make these problems even more severe ((O\ref{longprompt} and O\ref{kvcache2})). Chunking not only can mitigate the adverse effect from long prompts but also can help increase GPU compute utilization and throughput (O\ref{pivot}) while avoiding overflowing the KVC. In each iteration, based on O\ref{imbalance}, we need to select prompts or create prompt chunks to ensure that the GPU and memory are neithbor overloaded nor underloaded. 
}
Based on the Observations (Os) from \cref{sec:analysis}, we propose \Sys. \Sys consists of the following three methods:\looseness=-1 
\squishlist

\item[(1)] \textbf{SLO-guaranteed dynamic chunking (\cref{sec:ragged-processor})}.
Based on O\ref{DynamicChunking} and O\ref{kvcache2}, it determines the target forward size (i.e., token budget) and batches requests with
similar SLOs to maximize GPU compute utilization as much as possible while satisfying the iteration-level 
 SLOs of the requests in the batch. In addition, it limits concurrently running long-prompt requests to improve throughput. 



\item[(2)] \textbf{Iteration-level SLO-based task prioritization  (\cref{sec:SLO})}. Based on O\ref{pivot}, it orders requests by their iteration-level SLOs and groups those with similar SLOs for scheduling in each iteration.


\DEL{\item[(2)] \textbf{SLO-guaranteed temporal GPU load balancing (\cref{sec:ragged-processor})}.
Based on O\ref{pivot}, it dynamically distribute the processing of a long prompt across multiple iterations to avoid increasing iteration time and reach the token budget ($S_{f}=S_b$) to fully utilize GPU while satisfying SLOs of the requests in the batch.}


\DEL{\item[(3)] \textbf{Iteration-level KV cache full allocation (\cref{sec:Cache})}.
To fully utilize the KVC, this method selectively chooses requests to guarantee that the KVC demands of a batch fully fill the KVC in each iteration.}


\item[(3)] \textbf{Multi-resource-aware batching (\cref{sec:Cache})}.
Based on O\ref{EvenChunking}, it concurrently considers both GPU compute and KVC resources in selecting prompts or creating prompt chunks based on their demands and available resources to maximize
both GPU compute utilization and KVC utilization in the
subsequent iteration. \looseness=-1

\squishend

\DEL{\noindent\textbf{Architecture of \Sys.} After each request arrives (either from a user or from the execution engine after an iteration), it is put into the waiting queue. Its order in the queue depends on its priority based on (\emph{task prioritization} \circled{1} \cref{sec:SLO}) (O\ref{thm10}). 
}

Figure~\ref{fig:sys-architecture} shows the overview of \Sys.
In the figure, each block in the waiting queue is a token, and the tokens of the same color constitute a request. Method (2) is executed to order the requests based on their SLOs for each newly arrived request or when a portion of a long prompt is added to the batch (\circled{1}). After an iteration finishes, the generated tokens of uncompleted requests (i.e., the rightmost three blocks) are returned to the scheduler (\circled{2}). The scheduler then executes Methods (1) and (3) to select requests to maximize the GPU compute and KVC resources (\circled{3}). Method (1) selects tokens sequentially from the waiting queue to fully utilize the GPU compute resource while meeting the requests' SLOs (\circled{3.1}). Method (3) further simultaneously considers the KVC resources when selecting requests or creating prompt chunks, enabling the new batch to fully utilize both the GPU compute and KVC resources (\circled{3.2}). 
In this example, green and blue tokens, and 502 tokens from the 6k long prompt shown in red 
are selected. Finally, the newly formed batch is sent to the execution engine (\circled{4}).


\DEL{picks up requests from the waiting queue to reach the token budget (instead of batch size) and also automatically chunks long prompts . 
At the same time, it makes sure that the KVC demands of requests in a batch can be satisfied (iteration-level KVC guarantee \circled{4}) (O\ref{cache}). Finally, 
}

\begin{figure}[t]
\centering
\includegraphics[width=1\linewidth,height=0.17\textheight]{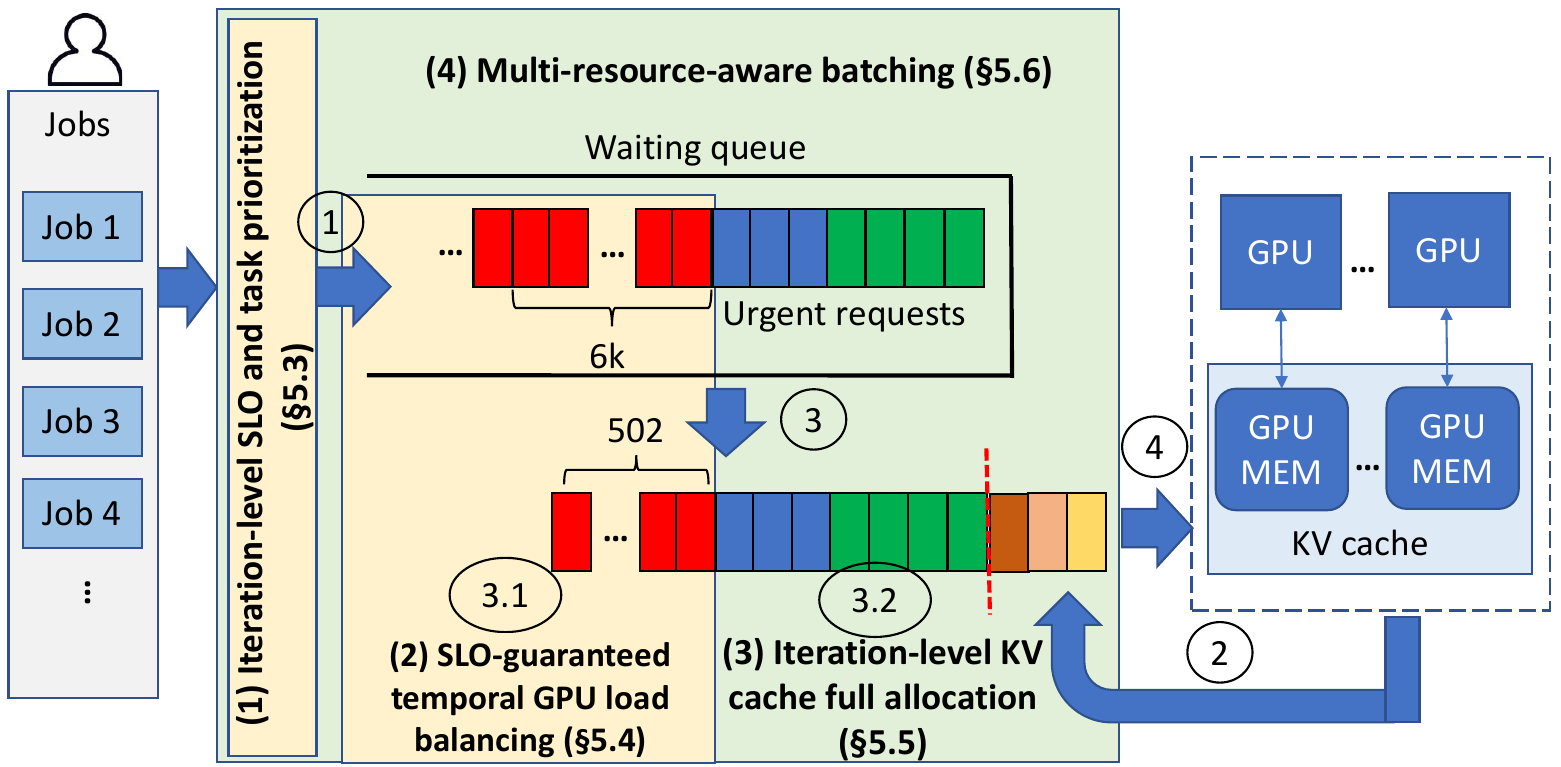}
   \vspace{-0.15in} \caption{System architecture of \Sys.}
    \label{fig:sys-architecture}\vspace{-0.00in}
\end{figure}

\DEL{\subsection{Problem Formulation}
(?revisit)After each iteration, \Sys
chooses the requests to be processed in a batch for the next iteration. Let there be a set of $K$ inference requests in the waiting queue and in the batch for the incoming iteration $j$: 
$\mathbf{R^j}=\{r_1^j,r_2^j,\cdots r_k^j,\cdots, r_K^j\}$.
Each request has its own iteration-level SLO denoted by $\mathbf{L^j}=\{L_1^j,L_2^j,\cdots,L_k^j, \cdots, L_K^j\}$. 
Now, the scheduler needs to form a new batch 
represented by $\mathbf{B}^j=\{b_1^j,b_2^j,\cdots, b_k^j,\cdots b_K^j\}$, 
where $b_k^j\in\{0,1\}$ is a binary variable indicating whether the $k^{th}$ request is included in the batch. 
The $b_k^j$ values for the requests in the returned batch from the last iteration are already equal to 1.
The goal of the optimization is to determine $\mathbf{B}^j$ to maximize the goodput, defined as the number of requests that satisfy their iteration-level SLOs during a certain time period $T$. 
Note that if a request only has partial tokens processed in an iteration, its iteration-level SLO is not considered satisfied.

For one execution of iteration $j$,   $\mathbf{c^{j}}=\{c_1^{j}, c_2^{j},\cdots, c_k^{j},\cdots, c_K^{j}\}$, where $c_k^{j}\in\{0,1\}$ denotes whether each request's iteration-level SLO is satisfied. 
Assume there are $J$ iterations during $T$, then our goal is formulated as follows: \looseness=-1\vspace{-0.1in}
\begin{equation}
          argmax \sum_{j=1}^{J} \sum_{k=1}^{K} b_k^j \cdot c_k^j
    \label{eq:optimization} \vspace{-0.1in}
\end{equation}
subject to constraints:\vspace{-0.1in}
 \begin{equation}
 \sum_{k=1}^K b_k^j \cdot M(r_k^j) \leq M_a^j, ~~j=1, 2, ..., J\label{eq:con-1} \vspace{-0.1in}
 \end{equation}
 where $M(r_k^j)$ represents the current KVC requirement of request $r_k^j$ and $M_a^j$ represents the overall available (i.e., unallocated) KVC space for the $j^{th}$ iteration. The current KVC requirement of a waiting prompt is the number of blocks needed to host its tokens, while that of a running request or a preempted request is one block if it has used up its allocated block and is zero otherwise. Equation~\eqref{eq:con-1} ensures that the requests in the batch at each iteration do not overflow the KVC.
 }


\DEL{Though Sarathi-serve can handle long prompts, in \cref{sec:analysis}, we observed that its static chunking 
fails to maximize the throughput (O\ref{DynamicChunking}) and its single, system-wide SLO setting cannot accommodate heterogeneous SLOs. Thus, we propose the \emph{SLO-guaranteed dynamic chunking} }

\subsection{SLO-guaranteed Dynamic Chunking}\label{sec:ragged-processor}

Based on O\ref{DynamicChunking}, we propose this method to maximize GPU compute utilization as much as possible while satisfying the heterogeneous SLOs of different requests. 
To achieve this objective, the method incorporates \emph{token budget determination}, \emph{limiting the number of concurrently running long-prompt requests} (O\ref{kvcache2}), and \emph{prompt selection with dynamic chunking}, as detailed below.


\noindent\textbf{Token budget determination.} 
For simplicity, unless otherwise specified, SLOs refer to iteration-level SLOs. The iteration time depends on the forward size. Therefore, we need to determine the token budget ($S_b$) for a batch to ensure high GPU compute utilization while satisfying the SLOs of the requests in the batch. 
Based on the GPU capacity of a server, we first find the pivot forward size, which saturates the GPU capacity. 
If the total number of {operations for generating one token is $x$ FLOPS (based on Equation~\eqref{eq:totop}) and the maximum allowable FLOP/s  by a GPU hardware is $X$, then the maximum
number of processed tokens at a time equals $X/x$ per second. 
GPU saturation throughput is usually lower than $X$~\cite{fastgen}, so we measure the pivot forward size ($S_{pf}$) by varying the forward size as described in \cref{sec:analysis}, and measure the corresponding batch execution time, denoted by $\bar{T}_{pf}$. We can also calculate it by dividing the number of operations for $S_{pf}$ by the GPU saturated throughput. In practice, the $S_{pf}$ is model-specific and based on the allowable number of FLOP/s on the GPU.
We can calculate $S_{pf}$ for different GPUs belonging to the same GPU family using the non-linear regression method~\cite{adainf}.
The model takes the $x$
and $X$ of the new GPU as inputs.

However, the inference time of a batch needs to satisfy the minimum SLO of the requests in a batch ($SLO_{min}$). Therefore, we determine the token budget by:\vspace{-0.1in}
\begin{equation}
\label{Eq:adjust}
S_b=S_{pf}\cdot \frac{SLO_{min}}{\bar{T}_{pf}}.\vspace{-0.05in}\end{equation} If the scheduler uses FIFO, adding queued requests to the batch may change $SLO_{min}$, requiring an adjustment to $S_b$ and the currently formed batch. Additionally, some requests in the batch may have SLOs significantly higher than $SLO_{min}$, resulting in their completion earlier than necessary and lower GPU compute utilization as loose-SLO requests can have a large batch. 
This is indicated in O\ref{pivot}.
\Sys's \emph{iteration-level SLO-based task prioritization} method (\cref{sec:SLO}) helps avoid the problem. It orders requests from the most stringent SLOs to the least stringent SLOs. As a result, the requests with similar SLOs will be together in the waiting queue. 
Since the first queued request has the most stringent SLO, we do not need to adjust $S_b$ as more requests are added to the batch. 
\DEL{This approach not only avoids frequent adjustment on the $S_b$ and the formed batch, but also increases GPU compute utilization 
compared to the case that the requests in a batch have largely deviated SLOs. 
}

\DEL{Subsequently, we  update $SLO_{min}$ to the minimum SLO in the current batch and then update $S_b$ accordingly to better align with each created batch. We stop until we reach a point where the forward size equals to $S_b$ calculated based on the $SLO_{min}$ in the current batch.} 


\DEL{In this method, requests are ordered in the waiting queue based on their iteration-level SLOs (\cref{{sec:SLO}}).}

\noindent \textbf{Limiting running long-prompt requests.} When there is token budget, Sarathi-Serve picks up one chunk for the existing requests in the batch, and then picks up the chunks from a new request as much as possible. However, this may lead to the case that multiple long-prompt requests occupy KVC simultaneously, leaving smaller KVC for processing requests and increasing JCT (O\ref{kvcache2}). To address this issue, \Sys uses ERA, ensuring that processing of chunks from a new long prompt does not begin until all currently served long-prompt requests are completed. \DEL{Figure~\ref{fig:NoConcurrency} shows a simple example for Sarathi-Serve and for \Sys, in which the sequence of the squares means the operation order and the square size does not represent KVC size, and the grey color means occupied KV cache. In Figure~\ref{fig:prompt-to-token-1}, in iteration 1, prompt 1's chunk 1 is picked into the batch. In iteration 2, another chunk of existing requests in the batch, which is prompt 1, is picked. Next, a chunk of a new request, prompt 2, is picked up. In iteration 3, another chunk of existing requests in the batch, which are prompts 1 and 2, are picked. In iteration 4, both requests generate the first generation token, and in iteration 5, both requests generate the second generation token. Thus, both requests finish at the 5th iteration and release their KV space. 
By finishing one request before picking up another request's chunks, \Sys avoids the case the multiple requests occupy the KV cache and decreases the average JCT. In Figure~\ref{fig:prompt-to-token-2}, in iteration 2, intead of picking up prompt 2's chunk, \Sys picks up prompt 1's chunk. Then, in iteration 3, since the KVC has available space for two chunks, the two chunks of prompt 2 are picked. In iteration 4, prompt 1 completes and releases it KVC, enabling to process many chunks of other prompts in iteration 5.} One might argue that this chunk selection strategy could limit the processing opportunities for waiting requests. However, in \Sys, this is not an issue, as the system aims to meet the heterogeneous SLOs of all requests and selects requests that must run in the next iteration to satisfy their SLOs.


\DEL{As mentioned in~\cref{subsec:background}, Logit values are the raw outputs of the FC2, prior to the TG. 
When each chunk generates the new token, that token is already existent in the next chunk. 
}

\noindent\textbf{Prompt selection with dynamic chunking.} After each iteration, 
\Sys selects requests in the ordered waiting queue in sequence, aiming to reach $S_b$. 
The static chunking cannot achieve the goal of exactly reaching $S_b$ since the number of uncompleted tasks in the batch ($S_B$) varies after an iteration, and hence cannot maximize the throughput as indicated in O\ref{DynamicChunking}. To address this problem,
we propose a dynamic chunking method. In this approach, 
when a batch completes an iteration, the \Sys scheduler calculates the new $S_b$ based on Equation~\eqref{Eq:adjust}. Subsequently, it sequentially selects tokens from the waiting queue until $S_{f}=S_b$, regardless of their
associated requests. 
These selected tokens can be from one request or from different requests. If the last picked token is in the middle of a prompt, the prompt is chunked after that token. In the next iteration, the selection starts from the next token in this prompt. As a result, after all the tokens of the prompt are picked, this prompt is automatically chunked into partitions to be processed in different iterations. 

\DEL{\begin{figure}[t]
\centering
    \subfloat[Sarathi-Serve: pick up chunks from different requests. P1 completse at iteration 5.\vspace{-0.05in}\label{fig:prompt-to-token-1}]{{\includegraphics[width=0.41\linewidth,height=0.14\textheight]{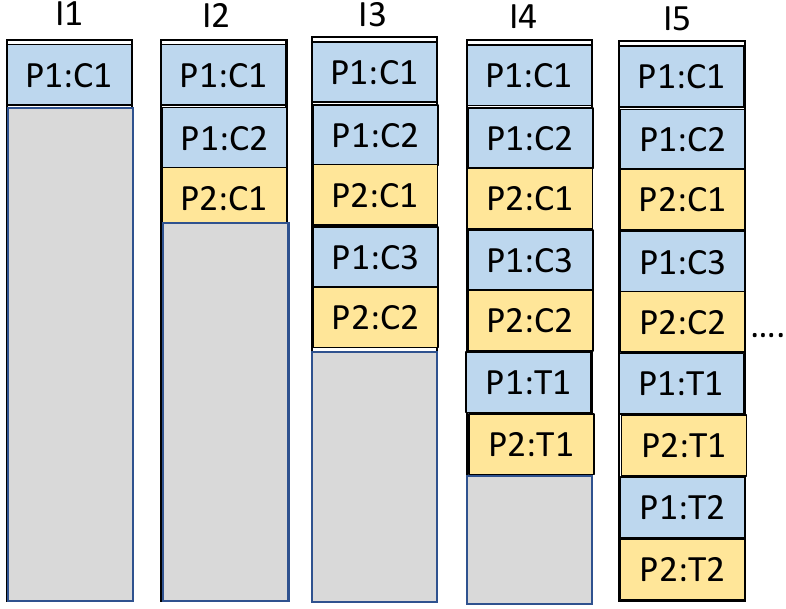} }}%
    \hfill
\subfloat[\Sys: pick up chunks from one prompt. P1 completes at iteration 4 and other prompts start earlier.\vspace{-0.05in}\label{fig:prompt-to-token-2}]{{\includegraphics[width=0.41\linewidth,height=0.14\textheight]{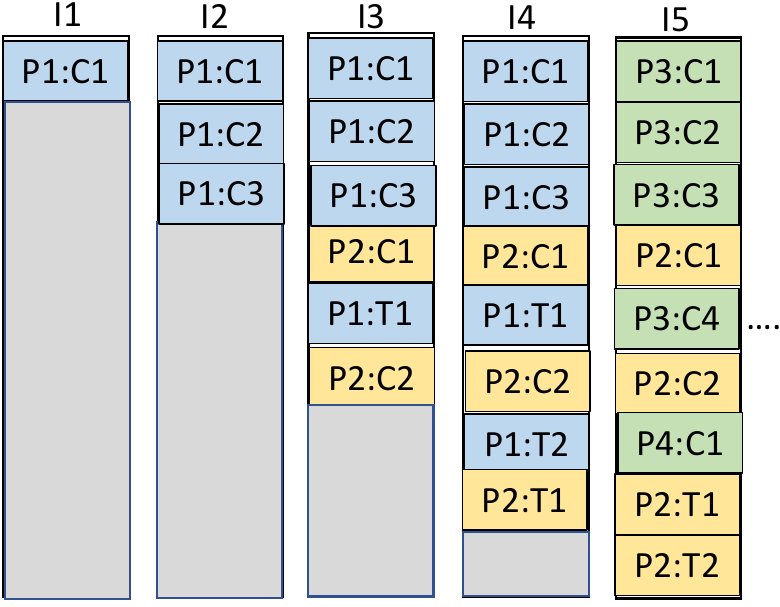} }}%
    \hfill
\vspace{-0.00in}
   \caption{Effect of chunk pick-up strategies. I: iteration, P: prompt, C: chunk, T: token. P1 has 3 chunks, P2 has 2 chunks and both have 2 generation tokens. Square size does not represent KVC size. Grey color means occupied KV cache.} 
    \label{fig:NoConcurrency}
\end{figure}
}

\DEL{--------------
When a batch returns in an iteration, if $m$ tokens are needed to fully use the token budget $S_b$, $m$ tokens will be sequentially selected from the queue, regardless of their associated requests. These $m$ tokens can be from different requests or from one request. For a long prompt, varying numbers of tokens may be added to the batch in different iterations, automatically partitioning the prompt into chunks of different lengths.}

\DEL{\begin{figure}[htbp]\vspace{0.05in}
\begin{minipage}[t]{0.48\linewidth}
\includegraphics[width=\linewidth,height=0.15\textheight]{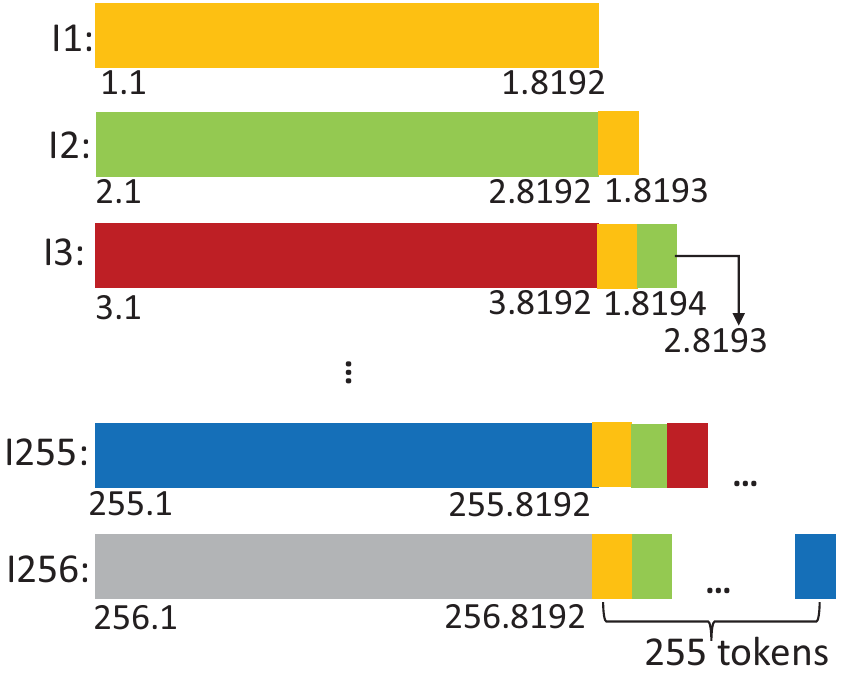}
\vspace{-0.01in}    \caption{No chunking: 256 concurrent running jobs, KVC= 8k $\times$ 256 = 2M tokens.} \vspace{-0.05in}
    \label{fig:chunk-motivation-2-1}
\end{minipage}%
    \hfill%
\begin{minipage}[t]{0.48\linewidth}
\includegraphics[width=\linewidth,height=0.15\textheight]{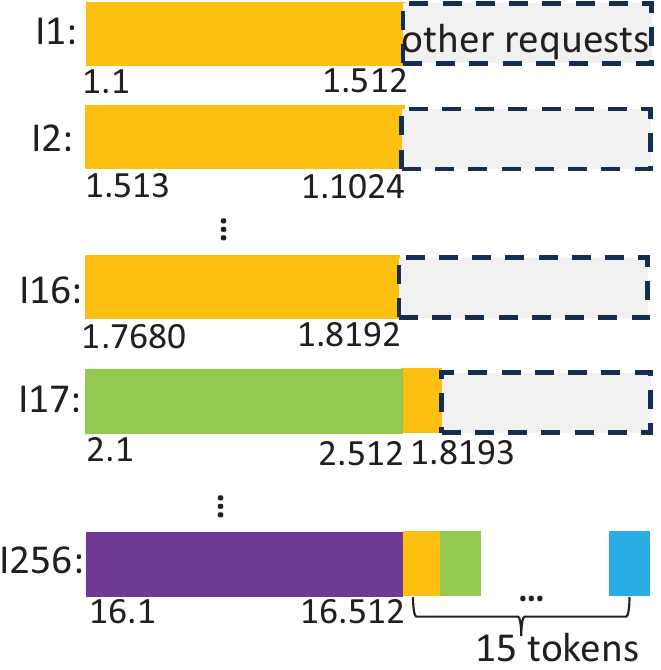}
\vspace{-0.01in} \caption{Chunking: 16 concurrent running jobs, KVC= 8k $\times$ 16 = 128K tokens. } \vspace{-0.05in}
    \label{fig:chunk-motivation-2-2}
\end{minipage}
\end{figure}
}

\DEL{\begin{figure}
    \centering
\includegraphics[width=0.85\columnwidth,height=0.15\textheight]{Fig/Component11.pdf}
    \caption{Dynamic chunking to fully utilize GPU compute resources. }
    \label{fig:dynaChunk}
\end{figure}
}

\DEL{Let us see an example in Figure~\ref{fig:dynaChunk}. There are two requests in the queue; request 1 (in green) has 96 tokens and request 2 (in yellow) is a long prompt with 8192 tokens. 
After batch 1 is processed, it has 100 available token budget. Then, 100 tokens are picked from the waiting queue; 96 tokens from request 1 and 4 tokens from request 2, which are within the available KVC. Then, after this batch is processed, batch 2 has 300 available token budget. The subsequent 300 tokens from request 2 are picked, which are within the available KCV. Next, after batch 2 is processed, batch 3 has 300 available token budget but there are only 128 available KVC. Then, only 128 subsequent tokens from request 2 are picked. As a result, request 2 is dynamically divided into chunks with 4, 300, 128 tokens and so on. Therefore, \Sys addresses the causes of resource underutilization depicted in Figure~\ref{fig:evenchunk}. 
}

The generation of a new token depends only on the logits of the preceding token. Consequently, when processing chunks, the logits operations for the entire prompt except the last token can be skipped. Consequently, only the last token of the final chunk is needed for calculating the logit value for generating the new token of the prompt.


For large models such as OPT-135B, model parallelism becomes necessary. However, adopting model parallelism introduces a tradeoff between peak compute throughput per GPU (which decreases with model parallelism) and forward size capacity (which increases). Consequently, the pivot forward size exceeds that of running the model on a single GPU.


\DEL{??what is the reason that the requests must be executed in sequence in \Orca, and how this method overcome this problem -- it is not in the above 1/11/2023)}

\subsection{Iteration-level SLO-based Task Prioritization }\label{sec:SLO}
\DEL{Unlike the general jobs for which users just need the job final results, for LLM inference service, a user needs to read the text output. The average reading speed of English readers is 4 words per second~\cite{OpenAIAPI}
and 0.1875 second per token~\cite{OpenAIAPI}\cite{jin2023s}. Therefore,}

\DEL{In LLM applications, a good user experience is guaranteed if (s)he receives the first token within her/his specified SLO latency ($SLO_{p}$), and then receives each token in the user's reading speed $SLO_{g}$. 
\Sys achieves this goal and also accommodates JCT SLO. 
}

\DEL{This iteration-level SLO approach also bypasses the obstacle of predicting the length of the response tokens in the JCT SLO approach.}

\DEL{Therefore, in \Sys, a user specifies $SLO_{p}$ based on how long (s)he wants to wait to start reading the text response, and $SLO_{g}$ is the average reading speed by default or set by the user based on his/her reading speed.}

An inference engine may receive both latency-sensitive online requests like chat, and throughput-oriented offline requests like article writing. For latency-sensitive requests, users typically expect a smooth and rapid token generation based on their reading speeds and application requirements, while for throughput-oriented requests, they prioritize receiving the full response within a specific time frame. Sarathi-serve's single system-wide SLO setting fails to satisfy heterogeneous SLOs. \Sys aims to meet the heterogeneous SLO requirements. 
It offers users the flexibility to define the iteration SLO for TTFT and for each TBT (with normal reading speed of 0.1875s/token as default) for online requests, and the JCT SLO for offline requests. In the following, we first demonstrate how \Sys manages the SLOs of online applications, followed by the JCT SLOs of offline applications.


\noindent \textbf{SLOs of online applications.} A request can be a long prompt, a short prompt, a preempted TG task or a returned TG task. The SLO of a request defines the time period allowed to generate a token, beginning at time ($t_0$) when the request enters the queue -- whether from a user or a preemption, or returns from a previous iteration. 
A request's remaining time is the period within which it must execute to meet its SLO. Requests in the waiting queue are ordered by ascending remaining time to prioritize requests with shorter remaining times.


\begin{wrapfigure}{c}{4.2cm}\vspace{-0.0in}
\centering
    \centering
\includegraphics[width=0.5\columnwidth,height=0.05\textheight]{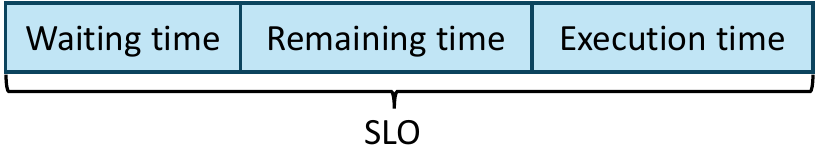}
   \vspace{-0.3in} \caption{Remaining time to execute to satisfy the iteration-level SLO or the JCT SLO. \vspace{-0.1in}}
    \label{fig:SLO}
\end{wrapfigure}


Now, let's see how to calculate a request's remaining time. 
Recall that \Sys may chunk a long prompt (details are in \cref{sec:ragged-processor}), so a long prompt completes processing only after the last chunk is processed. For a request with remaining token length $S_r$ to be processed, the number of its remaining chunks to be processed can be estimated by $N_{ck}\approx \lceil S_r/L_c \rceil$, where $L_c$ is the average chunk length. Note that for a TG task, $N_{ck}=1$. The iteration time of a request in a batch is the batch execution time. We use $T_{max}$ to denote the maximum batch execution time. Then, a request's remaining execution time can be estimated by $T_e \approx  N_{ck} \cdot T_{max}$. As shown in Figure~\ref{fig:SLO}, the remaining time for a request to meet its SLO equals: $T_r=SLO-T_w-T_e$, where $T_w$ is the waiting time since a new or preempted request entered the waiting queue and it equals to 0 for a returned TG task. 
When a request enters the waiting queue or a portion of a long prompt is added to the batch, its $T_r$ is calculated, and its position in the queue is adjusted to maintain ascending $T_r$ order among requests. 



\noindent \textbf{SLOs of offline applications.} Users may specify a JCT SLO (denoted as $SLO_{JCT}$) for their requests in offline applications. Although JCT SLO is not our primary focus, we ensure that \Sys can also accommodate it.
Unlike the iteration-level SLO, the JCT SLO applies to the entire job, encompassing prompt processing and all token generations. Consequently, \Sys must estimate the total number of tokens to be generated (denoted as $S_g$), potentially using methods like that in \cite{Zheng2023Response}. Accurately predicting token generation remains a challenging research problem, though it is beyond the scope of this paper. Moreover, JCT SLOs for offline applications typically do not require strict adherence, so an approximate prediction should be sufficient.

A TG task may experience preemption due to KVC overflow, resulting in a preemption period before resuming execution. Therefore, the maximum time for a single token generation can be estimated as $(T_{max}+P_{max}\cdot P)$, where $P_{max}$ represents the maximum duration of a preemption, and $P$ represents the probability of a request being preempted after a token generation process. When such a prompt enters the waiting queue, its JCT is then calculated as $T_e \approx N_{ck}\cdot T_{max}+S_{g}\cdot (T_{max}+P_{max}\cdot P)$, where $N_{ck}$ actually is the estimated total number of chunks. 
The total remaining waiting time is estimated as $SLO_{JCT}-T_e$. If this remaining waiting time is evenly distributed across all iterations of processing this prompt, the allowed waiting time per iteration is given by ${T}_r=(SLO_{JCT}-T_e)/(N_{ck}+S_g)$. For each iteration, the request's waiting time is initally set to ${T}_r$. 
However, this is not a hard deadline since the objective is to ensure that the JCT does not exceed the JCT SLO.


\DEL{To satisfy JCT SLO, a request must start running by a time that leaves enough time for the subsequent prompt processing and token generations. At a certain time, the subsequent execution time of a request is estimated as $T_e=N_{ck}^r\cdot T_{max}+S_{g}^r\cdot (T_{max}+P_{max}\cdot P)$, where $N_{ck}^r$ is the number of remaining chunks in $N_{ck}$, and $S_{g}^r$ is the number of remaining tokens in $S_{g}$.
Similarly, as shown in Figure~\ref{fig:SLO}, the remaining time for a request to start an iteration to meet its JCT SLO equals: $T_r=SLO_{JCT}-T_w-T_e.
}

Given that the JCT SLO applies to the entire job, any over-utilized or under-utilized waiting time is propagated to subsequent iterations to maintain the total allowable waiting time. If a request completes $d$ time units later than its allowed waiting time ${T}_r$, the allowed waiting time for the next iteration becomes $T_r={T}_r-d$. If the next iteration actually runs for $T_r+d_1$, the allowed waiting time for the next iteration becomes ${T}_r-d_1$. This excess waiting time is propagated across subsequent iterations until it reaches zero. Conversely, if a request completes $d$ time units earlier than ${T}_r$, the deadline for the next iteration is adjusted to ${T}_r+d$. This unused waiting time is also carried forward, allowing for extended waiting times in future iterations.  



\DEL{Such requests are marked, so \Sys can postpone running such requests in order to run other urgent requests. }

\Sys prioritizes requests in ascending order of remaining time, which helps meet SLOs and enhances throughput by batching requests with similar remaining times, as indicated by O\ref{pivot}. \Sys classifies the requests to two categories: urgent requests and non-urgent requests. Urgent requests are those that must be executed in the next iteration to satisfy their SLOs. If a request's remaining time $T_r$$\approx$${T}_{max}$, it is considered urgent. 


\subsection{Multi-resource-aware Batching
}\label{sec:Cache}
O\ref{EvenChunking} shows it is important to jointly maximize both GPU compute utilization and KVC utilization at each iteration to improve throughput.
O\ref{imbalance} highlights that the varying resource demands of requests present an opportunity to select prompts or prompt chunks that can fully utilize both resources. 
Selecting requests using FCFS may fail to reach $S_b$ due to KVC limits or may reach $S_b$ without fully allocate the KVC. Thus, instead of using FCFS, after each iteration, \Sys selectively chooses requests to be added to the batch to fully allocate KVC and also fully utilize GPU. \DEL{The cache space available to be allocated is the total KVC space minus the allocated cache space and the cache space that is required by the requests in the currently returned batch.} 



\DEL{It wastes 10\% of the allocated cache.
{\color{red} It reduces cache wastage because it preempts the requests that cannot be executed with the available memory with requests from the waiting queue that can be executed{\sh{the former only needs 1 block, and the latter needs no less than 1 block, they the latter would need less cache than the former?}}. Thus, no cache memory gets unused if there are available requests to be executed.?} 
{\sh{Mura still has preemption? If yes, when does preemption occur? since SLO is removed, should be no preemption, right? }}
}


Requests with the same GPU compute demands can still have varying KVC demands. We use $D_m^i$ to denote the KVC demand of a request $i$. If it is the first chunk or an entire prompt,
then $D_m^i=\lceil \frac{S_l^i}{b} \rceil \times b$, where $S_l^i$ denotes its sequence length and $b$ denotes the block size. If it is a non-first chunk, some of its tokens may already have allocated KVC in the block of previous tokens. Thus, only the remaining tokens require KVC allocation, calculated as $D_m^i=\lceil \frac{remaining~S_l^i}{b} \rceil \times b$. For a TG task (from a previous preemption), if a block is already assigned for its previous tokens including this token, then it does not need KVC allocation; otherwise, it needs one block of KVC. As a result,  some requests may require less KVC space than others even though they have the same sequence length (or GPU compute demand). Therefore, it is important to identify the requests that can better utilize both resources to improve throughput.

\DEL{When \Sys sequentially selects each request (a prompt, a prompt chunk or a preempted TG task) from the waiting queue to add to the batch to reach $S_b$, it ensures that the available KVC space is
no less than the KVC demand of the request. 
If this condition cannot be satisfied, then if the selected request is a prompt chunk, \Sys reduces the length of the chunk to fit the available KVC space;
otherwise, \Sys will pick the next request repeatedly until finding a request whose KVC demand can be satisfied.  The process repeats until the KVC is fully allocated. 
}


\DEL{{\color{gray}{Due to the autoregressive feature, the requests can be initiated and completed any time. Therefore, even though a request does not have enough KVC space for its generated tokens when it joins the batch, it may receive KVC space during processing after other requests leave. The details of this method is presented in the next section.}}}


\DEL{As per O\ref{imbalance}, it's imperative to prevent both overutilization and underutilization of GPU and KVC resources in each iteration. 
However, 
the above request selection method may miss better options that can more fully utilize both resources simultaneously. 
}


\DEL{After each iteration, \Sys needs to select requests to form a batch in order to achieve the goal. \DEL{We assume that each request cannot wait in the queue for more than $\tau$ time period.
Urgent requests are the requests that have waited in the queue for more than $\tau$ and must be processed in the next iteration.} The urgent requests must be in the batch, and the previous TG requests should be remained in the batch as much as possible in order to let them complete and release the KVC early. Then, among the non-urgent requests, \Sys sequentially selects requests in the queue. 
We present the details in the following.
}

\begin{algorithm}[t]
    \footnotesize
    \SetAlgoLined
    \LinesNumbered
    \SetCommentSty{small}
    \SetKwInOut{Input}{Input}
    \SetKwInOut{Output}{Output}

    \Input{$S_{b}$, $S_f$\\
    $A_{KV}$: available (i.e., unallocated) KVC space\\
    \textbf{U}: urgent requests in waiting queue\\
    \textbf{Q}: waiting queue}
    \Output{Batch $\mathbf{B}$ for the next iteration}

$\mathbf{U}\rightarrow \mathbf{B}$ //Assign all urgent requests to the batch\\
\While{Available GPU or KVC capacity $<0$}
{Preempt the request$\in \mathbf{B}$ with $\max{T_r}$}
\If{exists available GPU and KVC capacity}
{$\mathrm{SelectRequests(S_b-S_f, A_{KV},\mathbf{Q},\mathbf{B})}$\\}

\DEL{ Add \textbf{U} to batch \textbf{B}\\
$A_{KV}=\mathrm{CheckCache(\textbf{U})}$
 \If{$S_{f}<S_b$ or $A_{KV}>0$}
  {$\mathrm{SelectRequests(S_b-S_f, A_{KV},\mathbf{B})}$}
       \eIf{$S_{f}>S_b$ or $A_{KV}<0$}
        {
          \tcc{Preempt requests to solve GPU and MEM deficit}
         \eIf{deficit only in GPU}
            {Preempt minimum \# tasks with loosest SLOs and least KVC use to solve the deficit}
            {Preempt minimum \# of tasks with loosest SLOs and most KVC use to solve the deficits}
            } }
    \caption{Pseudocode of the batching algorithm.}
    \label{alg-partition}
    \end{algorithm}

\begin{algorithm}[t]
\footnotesize
    \SetAlgoLined
    \LinesNumbered
    \SetCommentSty{small}
    \SetKwInOut{Input}{Input}
    \SetKwInOut{Output}{Output}

    \Input{
    $A_{GPU}$: available GPU resource represented by \# of tokens, $A_{KV}$, $\textbf{Q}$ and $\textbf{B}$}
    \Output{
    Batch \textbf{B} to fully utilize $A_{KV}$ and $A_{GPU}$}
$A_c=A_{GPU}$\\
$A_m=A_{KV}$\\

Get waiting requests with $T_r$ within $T_r^1+\gamma$\\

\For{each long prompt in the selected requests}{Create a shorter chunk with length either making $S_f=S_b$ or fully allocate KVC}
\While{ exists a request or chunk $i$ with $A_c\geq D_c^i$ \& $A_m\geq D_m^i$}
{Choose the request with $\min \sqrt{(A_c-D_c^i)^2+(A_m-D_m^i)^2}$\\
Add the request to $\mathbf{B}$\\
Remove the request from $\textbf{Q}$\\
$A_c=A_c-D_c^i$\\
$A_m=A_m-D_m^i$
}
    \caption{\small{$\mathrm{SelectRequests(A_{GPU}, A_{KV}, \mathbf{Q}, \mathbf{B})}$.}}
    \label{alg-checkCache}
    \end{algorithm}

After completing an iteration of a batch, the PP tasks in the batch transition to TG tasks from the next iteration onward.  The \Sys scheduler picks requests in the waiting queue to achieve $S_f$$\approx$$ S_b$. 
Algorithm~\ref{alg-partition} shows the pseudocode of the multi-resource aware batching algorithm. 
Urgent requests must be in the batch, and the previous TG requests should be retained in the batch as much as possible to allow them to complete and release KVC earlier (based on O\ref{kvcache2}). 
We use \textbf{U} to denote the group of urgent requests. \Sys first selects the requests from the urgent requests to the batch (line 1). If there are not enough resources to support an urgent request, the request with the highest $T_r$ in the current batch will be preempted (Lines 2-4). After allocating urgent requests, if there are available GPU compute and KVC capacities, \Sys selects requests from non-urgent queuing requests by calling function  {\tt{SelectRequests()}} (lines 5-7). 

\DEL{\begin{algorithm}[t]
\footnotesize
    \SetAlgoLined
    \LinesNumbered
    \SetCommentSty{small}
    \SetKwInOut{Input}{Input}
    \SetKwInOut{Output}{Output}

    \Input{$A_{KV}$: available KVC space\\
    \textbf{G}: a group of requests }
    \Output{If adding \textbf{G} to the batch will not overflow the KVC}

\For{each $r \in \textbf{G}$}{
\If{it is an entire prompt or first chunk}{
$D_{KV}=\lceil S_l/S_b\rceil\times S_b$\\
}
\If{it is non-first chunk of a prompt}
{
$D_{KV}=\lceil (remaining~S_l)/S_b\rceil\times S_b$\\
}
\If{it is a TG task}
{
    \eIf{It is in assigned KVC}
{$D_{KV}=0$}
{$D_{KV}=S_b$}
}
$D_{sum}=D_{sum}+D_{KV}$
}
Return $A_{KV} \geq D_{sum}$
    \caption{Pseudocode of $\mathrm{CheckCache(\textbf{G})}$.}
    \label{alg-checkCache}
    \end{algorithm}
}



\DEL{To calculate KVC demand of a request, $D_m$, if it is the first chunk or an entire prompt,
then $D_m=\lceil \frac{S_l}{S_b} \rceil \times S_b$, where $S_l$ is its length. If it is a non-first chunk, some of its tokens may already have been allocated with KVC in the block of previous tokens, then only the remaining tokens need to be allocated to the KVC, and $D_m=\lceil \frac{remaining~S_l}{S_b} \rceil \times S_b$.
For a TG task (from a previous preemption), if a block is already assigned for its previous tokens including this token, then it does not need KVC allocation; otherwise, it needs one block of KVC. As a result,  some requests may require less KVC space than others even though they have the same sequence length (or GPU compute demand). 
}


In the {\tt{SelectRequests()}} function (shown in Algorithm~\ref{alg-checkCache}), its goal is to select requests from the requests with the least remaining time to maximize GPU compute and KVC utilization in the next iteration. Let's use $T_r^1$ to denote the remaining time for the first non-urgent request. \Sys selects several requests from the head of the queue that have $T_r$ within $T_r^1+\gamma$, where $\gamma$ is a small value (Line 3). Then, if any long prompts exist, a chunk is created for each using the \emph{SLO-guaranteed dynamic chunking method} described in \cref{sec:ragged-processor}, or by fully allocating KVC, whichever is shorter (Lines 4–6). This step aims to simultaneously maximize both GPU compute and KVC utilization. From this group, \Sys first selects requests whose GPU and KVC demands ($D_c^i$ and $D_m^i$) are the nearest to the available resources ($A_c$ and 
$A_m$), i.e., those with the least Euclidean distance (Line 8). This ensures that the request that can use most of both the GPU and KVC resources is selected first. \Sys repeats this process until the remaining resources cannot support more requests (Lines 7-12).


\section{Implementation}\label{sec:implementation}

We implemented \Sys based upon the published source code of vLLM~\cite{vllm}.  
The \Sys system} is written in 9K lines of Python and 3K lines of C++ code. We developed the prioritization method 
using Python, and developed the dynamic chunking method 
using C++. We replaced the default FCFS scheduler in vLLM with \Sys. In \Sys, 
the module for the multi-resource aware batching method communicates with the vLLM's block manager to obtain information regarding the current KVC state. \Sys forwards the newly formed batch 
to the vLLM backend for execution. 
We added the $flash\_attn\_qkvpacked\_func$ from the FlashAttention2~\cite{Dao2023FlashAttention2FA} library in the vLLM engine for the parallelization of the attention operation. 
We directly use vLLM to handle KVC overflow and preemptions. 

\DEL{We used the default vLLM strategy for storing the KVC meta-data at a table to keep track of the physical block in the host memory\sh{what is the purpose for this operation, what's the meaning of "following the strategy in the host memory"?}.}  

\DEL{When KVC is overflown, we preempt the request and move the request in a queue to the CPU memory following the swapping strategy from vLLM~\cite{vllm}. When enough space is available in the GPU memory for at least one request, the preempted requests are moved back one by one by communicating with \Sys. }

\DEL{We developed a scheduler that selects the requests based on the   and the block manager \tanbin{(i.e., the part of the code that allocates, deallocates and swaps block of memory)} \sh{what is the block manager?-done}in Python while
using C++ for operations including Method (2) \emph{Chunk} along side the cache management.
Method (1) \emph{SLO} was implemented on the Python scheduler end. Method (3) \emph{Cache} was built using the  PagedAttention~\cite{vllm} in C++. We implemented Method (4) \emph{Batch} using the \tanbin{ragged batching technique from Triton inference server on the top of vLLM}. \sh{above descrption for the methods is not organzied. When you said Method (1) is on the scheuler end, why didn't you comine this into the first sentence, why Method 2 is in the first sentence. pls re-organize.}  We stored the page-table metadata \tanbin{holding the block locations} in the host memory\sh{this sentence is too all of sudden. you need to let people know what "page-table" is first}. We used NVIDIA Collective Communication Library (NCCL) for communication across the distributed GPU machines.

Upon deciding which requests to batch together from the SLO and task prioritization, our Python scheduler \sh{what does it do?} (Method 2) sends the requests to the Triton Inference Server \sh{what does it do?} (Method 4), which forms the batch.}

\DEL{Instead of forming a 2-dimensional batch, \sh{using the approach in ?? reference}, \Sys forms all requests in a batch into one sequence as a ragged batch. The requests in a batch are distinguished through markers placed at the request boundaries. During batching, the available prompt requests are chunked using Method 2. These chunks are determined through communication with our Python scheduler\sh{communication between what and what?} and added to the batch to meet the forward size\sh{who does it?}. Then, we\sh{which component? Inference engine?} executes the iterations. During the iteration, the cache is maintained using Method 3, which involves inter-communication between the Python scheduler\sh{what does it do?} and the Triton Inference Server\sh{what does it do?}. Finally, the batched sequences are processed in parallel using multi-threading~\cite{Dao2022FlashAttentionFA,Dao2023FlashAttention2FA}. We used the $flash\_attn\_qkvpacked\_func$ from the FlashAttention2~\cite{Dao2023FlashAttention2FA} library to parallelize the attention operations. When KVC is overflown by a request, we preempt the request and move the request in a queue to the CPU memory following the swapping strategy from vLLM~\cite{vllm}. We let other requests occupy the KVC run\sh{correct English} and fill the empty cache keeping the iteration-level KVC full allocation in mind\sh{rewite English, in mind is not good}.
When enough space exists in the GPU memory for at least one request, the preempted requests are moved back one by one, following the strategy in~\cref{sec:SLO}\sh{this operation appears all of sudden, pls re-organize}.

\DEL{{\color{blue} Instead of forming a 2-dimensional batch, \Sys forms all requests in a batch into one sequence as a ragged batch. The requests in a batch are distinguished through markers placed at the request boundaries. The sequences are processed in parallel using multi-threading~\cite{Dao2022FlashAttentionFA,Dao2023FlashAttention2FA}. We used the $flash\_attn\_qkvpacked\_func$ from the FlashAttention2~\cite{Dao2023FlashAttention2FA} library to parallelize the attention operations. 
{\color{blue} When KVC is overflown by a request, we preempt the request and move the request in a queue to the CPU memory following the swapping strategy from vLLM~\cite{vllm}. We let other requests occupy the KVC run and fill the empty cache keeping the iteration-level cache guarantee in mind.
When enough space exists in the GPU memory for at least one request, the preempted requests are moved back one by one, following the strategy in~\cref{sec:SLO}.}}}

Currently, \Sys can only tested for the mentioned models and the GPU settings. We will make it more generalizable for different models and machine settings. } 

\DEL{Instead of forming a 2-dimensional batch, \Sys forms all requests in a batch into one sequence. The requests in a batch are distinguished through markers placed at the request boundaries. The sequences are processed in parallel using multi-threading~\cite{Dao2022FlashAttentionFA,Dao2023FlashAttention2FA}.  For a long sequence, \Sys parallelizes over the sequence length dimension using the approach in~\cite{Dao2023FlashAttention2FA}. \Orca executes requests in the attention layer in sequence since they are not in the same phase. In contrast, \Sys parallelizes the execution of attention layers within a batch
by using the process from FlashAttention2~\cite{Dao2023FlashAttention2FA}. Additional details and experimental results on the improvement from this parallelism can be found in Appendix~\ref{parallelAttention}.}


\section{Performance Evaluation} 
\label{sec:evaluation}

\begin{figure*}[t]
\centering
\subfloat[Throughput.\vspace{-0.01in}\label{fig:th-overall}]{{\includegraphics[width=0.32\linewidth,height=0.15\textheight]{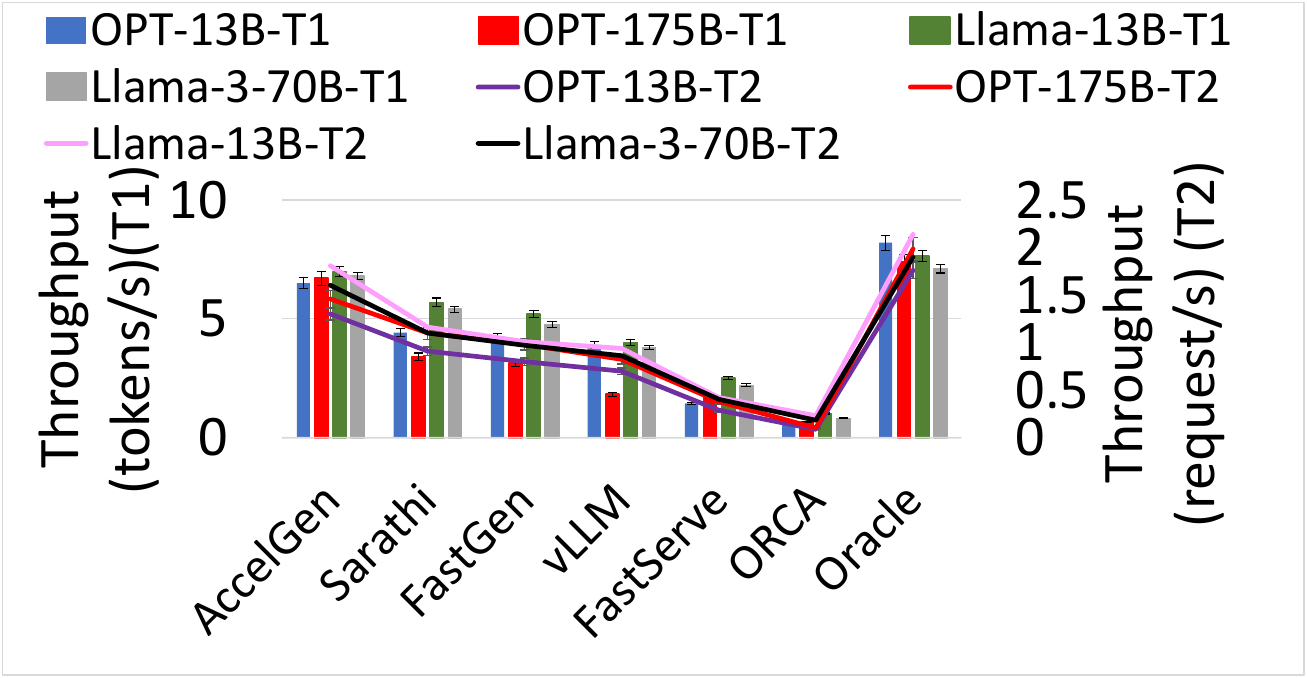} }}
    \hfill
\subfloat[Goodput.\vspace{-0.01in}\label{fig:gd-overall}]{{\includegraphics[width=0.32\linewidth,height=0.15\textheight]{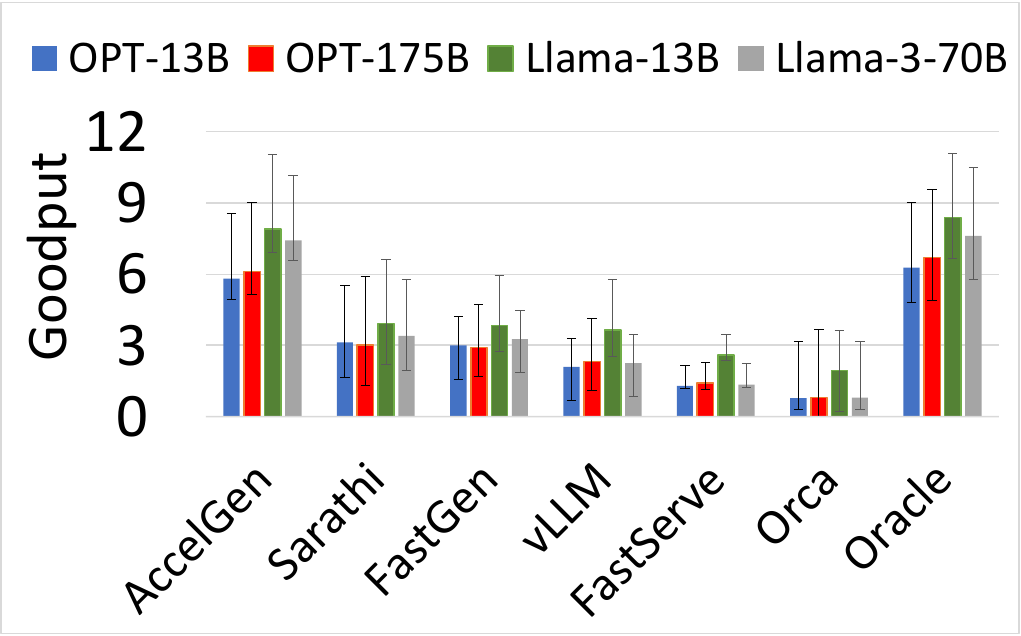} }}
    \hfill
\subfloat[SLO attainment.\vspace{-0.01in}\label{fig:slo-overall}]{{\includegraphics[width=0.32\linewidth,height=0.15\textheight]{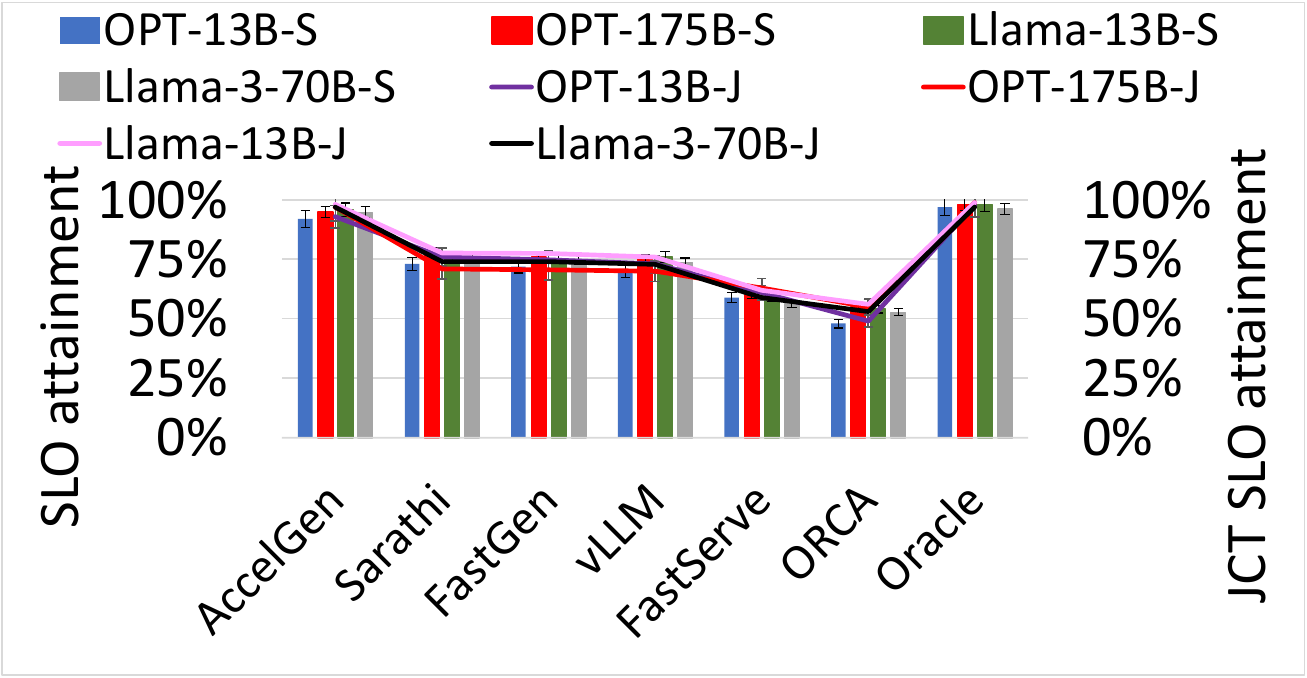} }}
    \hfill
\DEL{\subfloat[Iteration time.\vspace{-0.01in}\label{fig:iteration-time-overall}]{{\includegraphics[width=0.24\linewidth,height=0.15\textheight]{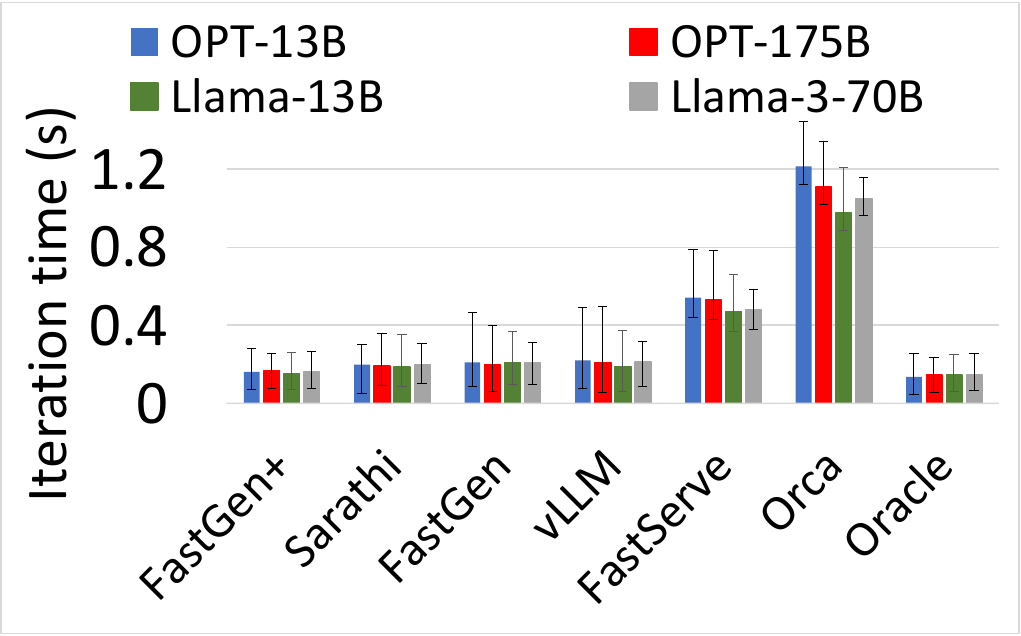} }}
\hfill}
\subfloat[JCT.\vspace{-0.01in}\label{fig:jct-overall}]{{\includegraphics[width=0.32\linewidth,height=0.15\textheight]{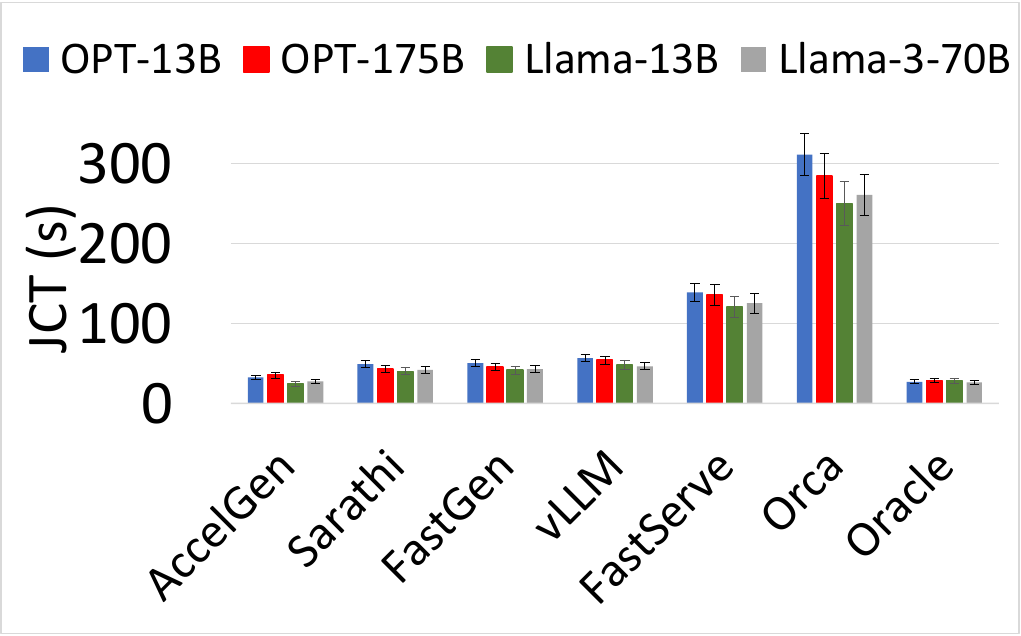} }}
\hfill
\DEL{\subfloat[Resource utilization.\vspace{-0.01in}\label{fig:th-gpu-overall}]{{\includegraphics[width=0.24\linewidth,height=0.15\textheight]{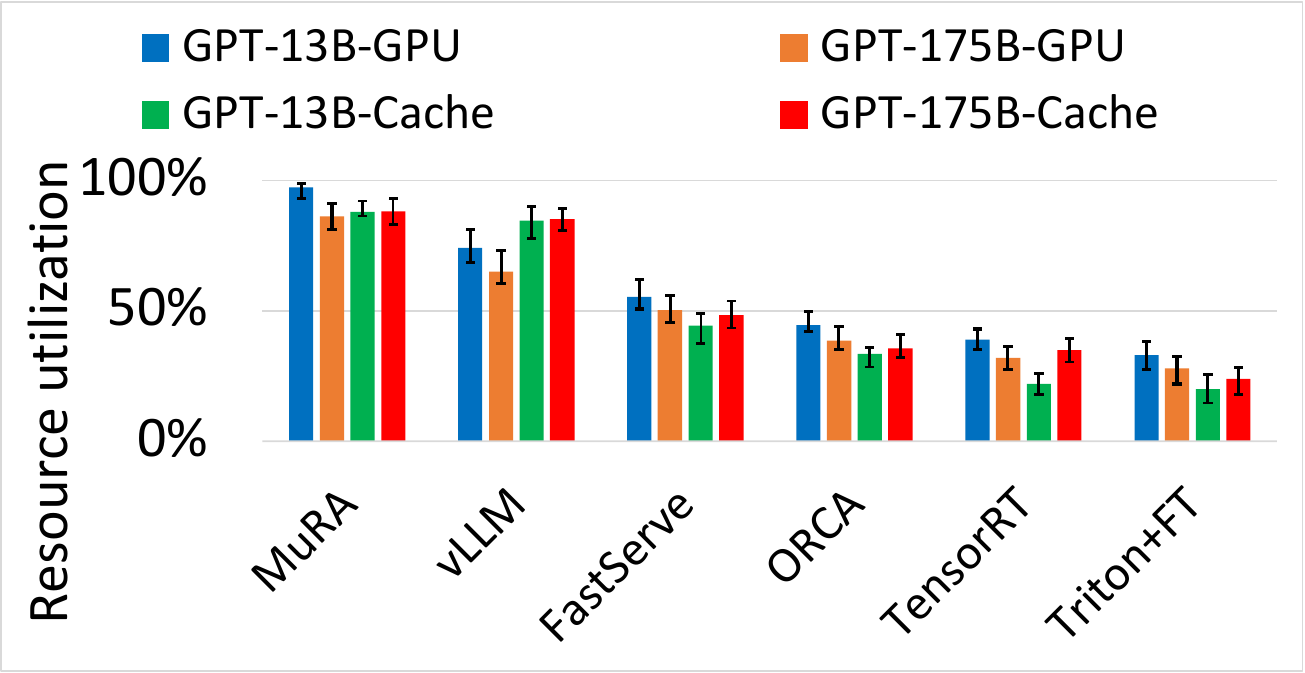} }}
    \hfill}
\DEL{\subfloat[KVC overflow ratio.\vspace{-0.01in}\label{fig:mem-overall}]{{\includegraphics[width=0.24\linewidth,height=0.15\textheight]{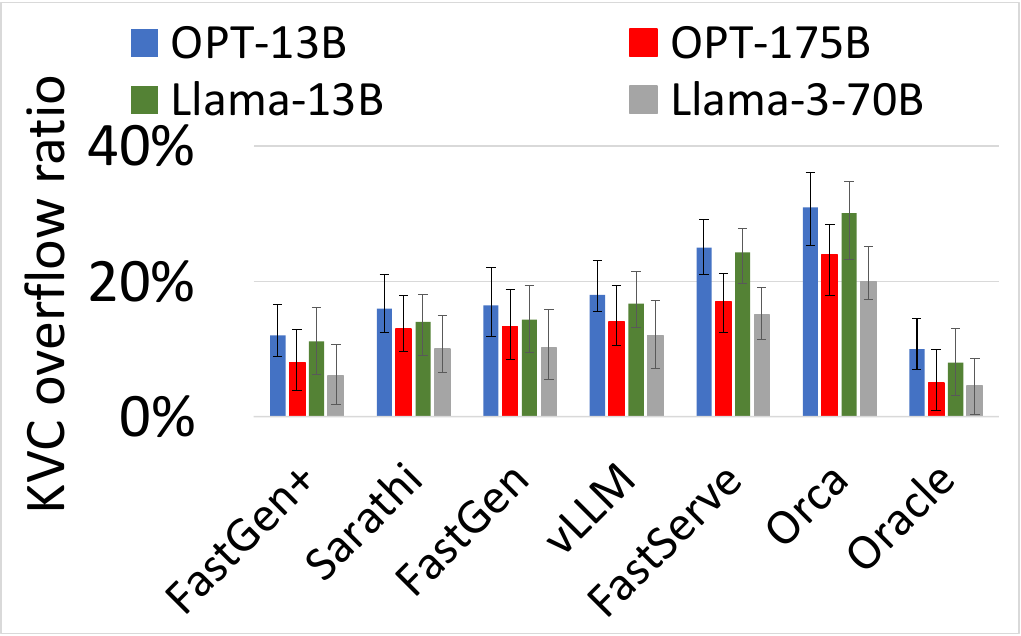} }}
    \hfill}
  \subfloat[GPU compute and unallocated KVC. 
\label{fig:time-mura}]{{\includegraphics[width=0.32\linewidth,height=0.15\textheight]{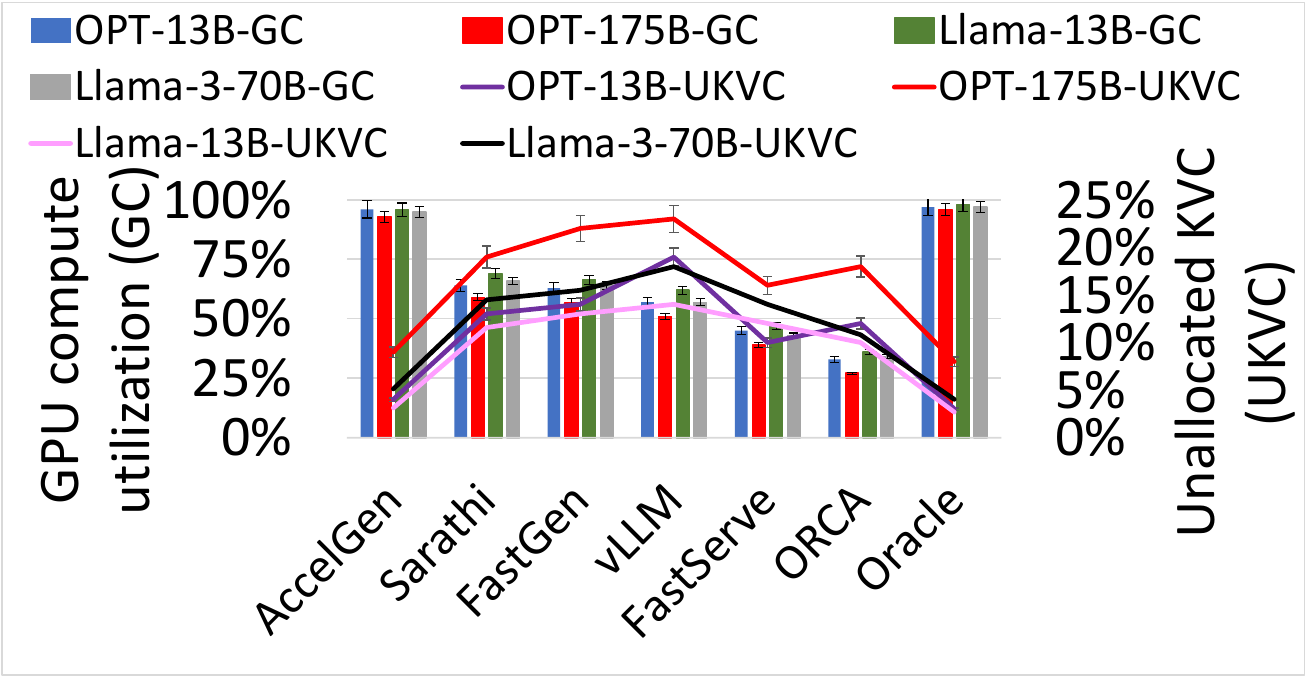}}}
      \hfill
   \subfloat[Time overhead. \label{fig:overhead-all}]{{\includegraphics[width=0.32\linewidth,height=0.15\textheight]{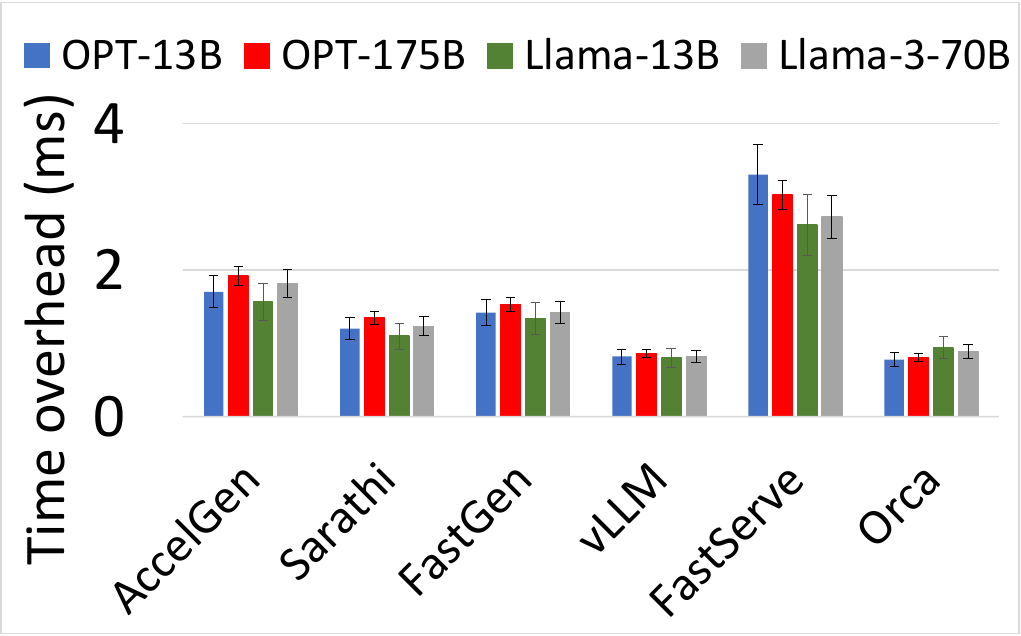} }}
\vspace{-0.05in}
   \caption{Results of \Sys and compared systems for different models.\vspace{-0.1in}}
    \label{fig:met-overall}
\end{figure*}

\subsection{Experiment Settings} The experiment settings are the same as in \cref{sec:analysis} unless otherwise specified. \DEL{We ran th used the OPT model with 175B parameters
~\cite{brown2020language}. It was executed using inter-layer parallelism across two machines and intra-layer parallelism among the GPUs within a single machine. We used two machines because at least 640GB of GPU memory is required to run the model with a sequence length of 1048, while one machine has only 320GB GPU memory. We evenly divided the model between the two machines. A total of 264GB GPU memory was used for the KVC.}
In addition to the mentioned LLM models in~\cref{sec:analysis}, we ran Llama-13B and Llama-3-70B model on one GPU and four GPUs of one machine, respectively. They used 12GB and 135GB of GPU memory for the KVC, respectively. 
We set $\gamma=0.75s$ empirically. 
\DEL{We fixed the same settings for all the methods. We reported the average with the 5th and 95th percentile values.}


\DEL{We grouped the prompts within the length range of 512 tokens, and measured the average prompt processing latency for each group using the Triton Inference Server with the Faster Transformer backend (Triton+FT)). As~\cite{Li2023AlpaServeSM} that uses SLO-scale to determine SLO, we set a PP task's SLO as the product of the average prompt processing latency of its group and an SLO-scale~\cite{Li2023AlpaServeSM} randomly selected from the range of [0.5,1.5].}
We set the TTFT SLO as explained in \cref{sec:analysis}. To set the TBT SLO of a request, we randomly chose the SLO-scale from [0.75,1.25] and multiplied it by 0.1875s considering that some people may have a faster reading speed than the normal speed while others may have a slower reading speed.
\DEL{We set the SLO for a TG task to 0.1875s, ensuring the generation of one token within the normal reading speed.}


\noindent{\textbf{Compared Methods.}}
We compared \Sys with vLLM, \Orca, FastGen and Sarathi-Serve (Sarathi in short). 
We also include FastServe~\cite{Wu2023FastDI} as a reference, which is built on \Orca. FastServe 
uses a skip-join multi-level feedback queue scheduler~\cite{skq} to preempt the longest-running job to a lower-priority queue to minimize the average JCT. We set the number of queues to five. We also include the \emph{Oracle} that knows the remaining time for job completion, the arrival times and output lengths of all requests, and schedules the chunking and batching to maximize the goodput. We use the Vidur simulator~\cite{Agrawal2024VidurAL} to find the \emph{Oracle}'s schedule solution. 

\DEL{order of the chunk selection\sh{what is the order of chunk selection? If you do not preknow it, then what happens?-done} beforehand. Knowing the chunk order can help schedule chunks in a way that minimizes KVC conflicts and avoids unnecessary preemptions or delays. To obtain the Oracle, we ran all the requests in the waiting queue beforehand in the Vidur simulator~\cite{Agrawal2024VidurAL}, and generated 8 possible combinations of scheduling the prompts by chunks and chose the combination that gave me the best goodput.}  

\DEL{However, none of these methods are targeted for long prompts.
We also measured the contribution of each component of \Sys in the overall performance improvement using the following variants of \Sys.
\squishlist
\item{\Sys/Chunk.} It is \Sys without the long prompt chunking.
\item{\Sys/Cache.} It is \Sys without iteration-level KVC allocation.
\item{\Sys/SLO.} It is \Sys without iteration-level SLO setting.
\item{\Sys/Batch.} It is \Sys without the dynamic batching policy.
\squishend
}

\DEL{
\noindent{\textbf{Metrics.}}
We report the following metrics.
\squishlist
\item{\textbf{JCT SLO attainment.}} SLO guarantee is the ratio of the number of jobs that meet the set JCT SLO to the total number of jobs for each experiment.

\item{\textbf{Goodput.}} It is the number of requests that satisfy their iteration-level SLOs during a certain time period.

\item{\textbf{Iteration time.}} It is measured as the time spent in seconds for generating each token in the TG phase for each job that met the JCT SLO.

\item{\textbf{GPU computation utilization.}} It is measured as the percentage of GPU compuation resources utilized using the gpustat~\cite{gpustat} library.

\item{\textbf{KVC overflow ratio.}} It is measured as the ratio of the requests for which allocated or used memory has overflown to the ratio of the total number of requests for each experiment.
\squishend
}

\subsection{Evaluation Results} 

In figures featuring both bars and lines, the bars represent the metric on the left axis, while the lines correspond to the metric on the right axis.

\DEL{\begin{figure}[]
\centering
    \subfloat[Overall.\vspace{-0.01in}\label{fig:jct-time-overall}]{{\includegraphics[width=0.45\linewidth,height=0.15\textheight]{Fig/} }}
    \subfloat[\Sys without a method.\vspace{-0.01in}\label{fig:jct-time-component}]{{\includegraphics[width=0.45\linewidth,height=0.15\textheight]{Fig/JCT-comp-err-2.png} }}
    \vspace{-0.05in}
    \hfill   \caption{\small{JCT (to be deleted).\vspace{-0.1in}}}%
    \label{fig:jct}
\end{figure}}
\DEL{\noindent\textbf{JCT and JCT SLO Guarantee.} Figure~\ref{fig:jct-time-overall} shows the  JCT, decomposed to the time on PP and on TG. For the long-prompt scenario, compared to \Orca and FastServe, \Sys has 11$\times$ and 4.8$\times$ lower JCT, 4.65$\times$ and 2.5$\times$ lower PP time, and 15.9$\times$ and 4.7$\times$ lower TG time.
For the mixed-prompt scenario, \Sys has 10.1$\times$ and 3.8$\times$ lower JCT, 4.54$\times$ and 2.3$\times$ lower PP time, and 14.8 $\times$ and 4.3$\times$
lower TG time.

Figure~\ref{fig:jct-time-component} shows the JCT of each variant of \Sys. ``/$Z$'' means it does not have method $Z$ in \Sys. \Sys/Chunk, \Sys/SLO, \Sys/Batch and \Sys/Cache
increase the JCT of \Sys by 5.6$\times$, 5.5$\times$, 4.1$\times$ and 32\%, respectively, on average for both scenarios (we report this average value for other metrics in the following). \Sys/Chunk has the highest impact because it greatly reduces negative impact from the long prompts while leverages them to improve throughput and JCT. \Sys/Cache has the second impact on throughput increases the JCT the least. Forming the resource aware batches reduces JCT, which is reflected with the increase of JCT for \Sys/Batch. \Sys/SLO helps ensure that each request completes its iteration within a certain time limit. }

\DEL{\begin{figure}[]
\centering
    \subfloat[Overall\vspace{-0.01in}.\label{fig:jct-overall}]{{\includegraphics[width=0.405\linewidth,height=0.15\textheight]{Fig/jct-221.png} }}
    \hfill
    \subfloat[\Sys without a method.\vspace{-0.01in}\label{fig:jct-comp}]{{\includegraphics[width=0.565\linewidth,height=0.122\textheight]{Fig/jct-comp-221.png} }}
    \hfill
\vspace{-0.05in}
   \caption{\small{JCT SLO attainment. (to be deleted)\vspace{-0.1in}}}%
    \label{fig:jct-guarantee}
\end{figure}}

\DEL{Though we do not focus on JCT SLO, we still would like to present the results as reference. We use $\sum (SLO_p+SLO_g\times S_g)$, where $S_g$ denotes the number of generation tokens as a job's JCT SLO. Figure~\ref{fig:jct-overall} shows the results of JCT SLO attainment.
In the long-prompt scenario, compared to \Orca and FastServe, \Sys has 45\% and 23\% higher ratios, while in the mixed-prompt scenario, it has 41\% and 21\% higher ratios.}

\DEL{Figure~\ref{fig:jct-comp} shows the JCT SLO satisfaction for the variants of \Sys. 
\Sys/Chunk, \Sys/SLO, \Sys/Batch, and \Sys/Cache
reduce \Sys's JCT SLO satisfaction by 18\%, 15\%, 14\% and 10\%, respectively.

The above results verify the effectiveness of \Sys's methods in dealing with long prompts and reducing JCT. Its effectiveness is more obvious in the long-prompt scenario.
FastServe outperforms \Orca because of it can solve the head-of-the-blocking problem. However, it still shows lower JCT performance than \Sys because of processing long prompts and failing to fully utilize resources. 

The results also show the effectiveness of individual methods in \Sys. The iteration-level SLO guarantee method and chunking method help reduce JCT and improve the JCT SLO guarantee the most. 
Batching helps by forming batches that ensure high GPU compute utilization, which ultimately reduces the job latency. \Sys/Cache's effectiveness is the least because it only makes sure that
a request won't be selected for a batch if its KVC demand cannot be satisfied instead of directly reducing the latency. }

\DEL{\noindent{\textbf{Throughput.}} 
Figure~\ref{fig:goodput-overall} shows the goodput results.
Compared to \Orca and FastServe, \Sys has 47\% and 24\% higher goodput in the long-prompt scenario, and 43\% and 22\% higher goodput for the mixed-prompt scenario, respectively. 

Figure~\ref{fig:goodput-comp} shows  the goodput for the variants of \Sys. \Sys/Chunk, \Sys/SLO, \Sys/Batch, and \Sys/Cache reduce \Sys's goodupt by 21\%, 16\%, 12\% and 9\%, respectively. 
The reasons for the performance improvement are the same as explained above. }

\DEL{\begin{figure}[]
\centering
    \subfloat[Overall.\vspace{-0.01in}\label{fig:goodput-overall}]{{\includegraphics[width=0.405\linewidth,height=0.15\textheight]{Fig/goodput-221.png} }}
    \hfill
    \subfloat[\Sys without a method.\vspace{-0.01in}\label{fig:goodput-comp}]{{\includegraphics[width=0.565\linewidth,height=0.15\textheight]{Fig/goodput-comp-221.png} }}
    \hfill
\vspace{-0.05in}
   \caption{\small{Goodput (to be deleted).\vspace{-0.1in}}}%
    \label{fig:goodput}
\end{figure}}

\DEL{\begin{figure}[]
\centering
    \subfloat[13B.\vspace{-0.01in}\label{fig:iteration-alpaca}]{{\includegraphics[width=0.48\linewidth,height=0.15\textheight]{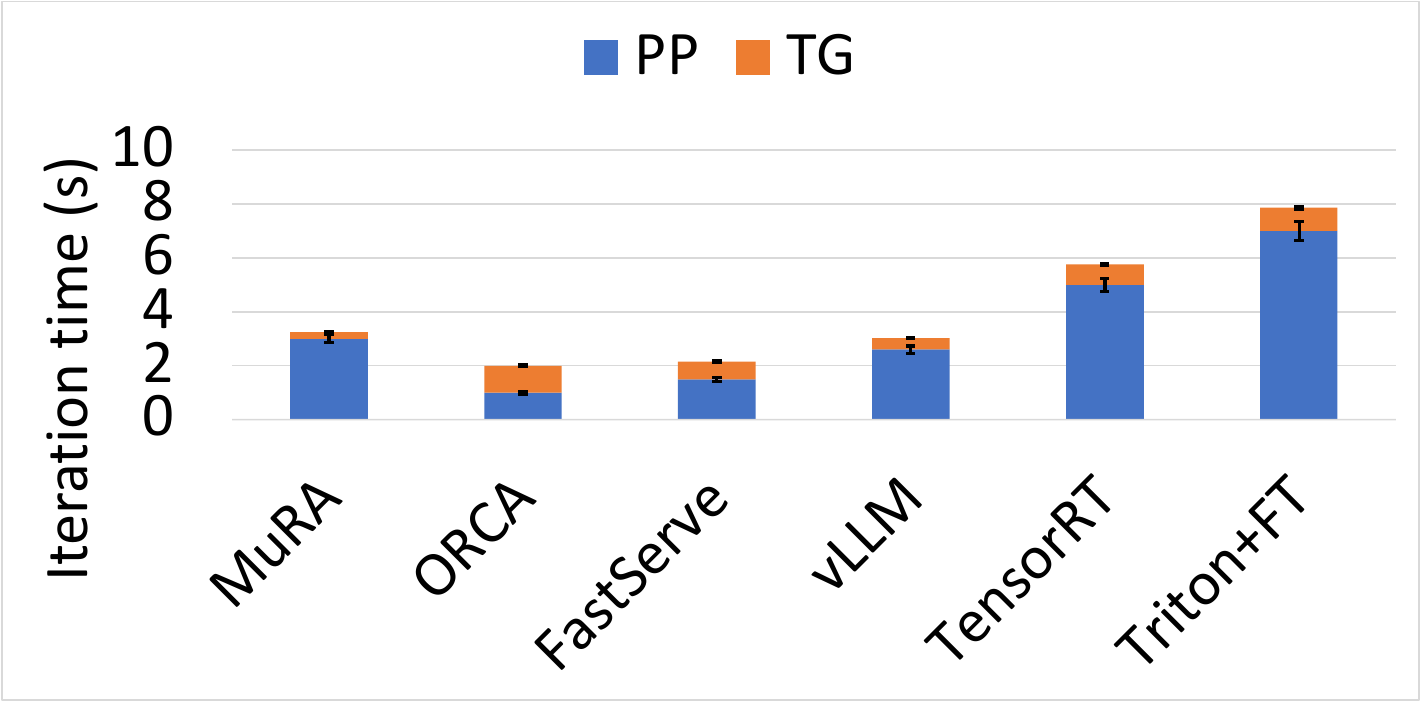} }}
    \hfill
    \subfloat[175B.\vspace{-0.01in}\label{fig:iteration-sh}]{{\includegraphics[width=0.48\linewidth,height=0.15\textheight]{Fig/iteration-time-up.pdf} }}
    \hfill
\vspace{-0.05in}
   \caption{\small{(Fake) Iteration time.\vspace{-0.1in}}}%
    \label{fig:iteration-time-overall}
\end{figure}}

\DEL{\begin{figure}[]
\centering
    \subfloat[13B.\vspace{-0.01in}\label{fig:jct-alpaca}]{{\includegraphics[width=0.48\linewidth,height=0.15\textheight]{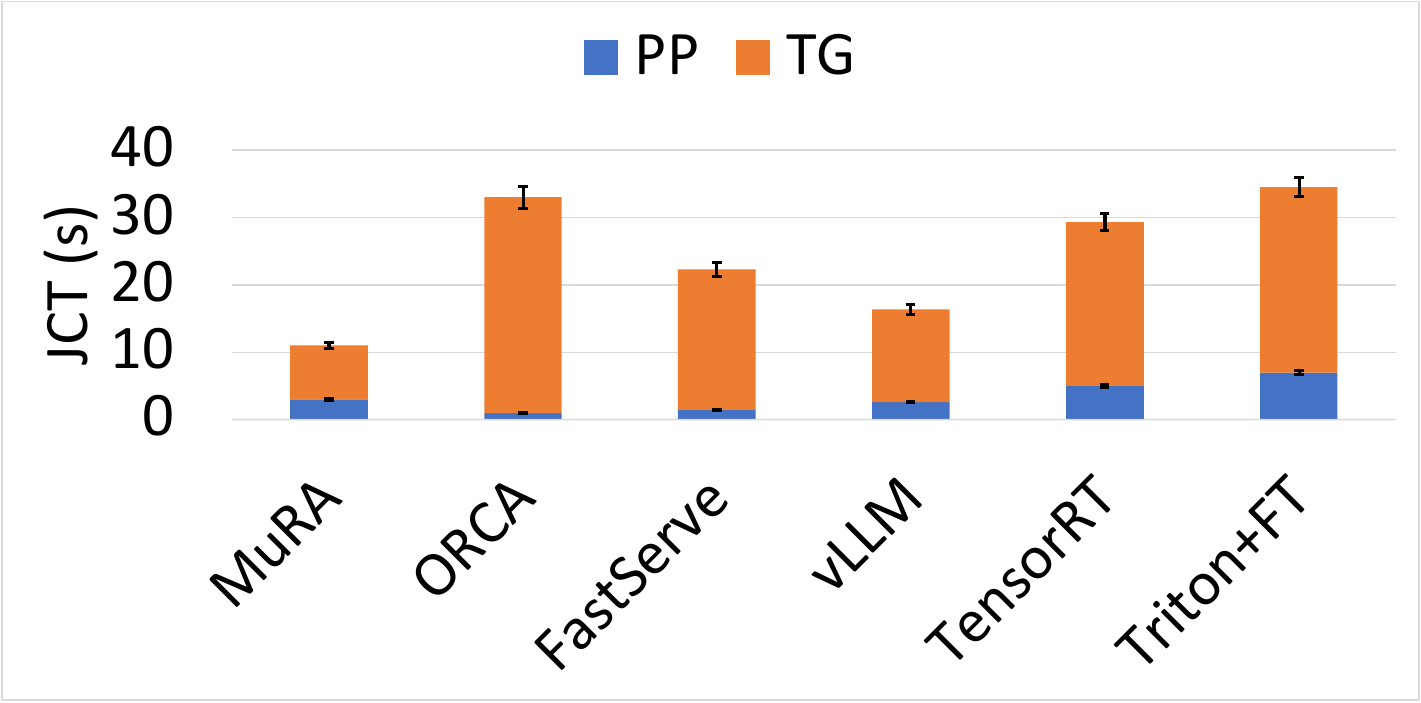} }}
    \hfill
    \subfloat[175B.\vspace{-0.01in}\label{fig:jct-sh}]{{\includegraphics[width=0.48\linewidth,height=0.15\textheight]{Fig/jct-up.pdf} }}
    \hfill
\vspace{-0.05in}
   \caption{\small{(Fake) Job Completion time.\vspace{-0.1in}}}
    \label{fig:jct-overall}
\end{figure}}

\DEL{-----------the 3 figures within ---below should be done when you are doing the experiment for the exp. section and when executing our chunking method------------
{\sh{9.excel and 10. excel. Figure~\ref{fig:9} shows the density of chunks versus cache size needed by each chunk (short prompt is considered as one chunk). Figure~\ref{fig:10} shows the density of chunks versus the chunk size in tokens. We see that the cache demand and the chunk size, i.e., GPU demand of different chunks are different. }}

{\sh{11.excel. Figure~\ref{fig:11} shows the cache needed by a chunk and the chunk size of each chunk over time. They exhibit the varience. So, after each iteration, there are many chunk options. There is a need to find chunks that more fully utilize GPU and memory resources.
}}
--------------------------------}

\noindent{\textbf{Throughput.}}
Figure~\ref{fig:th-overall} shows the throughput in tokens/s and request/s of different systems. \Sys has 1.42$\times$-11.21$\times$, 
higher throughput in tokens/s and 1.54$\times$-10$\times$ higher throughput in requests/s than other systems.  
\Orca, with iteration-level scheduling, and maximum memory allocation shows the minimum throughput. FastServe aims to handle head-of-line blocking in \Orca, thus improving its throughput. But its 
job preemption and swapping produce high latency and overhead. 
vLLM, FastGen, and Sarathi employ block-based KVC allocation to mitigate the KVC bottleneck and use chunking to enhance GPU compute utilization. However, by relying on FCFS and neglecting to optimize KVC utilization, they fail to consistently maximize GPU compute and KVC utilizations as shown in \cref{sec:analysis}. 
\Sys's SLO-guaranteed dynamic chunking method segments long prompts and batches requests with similar SLOs (facilitated by the iteration-level SLO-based task prioritization method) to fully utilize GPU while meeting the SLO requirements. 
Moreover, \Sys's multi-resource-aware batching method simultaneously considers both GPU and KV cache resources when selecting requests to fully utilize both resources, thus increasing throughput. \Sys shows 12\%  lower throughput than \emph{Oracle}.




\noindent{\textbf{Goodput and SLO attainment.}}
Figure~\ref{fig:gd-overall} shows the goodput with $T$=1s of different systems. Figure~\ref{fig:slo-overall} shows the iteration-level SLO and JCT SLO attainments of different systems. \Sys improves the goodput of other systems
by 1.43$\times$-13.71$\times$, improves their SLO attainment by 37\%-90\%, and improve their JCT SLO attainment by 34\%-93\%, 
\Sys accounts for heterogeneous SLOs in improving throughput, resulting in significantly higher goodput and improved SLO attainment, which deliver a superior user experience. \emph{Oracle} only shows 1.1\% higher SLO and JCT SLO  and 6\% higher goodput than \Sys. 


 
\noindent{\textbf{JCT.}} Figure~\ref{fig:jct-overall} shows the JCT of different systems. \Sys reduces reduces their JCT by 1.61$\times$-12.22$\times$. 
This is because \Sys's high throughput (explained in Figure~\ref{fig:th-overall}) helps reduce request waiting time. \emph{Oracle} shows 22\% lower JCT compared to \Sys.



\DEL{The iteration-level cache guarantee method ensures to satisfy the KVC demands in the next iteration, thus avoiding increasing the iteration time due to KVC overflow.

Forming resource-aware batches ensures high
GPU compute utilization, ultimately contributing to latency reduction.

?\Sys has a lower iteration time for TG because the chunking of the long prompts and the efficient batching to ensure GPU compute utilization also help to lower the iteration time for PP. Further, the iteration-level cache guarantee method ensures that the requests in the batch have enough memory to run and prevent the cases in which some requests cannot run due to  memory limit.\looseness=-1

}

\DEL{\noindent{\textbf{Iteration time.}} Figure~\ref{fig:iteration-time-overall} shows the iteration time in the PP and TG phase for all the compared methods. Compared to \Orca and FastServe, \Sys generates 4.7$\times$ and 2.6$\times$ lower PP time, and 16$\times$ and 5$\times$ lower TG time in the long-prompt scenario, and generates 4.5$\times$ and 2.3$\times$ lower PP time, and 15$\times$ and 4.5$\times$ lower TG time in the mixed-prompt scenario.
}

\DEL{\begin{figure}
    \centering
\includegraphics[width=0.48\linewidth,height=0.15\textheight]{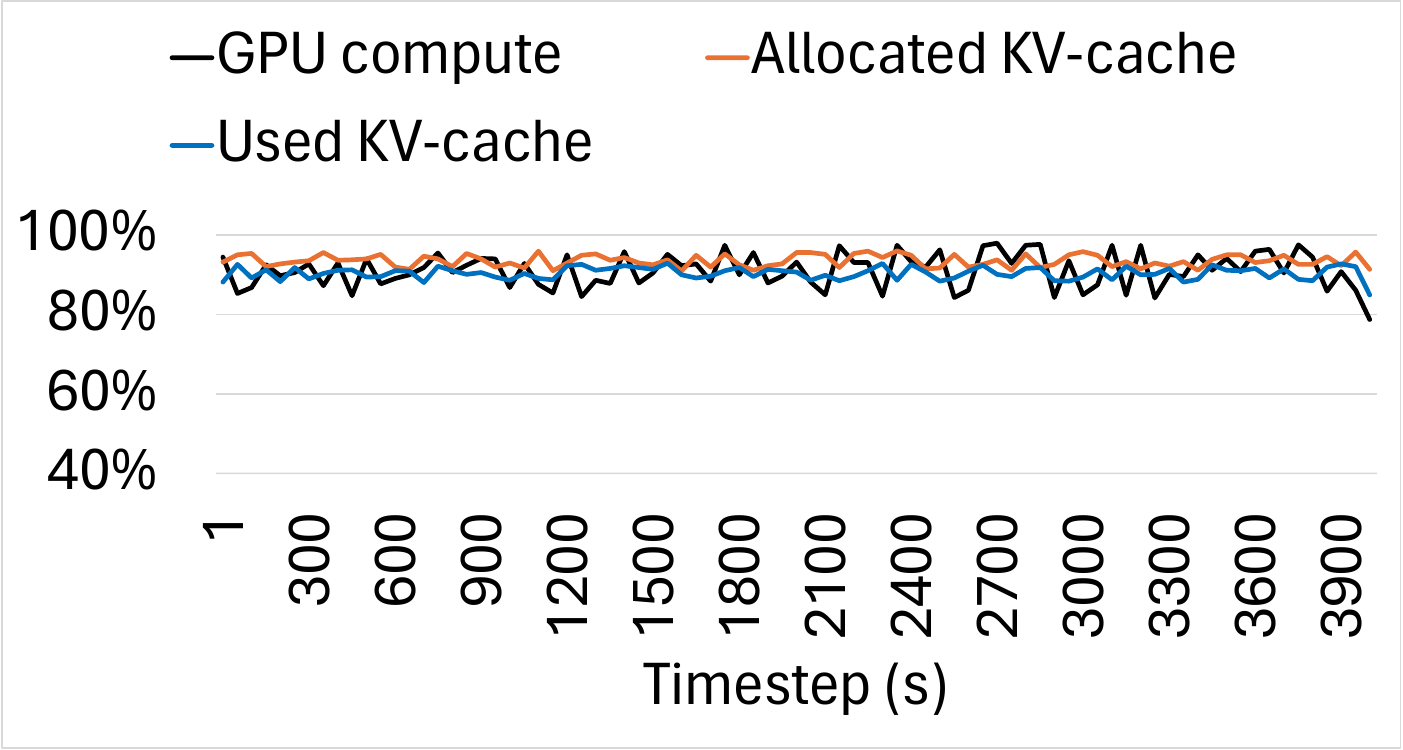}
    \caption{GPU compute and KVC resource over time in \Sys.}
    \label{fig:time-mura}
\end{figure}}

\DEL{\begin{figure}[]
\centering
\subfloat[13B.\vspace{-0.01in}\label{fig:gpu-overall}]{{\includegraphics[width=0.405\linewidth,height=0.15\textheight]{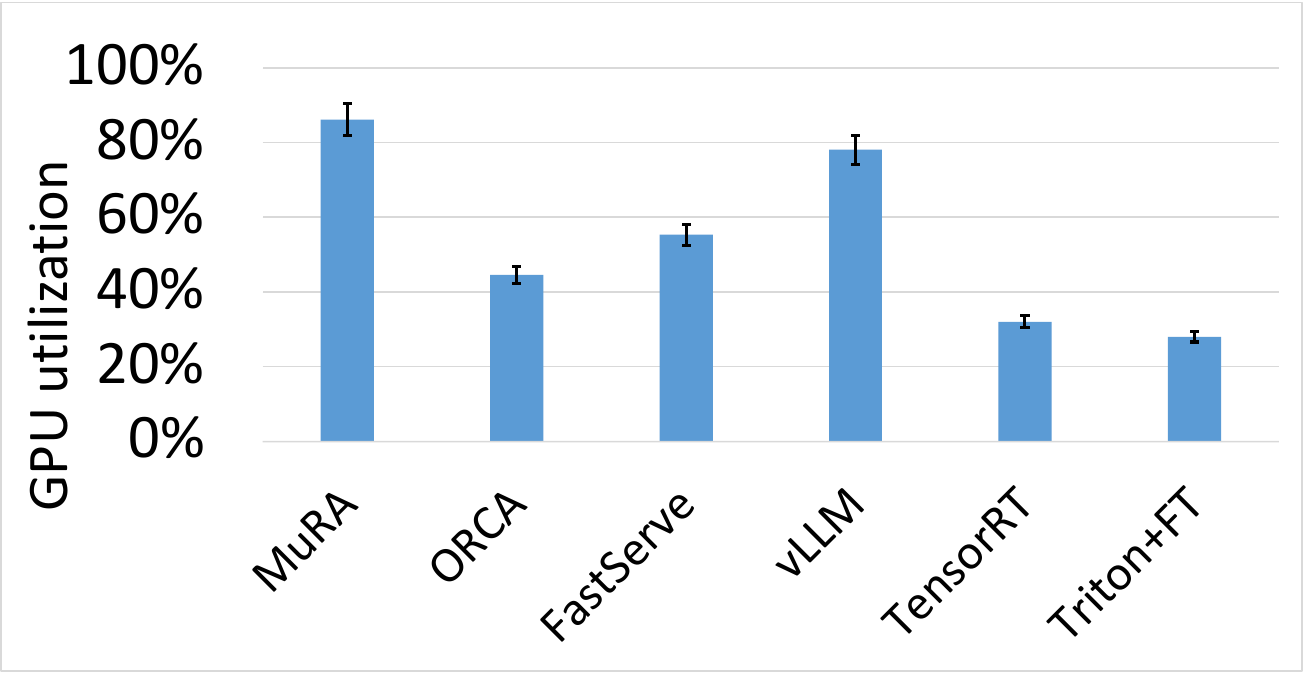} }}
    \hfill
    \subfloat[175B.\vspace{-0.01in}\label{fig:gpu-comp}]{{\includegraphics[width=0.565\linewidth,height=0.15\textheight]{Fig/gpu-overall.pdf} }}
    \hfill
\vspace{-0.05in}
   \caption{\small{(Fake) GPU compute utilization.\vspace{-0.1in}}}
    \label{fig:gpu-utilization}
\end{figure}}

\DEL{Figure~\ref{fig:gpu-overall} shows the GPU and KVC utilization. Compared to vLLM, FastServe, \Orca, ?? and ??, \Sys generates 93\% and 55\% higher GPU compute utilization for the long-prompt, and generates 75\%  and 35\%  higher GPU compute utilization for the mixed-prompt scenario, respectively.  
\Sys has the highest GPU compute utilization. Its dynamic chunking reduces the KVC requirement, which allows more GPU to be utilized.
Its resource-aware batch formation allows GPU to be fully utilized.
Its iteration-level KVC guarantee helps fully utilize the remaining cache to run requests instead of killing process due to lack of memory later. \Orca has the lowest GPU compute utilization because of the high memory allocation. FastServe mitigates this issue by swapping the KVC to the host memory, but jobs may still encounter memory overflows and also the swapping delay may reduce the GPU compute utilization.

Figure~\ref{fig:gpu-comp} shows the GPU compute utilization for the \Sys variants. Compared to \Sys, \Sys/Chunk, \Sys/Batch, \Sys/Cache and \Sys/SLO reduce the GPU compute utilization by 47\%, 42\%, 31\% and 2-3\%, respectively. The reasons of the effectiveness of each method is explained above. The iteration-level SLO methods does not have much effect on increasing the GPU compute utilization since it focuses on fulfilling SLO.\looseness=-1}

\DEL{\begin{figure}[]
\centering
    \subfloat[13B.\vspace{-0.01in}\label{fig:mem-overall}]{{\includegraphics[width=0.405\linewidth,height=0.15\textheight]{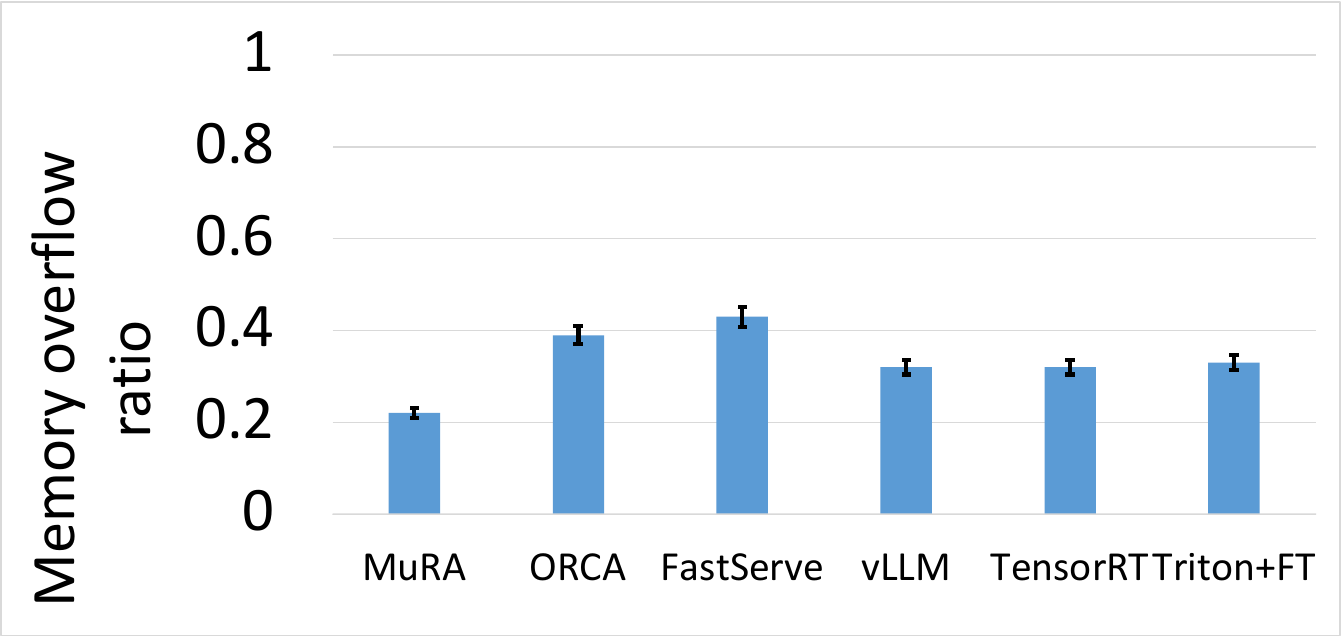} }}
    \hfill
    \subfloat[175B.\vspace{-0.01in}\label{fig:mem-comp}]{{\includegraphics[width=0.565\linewidth,height=0.15\textheight]{Fig/Memory-overall-2.pdf} }}
    \hfill
\vspace{-0.05in}
   \caption{\small{(Fake) KVC overflow ratio.\vspace{-0.1in}}}
    \label{fig:mem-overflow}
\end{figure}}

\DEL{\noindent{\textbf{KVC overflow ratio.}} 
The KVC overflow ratio represents the proportion of unsatisfied KVC demands from requests. 
For each job remaining in the queue (initially and after preemption) due to memory overflow, we counted the number of iterations it stayed in the waiting queue and calculated the ratio of this number to the total number of iterations it stayed in the system. Figure~\ref{fig:mem-overall} displays the KVC overflow ratio results. \Sys achieves 63\%, \tsr{66\%}, 78\%, 2.08$\times$, and 
lower ratio compared to Sarathi, \tsr{FastGen}, vLLM, FastServe, \Orca, TensorRT, and Triton+FT, respectively. The reasons are the same as those mentioned for Figure~\ref{fig:th-overall}. \DEL{TensorRT, Triton+FT, } \Orca and FastServe allocate KVC based on the maximum sequence length, leaving a smaller available KVC for usage. Therefore, they exhibit high KVC overflow ratios. \Orca increases throughput by using the iteration-level scheduling, so requests can complete sooner and release KVC space. FastServe preempts long-running jobs and releases their occupied large KVC space, further mitigating the overflow. 
vLLM and Sarathi use block-based KVC allocation, leaving much more KVC space. However, its employed FCFS scheduler stops picking up requests
when a long prompt at the queue head cannot be allocated. Therefore, it cannot guarantee the satisfaction of KVC demands of the long prompt and the subsequent prompts, increasing the overflow ratio. In this case, \Sys reduces the KVC requirements of long prompts by chunking or choosing other requests that can fit into the KVC, thus reducing the number of unsatisfied KVC demands from requests. Additionally, its higher throughput enables swift request completion and KVC release, reducing queuing requests and further alleviating the KVC overflow problem. \emph{Oracle shows only 5\% less KVC overflow ratio compared to \Sys.}
}


\looseness=-1


\DEL{To demonstrate sustained high resource utilization, we present the average GPU compute utilization and unallocated KVC at each time interval of 5s over time in Figure~\ref{fig:time-mura}. \Orca, TensorRT and Triton+FT 
underperform compared to vLLM as shown in~\cite{vllm} and the above, so to make the figures readable, we only include vLLM and \Sys in the figure. 
The GPU compute utilization and unallocated KVC for \Sys ranges of [82\%, 95\%] and [4\%, 9\%], respectively, while vLLM exhibit [18\%, 82\%] and [17\%, 89\%], respectively. Notably, \Sys maintains high and stable GPU compute utilization over time, in contrast to vLLM's fluctuating GPU compute utilization. \Sys significantly enhances the GPU compute utilization of vLLM and reduces unallocated KVC.}

\noindent{\textbf{GPU and KVC utilization.}} We present the average GPU compute utilization and unallocated KVC for all systems in Figure~\ref{fig:time-mura}. \Sys shows 31\%-63\% higher GPU compute utilization than other systems. Moreover, \Sys shows 7.40\%-13.20\% lower unallocated KVC compared to other systems. This is because \Sys jointly optimizes the utilization of both resources. Selecting the request that better utilizes both resources reduces instances where one resource is insufficient for a request, thereby increasing overall GPU compute utilization. Oracle shows 2\% higher GPU compute utilization and 1.5\% lower unallocated KVC compared to \Sys.



\DEL{\begin{figure}[]
\centering
\subfloat[13B.\vspace{-0.01in}\label{fig:overhead-all}]{{\includegraphics[width=0.41\linewidth,height=0.15\textheight]{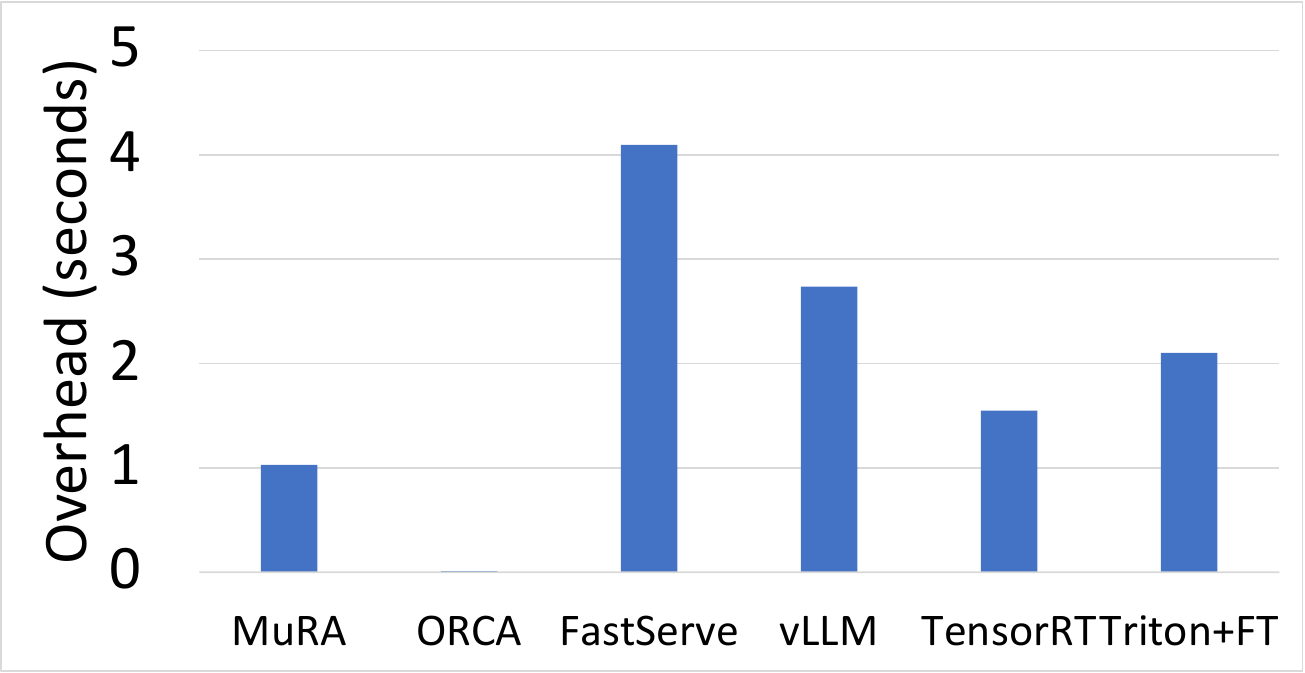} }}
    \hfill
    \subfloat[175B.\vspace{-0.01in}\label{fig:overhead-components}]{{\includegraphics[width=0.56\linewidth,height=0.15\textheight]{Fig/overhead-overall-new.pdf} }}
    \hfill
\vspace{-0.05in}
   \caption{\small{(Fake) Overhead.\vspace{-0.1in}}}
    \label{fig:overhead}
\end{figure}}
\noindent{\textbf{Time overhead.}} Figure~\ref{fig:overhead-all} shows the average time overhead per iteration for each system. Compared to \Sys, FastServe has 95\% higher time overhead, while other systems have 28\%-93\% lower time overhead. FastServe has a higher time overhead because it makes decisions on preemption and memory swapping. Other systems simply use FCFS. 
\Sys, building upon vLLM, introduces additional methods that contribute to slightly longer time overhead. Despite this, \Sys achieves a significant reduction of  1.61$\times$-12.22$\times$ in JCT compared to these methods, highlighting the effectiveness of \Sys's approaches. Notably, \Sys's time overhead remains 0.02\% of the average JCT.

\DEL{\begin{figure}[]
\centering
\subfloat[In a batch.\vspace{-0.01in}\label{fig:batch-all}]{{\includegraphics[width=0.41\linewidth,height=0.15\textheight]{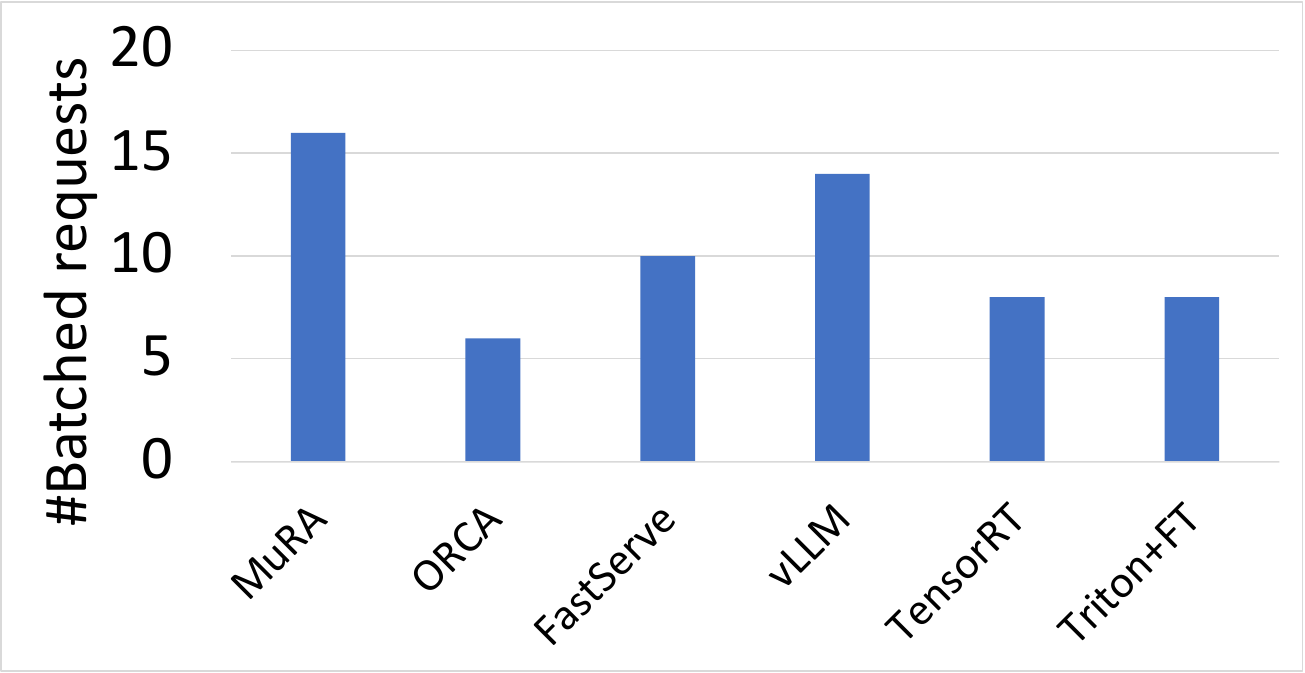} }}
    \hfill
    \subfloat[OOM.\vspace{-0.01in}\label{fig:batch-components}]{{\includegraphics[width=0.56\linewidth,height=0.15\textheight]{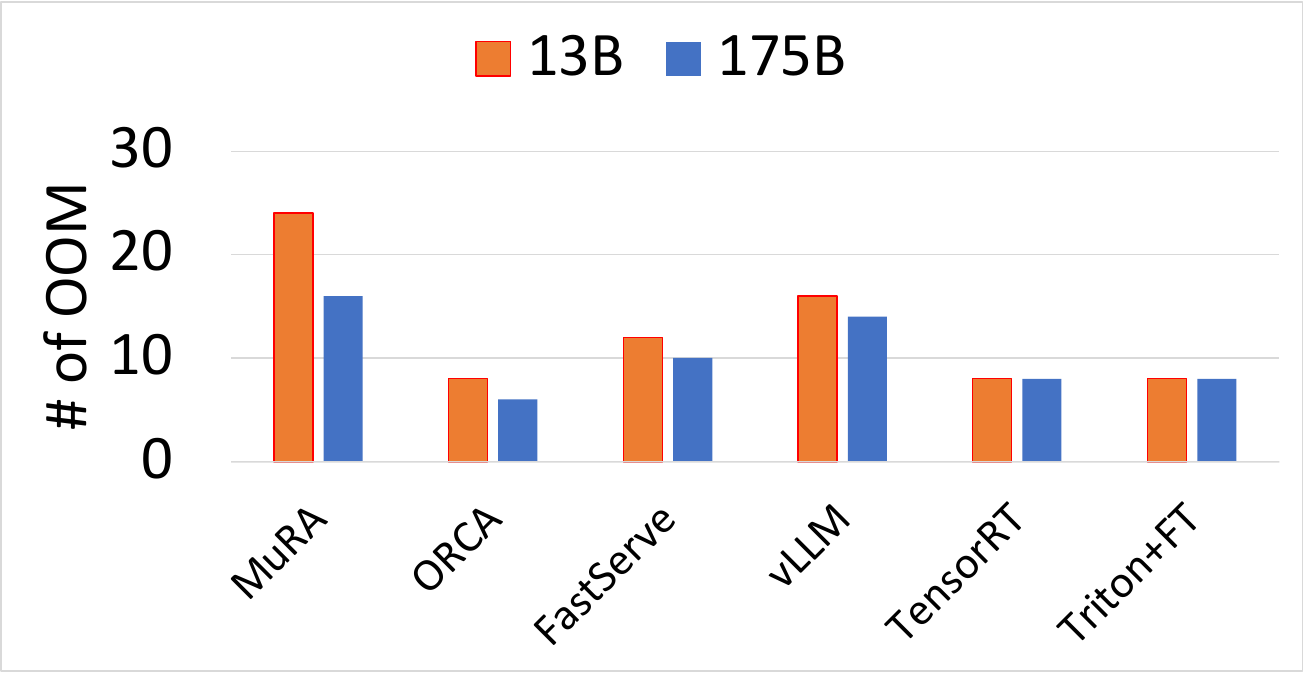} }}
    \hfill
\vspace{-0.05in}
   \caption{\small{(Fake) Number of requests.\vspace{-0.1in}}}
    \label{fig:overhead}
\end{figure}}

\DEL{\noindent{\textbf{Scalibility.}}
Figure~\ref{fig:fw1} and Figure~\ref{fig:fw2} show the impact of the forward size 
on the goodput, GPU compute utilization, JCT and throughput (in TFLOP/s). %
As the forward size increases, the GPU compute utilization, goodput and throughput keep increasing up to a certain point (1024), and then remain the same. 
The JCT keeps increasing with the forward size. \Sys is capable of running more than 4$\times$ longer forward size than \Orca as we see in Figure~\ref{fig:throughput}. Moreover, although JCT keeps increasing, the goodput doesn't drop because of the task prioritization. The results show that \Sys has high scalability.}


\DEL{The above results verify the effectiveness of \Sys's methods in dealing with long prompts and reducing JCT. 
FastServe outperforms \Orca because of it can solve the head-of-the-blocking problem. However, it still shows lower JCT performance than \Sys because of processing long prompts and failing to fully utilize resources. }

All systems perform better in OPT-175B than in OPT-13B due to the model parallelization. The parallelization allows memory-bound LLM models to access more available memory across eight GPUs, resulting in improved performance. However, the overall improvement for \Sys is lower for the OPT-175B model compared to the OPT-13B model because of more available KVC for the OPT-175B model. All systems perform similarly on Llama-13B and Llama-3-70B, and \Sys shows comparable performance improvements on both models due to the lower degree of parallelism. 




\DEL{\begin{figure}[]
\centering
    \subfloat[13B.\vspace{-0.01in}\label{fig:iteration-alpaca}]{{\includegraphics[width=0.48\linewidth,height=0.15\textheight]{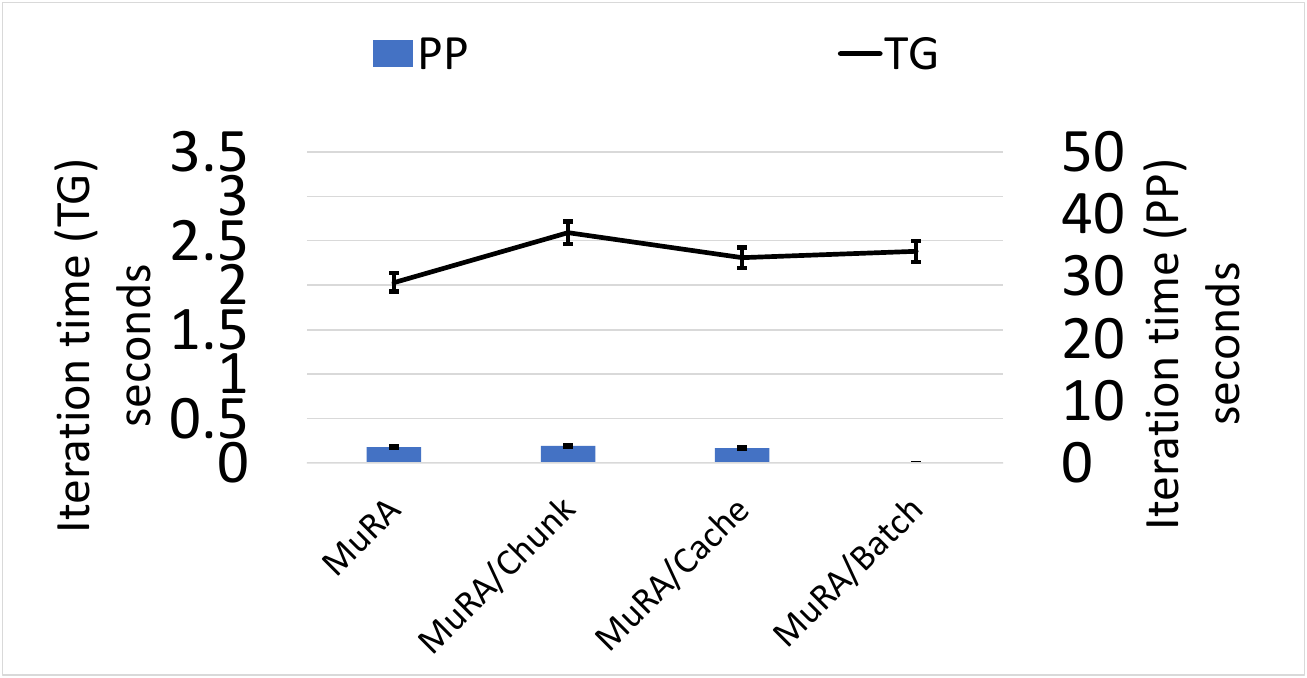} }}
    \hfill
    \subfloat[175B.\vspace{-0.01in}\label{fig:iteration-sh}]{{\includegraphics[width=0.48\linewidth,height=0.15\textheight]{Fig/iteration-time-componets} }}
    \hfill
\vspace{-0.05in}
   \caption{\small{(Fake) Iteration time of \Sys without a method.\vspace{-0.1in}}}
    \label{fig:iteration-time-comp}
\end{figure}}

\DEL{\begin{figure}[]
\centering
    \subfloat[Utilization and goodput.\vspace{-0.01in}\label{fig:fw1}]{{\includegraphics[width=0.48\linewidth,height=0.15\textheight]{Fig/fw-2.png} }}
    \hfill
    \subfloat[JCT and throughput.\vspace{-0.01in}\label{fig:fw2}]{{\includegraphics[width=0.48\linewidth,height=0.15\textheight]{Fig/scalability-th.png} }}
   \hfill
\vspace{-0.05in}
   \caption{\small{Scalability testing (to be deleted).\vspace{-0.1in}}}
    \label{fig:forward-size}
\end{figure}}

\DEL{\begin{figure*}[]
\centering
\subfloat[Iteration time.\vspace{-0.01in}\label{fig:iteration-comp}]{{\includegraphics[width=0.24\linewidth,height=0.15\textheight]{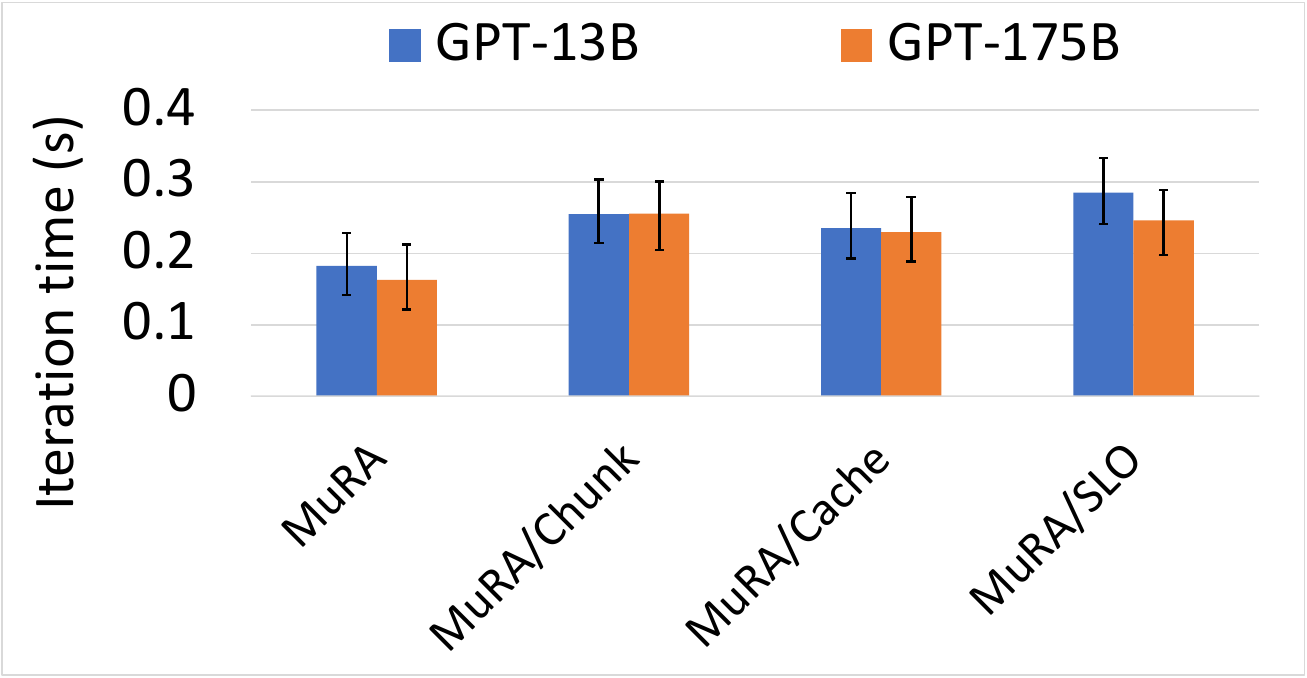} }}
    \hfill
    \subfloat[JCT.\vspace{-0.01in}\label{fig:jct-comp}]{{\includegraphics[width=0.24\linewidth,height=0.15\textheight]{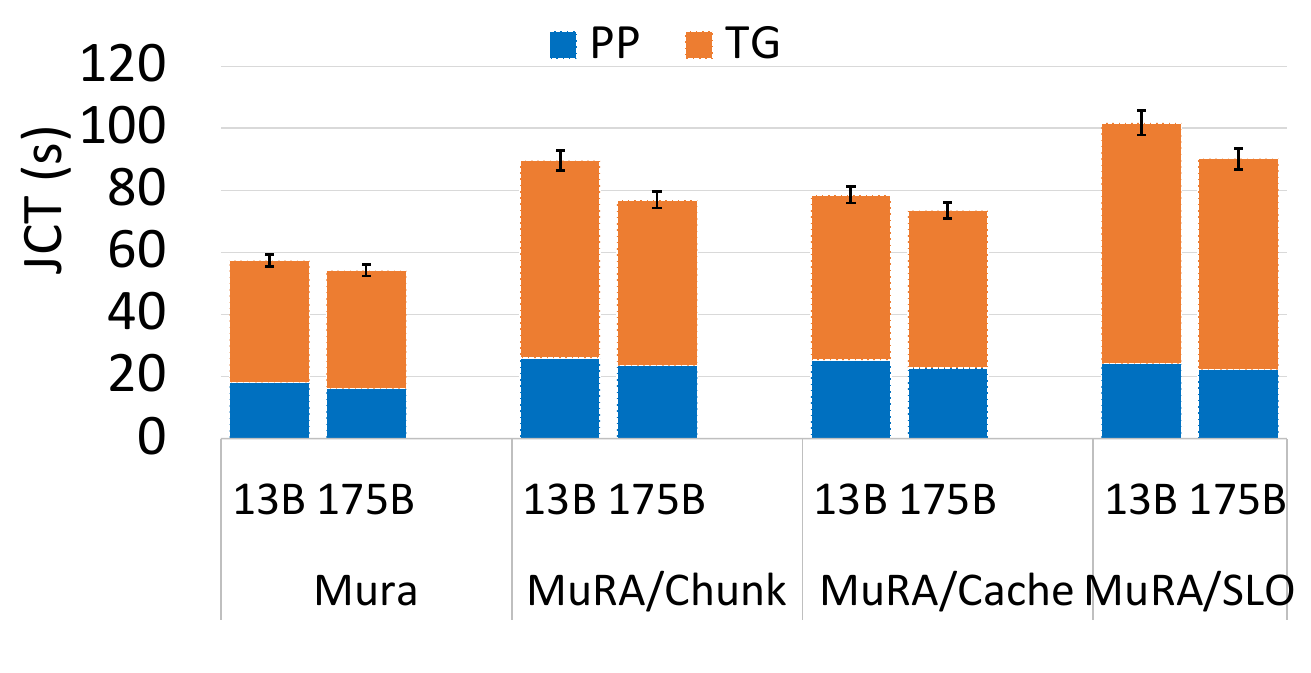} }}
    \hfill
    \subfloat[Throughput.\vspace{-0.01in}\label{fig:th-comp}]{{\includegraphics[width=0.24\linewidth,height=0.15\textheight]{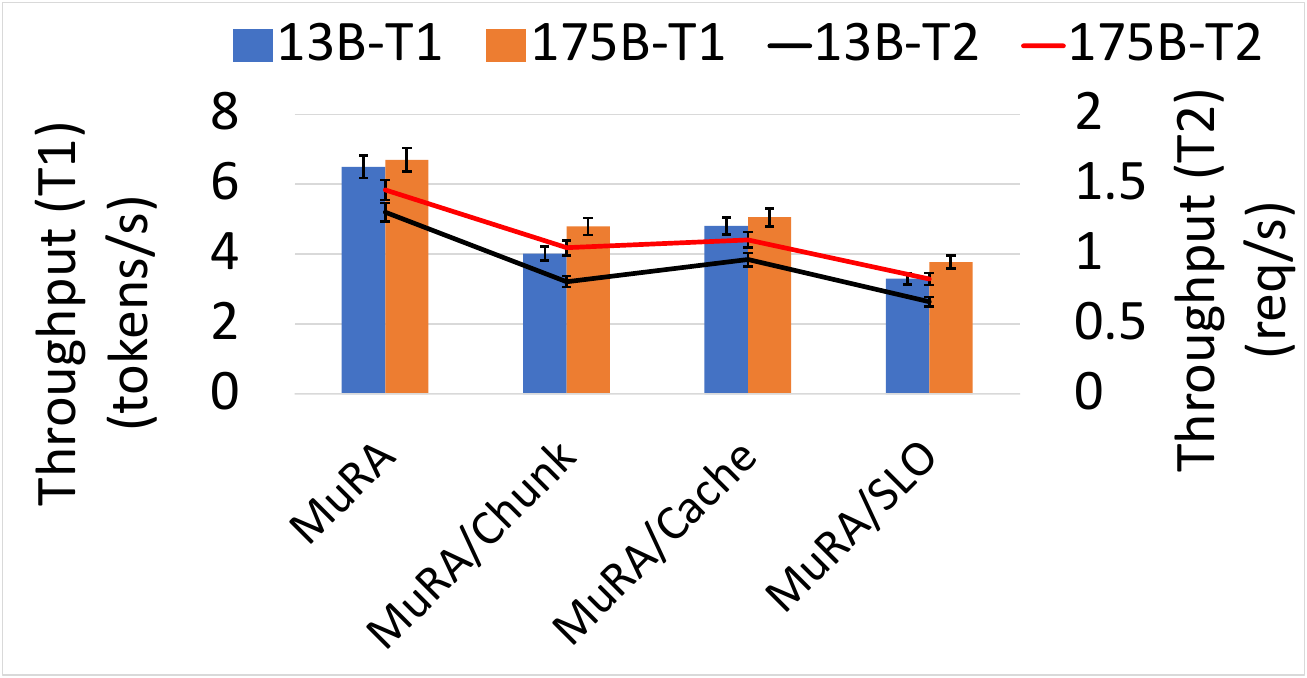} }}
    \hfill
    \subfloat[Goodput.\vspace{-0.01in}\label{fig:goodput-comp}]{{\includegraphics[width=0.24\linewidth,height=0.15\textheight]{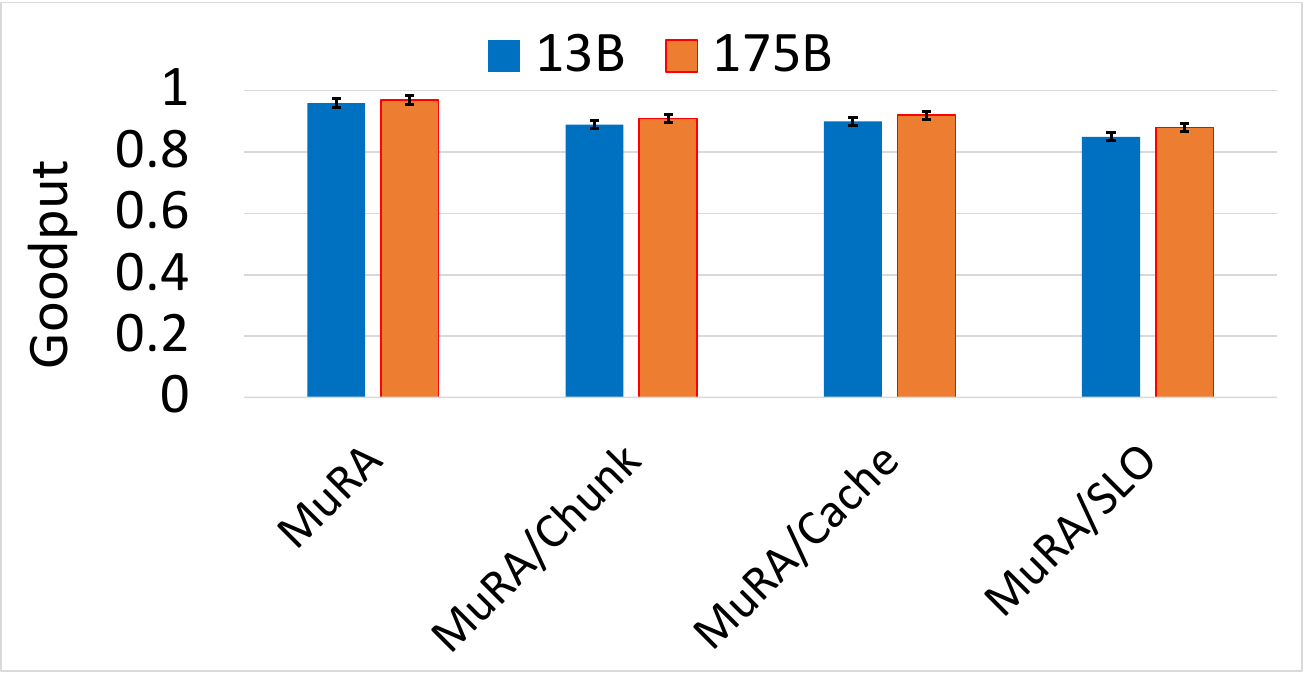} }}
    \hfill
    \subfloat[GPU compute utilization.\vspace{-0.01in}\label{fig:gpu-comp}]{{\includegraphics[width=0.24\linewidth,height=0.15\textheight]{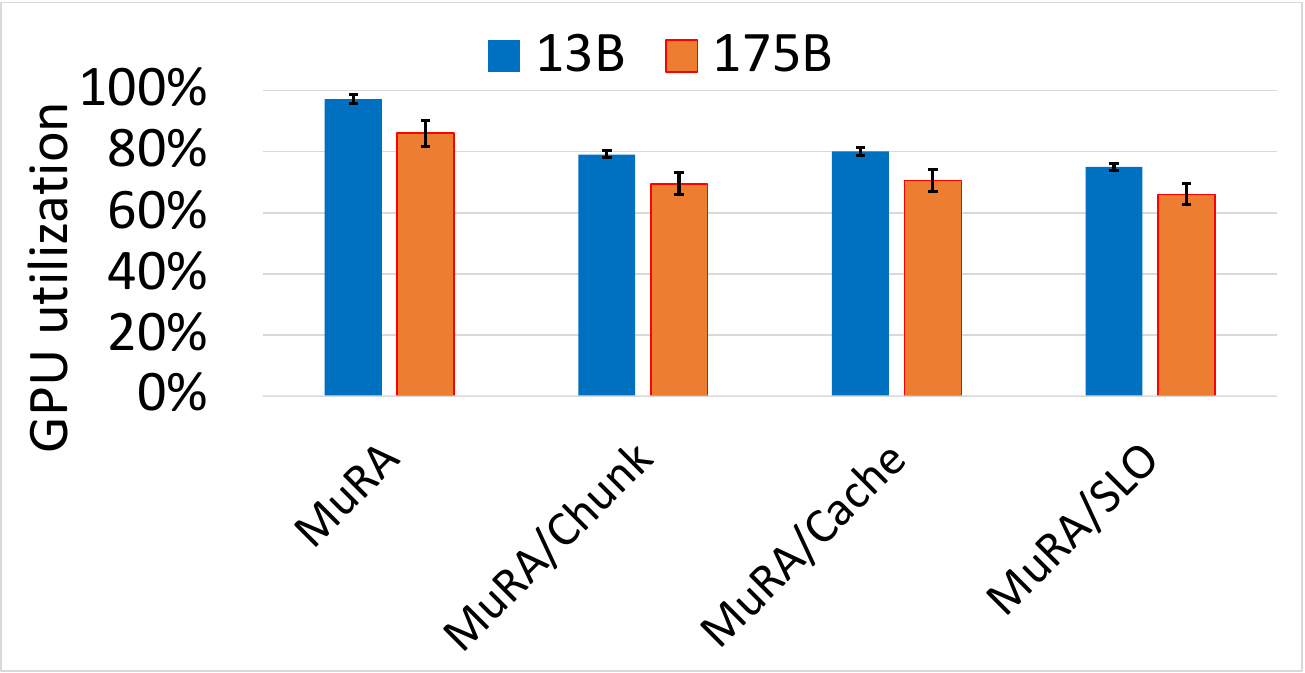} }}
    \hfill
    \subfloat[KVC overflow ratio.\vspace{-0.01in}\label{fig:mem-comp}]{{\includegraphics[width=0.24\linewidth,height=0.15\textheight]{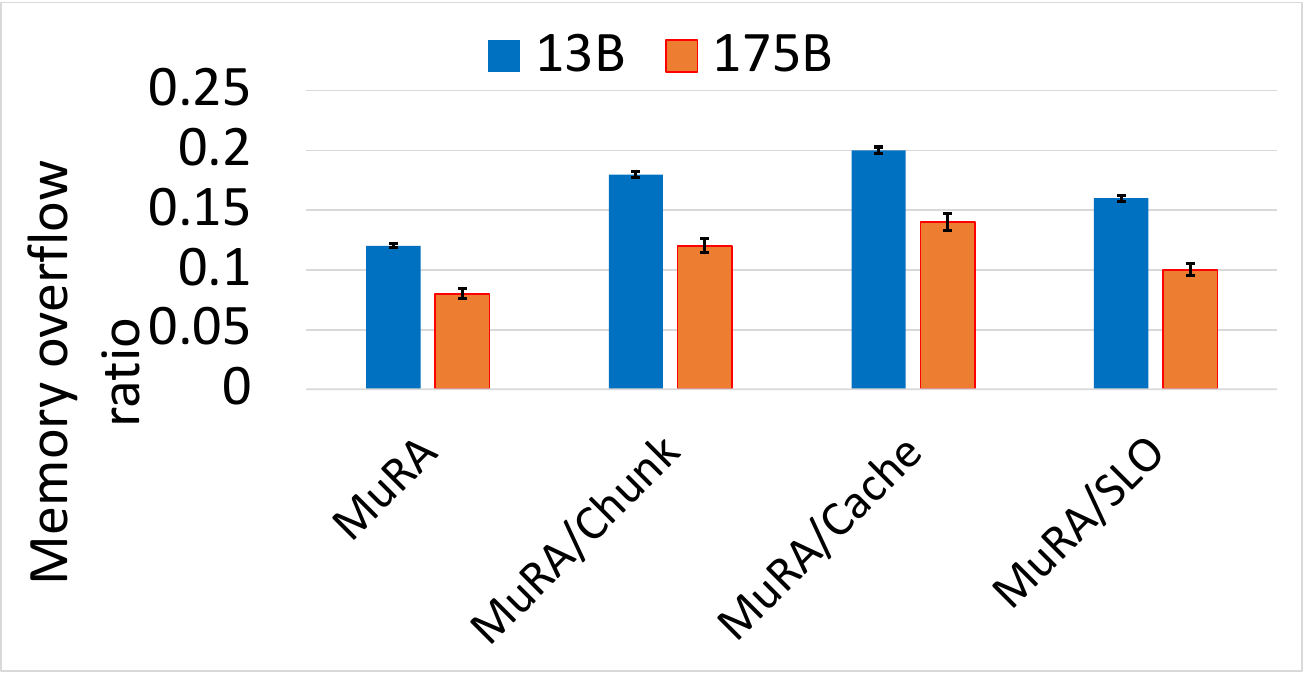} }}
    \hfill
    \subfloat[Time overhead. \label{fig:low_cpu1}]{{\includegraphics[width=0.24\linewidth,height=0.15\textheight]{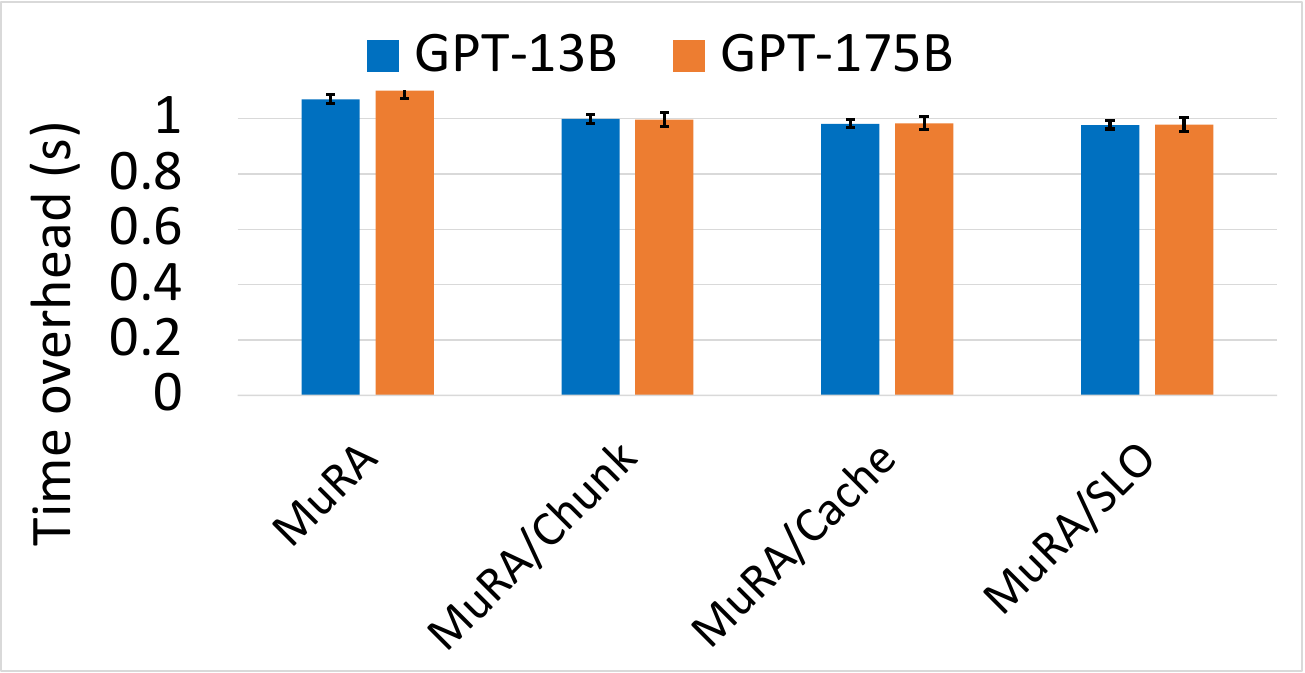} }}%
   \subfloat[Batched requests.\label{fig:low_bw1}]{{\includegraphics[width=0.24\linewidth,height=0.15\textheight]{Fig/batch-mura-comp.pdf} }}%
\vspace{-0.05in}
   \caption{\small{Results of \Sys without a method for both models.\vspace{-0.1in}}}
    \label{fig:met-comp}
\end{figure*}}

\begin{figure*}[]
\centering
\subfloat[Throughput.\vspace{-0.01in}\label{fig:th-comp}]{{\includegraphics[width=0.32\linewidth,height=0.15\textheight]{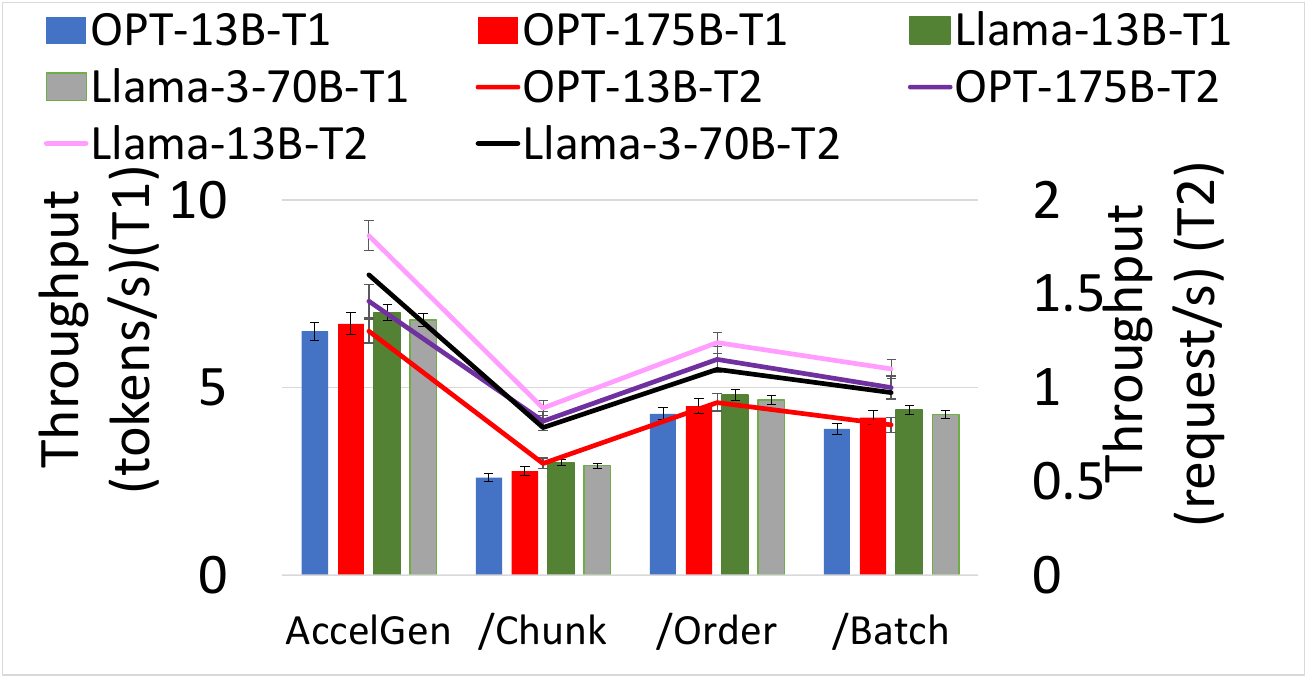} }}
    \hfill
\subfloat[Goodput.\vspace{-0.01in}\label{fig:goodput-comp}]{{\includegraphics[width=0.32\linewidth,height=0.15\textheight]{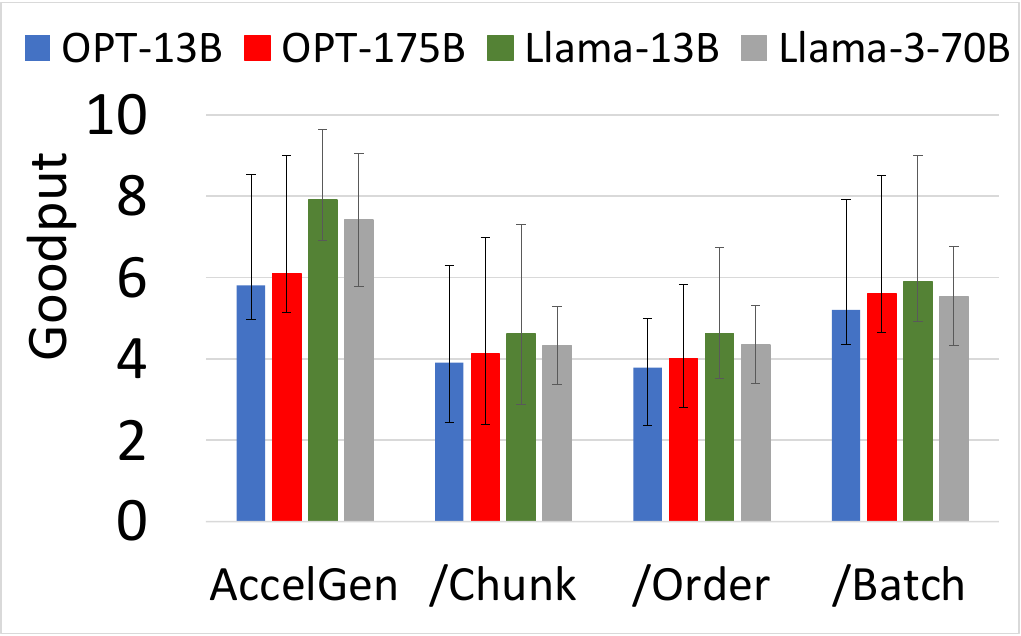} }}
    \hfill
\subfloat[SLO attainment.\vspace{-0.01in}\label{fig:slo-comp}]{{\includegraphics[width=0.32\linewidth,height=0.15\textheight]{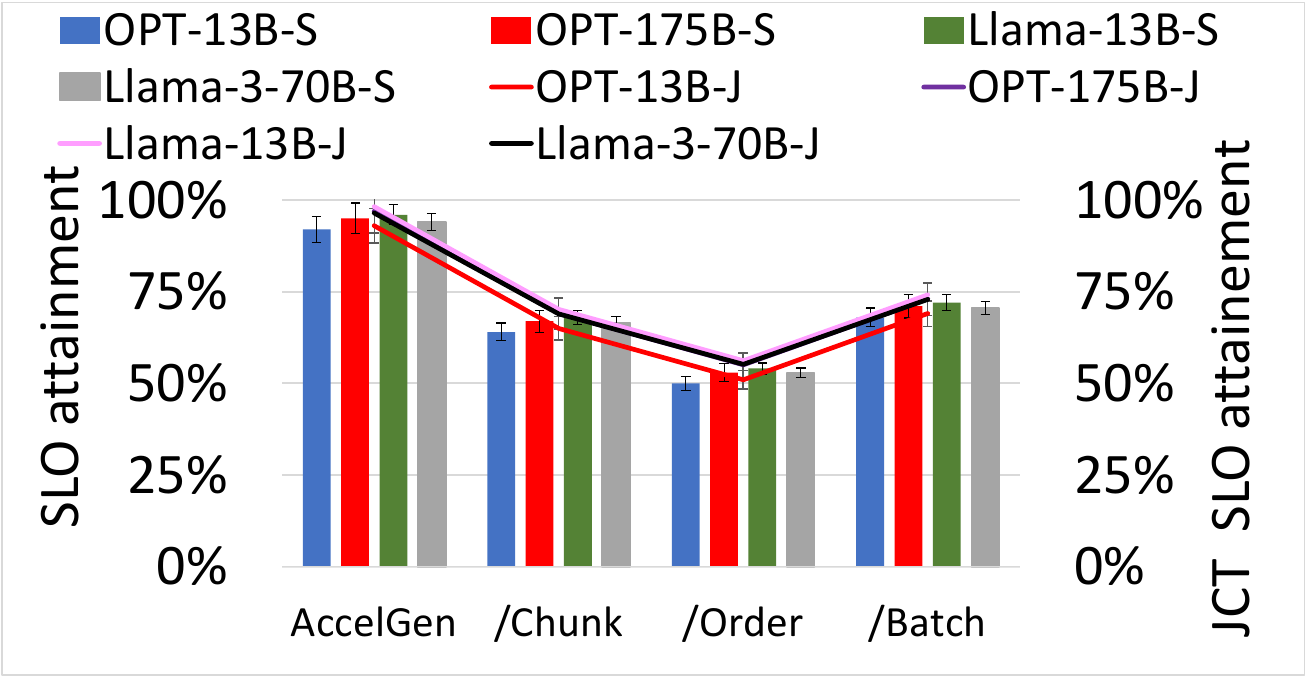} }}
    \hfill
\DEL{\subfloat[Iteration time.\vspace{-0.01in}\label{fig:iteration-time-comp}]{{\includegraphics[width=0.24\linewidth,height=0.15\textheight]{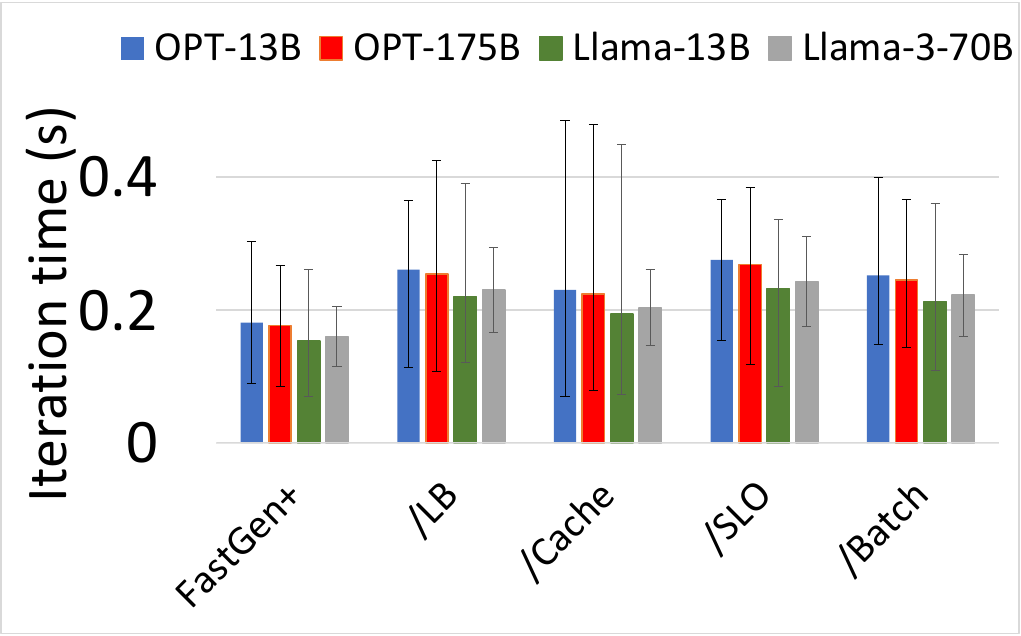} }}
\hfill}
\subfloat[JCT.\vspace{-0.01in}\label{fig:jct-comp}]{{\includegraphics[width=0.32\linewidth,height=0.15\textheight]{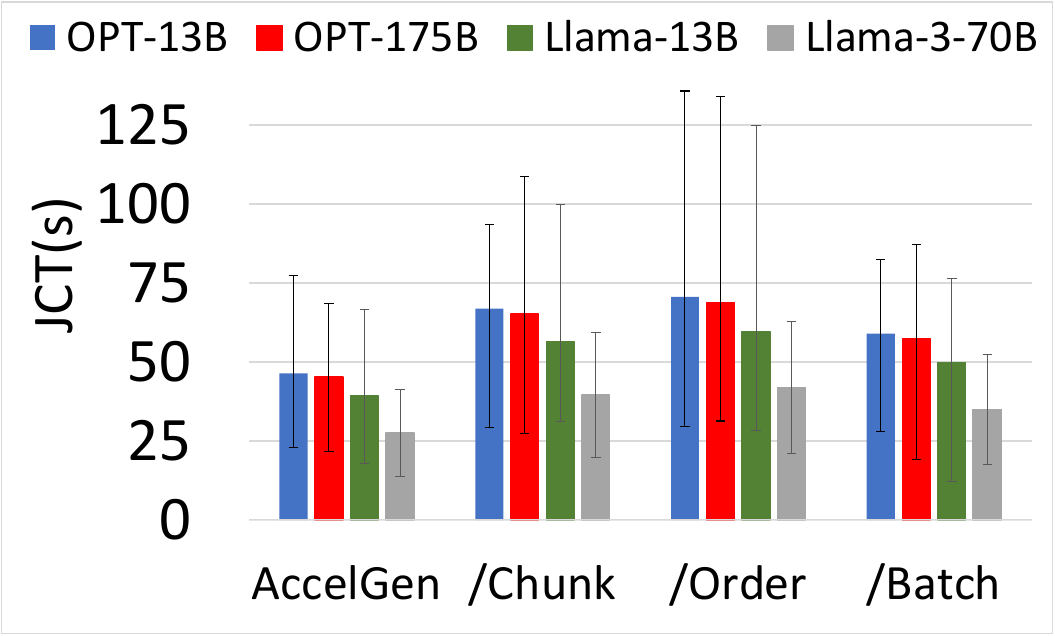} }}
\hfill
\DEL{\subfloat[KVC overflow ratio.\vspace{-0.01in}\label{fig:mem-comp}]{{\includegraphics[width=0.24\linewidth,height=0.15\textheight]{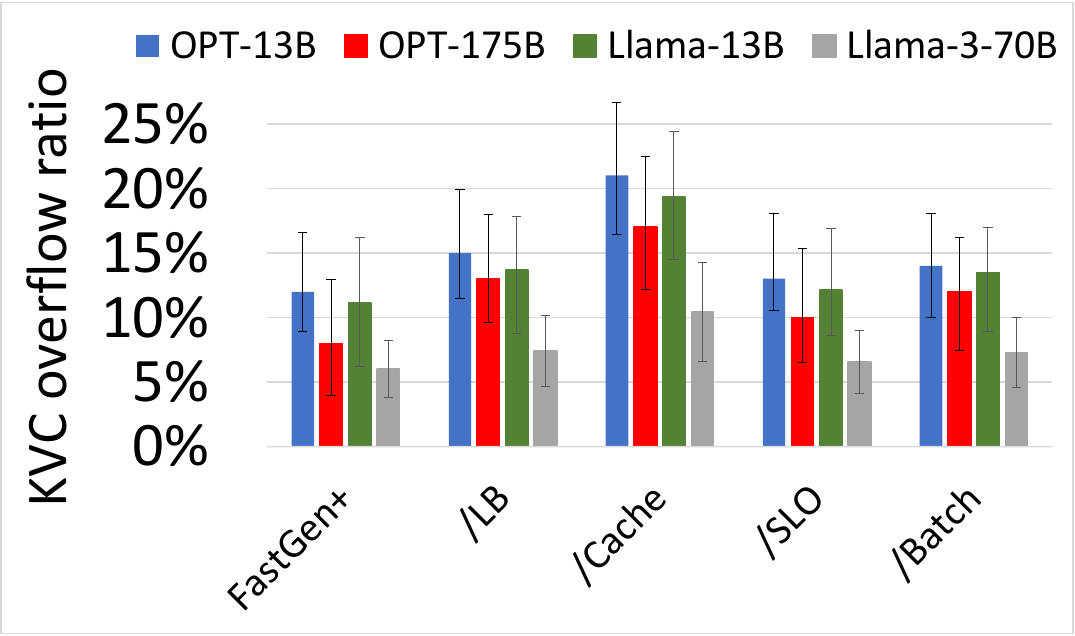} }}
    \hfill}
    \subfloat[Resource utilization.\vspace{-0.01in}\label{fig:th-gpu-comp}]{{\includegraphics[width=0.32\linewidth,height=0.15\textheight]{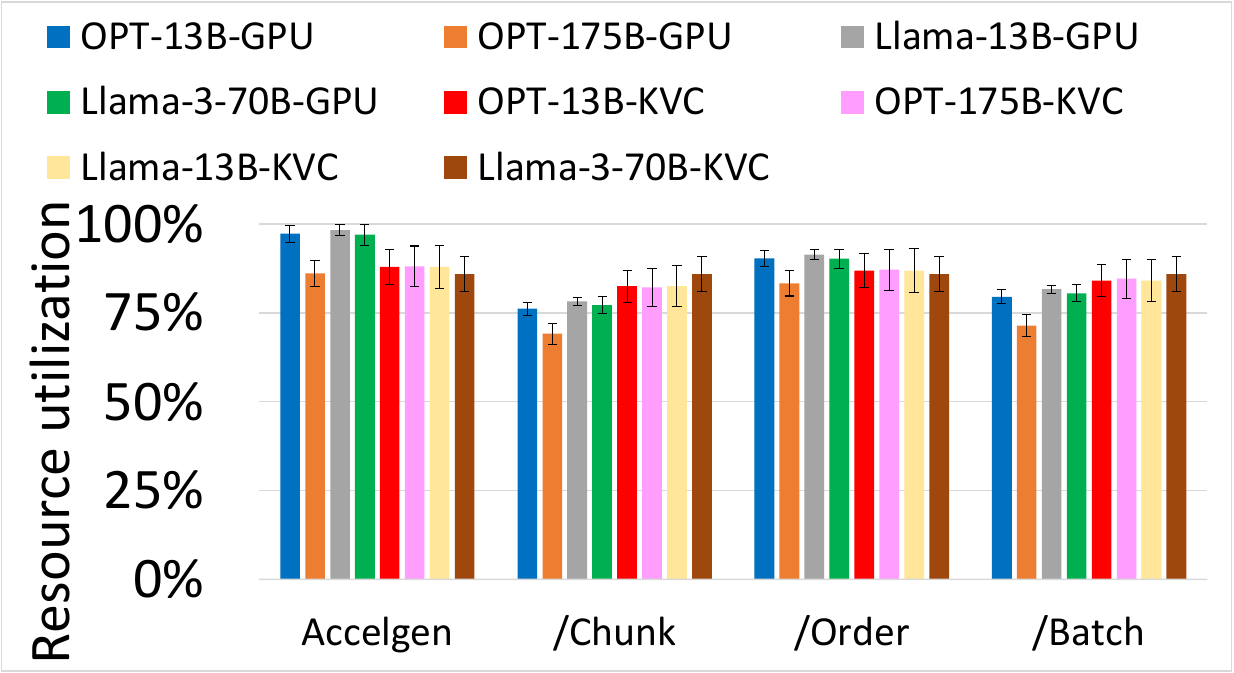} }}
    \hfill
   \subfloat[Time overhead. \label{fig:overhead-comp}]{{\includegraphics[width=0.32\linewidth,height=0.15\textheight]{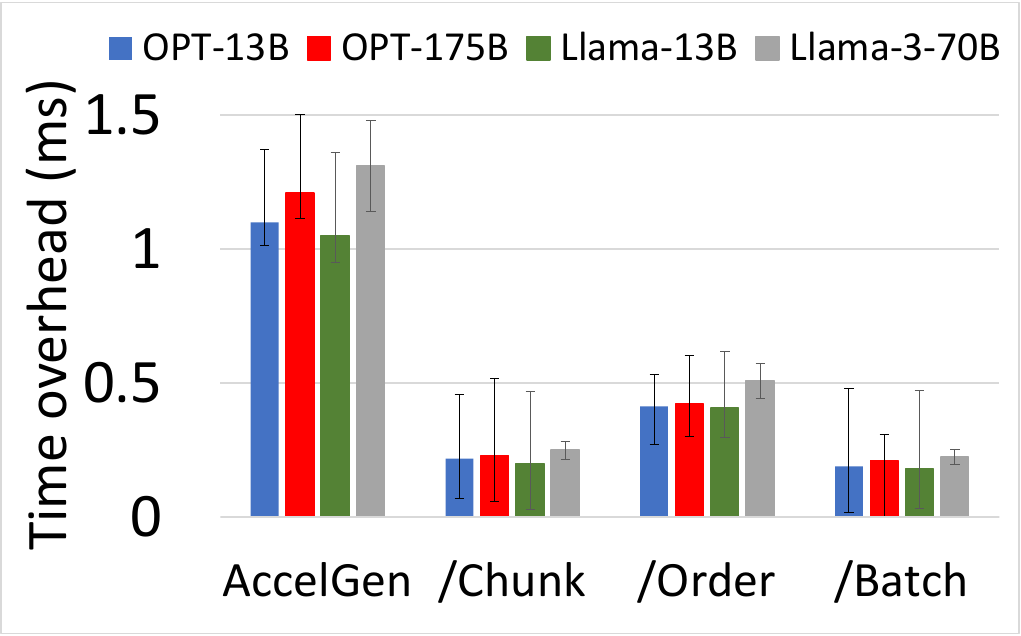} }}
    \hfill
\vspace{-0.05in}
   \caption{Results of \Sys without a method for different models. \vspace{-0.1in}}
    \label{fig:met-comp}
\end{figure*}

\subsection{Ablation Testing}

We use \emph{Chunk}, \emph{Order}, \emph{Batch} to represent the three methods in sequence in \Sys, respectively. We use /Chunk, /Order, and /Batch to represent \Sys without each method, respectively. 


Figure~\ref{fig:th-comp} illustrates the throughput of individual variants of \Sys. Notably, /Chunk has the most substantial effect, reducing throughput by 97\% since \emph{Chunk} leverages long prompts to mitigate their negative impact and enhance throughput. 
/Batch has the second-highest impact (61\%), as \emph{Batch} aims fully utilizing both resources. /Order exhibits the least impact, reducing throughput by 35\%, as \emph{Order} primarily focuses on ensuring that each request is completed within a designated time limit.

Figures~\ref{fig:goodput-comp}, and~\ref{fig:slo-comp} show /Order, /Chunk, /Batch decrease the goodput of \Sys by 52\%, 44\%, and 27\%, respectively, reduce \Sys's SLO attainment by 42\%, 28\%, and 24\%, respectively, and they increase \Sys's iteration time by 56\%,  39\% and 29\%, respectively. \emph{Order} has the highest impact as it helps to satisfy the SLOs. \emph{Chunk} has the second impact because it significantly improves GPU compute utilization and throughput while satisfying SLOs. \emph{Batch} has lesser impacts as it aids in more fully utilizing GPU compute and KVC resources simultaneously.

Figure~\ref{fig:jct-comp} depicts the JCT of each variant. /Order, /Chunk, and /Batch increase JCT by 17\%, 31\%, and 19\%, respectively. \emph{Chunk} stands out with the highest impact, followed by \emph{Batch}.

\DEL{Figure~\ref{fig:mem-comp} highlights the KVC overflow ratio impact, with \emph{Cache} demonstrating the highest impact (15\%) by guaranteeing full KVC allocation. 
/Order, /Chunk and /Batch increase the KVC overflow ratio by 3\%, 5\% and 6\%. \emph{Chunk}, by reducing KVC requirements from long prompts, and \emph{Batch}, by simultaneously considering GPU and KVC resources, help improve throughput and reduce memory overflows. \emph{SLO} has the least impact as it focuses primarily on SLO satisfaction.
}

In Figure~\ref{fig:th-gpu-comp}, /Order, /Chunk, and /Batch decrease GPU compute utilization by 8\%, 28\%, and 15\%, respectively. Similarly, they decrease KVC utilization by 1\%, 7\%, and 11\%, respectively. \emph{Chunk} is most effective at fully utilizing both resources, leveraging long prompts, while \emph{Batch} contributes significantly to resource utilization by finding requests that fully utilize both GPU and KVC resources simultaneously. \emph{Order} has the least impact, focusing solely on SLO considerations.\looseness=-1

Figure~\ref{fig:overhead-comp} shows the time overhead for \Sys variants. /Order, /Chunk and /Batch have 63\%, 81\%, and 82\% lower overhead than \Sys. \emph{Order} generates lower time overhead than others, as it primarily involves ordering requests in the queue. 
\emph{Chunk} and \emph{Batch} generate higher time overhead due to their more complex operations.

\DEL{The order of the overheads are as follows:/Batch</Cache</Chunk</Order<\Sys.
\emph{Batch} has the highest overhead because it needs to search requests considering both GPU and KVC resources. 

/Batch has the lowest overhead because the batching technique needs to search for the requests iteratively that utilize the GPU and memory efficiently. Cache finds requests from the waiting queue whose KVC demands can be met. \emph{Chunk} only needs to compute the token budget and fetch requests, and \emph{SLO} orders requests according to their SLOs, so they need a shorter time than /Batch and /Cache.

The results also show the effectiveness of individual methods in \Sys. The iteration-level SLO guarantee method and chunking method help reduce JCT and improve the JCT SLO guarantee the most. 
Batching helps by forming batches that ensure high GPU compute utilization, which ultimately reduces the job latency. /Cache's effectiveness is the least because it only makes sure that
a request won't be selected for a batch if its KVC demand cannot be satisfied instead of directly reducing the latency.
}

\DEL{The order of the overheads are as follows:/Batch</Cache</Chunk</Order<\Sys.
\emph{Batch} has the highest overhead because it needs to search requests considering both GPU and KVC resources. 

/Batch has the lowest overhead because the batching technique needs to search for the requests iteratively that utilize the GPU and memory efficiently. Cache finds requests from the waiting queue whose KVC demands can be met. \emph{Chunk} only needs to compute the token budget and fetch requests, and \emph{SLO} orders requests according to their SLOs, so they need a shorter time than /Batch and /Cache.

The results also show the effectiveness of individual methods in \Sys. The iteration-level SLO guarantee method and chunking method help reduce JCT and improve the JCT SLO guarantee the most. 
Batching helps by forming batches that ensure high GPU compute utilization, which ultimately reduces the job latency. /Cache's effectiveness is the least because it only makes sure that
a request won't be selected for a batch if its KVC demand cannot be satisfied instead of directly reducing the latency.
}


\DEL{\begin{figure*}[]
\centering
\subfloat[Iteration time.\vspace{-0.01in}\label{fig:jct}]{{\includegraphics[width=0.24\linewidth,height=0.15\textheight]{Fig/iteration-time-componets} }}
    \hfill
    \subfloat[JCT.\vspace{-0.01in}\label{fig:jct}]{{\includegraphics[width=0.24\linewidth,height=0.15\textheight]{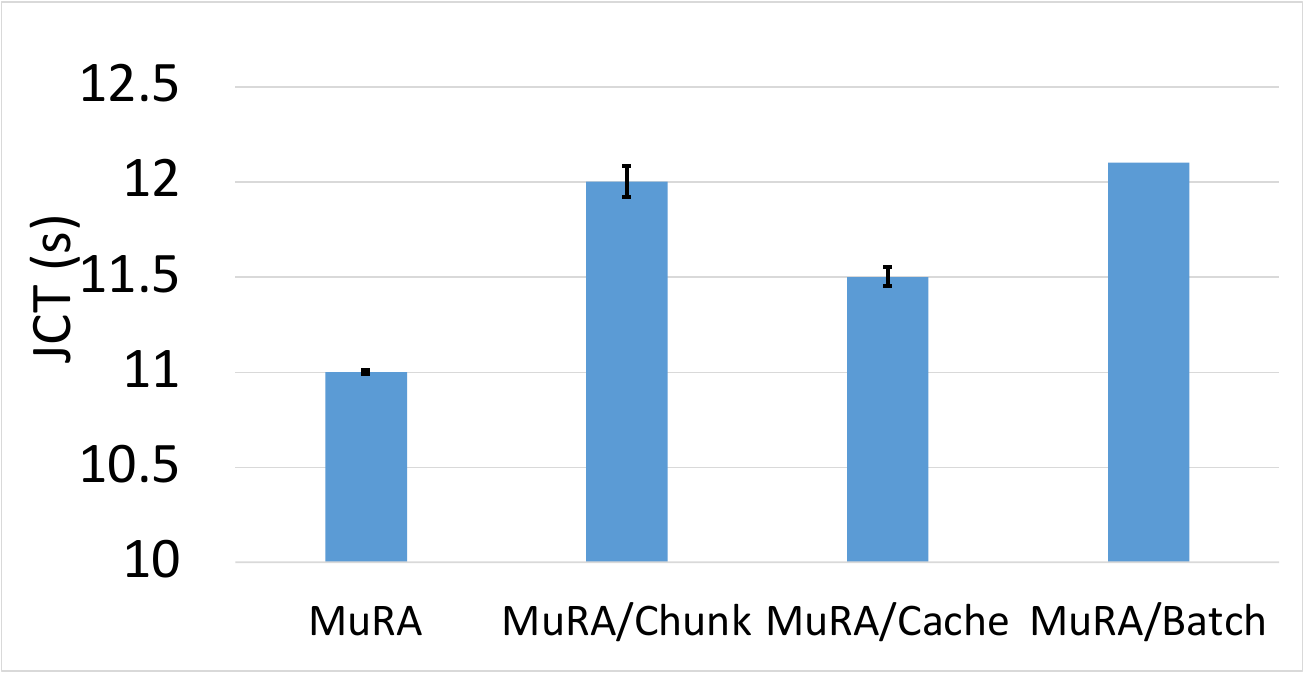} }}
    \hfill
    \subfloat[Throughput and normalized latency.\vspace{-0.01in}\label{fig:th-alpaca-c}]{{\includegraphics[width=0.24\linewidth,height=0.15\textheight]{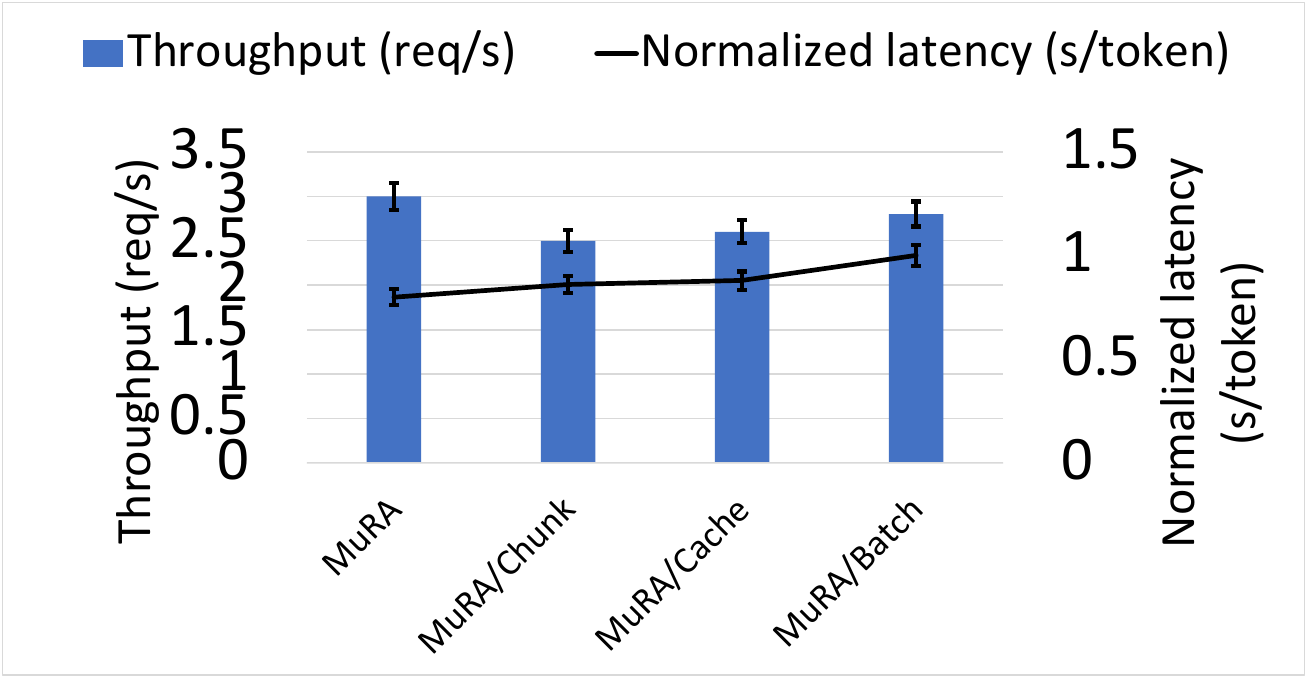} }}
    \hfill
    \subfloat[Throughput in terms of token.\vspace{-0.01in}\label{fig:th-comp-t}]{{\includegraphics[width=0.24\linewidth,height=0.15\textheight]{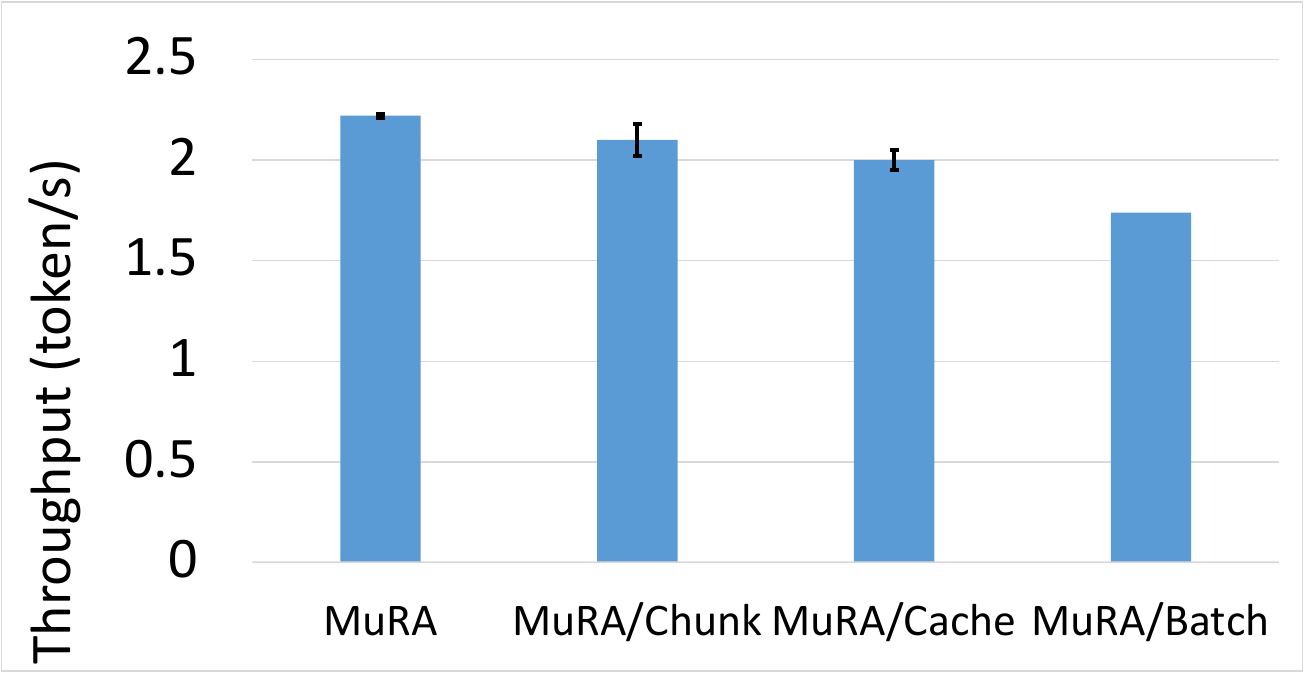} }}
    \hfill
    \subfloat[GPU compute utilization.\vspace{-0.01in}\label{fig:th-gpu-comp}]{{\includegraphics[width=0.24\linewidth,height=0.15\textheight]{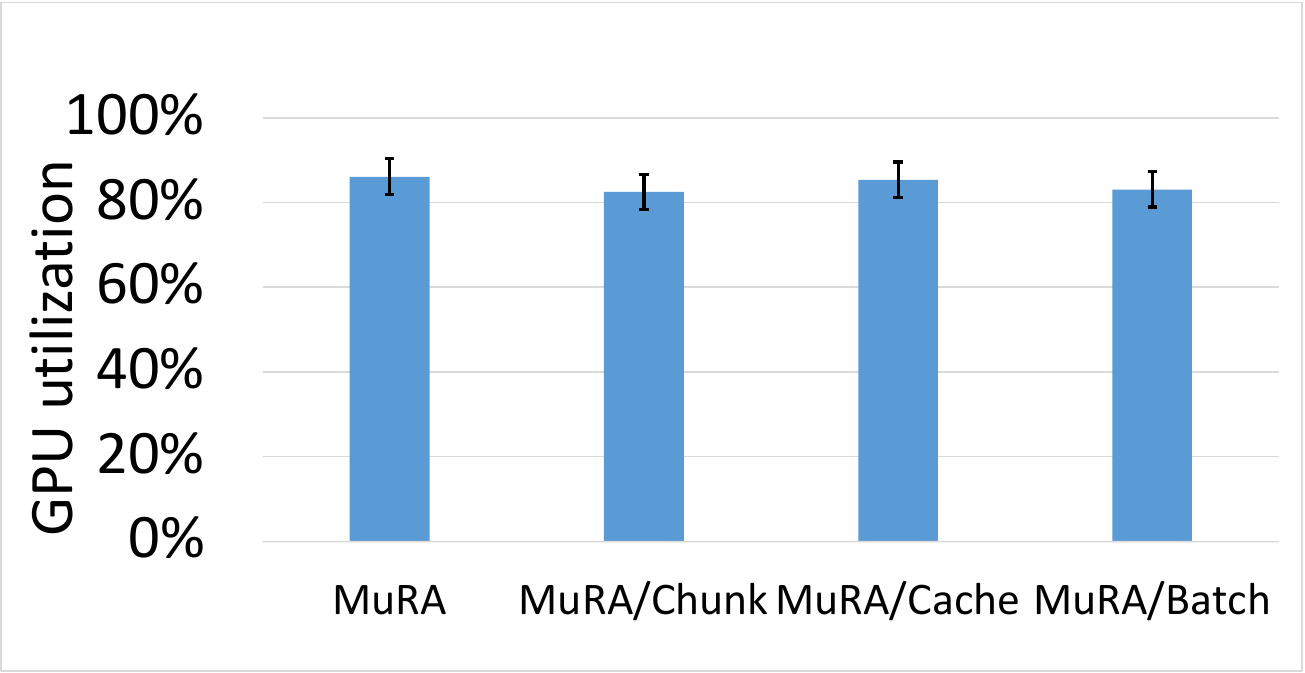} }}
    \hfill
    \subfloat[KVC overflow ratio.\vspace{-0.01in}\label{fig:mem-comp}]{{\includegraphics[width=0.24\linewidth,height=0.15\textheight]{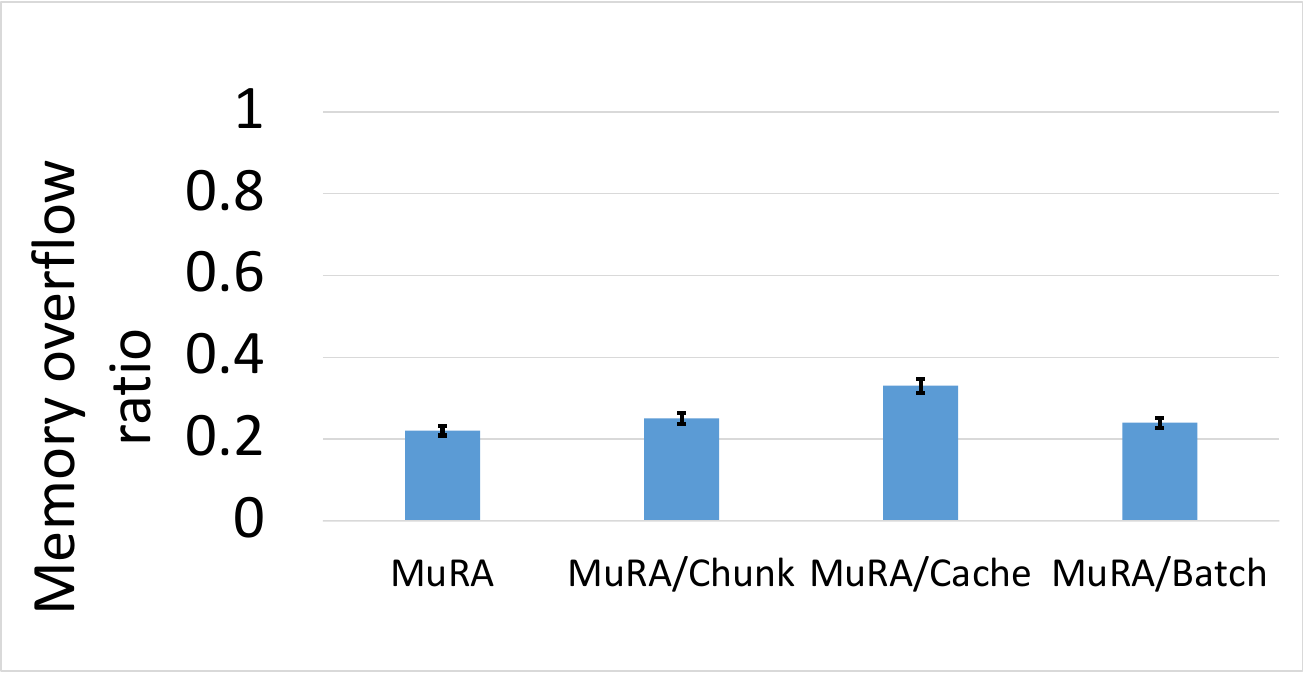} }}
    \hfill
    \subfloat[Overhead. \label{fig:low_cpu1}]{{\includegraphics[width=0.24\linewidth,height=0.15\textheight]{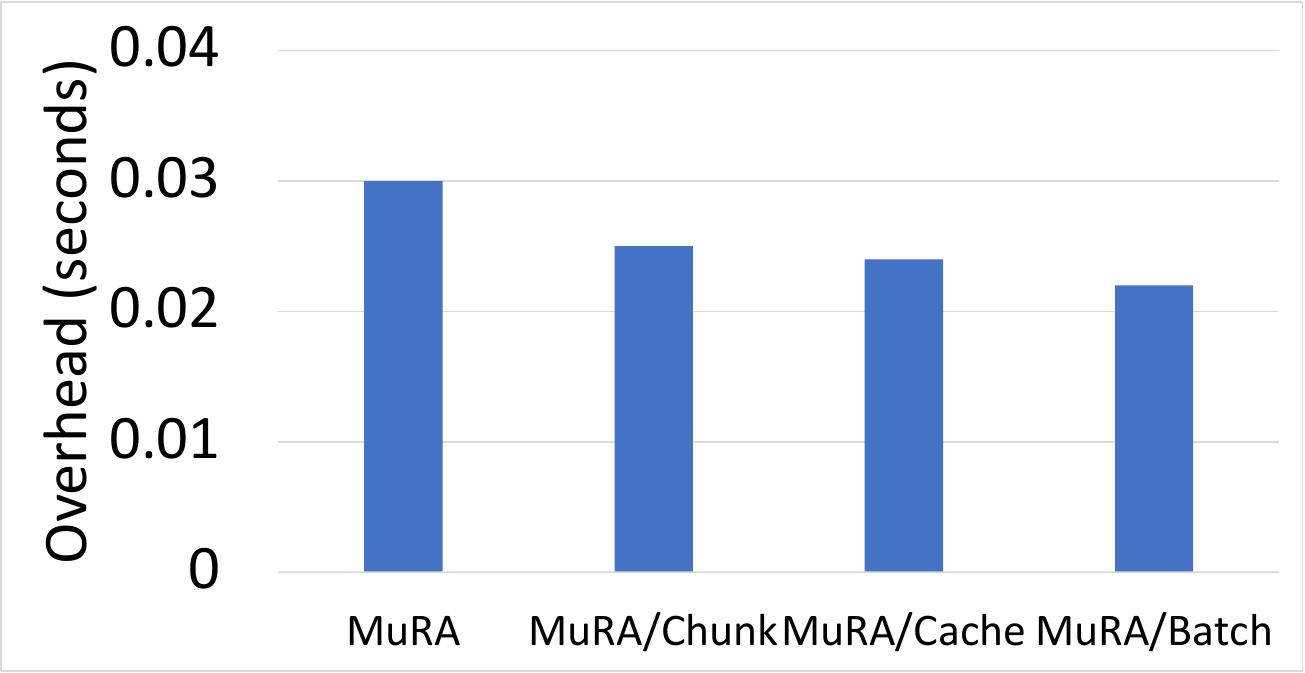} }}%
   \subfloat[Batched requests.\label{fig:low_bw1}]{{\includegraphics[width=0.24\linewidth,height=0.15\textheight]{Fig/batch-mura-components-2.pdf} }}%
\vspace{-0.05in}
   \caption{\small{(Fake) Results of \Sys without a method for 175B model.\vspace{-0.1in}}}
    \label{fig:met-comp-2}
\end{figure*}}


\begin{figure}[]
\vspace{-0.15in}
\centering
\DEL{\subfloat[Density of prompt length upon allocation failure.\vspace{-0.01in}\label{fig:scala-1}]{{\includegraphics[width=0.32\linewidth,height=0.15\textheight]{Fig/14-a.pdf} }}
    \hfill}
\subfloat[Percentage of long prompts.\vspace{-0.01in}\label{fig:scala-1}]{{\includegraphics[width=0.48\linewidth,height=0.15\textheight]{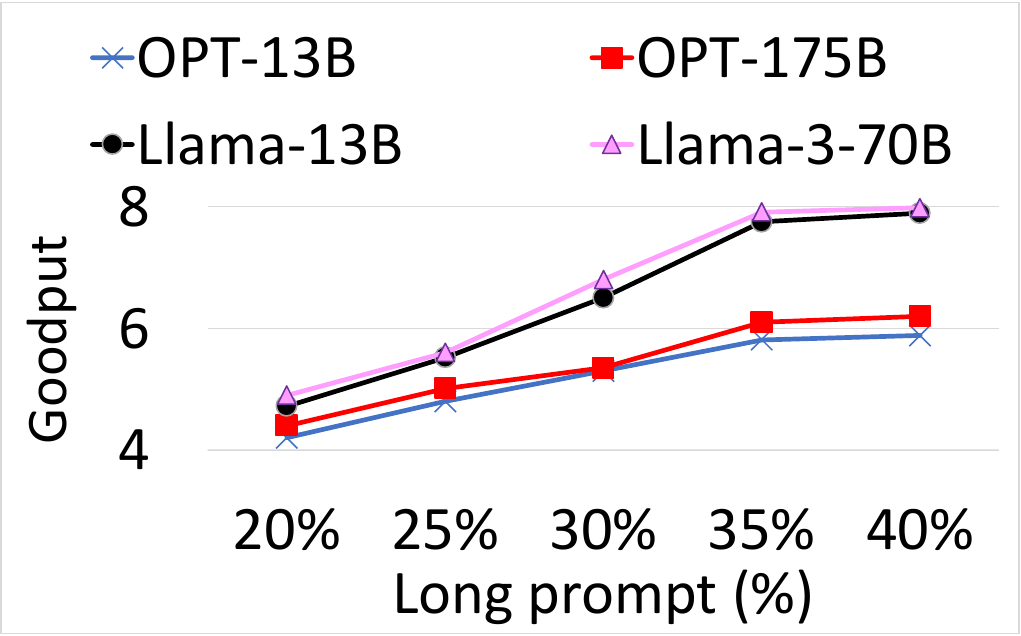} }}
    \hfill
    \subfloat[Arrival rate.\vspace{-0.01in}\label{fig:scala-2}]{{\includegraphics[width=0.48\linewidth,height=0.15\textheight]{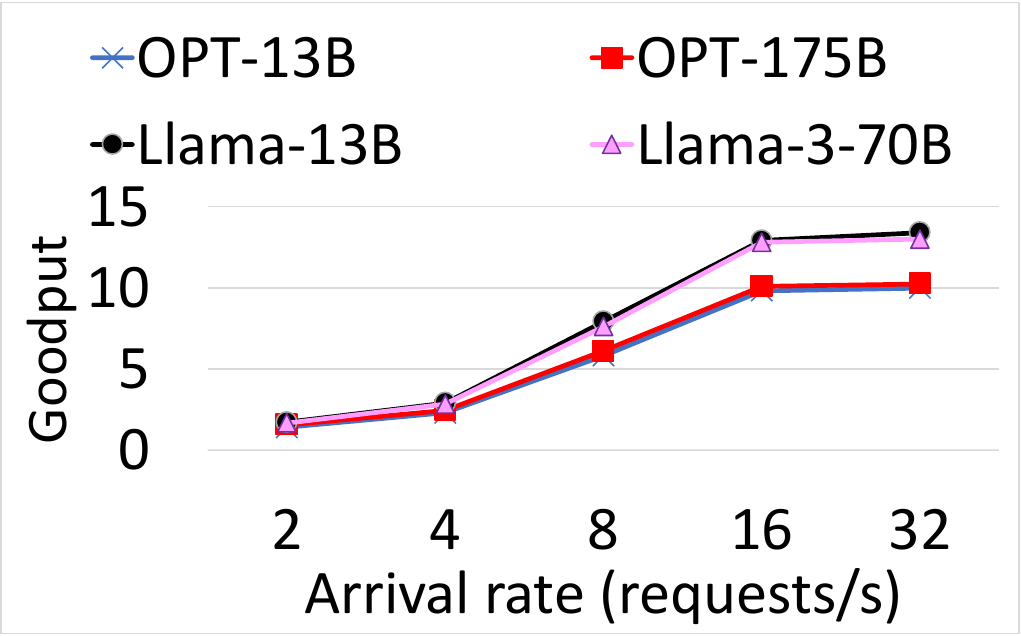} }}
    \hfill
  %
\vspace{-0.0in}
   \caption{Sensitivity testing.}\vspace{-0.15in}%
    \label{fig:scalability-2} \vspace{0.15in}
\end{figure}

\DEL{
\begin{figure}[]
\centering
    \subfloat[Throughput.\vspace{-0.01in}\label{fig:fw2}]{{\includegraphics[width=0.65\linewidth,height=0.15\textheight]{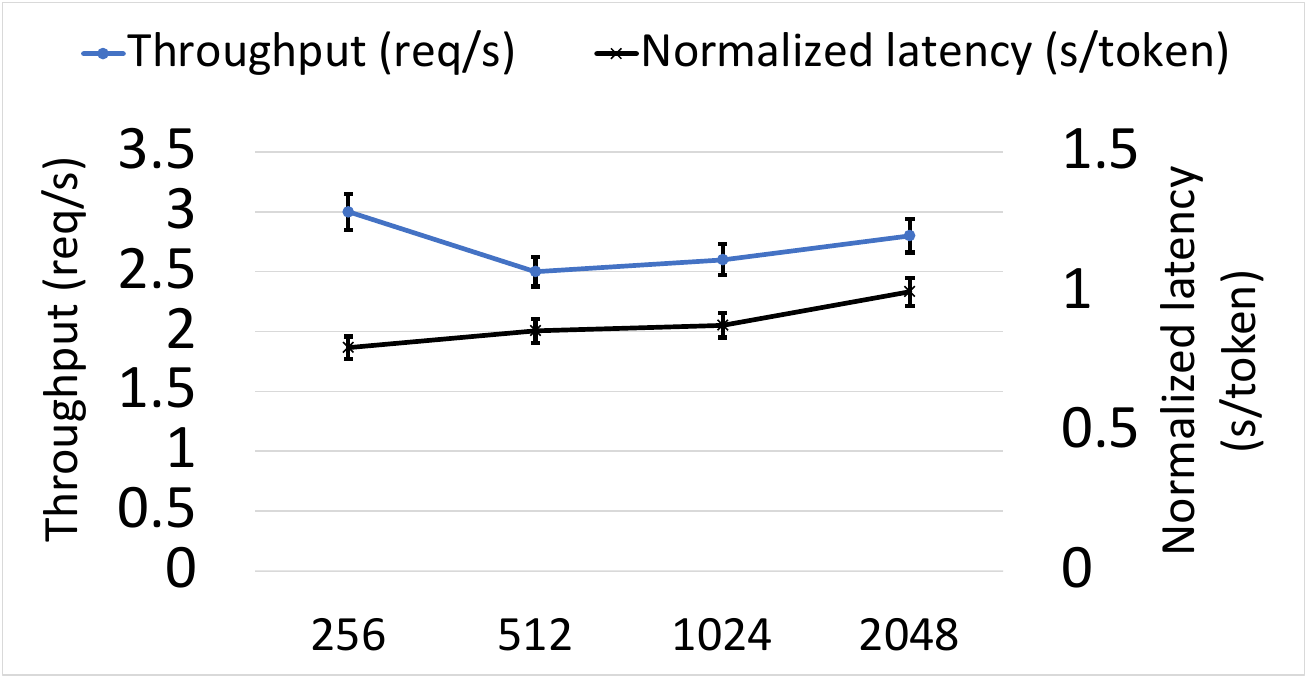} }}
   \hfill

\vspace{-0.05in}
   \caption{\small{(Fake) Scalability testing s(??what is this exp-done, scalability testing for forward size).\vspace{-0.1in}}}
    \label{fig:forward-size}
\end{figure}}

\begin{figure*}[t]
\centering
\subfloat[Alpaca.\vspace{-0.01in}\label{fig:gp-alpaca}]{{\includegraphics[width=0.32\linewidth,height=0.15\textheight]{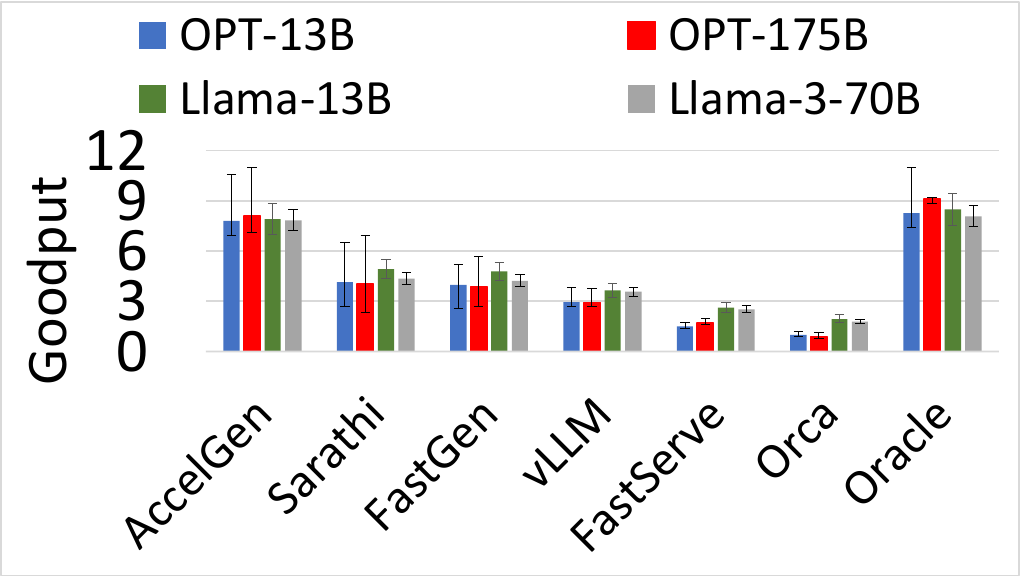} }}
    \hfill
\subfloat[ShareGPT.\vspace{-0.01in}\label{fig:goodput-sharegpt}]{{\includegraphics[width=0.32\linewidth,height=0.15\textheight]{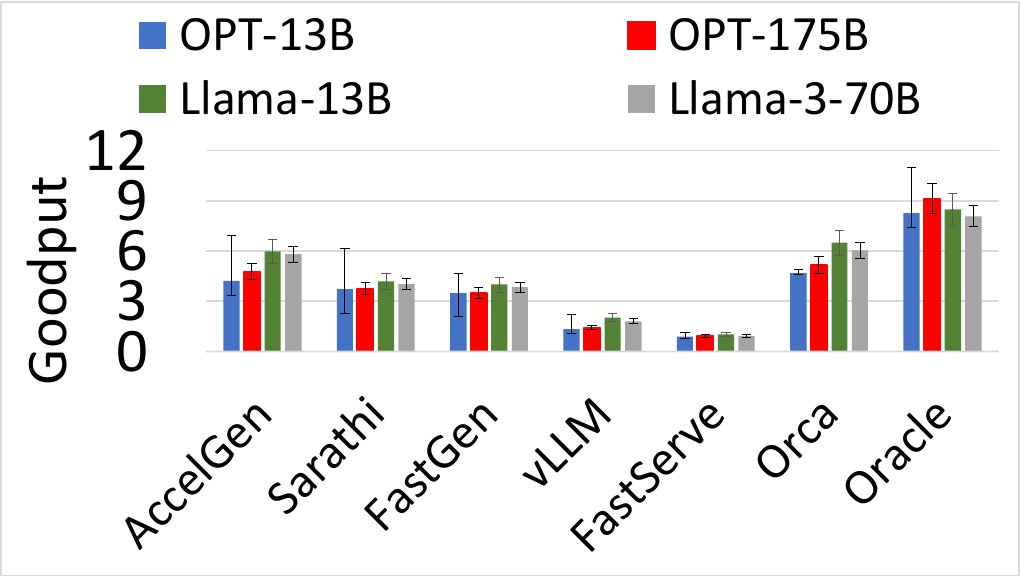} }}
    \hfill
\subfloat[BookCorpus.\vspace{-0.01in}\label{fig:gp-book}]{{\includegraphics[width=0.32\linewidth,height=0.15\textheight]{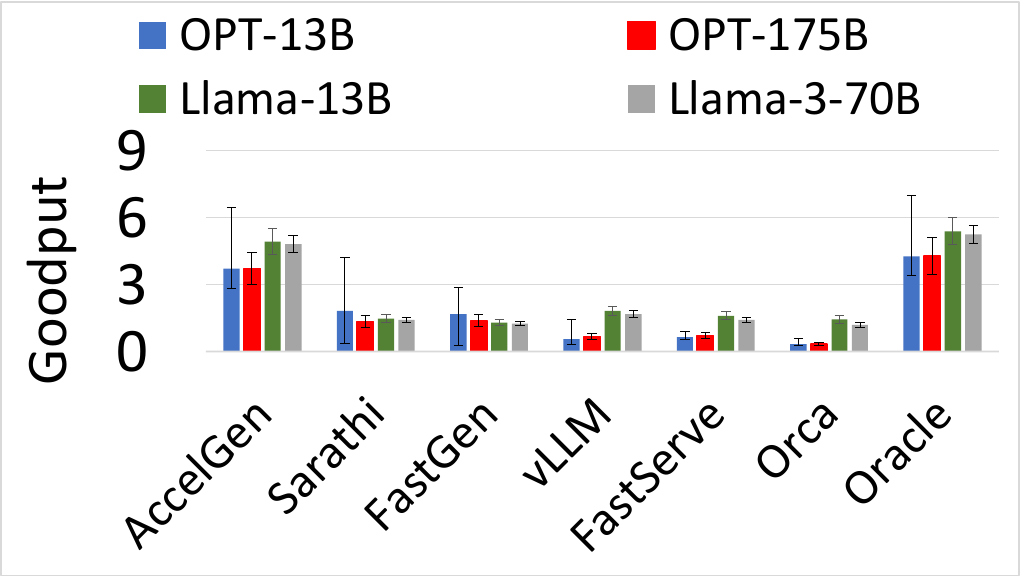} }}
    \hfill
\vspace{-0.05in}
   \caption{Results of \Sys for individual dataset. \vspace{-0.1in}}
    \label{fig:ind-comp}
\end{figure*}

\subsection{Sensitivity Testing}  
Figure~\ref{fig:scalability-2} illustrates the influence of the percentage of long prompts and the arrival rate on the goodput. As the percentage of long prompts and the arrival rate increase, the goodput exhibits a continuous rise, with the increase rate decreasing toward the end as the system approaches its capacity limits. The results indicate \Sys's ability to handle long prompts and high request arrival rates. This is achieved through effective utilization of various prompts, ensuring the full use of both GPU and KVC resources to enhance goodput\looseness=-1. 

\subsection{Performance on Individual Dataset}
Figure~\ref{fig:ind-comp} shows the goodput performance for each of the three datasets. \Sys shows the highest goodput improvement for the long-prompt BookCorpus dataset compared to Alpaca and ShareGPT. \Sys improves the goodput of other systems by 2.57$\times$-33$\times$ for BookCorpus 1.41-18$\times$ and 1.45-19$\times$ for Alpaca and ShareGPT, respectively. 
Long prompts offer greater opportunities to enhance goodput.


\DEL{vLLM by 2.57$\times$ for BookCorpus compared to 1.41$\times$  and 1.45$\times$ for Alpaca and ShareGPT, respectively. It improves the goodput of other methods by 6-33$\times$ for BookCorpus compared to 4.1-18$\times$  and 4.3-19$\times$ for Alpaca and ShareGPT, respectively. }



\section{Limitations and Discussion}

\squishlist


\item  If the number of generated tokens can be accurately predicted, \Sys can be further optimized by fully leveraging this information to approach the performance of \emph{Oracle}. For instance, it can avoid preempting jobs that are nearing completion, leading to better long-term scheduling of KVC usage. We will study this topic.



\item Chunking long prompts has the potential to introduce delays in their processing. We aim to explore adaptive methods for determining the chunking size, ensuring that SLOs for long prompts are met.

\item \Sys currently incurs certain scheduling latency, and we plan to investigate approaches to minimize this latency.

\item Exploring the impact of preempting long prompts and designing a strategy to select requests for preemption based on factors like sequence length and SLO is an important aspect that has not been thoroughly studied in \Sys. We will investigate these topics.

\squishend
\section{Related Work}
\label{sec:related-work}


Job scheduling for transformer-based generative LLM inference systems has recently garnered significant attention.
\Orca~\cite{280922} integrates the scheduler and execution engine, employing iteration-level scheduling and selective batching. Despite these features, \Orca still faces challenges related to memory bottlenecks and GPU under-utilization. 
\DEL{Romero \emph{et al.}~\cite{infaas} proposed INFaaS to automate the model selection, deployment, and serving process. Besides, there are systems such as Clockwork~\cite{clockwork}, and
Shepherd~\cite{shepherd}, specializing in serving relatively small models like ResNet in a cluster. These systems support latency-aware resource allocation to maximize overall throughput.}
Kwon \emph{et al.} proposed vLLM~\cite{vllm}, which uses a block-based approach to overcome the problem of memory bottlenecks. The new vLLM version adopts the chunking method~\cite{vLLM-long}. Sarathi-Serve~\cite{298679} introduces chunked-prefill which splits a prefill request into near equal sized chunks and stall-free schedule that adds new requests into a batch without pausing ongoing decodes. Jin \emph{et al.}~\cite{jin2023s}  addressed GPU under-utilization by proposing $S^3$, a method that predicts the output sequence length for memory allocation. Wu \emph{et al.}~\cite{Wu2023FastDI} proposed FastServe, which uses preemptive scheduling to minimize JCT with a skip-join multi-level feedback queue scheduler. Zheng \emph{et al.}~\cite{Zheng2023Response} proposed a system that predicts the output sequence length of an input sequence and then groups queries with similar response lengths into micro-batches.  
Oh \emph{et al.}~\cite{3640383} proposed \emph{ExeGPT} which 
finds the execution schedule that maximizes
inference throughput under a given latency constraint. The execution schedule includes the batch size,  GPU count, and tensor parallelism degree. Splitwise~\cite{Patel2023SplitwiseEG} and DistServe~\cite{distserve} assign prefill and decoding computation to different GPUs in order to maximize the throughput. Sheng \emph{et al.}~\cite{Fairness-OSDI-2024} considered fairness in scheduling based on a cost function that accounts for the number of input and output tokens processed in the inference server from each client. Zhao \emph{et al.}~\cite{Zhao2024ALISAAL} introduced ALISA, a solution that combines a Sparse Window Attention algorithm to reduce the memory footprint of KVC with a three-phase token-level dynamic scheduling system. Llumnix~\cite{llumnix} reschedules heterogeneous and unpredictable requests among multiple model instances to reduce tail latency.


Many other approaches have been proposed to improve performance. Sheng \emph{et al.}~\cite{sheng2023high} proposed FlexGen, which addresses memory bottlenecks by solving a linear programming problem to identify efficient patterns for storing and accessing tensors. Liu \emph{et al.}~\cite{pmlr-v202-liu23am} proposed asynchronous lookahead predictors that predict the sparsity
for the attention head at the next layer and
optimize the operations to speed up LLM inference in
real time. Zheng~\emph{et al.}\cite{Radix-attention} proposed KVC reuse by storing KVC in a radix tree for multiple requests. Liu \emph{et al.}\cite{liu2023scissorhands} proposed compressing KVC by storing pivotal tokens based on attention scores. Compiler systems like those in~\cite{pit,280848} address dynamic sparsity in LLM models by constructing GPU-efficient tiles to increase GPU compute utilization and low memory wastage. 



However, none of the aforementioned schedulers handle the heterogeneous SLOs in the mixed-prompt scenarios. \Sys addresses this and aims to simultaneously maximize both the GPU compute and KVC utilization to improve throughput and goodput. 





\DEL{\noindent{\textbf{Processors of Transformer Models.}}
\tannsdi{Specialized execution engines~\cite{fastertransformer,deepspeed-inf,megatron,fang-turbo} for transformer models serve the inference requests scheduled by the existing systems~\cite{fang2021deployment,46801,mii}.\looseness=-1}}

\DEL{FasterTransformer~\cite{fastertransformer,fang-turbo}, DeepSpeed-Inference~\cite{deepspeed-inf}, Megatron-LM~\cite{megatron}  and  TurboTransformers~\cite{fang-turbo} serve the inference requests scheduled by the existing scheduling systems~cite{}.
behave as the backend execution engines or model processors for existing scheduling systems and serve the inference requests. 
Nvidia Triton Inference Server~\cite{fang2021deployment},  TensorFlow Serving~\cite{46801}, and Deepspeed-MII~\cite{mii} are scheduling systems that use request-level scheduling as the scheduling mechanism.}

\DEL{\section{Limitations and Future Work}

\begin{itemize}
\item This work is for the case that the number of generated tokens for a request cannot be accurately predicted. If it can be accurately predicted, then \Sys can be enhanced to fully utilize this information. For example, this information can be used to allocate KVC for a request, to enable the request-level SLO.

\tannsdi{\item A customized cuda kernel could help reduce the scheduling latency. We will fuse the cuda kernel in the future to make the system more robust.}

\end{itemize}}

\section{Conclusion}\label{sec:conclusion}
Existing schedulers for transformer-based LLM inference serving systems often fail to meet heterogeneous SLOs of different requests and maximize throughput in the mixed-prompt scenarios. To tackle this issue, we conducted a thorough measurement analysis of the performance of existing LLM inference systems in this scenario. Drawing from our observations, we propose \Sys, which offers iteration-level SLO guarantees and maximizes both GPU compute and KVC resources in scheduling. Our trace-driven experiments show the superior performance of \Sys in meeting heterogeneous SLOs hence delivering satisfactory user experience and improving throughput, goodput and JCT compared to the state-of-the-art approaches. Currently, \Sys is only tested for the mentioned models and the GPU settings. We will test it on more different models and machine settings.



\phantomsection
\label{EndOfPaper}

\bibliographystyle{unsrt}
\bibliography{myBib,myBib-2,nlp}



\end{document}